\documentclass[pdflatex,sn-apa]{sn-jnl-preprint}

\usepackage{xspace}
\usepackage{thmtools}
\usepackage{subcaption}
\newcommand{\Bbf}{\ensuremath{\mathbf{B}}}
\newcommand{\Cbf}{\ensuremath{\mathbf{C}}}
\newcommand{\Ibf}{{\mathbf I}}
\newcommand{\Pbf}{\textbf{\bf P}}
\newcommand{\vbf}{\mathbf{v}}
\newcommand{\wbf}{\mathbf{w}}
\newcommand{\xbf}{\ensuremath{\mathbf{x}}}
\newcommand{\epsilonbf}{\boldsymbol{\epsilon}}
\newcommand{\omegabf}{\boldsymbol{\omega}}
\newcommand{\zerobf}{{\mathbf 0}}

\newcommand{\Acal}{\ensuremath{\mathcal A}}
\newcommand{\Dcal}{\ensuremath{\mathcal D}}
\newcommand{\Hcal}{\ensuremath{\mathcal H}}
\newcommand{\Ncal}{\ensuremath{\mathcal N}}
\newcommand{\Mcal}{\ensuremath{\mathcal M}}
\newcommand{\Pcal}{\ensuremath{\mathcal P}}
\newcommand{\Qcal}{\ensuremath{\mathcal Q}}
\newcommand{\Scal}{\ensuremath{\mathcal S}}
\newcommand{\Tcal}{\ensuremath{\mathcal T}}
\newcommand{\Ucal}{\ensuremath{\mathcal U}}
\newcommand{\Xcal}{\ensuremath{\mathcal X}}
\newcommand{\Ycal}{\ensuremath{\mathcal Y}}
\newcommand{\Zcal}{\ensuremath{\mathcal Z}}

\newcommand{\Rbb}{\R}
\newcommand{\Nbb}{\mathbb{N}}

\DeclareMathOperator*{\argmin}{\mathrm{argmin}}
\DeclareMathOperator*{\esssup}{\mathrm{esssup}}
\DeclareMathOperator*{\EE}{\mathbb{E}}
\DeclareMathOperator*{\PP}{\mathbb{P}}

\newcommand{\ie}{{\em i.e.\/}}
\newcommand{\eg}{{\em e.g.\/}}

\newcommand{\LB}{\left[}
\newcommand{\LC}{\left\{}

\newcommand{\LN}{\left\|}
\newcommand{\LP}{\left(}
\newcommand{\LV}{\left\vert}

\newcommand{\RB}{\right]}
\newcommand{\RC}{\right\}}

\newcommand{\RN}{\right\|}
\newcommand{\RP}{\right)}
\newcommand{\RV}{\right\vert}

\newcommand{\prior}{\Pcal}
\renewcommand{\Q}{\Qcal}
\renewcommand{\P}{\Pcal}
\newcommand{\AQ}{\Q_{\Scal}}
\newcommand{\AQprime}{\Q_{\Scal'}}
\renewcommand{\R}{\mathbb{R}}
\newcommand{\D}{D}
\newcommand{\Risk}{{R}}
\newcommand{\KL}{{\rm KL}}
\newcommand{\kl}{{\rm kl}}

\newcommand{\Rpe}{\R_{+}^{*}}

\newcommand{\ours}{{\small\sc ours}\xspace}
\newcommand{\rivasplata}{{\small\sc rivasplata}\xspace}
\newcommand{\blanchard}{{\small\sc blanchard}\xspace}
\newcommand{\catoni}{{\small\sc catoni}\xspace}
\newcommand{\stoNN}{{\small\sc stochastic}\xspace}

\makeatletter
\newcommand{\customlabel}[2]{%
 \@bsphack\begingroup
 \def\@currentlabel{#2}%
 \label{#1}%
 \endgroup\@esphack
}
\makeatother
    
\jyear{2021}%

\theoremstyle{thmstyleone}%
\newtheorem{theorem}{Theorem}
\newtheorem{corollary}[theorem]{Corollary}%
\newtheorem{lemma}[theorem]{Lemma}

\theoremstyle{thmstyletwo}%

\theoremstyle{thmstylethree}%

\begin{document}

\title[A General Framework for the Disintegration of PAC-Bayesian Bounds]{A General Framework for the Practical Disintegration of PAC-Bayesian Bounds}


\author*[1]{\fnm{Paul} \sur{Viallard}}\email{paul.viallard@inria.fr}
\equalcont{This work was done when P. Viallard  was affiliated to Laboratoire Hubert Curien}

\author[2]{\fnm{Pascal} \sur{Germain}}\email{pascal.germain@ift.ulaval.ca}

\author[3,4]{\fnm{Amaury} \sur{Habrard}}\email{amaury.habrard@univ-st-etienne.fr}

\author[3]{\fnm{Emilie} \sur{Morvant}}\email{emilie.morvant@univ-st-etienne.fr}

\affil[1]{\orgdiv{Inria, CNRS, Ecole Normale Supérieure}, \orgname{PSL Research University}, \orgaddress{Paris, \country{France}}}

\affil[2]{\orgdiv{D\'epartement d'informatique et de g\'enie logiciel}, \orgname{Universit\'e Laval}, \orgaddress{\state{Qu\'ebec}, \country{Canada}}}

\affil[3]{\orgdiv{Université Jean Monnet Saint-Étienne, CNRS, Institut d Optique Graduate School}, \orgname{Laboratoire Hubert Curien UMR 5516}, \orgaddress{F-42023, SAINT-ÉTIENNE, \country{FRANCE}}}

\affil[4]{\orgdiv{Institut Universitaire de France (IUF)}}

\abstract{PAC-Bayesian bounds are known to be tight and informative when studying the generalization ability of randomized classifiers.
However, they require a loose and costly derandomization step when applied to some families of deterministic models such as neural networks.
As an alternative to this step, we introduce new PAC-Bayesian generalization bounds that have the originality to provide \textit{disintegrated} bounds, {\it i.e.}, they give guarantees over one \textit{single} hypothesis instead of the usual {\it averaged} analysis.
Our bounds are easily optimizable and can be used to design learning algorithms.
We illustrate this behavior on neural networks, and we show a significant practical improvement over the state-of-the-art framework.}


\maketitle

\section{Introduction}
\label{sec:intro}

In statistical learning theory, PAC-Bayesian theory\footnote{The reader can refer to~\citet{Guedj2019}~or~\citet{Alquier2021} for recent surveys on PAC-Bayes.}~\citep{ShaweTaylorWilliamson1997,McAllester1998}
provides a powerful framework 
for analyzing the generalization ability of machine learning models such as linear classifiers~\citep{GermainLacasseLavioletteMarchand2009}, SVM~\citep{AmbroladzeHernandezShaweTaylor2006}, or neural networks~\citep{DziugaiteRoy2017,PerezOrtizRivasplataShaweTaylorSzepesvari2020}.
In the PAC-Bayesian theory, the machine learning models are considered {\it randomized} (or {\it stochastic}), \ie, a model is sampled from a \textit{posterior} probability distribution for each prediction.
The analysis of such a randomized classifier usually takes the form of bounds on the average risk with respect to a learned \textit{posterior} distribution given a learning sample and a chosen \textit{prior} distribution defined over a set of hypotheses.
Note that the prior distribution can encode an \textit{a priori} belief on the set of hypotheses, or if we have no belief, it can be set to a non-informative distribution, such as the uniform distribution.
While such bounds are very effective for analyzing randomized/stochastic classifiers, the vast majority of machine learning methods nevertheless need guarantees on deterministic models. 
In this case, a {\it derandomization step} of the bound is required to get a bound on the risk of the deterministic model.
In general, the {\it derandomization step} consists in obtaining a bound on the risk of a deterministic model from a bound that is originally for randomized/stochastic models.
Different forms of derandomization have been introduced in the literature for specific settings.
Among them, \citet{LangfordShaweTaylor2002} proposed a derandomization for Gaussian posteriors over linear classifiers: thanks to the Gaussian symmetry, a bound on the risk of the \emph{maximum a posteriori} (deterministic) classifier is obtainable from the bound on the average risk of the randomized classifier. 
Also relying on Gaussian posteriors, \citet{LetarteGermainGuedjLaviolette2019} derived a PAC-Bayesian bound for a very specific deterministic network architecture using sign functions as activations; this approach has been further extended by \citet{BiggsGuedj2021,BiggsGuedj2022}.
Another line of works derandomizes neural networks~\citep{NeyshaburBhojanapalliSrebro2018, NagarajanKolter2019}.
While technically different, it starts from PAC-Bayesian guarantees on the randomized classifier and uses an ``output perturbation'' bound to convert
a guarantee from a random classifier to the mean classifier.
These works highlight the need for a general framework for the derandomization of classic PAC-Bayesian bounds.\\

\noindent{}In this paper, we focus on another kind of derandomization, sometimes referred to as {\it disintegration of the PAC-Bayesian bound}, and first proposed by \citet[Th.1.2.7]{Catoni2007} and \citet{BlanchardFleuret2007}: 
instead of bounding the {\it average risk of a randomized} classifier with respect to the posterior distribution, the {\it disintegrated PAC-Bayesian bounds} upper-bound the {\it risk of a sampled} (unique) classifier from the posterior distribution.
Despite their interest in derandomizing PAC-Bayesian bounds, these kinds of bounds have only received little study in the literature; especially, we can cite the recent work of~\citet[Th.1{\it(i)}]{RivasplataKuzborskijSzepesvariShaweTaylor2020} who have derived a general disintegrated PAC-Bayesian theorem.
It is important to note that these bounds have never been used in practice.
Driven by machine learning practical purposes, our objective is thus twofold.
We derive new tight and usable {\it disintegrated} PAC-Bayesian bounds {\it (i)} that directly derandomize any classifiers without any other additional step and with \textit{almost} no impact on the guarantee,  and {\it (ii)} that can be easily optimized to learn classifiers with strong guarantees.
To achieve this objective, our contribution consists in providing a new general disintegration framework based on the Rényi divergence (in Theorem~\ref{theorem:disintegrated}), allowing us to meet the practical goal of efficient learning. 
From the theoretical standpoint, due to the Rényi divergence term, our bound is expected to be looser than the one of \citet[Th.1{\it(i)}]{RivasplataKuzborskijSzepesvariShaweTaylor2020} in which the divergence term is ``disintegrated'' but depends on the sampled hypothesis only.
However, as we show in our experimental evaluation on neural networks, their ``disintegrated'' term is, in practice, subject to high variance, making their bound harder to optimize. 
This variance arises because the sampled hypothesis does not influence our Rényi divergence term.
Our bound has then the main advantage of leading to a more stable learning algorithm  with better empirical results.
In addition, we derive a new theoretical result in the form of an information-theoretic bound, giving new insights into disintegration procedures.\\

\noindent{}The rest of the paper is organized as follows.
Section~\ref{sec:setting} introduces the notations we follow and recalls some basics on generalization bounds.
In Section~\ref{sec:contrib}, we derive our main contribution relying on {\it disintegrated} PAC-Bayesian bounds.
Then, we illustrate the practical usefulness of this disintegration on deterministic neural networks in Section~\ref{sec:expe}.
Before concluding in Section~\ref{sec:conclu}, we discuss in Section~\ref{sec:info-theoretic} another point of view of the disintegrated through an information-theoretic bound.
For readability, we deferred the proofs of our theoretical results to the Appendix.
\section{Setting and basics}
\label{sec:setting}
\subsection{General notations} 
We denote by $\Mcal(\Acal)$ the set of probability densities on the measurable space $(\Acal, \Sigma_{\Acal})$ with respect to a reference measure\footnote{The measure considered for $(\Acal, \Sigma_{\Acal})$ is usually the Lebesgue or the counting measure.} where $\Sigma_{\Acal}$ is the $\sigma$-algebra on the set $\Acal$.
In this paper, we consider supervised classification tasks with $\Xcal$  the \emph{input space}, $\Ycal$  the \emph{label set}, and $\Dcal\in\Mcal(\Xcal{\times}\Ycal)$  an unknown \emph{data distribution} on $\Xcal {\times} \Ycal{=}\Zcal$.
An \emph{example} is denoted by $z{=} (\xbf,y)\! \in\! \Zcal$, and  the \emph{learning sample}  $\Scal{=} \{z_i\}_{i=1}^{m}$ is constituted by $m$ examples drawn {\it i.i.d.} from $\Dcal$; the distribution of such an \mbox{$m$-sample} being $\Dcal^m\in\Mcal(\Zcal^m)$.
We consider a \emph{hypothesis set} $\Hcal$ of functions $h\!:\!\Xcal {\to} \Ycal$.
The learner aims to find  $h\!\in\!\Hcal$ that assigns a label $y$ to an input $\xbf$ as accurately as possible. 
Given an example $z$ and a hypothesis $h$, we assess the quality of the prediction of $h$ with a {\it loss function} $\ell\!:\! \Hcal{\times} \Zcal {\to} [0, 1]$ evaluating to which extent the prediction is accurate.
Given a loss function $\ell$, the {\it true risk} $\Risk_{\Dcal}(h)$ of a hypothesis $h$ on the distribution $\Dcal$ and its empirical counterpart, \textit{the empirical risk}, $\Risk_{\Scal}(h)$ estimated on $\Scal$ are defined as
\[
\Risk_{\Dcal}(h) 
\triangleq \EE_{z\sim \Dcal}\ell(h, z)\,, \quad\text{ and } \quad 
\Risk_{\Scal}(h) 
\triangleq \frac{1}{m}\sum_{i=1}^{m} \ell(h, z_{i})\,.
\]
Then, the learner wants to find the hypothesis $h$ from $\Hcal$ that minimizes $\Risk_{\Dcal}(h)$. 
However, we cannot compute $\Risk_{\Dcal}(h)$ since $\Dcal$ is unknown.  
In practice, one could work under the Empirical Risk Minimization  principle ({\sc erm}) that looks for a hypothesis minimizing $\Risk_{\Scal}(h)$.
Generalization guarantees over unseen data from $\Dcal$ can be obtained by quantifying how much the empirical risk $\Risk_{\Scal}(h)$ is a good estimate of $\Risk_{\Dcal}(h)$. 
Statistical machine learning theory~\citep[see, \eg,][]{Vapnik2000} studies the conditions of consistency and convergence of {\sc erm} towards the true risk.
This kind of result is called \textit{generalization bound}, often referred to as PAC (Probably Approximately Correct) bound~\citep{Valiant1984}, and takes the form:
\begin{align*}
\PP_{\Scal\sim\Dcal^{m}} \Big[ \big\vert \Risk_{\Dcal}(h) -\Risk_{\Scal}(h)
\big\vert 
\leq 
\varepsilon\big( \tfrac1\delta, \tfrac1m\big) \Big]
\geq 1-\delta.
\end{align*}
Put into words, with high probability (at least $1{-}\delta$) on the random choice of the learning sample $\Scal$, good generalization guarantees are obtained when the deviation between the true risk $\Risk_{\Dcal}(h)$  and its empirical estimate $\Risk_{\Scal}(h)$ is low, {\it i.e.}, $ \varepsilon\big(\tfrac{1}{\delta}, \tfrac{1}{m}\big)$ should be as small as possible. 
The function $\varepsilon$  depends mainly on two quantities: {\it (i)} the number of examples $m$ for statistical precision, and \mbox{{\it (ii)} the} confidence parameter $\delta$.
We now recall three classical bounds while focusing on the PAC-Bayesian theory at the heart of our contribution. 
By abuse of notation, in the following, we use the function $\varepsilon$ for the different presented frameworks: we consider an additional argument of $\varepsilon$ to pinpoint the differences between the frameworks.

\subsection{Uniform convergence bound} 
A first classical type of generalization bounds is referred to as {\it Uniform Convergence} bounds based on a measure of complexity of the set $\Hcal$ (such as the VC-dimension or the Rademacher complexity) and hold for all the hypotheses of $\Hcal$.
This type of bound takes the form:
\begin{align*}
\PP_{\Scal\sim\Dcal^{m}}\LB\,  \sup_{h\in\Hcal}\LV \Risk_{\Dcal}(h)-\Risk_{\Scal}(h)\RV \le \varepsilon\big(\tfrac{1}{\delta}, \tfrac{1}{m},\Hcal\big) \RB\ge 1-\delta.
\end{align*}
Due to $\sup_{h\in\Hcal}$, this bound can be seen as a {\it worst-case} analysis.
Indeed, it means that the bound $\LV\Risk_{\Dcal}(h)-\Risk_{\Scal}(h)\RV \le \varepsilon\big(\tfrac{1}{\delta}, \tfrac{1}{m},\Hcal\big)$  holds with a high probability for all $h\!\in\!\Hcal$, including the best but also the worst.
This {\it worst-case} analysis makes it hard to obtain a non-vacuous bound
\ie, with \mbox{$\varepsilon(\frac{1}{\delta}, \frac{1}{m}, \Hcal)<1$}. 
Note that the ability of such bounds to explain the generalization of deep learning has been recently \mbox{challenged~\citep{NagarajanKolter2019b}}.

\subsection{Algorithmic-dependent bounds}
A potential drawback of the Uniform Convergence bounds is that they are independent of the learning algorithm, \ie, they do not take into account the way the hypothesis space is explored.
To tackle this issue, algorithmic-dependent bounds have been proposed to take advantage of some particularities of the learning algorithm, such as its uniform stability~\citep{BousquetElisseeff2002} or robustness~\citep{XuMannor2012}.
In this case, the bounds obtained hold for a single hypothesis $h_{L(\Scal)}$, the one learned with the algorithm $L$ from the learning sample $\Scal$. 
The form of such bounds is:
\begin{align*}
\PP_{\Scal\sim\Dcal^{m}} \Big[ \LV \Risk_{\Dcal}(h_{L(\Scal)}){-}\Risk_{\Scal}(h_{L(\Scal)})\RV \le \varepsilon\big(\tfrac{1}{\delta}, \tfrac{1}{m}, L
\big) \Big]\ge 1-\delta.
\end{align*}
For example, this approach has been used by~\citet{HardtRechtSinger2016} to derive generalization bounds for hypotheses learned by stochastic gradient descent.

\subsection{PAC-Bayesian bound} 
This paper leverages PAC-Bayesian bounds that stand in the PAC framework but borrows inspiration from the Bayesian probabilistic view that deals with randomness and uncertainty in machine learning~\citep{McAllester1998}.
In the PAC-Bayesian setting, we consider a {\it prior} distribution \mbox{$\P\!\in\!\Mcal^{*}(\Hcal)\subseteq\Mcal(\Hcal)$} on $\Hcal$, with $\Mcal^{*}(\Hcal)$ the set of strictly positive probability densities.
This distribution encodes an \textit{a priori} belief on $\Hcal$ before observing the learning sample~$\Scal$.
Then, given $\Scal$ and the prior $\P$, we learn a {\it posterior} distribution \mbox{$\Q\!\in\!\Mcal(\Hcal)$}.
In this case, the bounds take the form:
\begin{align*}
\PP_{\Scal\sim\Dcal^{m}}\!\Big[\forall  \Q  \in\Mcal(\Hcal),\quad \EE_{h\sim \Q}\!\LV \Risk_{\Dcal}(h){-}\Risk_{\Scal}(h)\RV{\le}\,\varepsilon\big(\tfrac{1}{\delta}{,}\tfrac{1}{m}{,}\Q\big) \Big]\ge 1-\delta.
\end{align*}
A key notion is that the function $\varepsilon()$ upper-bounds a {\it $\Q$-weighted expectation} over the risks of all classifiers in $\Hcal$.
Hence, it upper-bounds the risk of a {\it randomized classifier}.\footnote{The risk of the randomized classifier $\EE_{{h\sim\Q}}\Risk_{\Dcal}(h)$ is sometimes referred to as the Gibbs risk in the PAC-Bayes literature.}
Such a randomized classifier can be described as follows: to predict the label of an input $\xbf\in\Xcal$, {\it (i)}~a hypothesis $h\in\Hcal$ is sampled from~$\Q$ and {\it (ii)}~the classifier predicts the label given by $h(\xbf)$.

\noindent{}We recall below the classical PAC-Bayesian bounds in a general form as proposed by \citet{GermainLacasseLavioletteMarchand2009,BeginGermainLavioletteRoy2016}.
The idea is to express the bound in terms of a generic function $\phi\!:\! \Hcal{\times}\Zcal^{m}{\rightarrow} \Rpe$ that is meant to capture the 
the deviation between the true and the empirical risks, instead of deriving a theorem by settling on a specific measure of deviation such as $\vert \Risk_{\Dcal}(h){-}\Risk_{\Scal}(h)\vert$.
Note that, Theorem~\ref{theorem:general-classic} is expressed in a slightly different form than the original ones; we prove Theorem~\ref{theorem:general-classic} in Appendix~\ref{ap:proof-classic} for the sake of completeness.

\begin{restatable}[General PAC-Bayes bounds]{theorem}{theoremclassical}
For any distribution $\Dcal$ on $\Zcal$, for any hypothesis set $\Hcal$, for any prior distribution $\P\in\Mcal^{*}(\Hcal)$ on $\Hcal$, for any measurable function $\phi\!:\! \Hcal{\times}\Zcal^{m}{\rightarrow} \Rpe$, for any $\delta\in(0,1]$ we have 
\allowdisplaybreaks[4]

\begin{align}
&\underbrace{\PP_{\Scal{\sim}\Dcal^{m}}\!\!\LP\!\begin{array}{l}
\forall \Q\in\Mcal(\Hcal),\\
{\displaystyle\EE_{h\sim\Q}\!\ln(\phi(h,\!\Scal)) \le \KL( \Q\|\P) 
     \!+\! \ln\!\left[\frac{1}{\delta}\!\EE_{\Scal{\sim}\Dcal^{m\!}}\EE_{h{\sim}\P}\phi(h,\!\Scal)\right]}
\end{array}
\!\RP \!\ge\! 1{-}\delta}_{ \text{\small\citep{GermainLacasseLavioletteMarchand2009}}},\label{eq:kl-bound-classic}\\
&\mbox{and}\nonumber \\
&\underbrace{\PP_{\Scal{\sim}\Dcal^{m}}\!\!\LP\!\begin{array}{l}
\forall \Q\in\Mcal(\Hcal),\\
{\displaystyle\tfrac{\alpha}{\alpha{-}1}\!
   \ln\! \left[\EE_{h{\sim}\Q}\!\phi(h,\!\Scal)\right]\le \D_{\alpha}(\Q\|\P) 
    \!+\! \ln\!\left[\frac{1}{\delta}\!\EE_{\Scal{\sim}\Dcal^{m\!}}\EE_{h{\sim}\P}\!\phi(h,\!\Scal)^\frac{\alpha}{\alpha{-}1}\right]\!}
\end{array}
\RP \! \ge\! 1{-}\delta}_{\text{\small\citep{BeginGermainLavioletteRoy2016}}},\nonumber\\
\label{eq:renyi-bound-classic}
\end{align}
with $\KL(\Q\|\P){\triangleq}\text{\small${\displaystyle \EE_{h{\sim}\Q}}$} \ln\tfrac{\Q(h)}{\P(h)}$  the Kullback-Leibler (KL-)divergence between \mbox{$\Q$ and $\P$}, and $D_{\alpha}(\Q\|\P) {\triangleq} \frac{1}{\alpha{-}1}\ln\!\LB \text{\small${\displaystyle \EE_{h{\sim}\P}}$}\!\LB\!\frac{ \Q(h)}{\P(h)}\RB^{\!\alpha}\RB$  the Rényi divergence  between $\Q$ and $\P$ $(\alpha{>}1)$.
\label{theorem:general-classic}
\end{restatable}

\noindent Note that Equation~\eqref{eq:renyi-bound-classic} is more general than Equation~\eqref{eq:kl-bound-classic}. Indeed, the former is obtained from the latter by the three following steps: {\it (i)}~substituting $\phi(h, \Scal)$ by $\phi(h, \Scal)^{\frac{\alpha-1}{\alpha}}$ in Equation~\eqref{eq:renyi-bound-classic}, 
 {\it (ii)}~applying Jensen's inequality in order to move the expectation over~$\Q$ in front of the logarithm, and {\it (iii)}~taking the limit when $\alpha$ tends to $1$.
Note also the original bound statements of \citet{GermainLacasseLavioletteMarchand2009,BeginGermainLavioletteRoy2016} are recovered by choosing a convex function $\Delta: [0,1]^2 {\rightarrow} \mathbb{R}$ that captures a deviation between the true risk $\Risk_{\Dcal}(h)$ and the empirical risk $\Risk_{\Scal}(h)$.
Then, two steps are required: 
{\it (i)} setting $\phi(h,\!\Scal){=}\exp(m\Delta(\Risk_{\Scal}(h), \Risk_{\Dcal}(h)))$ in Equation~\eqref{eq:kl-bound-classic}, or  $\phi(h,\!\Scal){=}\Delta(\Risk_{\Scal}(h), \Risk_{\Dcal}(h))$ in Equation~\eqref{eq:renyi-bound-classic}, and then {\it (ii)} applying Jensen's inequality on the left-hand side of the inequation.
In fact, our proofs follow the exact same steps than those of \citet[Th.2.1]{GermainLacasseLavioletteMarchand2009} and \citet[Th.9]{BeginGermainLavioletteRoy2016}, but instead of starting from $\Delta(\Risk_{\Scal}(h), \Risk_{\Dcal}(h))$, we consider the slightly more general expression $\phi(h, \Scal)$ from the beginning.
\footnote{We refer the reader to the proof sketches given by Figure 1 of \citet{BeginGermainLavioletteRoy2016} for more insights.}

\noindent The advantage of Theorem~\ref{theorem:general-classic} is that it can be used as a starting point for deriving different forms of bounds.
For instance, for a loss function \mbox{$\ell\!:\! \Hcal{\times} \Zcal {\to} [0, 1]$} with $\phi(h,\!\Scal){=} 
\exp\left(m\Delta(\Risk_{\Scal}(h), \Risk_{\Dcal}(h))\right)$ and $\Delta(\Risk_{\Scal}(h), \Risk_{\Dcal}(h))=2[\Risk_{\Scal}(h){-}\Risk_{\Dcal}(h)]^2$
we retrieve from Equation~\eqref{eq:kl-bound-classic} the bound proposed by~\citet{McAllester1998}:
\begin{align*}
&\PP_{\Scal{\sim}\Dcal^{m}}\!\!\left( \forall \Q,\ 
\LV\EE_{{h\sim \Q}}\Risk_{\Scal}(h)-\EE_{{h\sim\Q}}\Risk_{\Dcal}(h)\RV \le \sqrt{
\frac{\KL( \Q\|\P)+\ln\tfrac{2\sqrt{m}}{\delta}}{2m}}
    \right) \! \ge\! 1{-}\delta\\
    \Longrightarrow & \PP_{\Scal{\sim}\Dcal^{m}}\!\!\left( \forall \Q,\ 
\EE_{{h\sim\Q}}\Risk_{\Dcal}(h) \le \EE_{{h\sim \Q}}\Risk_{\Scal}(h) +
\sqrt{
\frac{\KL( \Q\|\P)+\ln\tfrac{2\sqrt{m}}{\delta}}{2m}}
    \right) \! \ge\! 1{-}\delta.\\
\end{align*}
This bound illustrates the trade-off between the average empirical risk  and $\textstyle\varepsilon\big(\tfrac{1}{\delta}{,}\tfrac{1}{m}{,}\Q \big) = \sqrt{
\frac{1}{2m} (\KL( \Q\|\P)+\ln\tfrac{2\sqrt{m}}{\delta})}$.
More precisely, the higher $m$ is, the lower $\textstyle \varepsilon\big(\tfrac{1}{\delta}{,}\tfrac{1}{m}{,}\Q \big)$
is therefore the smaller the difference between the true risk $\EE_{h{\sim}\Q}\Risk_{\Dcal}(h)$ and the empirical risk $\EE_{{h\sim \Q}}\Risk_{\Scal}(h)$.

\noindent Another example leading to a slightly tighter but less interpretable bound is the \citet{Seeger2002,Maurer2004}'s bound that we retrieve with $\phi(h,\!\Scal){=}
\exp\left(m\,\Delta(\Risk_{\Scal}(h),\Risk_{\Dcal}(h))]\right)$ and $\Delta(\Risk_{\Scal}(h),\Risk_{\Dcal}(h))=\kl[\Risk_{\Scal}(h)\|\Risk_{\Dcal}(h)]$:
\begin{align}
\PP_{\Scal{\sim}\Dcal^{m}}\!\!\left( \forall \Q,\  
\EE_{{h\sim \Q}} \kl(\Risk_{\Scal}(h)\| \Risk_{\Dcal}(h))\le 
\frac{\KL( \Q\|\P)+\ln\tfrac{2\sqrt{m}}{\delta}}{m}
\right) \! \ge\! 1{-}\delta,
\label{eq:seeger}
\end{align}
where 
\begin{align}
\label{eq:small_kl}
\kl(q\|p)= q\ln\tfrac{q}{p}{+}(1{-}q)\ln\tfrac{1{-}q}{1{-}p}
\end{align}
is the KL divergence between two Bernoulli distributions of parameters $q$ and~$p$.

\noindent Such PAC-Bayesian bounds are known to be tight (\eg, \citet{PerezOrtizRivasplataShaweTaylorSzepesvari2020,ZantedeschiViallardMorvantEmonetHabrardGermainGuedj2021}), but they hold for a randomized classifier by nature (due to the expectation \mbox{on $\Hcal$}).
A key issue for usual machine learning tasks is then the derandomization of the PAC-Bayesian bounds to obtain a guarantee for a deterministic classifier instead of a randomized one (by removing the expectation on $\Hcal$).
In some cases, this derandomization results from the structure of the hypotheses, such as for randomized linear classifiers that can be directly expressed as one deterministic linear classifier~\citep{GermainLacasseLavioletteMarchand2009}.
However, in other cases, the derandomization is much more complex and specific to the class of hypotheses, such as for neural networks ({\it e.g.}, \citet{NeyshaburBhojanapalliSrebro2018}, \citet[Ap. J]{NagarajanKolter2019b}, \citet{BiggsGuedj2022}).

\noindent The next section states our main contribution, which is a general derandomization framework (based on the Rényi divergence) for disintegrating PAC-Bayesian bounds into a bound for a single hypothesis from $\Hcal$.

\section{Disintegrated PAC-Bayesian theorems}
\label{sec:contrib}

\subsection{Form of a disintegrated PAC-Bayes bound}

First, we recall another kind of bound introduced by \citet{BlanchardFleuret2007} and \citet[Th.1.2.7]{Catoni2007} and referred to \textit{as the disintegrated PAC-Bayesian bound}. Its form is:
\begin{align}
\label{eq:disintegration}
\PP_{\Scal\sim\Dcal^{m},\, h\sim \AQ}\ \Big( \LV \Risk_{\Dcal}(h)-\Risk_{\Scal}(h)\RV \le \varepsilon\big(\tfrac{1}{\delta}, \tfrac{1}{m},\AQ
\big) \Big)\ge 1-\delta,
\end{align}
where $\AQ{\triangleq}A(\Scal, \P)$ with  $A\!:\!\Zcal^{m}{\times}\Mcal^{*}(\Hcal){\to} \Mcal(\Hcal)$ a \textit{deterministic} algorithm chosen {\it a priori} which {\it (i)}~takes a learning sample  $\Scal\!\in\!\Zcal^{m}$ and a prior distribution $\P$ as inputs, and \mbox{{\it (ii)}}~outputs  a {\it data-dependent} distribution $\AQ{\triangleq}A(\Scal, \P)$ from the set $\Mcal(\Hcal)$ of all possible probability densities on $\Hcal$.
Concretely, this kind of generalization bound allows one to derandomize the usual PAC-Bayes bounds as follows.
Instead of considering a bound holding for all the posterior distributions on $\Hcal$ as usually done in PAC-Bayes (the ``$\,\forall \Q\,$'' in Theorem~\ref{theorem:general-classic}), we consider only the posterior distribution $\AQ$ obtained through a deterministic algorithm $A$ taking the learning sample $\Scal$ and the prior $\P$ as inputs.
Then, the above bound holds for a unique hypothesis $h{\sim}\AQ$ instead of the randomized classifier: the individual risks are no longer averaged with respect to $\AQ$; this is the {\bf PAC-Bayesian bound disintegration}.
The dependence in probability on $\AQ$ means that the bound is valid with probability at least $1{-}\delta$ over the random choice of the learning sample $\Scal{\sim}\Dcal^m$ and the hypothesis $h{\sim}\AQ$.
Under this principle, we introduce in Theorems~\ref{theorem:disintegrated} and~\ref{theorem:disintegrated-lambda} below two new general disintegrated PAC-Bayesian bounds.
A key asset of our results is that the bounds are instantiable to specific settings as for the ``classical'' PAC-Bayesian bounds (\eg, with {\it i.i.d.}/non-{\it i.i.d.} data, unbounded losses, etc.): to instantiate the bound, one has to instantiate the function $\phi$.
Note that, except our bound and the one of~\citet[Th.1(i)]{RivasplataKuzborskijSzepesvariShaweTaylor2020}, the disintegrated bounds of the literature introduced by \citet{BlanchardFleuret2007} and \citet[Th.1.2.7]{Catoni2007} do not depend on such a general function $\phi$.
With an appropriate instantiation, we obtain an easily optimizable bound, leading to a self-bounding\footnote{A self-bounding algorithm minimizes a generalization bound to obtain a model with a generalization guarantee.} algorithm~\citep{Freund1998} with theoretical guarantees.
As an illustration of the usefulness of our results, we provide, in Section~\ref{sec:desintegration}, such an instantiation for neural networks.

\subsection{Disintegrated PAC-Bayesian bounds with the Rényi divergence}

\subsubsection{Our main contribution: a general disintegrated bound}

In the same spirit as Equation~\eqref{eq:renyi-bound-classic} our main result stated in Theorem~\ref{theorem:disintegrated} is a general bound involving the Rényi divergence $D_{\alpha}(\AQ\|\P)$ of order \mbox{$\alpha\!>\!1$}.
\begin{restatable}[General Disintegrated PAC-Bayes Bound]{theorem}{theoremdisintegrated}\label{theorem:disintegrated} For any distribution $\Dcal$ on $\Zcal$, for any hypothesis set $\Hcal$, for any prior distribution $\P\in\Mcal^{*}(\Hcal)$, for any measurable function \mbox{$\phi\!:\! \Hcal{\times}\Zcal^{m}{\to} \Rpe$}, for any \mbox{$\alpha\!>\!1$}, for any $\delta\in(0,1]$, for any algorithm \mbox{$A\!:\!\Zcal^{m}{\times}\Mcal^{*}(\Hcal){\to} \Mcal(\Hcal)$}, we have
\begin{align*}
    \PP_{\Scal\sim\Dcal^{m},h\sim \AQ}\!\Bigg(\! &\frac{\alpha}{\alpha{-}1}\ln\LP\phi(h,\!\Scal)\RP\\
    &\le  \ {\frac{2\alpha{-}1}{\alpha{-}1}}\ln\frac{2}{\delta}
 +D_{\alpha}(\AQ\|\P){+} \ln\LB\EE_{\Scal'{\sim}\Dcal^{m}}\EE_{h'{\sim}\P}\LP\phi(h'\!, \Scal')^{\frac{\alpha}{\alpha{-}1}}\RP\RB \!\Bigg)\!\!\ge\! 1{-}\delta,
\end{align*}
\mbox{where $\AQ{\triangleq}A(\Scal, \P)$ is output by the deterministic algorithm $A$}. 
\end{restatable}
\begin{proof}[Proof sketch (see Appendix~\ref{ap:proof-disintegrated} for details)]
Recall that $\AQ$ is obtained 
with the algorithm $A(\Scal, \P)$.
Applying Markov's inequality on $\phi(h,\!\Scal)$ with the random variable $h$ and using Hölder's inequality to introduce $D_{\alpha}(\AQ\|\P)$, we have,  with  probability at least $1{-}\tfrac\delta2$ on \mbox{$\Scal\!\sim\! \Dcal^m$ and $h\!\sim\!\AQ$}, 
\begin{align*}
   \frac{\alpha}{\alpha{-}1}\ln\left[\phi(h,\!\Scal)\right] \ &\le\ \frac{\alpha}{\alpha{-}1}\ln\left[\frac{2}{\delta}\EE_{h'{\sim} \AQ}\phi(h'\!, \Scal)\right]\\   
   &\le\  D_{\alpha}(\AQ\|\P)  + \frac{\alpha}{\alpha{-}1}\ln\frac{2}{\delta} +\ln\left[\EE_{h'{\sim}\P}\LP\phi(h'\!, \Scal)^{\frac{\alpha}{\alpha-1}}\RP\right].
\end{align*}
By applying again Markov's inequality on $\phi(h,\!\Scal)$ with the random variable $\Scal$, we have, with probability at least $1{-}\tfrac\delta2$ on \mbox{$\Scal\!\sim\! \Dcal^m$ and $h\!\sim\!\AQ$},
\begin{align*}
\ln \left[\EE_{h'{\sim}\P}
\LP\phi(h'\!, \Scal)^{\frac{\alpha}{\alpha{-}1}}\RP\right]
\le 
\ln\left[\frac{2}{\delta}\EE_{\Scal'{\sim}\Dcal^{m}}\EE_{h'{\sim}\P}\LP\phi(h'\!, \Scal')^{\frac{\alpha}{\alpha{-}1}}\RP\right].
\end{align*}
Lastly, we combine the two bounds with a union-bound argument.
\end{proof}

\noindent As for the general classical PAC-Bayesian bounds (Theorem~\ref{theorem:general-classic}), the above theorem can be seen as the starting point of the derivation of generalization bounds depending on the choice of the function~$\phi$, as done in Corollary~\ref{corollary:nn} in Section~\ref{sec:training-method};
this property makes it the main result of our paper. 

\noindent{}In its proof, Hölder's inequality is used differently than in the classic PAC-Bayes bound's proofs.
Indeed, in \citet[Th. 8]{BeginGermainLavioletteRoy2016}, the change of measure based on Hölder's inequality is key for deriving a bound that holds for all posteriors $\Q$ with high probability, while our bound holds for a unique \mbox{posterior $\AQ$} dependent on the sample $\Scal$ and the prior $\P$.
In fact, we use Hölder's inequality to introduce a prior $\P$ independent from $\Scal$: a crucial point for our bound instantiated in Corollary~\ref{corollary:nn}.

\noindent{}Compared to Equation~\eqref{eq:renyi-bound-classic}, our bound involves the term ${\frac{2\alpha{-}1}{\alpha{-}1}}\ln\frac{2}{\delta}$ instead of $\ln\frac{1}{\delta}$, that is an additional constant value of ${\frac{2\alpha{-}1}{\alpha{-}1}}\ln\frac{2}{\delta}-\ln\frac{1}{\delta}=\ln2{+}\frac{\alpha}{\alpha{-}1}\ln\tfrac{2}{\delta}$.
When $\alpha=2$, this constant equals  $\ln\frac{8}{\delta^2}$, which turns out to be a reasonable cost to ``derandomize'' a bound into a disintegrated one, as typical choices for $\phi(h,\!\Scal)$ will make the constant imprint on the bound value decay with $m$.
\noindent This is similar to the bounds of Theorem~\ref{theorem:disintegrated} that tighten as $m$ increases, provided that $\phi(h,\Scal)$ is chosen wisely.
For instance, by setting  $\phi(h,\!\Scal)=\exp(\tfrac{\alpha-1}{\alpha}m\,\kl(\Risk_{\Scal}(h)\|\Risk_{\Dcal}(h))$ with $\kl(\cdot\|\cdot)$ defined by Equation~\eqref{eq:small_kl}, the bound depends on $m$ and converges as $m$ increases (see Section~\ref{sec:desintegration}).
Moreover, the tightness of the bound depends also on the deviation between $\AQ$ and $\P$, which makes the bound tighter when $\AQ=\P$.

\noindent We instantiate below Theorem~\ref{theorem:disintegrated} for $\alpha{\rightarrow}1^+$ and $\alpha{\rightarrow}{+}{\infty}$ showing that the bound converges when $\alpha{\to}1^+$ and $\alpha{\to}{+}\infty$.
\begin{restatable}{corollary}{corollarydisintegrated}
Under the assumptions of  Theorem~\ref{theorem:disintegrated}, when $\alpha{\to}1^+$,  we have
\begin{align*}
   \PP_{\Scal\sim\Dcal^{m},h\sim \AQ}\!\Bigg(
  \ln\phi(h{,}\Scal) \le \ln\frac{2}{\delta} + \ln\left[\esssup_{\Scal'\in\Zcal, h'\in\Hcal}\phi(h'{,} \Scal')\right]
  \Bigg)\!\ge\! 1{-}\delta,
\end{align*}
when $\alpha{\to}+\infty$,
we have
\begin{align*}
 \PP_{\Scal\sim\Dcal^{m},h\sim \AQ}\!\Bigg(    \ln\phi(h{,} \Scal)\le\ln{\displaystyle\esssup_{h'\in\Hcal}}\,\frac{\AQ(h')}{\P(h')}{+}\ln\!\left[\frac{4}{\delta^2} {\displaystyle \EE_{\Scal'{\sim}\Dcal^{m}}\EE_{h'{\sim}\P}\phi(h'{,}\Scal')}\right] \Bigg)\!\ge\! 1{-}\delta,
\end{align*}
where $\esssup$ is the essential supremum defined as the supremum on a set with non-zero probability measures, \ie, 
\begin{align*}
    &\esssup_{\Scal'\in\Zcal, h'\in\Hcal}\phi(h'{,} \Scal')=\inf\LC\tau\in\Rbb, \PP_{\Scal\sim\Dcal^{m},h\sim \AQ}\!\Big[ \phi(h{,} \Scal) > \tau\Big] = 0\RC,\\
    \text{and}\quad &\esssup_{h'\in\Hcal}\,\frac{\AQ(h')}{\P(h')}=\inf\LC\tau\in\Rbb, \PP_{h\sim\AQ}\!\Big[ \tfrac{\AQ(h)}{\P(h)} > \tau\Big] = 0\RC.
\end{align*}
\label{corollary:disintegrated}
\end{restatable}
\noindent{}This corollary illustrates that the parameter $\alpha$ controls the trade-off between the Rényi divergence 
$D_{\alpha}(\AQ\|\P)$ and 
$\ln\LB\EE_{\Scal'{\sim}\Dcal^{m}}\EE_{h'{\sim}\P}\phi(h'\!, \Scal')^{\frac{\alpha}{\alpha{-}1}}\RB$. 
Indeed, when $\alpha{\rightarrow}1^+$, the Rényi divergence vanishes while the other term converges toward $\ln\!\left[\esssup_{\Scal'\in\Zcal, h'\in\Hcal}\phi(h'\!, \Scal')\right]$, roughly speaking the maximal value possible for the second term.
On the other hand, when $\alpha{\rightarrow}{+}\infty$, the Rényi divergence increases and converges toward $\ln\esssup_{h'\in\Hcal}\frac{\AQ(h')}{\P(h')}$ and the other term decreases toward $\ln\!\left[\EE_{\Scal'{\sim}\Dcal^{m}}\EE_{h'{\sim}\P}\phi(h'{,}\Scal')\right]$.

\subsubsection[Comparison with the bound of Rivasplata et al. (2020)]{Comparison with the bound of~\texorpdfstring{\citet{RivasplataKuzborskijSzepesvariShaweTaylor2020}}{Rivasplata et al. (2020)}}

For the sake of comparison, we recall in Equation~\eqref{eq:rivasplata} the bound proposed by~\citet[Th.1{\it (i)}]{RivasplataKuzborskijSzepesvariShaweTaylor2020}, that is more general than the bounds of \citet{BlanchardFleuret2007} and \citet[Th.1.2.7]{Catoni2007}: 
\begin{align}
\PP_{\Scal\sim\Dcal^{m}, h\sim \AQ}\Bigg(\!
\ln(\phi(h,\!\Scal)) \le \ln\frac{\AQ(h)}{\P(h)} +\ln\LP\frac1\delta \displaystyle\EE_{\Scal'{\sim}\Dcal^{m}}\EE_{h'{\sim}\P}\phi(h'{,} \Scal')\RP\!\Bigg) \geq 1{-}\delta.\label{eq:rivasplata}
\end{align}

\noindent The term $\ln\!\tfrac{\AQ(h)}{\P(h)}$ (also involved in~\citet{Catoni2007,BlanchardFleuret2007}) can be seen as a ``disintegrated\footnote{We say that the KL divergence is ``disintegrated'' since the log term is not averaged in contrast to the KL divergence.} KL divergence'' depending only on the sampled  $h{\sim}\AQ$.
In contrast, our bound involves the Rényi divergence $D_\alpha(\AQ\|\P)$ between the \mbox{prior $\P$} and the posterior $\AQ$, meaning our bound involves only one term that depends on the sampled hypothesis (the risk): the divergence value is the same whatever the hypothesis.
Our bound is expected to be looser because of the Rényi divergence~\citep[see][]{VanErvenHarremos2014} and the dependence in $\delta$ (which is worse than Equation~\eqref{eq:rivasplata}).
Nevertheless, our divergence term is the main advantage of our bound.
Indeed, as confirmed by our experiments (Section~\ref{sec:expe}), our bound with $\D_{\alpha}(\AQ\|\P)$ makes the learning procedure (in our self-bounding algorithm) more stable and efficient compared to the optimization of Equation~\eqref{eq:rivasplata} with $\ln\tfrac{\AQ(h)}{\P(h)}$ that is subject to high variance.
  
\subsubsection{A parameterizable general disintegrated  bound} 

In the PAC-Bayesian literature, parametrized bounds have been introduced ({\it e.g.}, \citet{Catoni2007, ThiemannIgelWintenbergerSeldin2017}) to control the trade-off  between the empirical risk and the divergence along with the additional term. 
For the sake of completeness, we now provide a parametrized version of our bound, enlarging its practical scope.
 We follow a similar approach to introduce a version of a disintegrated Rényi divergence-based bound that has the advantage of being parameterizable.
\begin{restatable}[Parametrizable Disintegrated PAC-Bayes Bound]{theorem}{theoremdisintegratedlambda} For any distribution $\Dcal$ on $\Zcal$, for any hypothesis set $\Hcal$, for any prior distribution $\P\in\Mcal^{*}(\Hcal)$, for any measurable function \mbox{$\phi\!:\! \Hcal{\times}\Zcal^{m}{\to} \Rpe$}, for any $\delta\in(0,1]$, for any algorithm \mbox{$A\!:\!\Zcal^{m}{\times}\Mcal^{*}(\Hcal){\to} \Mcal(\Hcal)$}, we have
\begin{align*}
&\PP_{\substack{\Scal\sim\Dcal^{m},\\h\sim \AQ}} \!\Bigg(\!\forall\lambda{>}0,\,\ln\LP\phi(h,\!\Scal)\RP{\le} \ln\!\bigg[
\frac\lambda 2\displaystyle e^{D_2(\AQ\|\P)}{+} \frac{8}{2\lambda\delta^3}\EE_{\Scal'{\sim}\Dcal^{m}}\EE_{h'{\sim}\P}\!\Big[\phi(h'\!, \Scal')^2\Big]
\bigg]\!\Bigg)\ge1{-}\delta,
\end{align*}
\mbox{where $\AQ{\triangleq}A(\Scal,\P)$ is output by the deterministic algorithm $A$}. 
\label{theorem:disintegrated-lambda} 
\end{restatable}
\noindent Note that $e^{D_2(\AQ\|\P)}$ is closely related to the $\chi^2$-distance. Indeed we have:  
\mbox{$\chi^2(\AQ\|\P) \triangleq \EE_{h\sim\P}\LB\tfrac{\AQ(h)}{\P(h)}\RB^2\!\!{-}1 = e^{D_2(\AQ\|\P)} {-}1$.}
An asset of Theorem~\ref{theorem:disintegrated-lambda} is the parameter $\lambda$ controlling the trade-off between the exponentiated Rényi divergence $e^{D_2(\AQ\|\P)}$ and $\frac{1}{\delta^3}{\EE}_{\Scal'{\sim}\Dcal^{m}}{\EE}_{h'{\sim}\P}\phi(h'\!, \Scal')^2$.
Our bound is valid for all \mbox{$\lambda\!>\!0$}, thus, from a practical view, we can learn/tune the parameter $\lambda$ to minimize the bound and control the possible numerical instability due to  $e^{D_2(\AQ\|\P)}$.
Indeed, if $D_2(\AQ\|\P)$ is large, the numerical computation can lead to an infinite value due to finite precision arithmetic.
It is important to notice that, like other parametrized bounds \citep[\eg,][]{ThiemannIgelWintenbergerSeldin2017}, there exists a closed-form solution of the optimal parameter $\lambda$ (for a fixed $\P$ and $ \AQ$); the solution is derived in Proposition~\ref{prop:lambda-min} and shows that the optimal bound of Theorem~\ref{theorem:disintegrated-lambda} corresponds to the bound of Theorem~\ref{theorem:disintegrated}.

\begin{restatable}{proposition}{proplambdamin}\label{prop:lambda-min} For any distribution $\Dcal$ on $\Zcal$, for any hypothesis set $\Hcal$, for any prior distribution
$\P$ on $\Hcal$, for any $\delta{\in}(0,1]$, for any measurable function \mbox{$\phi\!:\! \Hcal{\times}\Zcal^{m}{\to} \Rpe$}, for any algorithm $A\!:\!\Zcal^{m}{\times}\Mcal^{*}(\Hcal){\to} \Mcal(\Hcal)$, let 
\begin{align*}
\lambda^* {=} \argmin_{\lambda>0} \, \ln\! \left[\frac{\lambda}{2} e^{D_2(\AQ\|\P)}\!+\!
\frac{\displaystyle \EE_{{\Scal'{\sim}\Dcal^{m}}} \EE_{h'{\sim}\P}\,\LB8\phi(h'\!, \Scal')^2\RB}{2\lambda\delta^3}  \right]\!,
\end{align*}
\begin{align*}
\mbox{then, we have}\quad & \overbrace{2\ln\!\LB\frac{\lambda^*}{2\ }e^{D_2(\AQ\|\P)}\!+\! \EE_{\Scal'{\sim}\Dcal^{m}}\EE_{h'{\sim} \P}
\LP\frac{8\phi(h'\!, \Scal')^2}{2\lambda^*\delta^3}\RP\RB}^{\text{Theorem \ref{theorem:disintegrated-lambda}}}   \\ &= \ \underbrace{D_{2}(\AQ\|\P) + \ln\LB\EE_{\Scal'{\sim}\Dcal^{m}}\EE_{h'{\sim} \P}
\LP\frac{8\phi(h'\!, \Scal')^{2}}{\delta^3}\RP\RB}_{\text{Theorem \ref{theorem:disintegrated} with $\alpha=2$.} },
\end{align*}
where  $\displaystyle \lambda^* = \sqrt{\frac{\EE_{\Scal'{\sim}\Dcal^{m}}{\EE}_{{h'{\sim}\P}}\LB8\phi(h'\!, \Scal')^2\RB}{\delta^3 \exp(D_2(\AQ\|\P))}}$. \\[1mm]
Put into words: the optimal $\lambda^*$ gives the same bound for Theorem~\ref{theorem:disintegrated} and Theorem~\ref{theorem:disintegrated-lambda}.
\end{restatable}

\section{The disintegration in action}
\label{sec:desintegration}

So far, we have introduced theoretical results to derandomize PAC-Bayesian bounds through a disintegration approach.
Indeed, the disintegration allows us to obtain a bound for a unique model sampled from the distribution $\AQ$ instead of having a bound on the averaged risk of the models.
We propose in this section to illustrate the instantiation and the usefulness of Theorem~\ref{theorem:disintegrated} on neural networks compared to the classical PAC-Bayesian bounds.

\subsection{Specialization to neural network classifiers}
\label{sec:training-method}
We consider Neural Networks (NN) parametrized by a weight vector \mbox{$\wbf\!\in\!\Rbb^{d}$} and overparametrized, \ie, $d\!\gg\!m$. 
We aim to learn the weights of the NN leading to the lowest true risk.
Practitioners usually proceed by epochs\footnote{One epoch corresponds to one pass of the entire learning set during the optimization process.} and obtain one ``intermediate''  NN after each epoch.
Then, they select the ``intermediate''  NN associated with the lowest validation risk.
We propose translating this practice into our PAC-Bayesian setting by considering one prior per epoch.
Given $T$ epochs, we hence have
$T$ priors $\Pbf{=}\{\P_t\}_{t=1}^T$, where \mbox{$\forall t\!\in\!\{1,\ldots,T\}, \P_t=\Ncal(\vbf_t, \sigma^2{\bf I}_{d})$} is a Gaussian distribution centered \mbox{at $\vbf_t$} (the weights associated with the \mbox{$t$-th} ``intermediate''  NN) with a covariance matrix of $\sigma^2{\bf I}_{d}$ (where ${\bf I}_{d}$ is the $d{\times}d$-dimensional identity matrix).
Assuming the $T$ priors are learned from a set $\Scal_{\text{prior}}$ such that \mbox{$\Scal_{\text{prior}} \bigcap \Scal {=} \emptyset$}, then 
Corollaries~\ref{corollary:nn}~and~\ref{corollary:nn-rbc} will guide us to learn a posterior \mbox{$\AQ{=}\Ncal(\wbf, \sigma^2{\bf I}_{d})$} from a prior $\P\in\Pbf$ minimizing the empirical risk on $\Scal$ (we give more details on the procedure after the forthcoming corollaries).
Note that considering Gaussian distributions has the advantage of simplifying the expression of the KL divergence, and thus is commonly used in the PAC-Bayesian literature for neural networks \citep[\eg,][]{DziugaiteRoy2017,LetarteGermainGuedjLaviolette2019,ZhouVeitchAusternAdamsOrbanz2019}.\footnote{Gaussian distributions have been first studied in PAC-Bayes in  the context of linear classifiers \citep[\eg,][]{AmbroladzeHernandezShaweTaylor2006,GermainLacasseLavioletteMarchand2009,GermainHabrardLavioletteMorvant2020}, but in this context, the symmetry of the Gaussian distribution also ease the derandomization.}

\noindent Corollary~\ref{corollary:nn} below instantiates Theorem~\ref{theorem:disintegrated} to this neural networks setting.
Then, for the sake of comparison, Corollary~\ref{corollary:nn-rbc} instantiates other disintegrated bounds from the literature; more precisely, Equation~\eqref{eq:nn-rivasplata} corresponds to \citet{RivasplataKuzborskijSzepesvariShaweTaylor2020}'s bound of Equation~\eqref{eq:rivasplata}, Equation~\eqref{eq:nn-blanchard} to \citet{BlanchardFleuret2007}'s one,  and Equation~\eqref{eq:nn-catoni} to \citet{Catoni2007}'s one.

\begin{restatable}{corollary}{corollarynn}\label{corollary:nn} For any distribution $\Dcal$ on $\Zcal$, for any hypothesis set $\Hcal$, for any set \mbox{$\Pbf=\{\Pcal_1,\dots, \Pcal_T\}$} of $T$ priors on $\Hcal$ where $\Pcal_t=\Ncal(\vbf_t, \sigma^2\Ibf_d)$, for any algorithm $A\!:\!\Zcal^{m}\times\Mcal^{*}(\Hcal) {\rightarrow} \Mcal(\Hcal)$, for any loss $\ell\!:\! \Hcal{\times} \Zcal {\to} [0, 1]$, for any $\delta{\in}(0,1]$,
we have 
\begin{align*}
 \PP_{\Scal\sim\Dcal^{m}, h\sim \AQ}\Bigg(\forall& \P_t\in \Pbf,\ \kl(\Risk_{\Scal}(h)\| \Risk_{\Dcal}(h))\le \frac{1}{m}\left[ \frac{\|\wbf{-}\vbf_t\|_{2}^{2}}{\sigma^2}+\ln\frac{16T\sqrt{m}}{\delta^3}\right]\!\Bigg)\geq 1{-}\delta,
\end{align*}
where $\kl(a\|b) = a\ln\tfrac{a}{b}+(1{-}a)\ln\tfrac{1-a}{1-b}$, $\AQ = \Ncal(\wbf, \sigma^2\Ibf_d)$, and the hypothesis $h\sim\AQ$ is parametrized by $\wbf{+}\epsilonbf$.
\end{restatable}

\begin{restatable}{corollary}{corollarynnrbc}\label{corollary:nn-rbc} For any distribution $\Dcal$ on $\Zcal$, for any set $\Hcal$, for any set \mbox{$\Pbf=\{\Pcal_1,\dots,\Pcal_T\}$} of $T$ priors on $\Hcal$ where $\Pcal_t=\Ncal(\vbf_t, \sigma^2\Ibf_d)$, for any algorithm $A\!:\!\Zcal^{m}\times\Mcal^{*}(\Hcal){\rightarrow} \Mcal(\Hcal)$, for any loss $\ell\!:\! \Hcal{\times} \Zcal {\to} \{0, 1\}$, for any $\delta{\in}(0,1]$,  with probability at least $1{-}\delta$ over the learning sample $\Scal{\sim}\Dcal^{m}$ and the hypothesis $h{\sim} \AQ$ parametrized by $\wbf{+}\epsilonbf$, we have $\forall \P_t\in \Pbf$
\begin{align}
&\kl(\Risk_{\Scal}(h)\|\Risk_{\Dcal}(h))\!\le\, \frac{1}{m}\! \Bigg[\! 
\frac{\| \wbf{+}\epsilonbf{-}\vbf_t\|^2_{2} {-}\|\epsilonbf\|^2_{2}}{2\sigma^2} {+}  \ln\! \frac{2T\!\sqrt{m}}{\delta}\!\Bigg]\!,\!\label{eq:nn-rivasplata} \\
\forall b\!\in\!\Bbf,\quad &\kl_{+}(\Risk_{\Scal}(h)\|\Risk_{\Dcal}(h))\!\le \frac{1}{m} \! \Bigg[
\frac{b{+}1}{b}\!\LB\frac{\| \wbf{+}\epsilonbf{-}\vbf_t\|^2_{2} {-}\|\epsilonbf\|^2_{2}}{2\sigma^2}\RB_{\!+}\!{+}  \ln\! \frac{(b{+}1)T\vert\Bbf\vert}{\delta} \Bigg]\!,\!\label{eq:nn-blanchard}\\
\forall c\!\in\!\Cbf,\quad &\Risk_{\Dcal}(h) \!\le\, \frac{\displaystyle1{-}\exp\left({\displaystyle\!\!{-}c\Risk_{\Scal}(h) {-}\frac{1}{m}\!\!\left[\! \frac{\| \wbf{+}\epsilonbf{-}\vbf_t\|^2_{2} {-}\|\epsilonbf\|^2_{2}}{2\sigma^2} {+} \ln\!\frac{T\vert\Cbf\vert}{\delta}\!\right]\!}\right)}{1{-}e^{{-}c}}\!,\!\label{eq:nn-catoni}
\end{align}
with $\LB x\RB_{+}\!{=}\max(x,0)$, and   $\kl_{+}(\Risk_{\Scal}(h)\|\Risk_{\Dcal}(h)){=}\kl(\Risk_{\Scal}(h)\|\Risk_{\Dcal}(h))$ if $\Risk_{\Scal}(h){<}\Risk_{\Dcal}(h)$ and 0 otherwise.
Moreover, $\epsilonbf{\sim}\Ncal(\zerobf, \sigma^2{\bf I}_{d})$ is a Gaussian noise such that $\wbf{+}\epsilonbf$ are the weights of $h{\sim}\AQ$ with \mbox{$\AQ{=}\Ncal(\wbf, \sigma^2{\bf I}_{d})$}, and $\Cbf$, $\Bbf$ are two sets of hyperparameters fixed a priori.
\end{restatable}

\noindent As the parameter $\lambda$ of the Theorem~\ref{theorem:disintegrated-lambda}, $c\!\in\!\Cbf$ is a hyperparameter that controls a trade-off between the empirical risk $\Risk_{\Scal}(h)$ and the term $\frac{1}{m}\!\!\left[\! \frac{\| \wbf{+}\epsilonbf{-}\vbf_t\|^2_{2} {-}\|\epsilonbf\|^2_{2}}{2\sigma^2} {+} \ln\!\frac{T\vert\Cbf\vert}{\delta}\!\right]$.
Besides, the parameter $b\!\in\!\Bbf$ controls the tightness of the bound. 
In general, these parameters can be tuned to minimize the bound of Equation~\eqref{eq:nn-blanchard} and Equation~\eqref{eq:nn-catoni}; however, there is no closed-form solution for the expression of the minimum of this equation.
In consequence, our experimental protocol requires minimizing the bounds by gradient descent for each $b\!\in\!\Bbf$, respectively $c\!\in\!\Cbf$, in order to learn the distribution $\AQ$ leading to the lowest bound value.
To obtain a tight bound, the divergence between one prior $\P_t\!\in\!\Pbf$ and $ \AQ$ must be low, \ie,  $\| \wbf{-}\vbf_t\|_{2}^2$ (or $\| \wbf{+}\epsilonbf{-}\vbf_t\|_{2}^2{-}\|\epsilonbf\|^2_{2}$) has to be small.
One solution 
is to split the learning sample into $2$ non-overlapping subsets  $\Scal_{\text{prior}}$ and $\Scal$, where $\Scal_{\text{prior}}$ is used to learn the prior, while $\Scal$ is used both to learn the posterior and compute the bound.
Hence, if we ``pre-learn'' a good enough prior $\P_t\!\in\!\Pbf$ from $\Scal_{\text{prior}}$, then we can expect to have a low \mbox{$\| \wbf{-}\vbf_t\|_{2}$}.

\begin{algorithm}[H]
  \caption{Training Method}
  \begin{algorithmic}
  \State{The original training set is split into two distinct subsets: $\Scal_{\text{prior}}$  and $\Scal$ (respectively of size $m_{\text{prior}}$ and $m$, that can be different).}
\State{The training has two phases.}
\State{{\bf 1)} The prior distribution $\P$ is ``pre-learned'' with $\Scal_{\text{prior}}$ and selected by early stopping,  with $\Scal$ as validation set, using the algorithm $A_{\text{prior}}$ (an arbitrary learning algorithm).}
\State{{\bf 2)} Given $\Scal$ and $\P$, we learn the posterior $\AQ$ with the algorithm $A$ (defined {\it a priori}).}
  \end{algorithmic}
\end{algorithm}

\noindent At first sight, the selection of the prior weights \mbox{with $\Scal$} by early stopping may appear to be ``cheating''.
However, this procedure can be seen as: \mbox{{\bf 1)}~first} constructing $\Pbf$ from the $T$ ``intermediate''  NNs learned after each epoch \mbox{from $\Scal_{\text{prior}}$}, then  \mbox{{\bf 2)}~optimizing} the bound with the prior that leads to the best risk \mbox{on $\Scal$}.
This gives a statistically valid result: since Corollary~\ref{corollary:nn} is valid for every $\prior_t\!\in\!\Pbf$, we can select the one we want, in particular the one minimizing  $\Risk_{\Scal}(h)$ for a sampled $h\sim\prior_t$.
This heuristic makes sense: it allows us to detect if a prior is concentrated around hypotheses that potentially overfit the learning sample $\Scal_{\text{prior}}$. 
Usually, practitioners consider this ``best'' prior as the final  NN.
In our case, the advantage is that we refine this ``best'' prior with $\Scal$ to learn the posterior $ \AQ$.
Note that \citet{PerezOrtizRivasplataShaweTaylorSzepesvari2020} have already introduced tight generalization bounds with data-dependent priors for---non-derandomized---stochastic NNs.\footnote{Stochastic NNs were introduced in the PAC-Bayesian literature by \citet{LangfordCaruana2001}.}
Nevertheless, the weights of the stochastic NNs are, by definition, sampled from the posterior distribution $\Q$ for each prediction.
In that sense, it is important to mention that stochastic NNs differ from  {\it derandomized} NNs where only one model is sampled from $\AQ$.
Moreover, our training method to learn the prior differs greatly since {\bf 1)}~we learn $T$  NNs (\ie, $T$ priors) instead of only one, {\bf 2)}~we fix the variance of the Gaussian in the posterior  $\AQ$.
Note that, as illustrated in Section~\ref{sec:expe}, fixing the variance is not restrictive: the advantage is that it simplifies the expression of the KL divergence while keeping the bounds tight.
To the best of our knowledge, our training method for the prior is new.

\subsection{A note about stochastic neural networks}

Due to its stochastic nature, PAC-Bayesian theory has been explored to study stochastic NNs (\eg,~\citet{LangfordCaruana2001,DziugaiteRoy2017, DziugaiteRoy2018, ZhouVeitchAusternAdamsOrbanz2019, PerezOrtizRivasplataShaweTaylorSzepesvari2020}).
In Corollary~\ref{corollary:nn-sto} below, we instantiate the bound of Equation~\eqref{eq:kl-bound-classic} for stochastic NNs to empirically compare the stochastic and the deterministic  NNs associated to the same prior and posterior distributions.
We recall that, in this paper, a deterministic  NN is a \emph{single} $h$ sampled from the posterior distribution $\AQ{=}\Ncal(\wbf, \sigma^2{\bf I}_{d})$ output by the algorithm $A$.
This means that for each example, the label prediction is performed by the same deterministic  NN: the one parametrized by the weights $\wbf+\epsilonbf\!\in\! \R^d$.
Conversely, the stochastic  NN associated with a posterior distribution $\Q{=}\Ncal(\wbf, \sigma^2{\bf I}_{d})$ predicts the label of a given example by {\it(i)} first sampling $h$ according to $\Q$, {\it (ii)} then returning the label predicted \mbox{by $h$.}
Thus, the risk of the stochastic  NN is the expected risk value $\EE_{h{\sim}\Q}\Risk_{\Dcal}(h)$, where the expectation is taken over \mbox{\emph{all} $h$ sampled from $\Q$}.
We compute the empirical risk of the stochastic  NN from a Monte Carlo approximation:  {\it (i)} we sample $n$ weight vectors, and {\it (ii)} we average the risk over the $n$ associated NNs;
we denote by $\Q^n$ the distribution of such $n$-sample.
In this context, we obtain the following PAC-Bayesian bound.

\begin{restatable}{corollary}{corollarynnsto}\label{corollary:nn-sto}
For any distribution $\Dcal$ on $\Zcal$, for any  $\Hcal$, for any set $\Pbf=\{\Pcal_1,\dots, \Pcal_T\}$ of $T$ priors on $\Hcal$ where $\Pcal_t=\Ncal(\vbf_t, \sigma^2\Ibf_d)$, for any loss $\ell\!:\! \Hcal{\times} \Zcal {\to} \{0, 1\}$, for any $\delta{\in}(0,1]$, with probability at least $1{-}\delta$ over $\Scal{\sim}\Dcal^m$ and $\{h_1,\dots,h_n\}{\sim}\Q^n$, we have simultaneously $\forall \P_t\!\in\!\Pbf,$
\begin{align}
&\kl\!\left(\EE_{h{\sim}\Q}\!\!\Risk_{\Scal}(h)\|\! \EE_{h{\sim}\Q}\!\!\Risk_{\Dcal}(h)\!\right)\!
{\le} \frac{1}{m}\!\!\LB 
\frac{\|\wbf\!{-}\vbf_t\|_{2}^{2}}{2\sigma^2}
{+}\ln\!\frac{4T\sqrt{m}}{\delta}\RB\!,\!\label{eq:nn-sto-seeger}\\
\!\!\mbox{and }\quad  &\kl\LP{\frac{1}{n}\sum_{i=1}^{n}\!\Risk_{\Scal}(h_i)}\|\EE_{h{\sim}\Q}\!\Risk_{\Scal}(h)\RP \le \frac1n \ln\frac{4}{\delta}, \label{eq:nn-sto-sample}
\end{align}
where $\Q=\Ncal(\wbf, \sigma^2\Ibf_d)$ and the hypothesis $h$ sampled from $\Q$ is parametrized by $\wbf+\epsilonbf$ with $\epsilonbf\sim\Ncal(\zerobf, \sigma^2\Ibf_d)$.
\end{restatable}

\noindent This result shows two key features that allow considering it as an adapted baseline for a {\it fair} comparison 
between disintegrated and classical PAC-Bayesian bounds,  thus between deterministic and stochastic  NNs.
On the one hand, it involves the same terms as Corollary~\ref{corollary:nn}.
On the other hand, it is close to the bound of~\citet[Sec.~6.2]{PerezOrtizRivasplataShaweTaylorSzepesvari2020}, since {\it (i)} we adapt the KL divergence to our setting (\ie, $\KL(\Q\|\P){=}\tfrac{1}{2\sigma^2}\|\wbf{-}\vbf_t\|_2^2$), {\it (ii)} the bound holds for $T$ priors thanks to a union-bound argument.

\section[Experiments with neural networks]{Experiments with neural networks\protect\footnote{The source code of our experiments is available at \href{https://github.com/paulviallard/MLJ-Disintegrated-PB}{https://github.com/paulviallard/MLJ-Disintegrated-PB}. We used the PyTorch framework~\citep{Pytorch2019}.}}
\label{sec:expe}

In this section, we do not seek state-of-the-art performance; in fact, we have a threefold objective:
{\bf (a)}~we check if $50\%/50\%$ is a good choice for splitting the original train set into $(\Scal_{\text{prior}}, \Scal)$ (which is the most common split in the PAC-Bayesian literature~\citep{GermainLacasseLavioletteMarchand2009,PerezOrtizRivasplataShaweTaylorSzepesvari2020});
{\bf (b)}~we highlight that our disintegrated bound associated with the deterministic NN is tighter than the randomized bound associated with the stochastic NN (Corollary~\ref{corollary:nn-sto});
{\bf (c)}~we show that our disintegrated bound (Corollary~\ref{corollary:nn}) is tighter and more stable than the ones based on \citet{RivasplataKuzborskijSzepesvariShaweTaylor2020}, \citet{BlanchardFleuret2007} and \citet{Catoni2007} (Corollary~\ref{corollary:nn-rbc}).

\subsection{Training method}
We follow our Training Method (Section~\ref{sec:training-method}) in which we integrate the direct minimization of all the bounds.
We refer as \ours the training method based on the minimization of our bound in Corollary~\ref{corollary:nn}, as \rivasplata the one based on Equation~\eqref{eq:nn-rivasplata}, as \blanchard the one based on Equation~\eqref{eq:nn-blanchard}, and as \catoni the one based on Equation~\eqref{eq:nn-catoni}.
\stoNN denotes the PAC-Bayesian bound with the prior and posterior distributions obtained from \ours.
To optimize the bound with gradient descent, we replace the non-differentiable 0-1 loss with a surrogate: the bounded cross-entropy loss \citep{DziugaiteRoy2018}.
We made this replacement since cross-entropy minimization works well in practice for neural networks \citep{GoodfellowBengioCourville2016} and because this loss is bounded between 0 and 1, which is required for the $\kl()$ function.
The cross-entropy is defined in a multiclass setting with \mbox{$y\!\in\!\{1,2,\ldots\}$} by \mbox{$\ell(h, (\xbf, y)) {=} -\frac{1}{Z}\ln(\Phi(h(\xbf)[y]))\!\in\! [0, 1]$} where $h(\xbf)[y]$ is the \mbox{$y$-th} output of the  NN, and \mbox{$\forall p\!\in\![0, 1], \Phi(p) {=} e^{-Z}{+}(1{-}2e^{-Z})p$}
(we set $Z{=}4$, the default parameter of \citet{DziugaiteRoy2018}).
That being said, to learn a good enough prior $\P\!\in\!\Pbf$ and the posterior $\AQ$, we run our Training Method with two stochastic gradient descent-based algorithms $A_{\text{prior}}$ and $A$.
Note that the randomness in the stochastic gradient descent algorithm is fixed to have deterministic algorithms.
In phase {\bf 1)} algorithm $A_{\text{prior}}$ learns from $\Scal_{\text{prior}}$  the $T$ priors \mbox{$\P_1,\dots,\P_T\!\in\!\Pbf$} (\ie, during $T$ epochs)  by minimizing the bounded cross-entropy loss.
In other words, at the end of the epoch $t$, the weights $\wbf_t$ of the classifier are used to define the prior $\P_t=\Ncal(\wbf_t, \sigma^2\Ibf_{d})$.
Then, the best prior $\P\!\in\!\Pbf$ is selected by early stopping on $\Scal$.
In phase {\bf 2)}, given $\Scal$ and $\P$, 
algorithm $A$ integrates the direct optimization of the bounds with the bounded cross-entropy loss.
\subsection[Optimization procedure]{Optimization procedure in algorithms $A$ and $A_{\text{prior}}$\protect\footnote{The details of the optimization and the evaluation of the bounds are described in Appendix~\ref{ap:evaluation-minimization}.}}
Let $\omegabf$ be the mean vector of a Gaussian distribution used as NN weights that we are optimizing.
In algorithms $A$ and $A_{\text{prior}}$, we use the Adam optimizer~\citep{KingmaBa2015}, and we sample a noise ${\epsilonbf}\!\sim\! \Ncal({\bf 0}, \sigma^2{\bf I}_{d})$ at each iteration of the optimizer.
Then, we forward the examples of the mini-batch to the NN parametrized by the weights $\omegabf{+}\epsilonbf$, and we update $\omegabf$ according to the bounded cross-entropy loss. 
Note that during phase {\bf 1)}, at the end of each epoch $t$, $\Pcal_{t} {=} \Ncal(\omegabf, \sigma^2\Ibf_{d}) {=} \Ncal(\vbf_t, \sigma^2\Ibf_{d})$ and finally at the end of phase {\bf 2)} we have $\AQ {=} \Ncal(\omegabf, \sigma^2\Ibf_{d}) {=} \Ncal(\wbf, \sigma^2\Ibf_{d})$.

\subsection{Experimental setting}

\subsubsection{Datasets}
We perform our experimental study on three datasets: MNIST~\citep{LeCunCortesBurges1998},  Fashion-MNIST~\citep{XiaoRasulVollgraf2017}, and  CIFAR-10~\citep{Krizhevsky2009}. 
We divide each original train set into two independent subsets $\Scal_{\text{prior}}$ of size $m_{\text{prior}}$ and $\Scal$ of size $m$ with varying split ratios defined as $\tfrac{m_{\text{prior}}}{m+m_{\text{prior}}}\in\{0, .1, .2, .3, .4, .5, .6, .7, .8, .9\}$.
The test sets denoted by $\Tcal$ remain the original ones.

\subsubsection{Models}
\label{sec:models}

For the (Fashion-)MNIST datasets, we train a variant of the All Convolutional  Network~\citep{SpringenbergDosovitskiyBroxRiedmiller2014}.
The model is a $3$-hidden layers convolutional network with $96$ channels.
We use $5\times 5$ convolutions  with a padding of size $1$, and a stride of size 1 everywhere except on the second convolution where we use a stride of size $2$.
We adopt the Leaky ReLU activation functions after each convolution. Lastly, we use a global average pooling of size $8\times 8$ to obtain the desired output size. 
Furthermore, the weights are initialized with Xavier Normal initializer~\citep{GlorotBengio2010} while each bias of size $l$ is initialized uniformly between $-1/{\sqrt{l}}$ and $1/\sqrt{l}$.\\
For the CIFAR-10 dataset, we train a ResNet-20 network, \ie, a ResNet network from \citet{HeZhangRenSun2016} with $20$ layers.
The weights are initialized with Kaiming Normal initializer~\citep{HeZhangRenSun2015} and each bias of size $l$ is initialized uniformly between $-1/{\sqrt{l}}$ and $1/\sqrt{l}$.

\subsubsection{Optimization}
\label{sec:optimization}

For the (Fashion-)MNIST datasets, we learn the parameters of our prior distributions $\Pcal_1,\dots,\Pcal_T$ by using Adam optimizer for $T=10$ epochs with a learning rate of $10^{-3}$ and a batch size of $32$ (the other parameters of Adam are left by default).
Moreover, the parameters of the posterior distribution $\AQ$ are learned for one epoch with the same batch size and optimizer (except that the learning rate is either $10^{-4}$ or $10^{-6}$).
For the CIFAR-10 dataset, the parameters of the priors $\Pcal_1,\dots,\Pcal_T$ are learned for $T=100$ epochs, and the posterior distribution $\AQ$ for $10$ epochs with a batch size of $32$ by using Adam optimizer as well. 
Additionally, the learning rate to learn the prior for CIFAR-10 is $10^{-2}$.

\subsubsection{Bounds}
For \blanchard's bounds, the set of hyperparameters is defined as \mbox{$\Bbf{=}\{ b{\in}\Nbb \;\vert\;  b{=}\sqrt{x},\ (x{+}1){\le}2\sqrt{m} \}$}, \ie, such that \blanchard's bounds can be tighter than \rivasplata's ones.
We fixed the set of hyperparameters for \catoni as $\Cbf{=}\left\{10^{k} \vert k{\in}\{-3, -2, \dots, +3\}\right\}$.
We try different values for $\sigma^2 {\in} \{10^{-3}, 10^{-4}, 10^{-5}, 10^{-6}\}$ associated with the disintegrated KL divergence $\ln\frac{\AQ(h)}{\P(h)}= \frac{1}{2\sigma^2}(\| \wbf{+}\epsilonbf{-}\vbf_t\|^2_{2} {-}\|\epsilonbf\|^2_{2})$, the ``normal'' Rényi divergence $\D_2(\Q\|\P){=}\tfrac{1}{\sigma^2}\|\wbf{-}\vbf_t\|_{2}^{2}$ and the KL divergence $\KL(\Q\|\P){=}\tfrac{1}{2\sigma^2}\|\wbf{-}\vbf_t\|_{2}^{2}$.
For all the figures, the values are averaged over $400$ deterministic NNs sampled from $\AQ$ (the standard deviation is small and presented in the Appendix~\ref{ap:details}).
We additionally report as \stoNN (Corollary~\ref{corollary:nn-sto}) the randomized bound value and KL divergence $\KL(\Q\|\P){=}\tfrac{1}{2\sigma^2}\|\wbf{-}\vbf_t\|_{2}^{2}$ associated with the model learned by \ours, meaning that $n{=}400$ and that the test risk reported for \ours also corresponds to the risk of the stochastic NN approximated with these \mbox{$400$ NNs}.

\subsection{Results}

\begin{figure}[!t]
    \centering
    \includegraphics[width=1.0\linewidth]{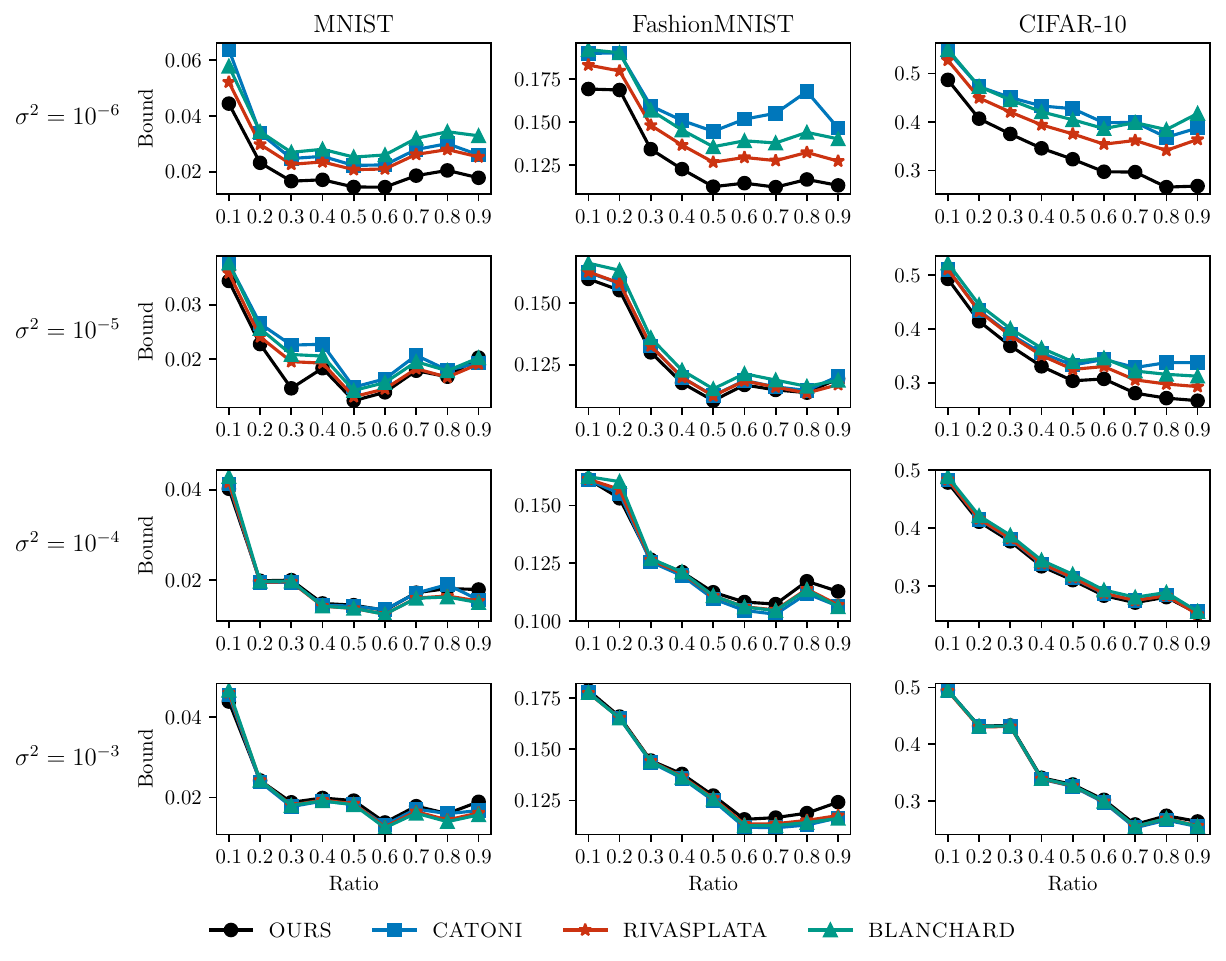}
    \caption{
    Evolution of the bound values in terms of the split ratio. 
    The \mbox{x-axis} represents the different split ratios, and the y-axis represents the bound values obtained after their optimization using our Training Method.
    Each row corresponds to a given variance $\sigma^2$, and each column corresponds to a dataset (MNIST, Fashion-MNIST, or CIFAR-10).
    In this figure, we consider a learning rate of $10^{-6}$.
    }
    \label{figure:exp-4}
\end{figure}

\begin{figure}[!h]
    \centering
    \includegraphics[width=1.0\linewidth]{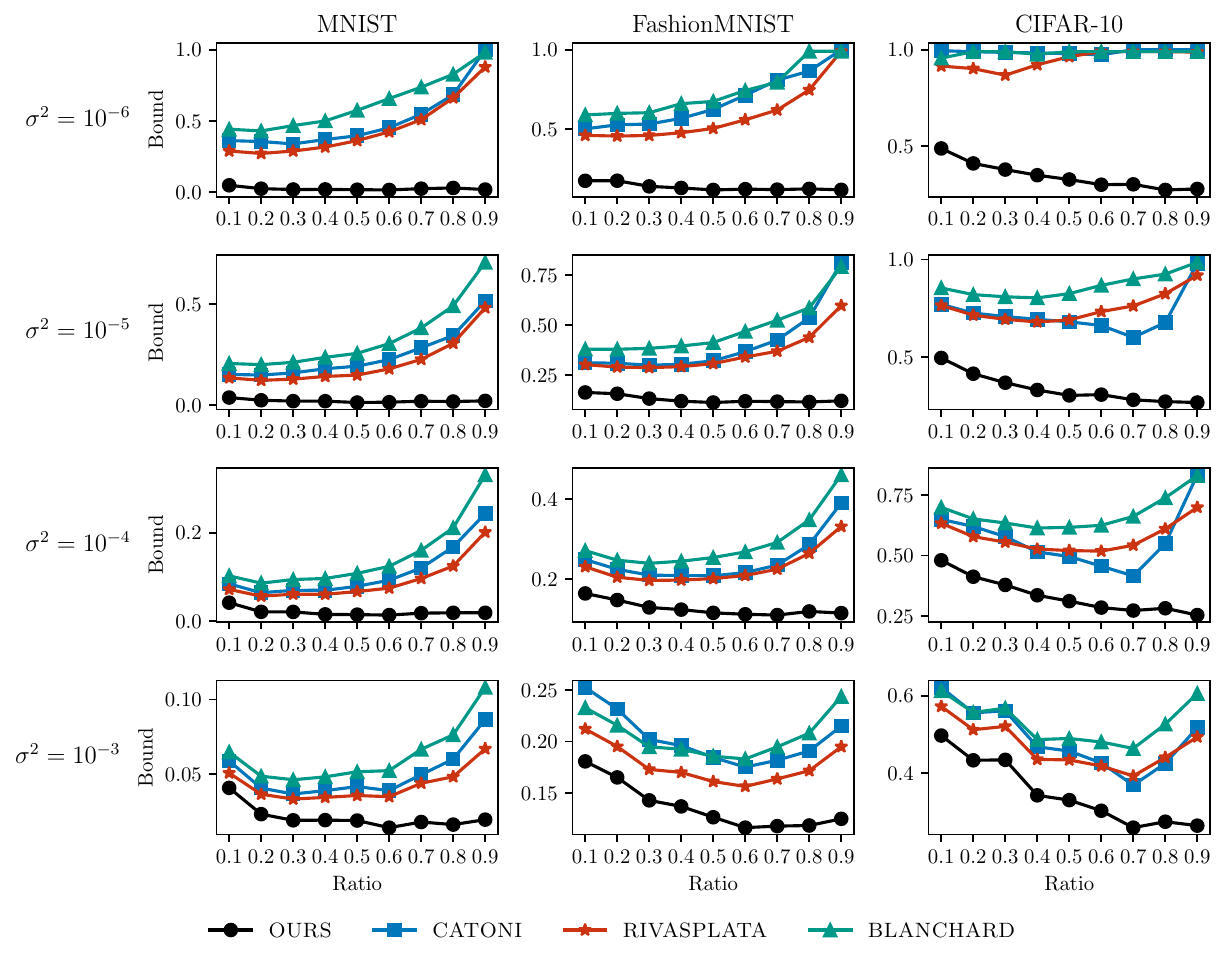}
    \caption{
    Evolution of the bound values in terms of the split ratio. 
    The x-axis represents the different split ratios, and the y-axis represents the bound values obtained after their optimization using our Training Method.
    Each row corresponds to a given variance $\sigma^2$, and each column corresponds to a dataset (MNIST, Fashion-MNIST, or CIFAR-10).
    In this figure, we consider a learning rate of $10^{-4}$.
    }
    \label{figure:exp-5}
\end{figure}

\subsubsection[Analysis of the influence of the split ratio]{Analysis of the influence of the split ratio between $\Scal_{\text{prior}}$ and $\Scal$}
Figures~\ref{figure:exp-4} and~\ref{figure:exp-5} study the evolution of the bound values after optimizing the bounds with our Training Method for different parameters.
Specifically, the split ratio of the original train set varies from $0.1$ to $0.9$ ($0.1$ means that $m_{\text{prior}}= 0.1(m+m_{\text{prior}})$), for four variances values $\sigma^2$ and the two learning rates ($10^{-6}$ and $10^{-4}$).
For the sake of readability, we present detailed results when the split ratio is $0$ in Table~\ref{table:1_prior_0.0}.
We first remark that the behavior is different for the two learning rates. 
On the one hand, for lr=$10^{-6}$, the mean bound values are close to each other, which is not surprising since the disintegrated KL divergences and the Rényi divergences are close to zero (see Tables~\ref{table:1_prior_0.1} to~\ref{table:1_prior_0.9}). 
Moreover, for MNIST and Fashion-MNIST, there is a trade-off between learning a good prior with $\Scal_{\text{prior}}$ and minimizing a generalization bound with $\Scal$.
In this case, the split ratio $0.5$ appears to be a good choice to obtain a tight (disintegrated) PAC-Bayesian bound. 
This ratio is widely used in the PAC-Bayesian literature (see, \eg, in the context of linear classifiers \citep{GermainLacasseLavioletteMarchand2009}, majority votes \citep{ZantedeschiViallardMorvantEmonetHabrardGermainGuedj2021}, and neural networks
\citep{LetarteGermainGuedjLaviolette2019,PerezOrtizRivasplataShaweTaylorSzepesvari2020}).
On the other hand, when lr=$10^{-4}$, the mean bound values tend to increase when the split ratio increases as well for the bounds introduced in the literature (\ie, for \blanchard, \catoni, and \rivasplata), while the mean bound values of our bound remain low.
Indeed, $m$ decreases as long as the split ratio increases, which has the effect of increasing the bound value drastically when the disintegrated KL divergence is high.
We further explain why the disintegrated KL divergence can become high for the disintegrated bounds of the literature.
To do so, we will now restrict our study to a split ratio of $0.5$ in order to {\it (i)} compare the tightness of the bounds, {\it (ii)} understand why the disintegrated bounds of the literature diverge.

\subsubsection[Comparison between disintegrated and "classic" bounds]{Comparison between disintegrated and ``classic'' bounds}

\begin{figure}[!t]
    \centering
    \includegraphics[width=1.0\linewidth]{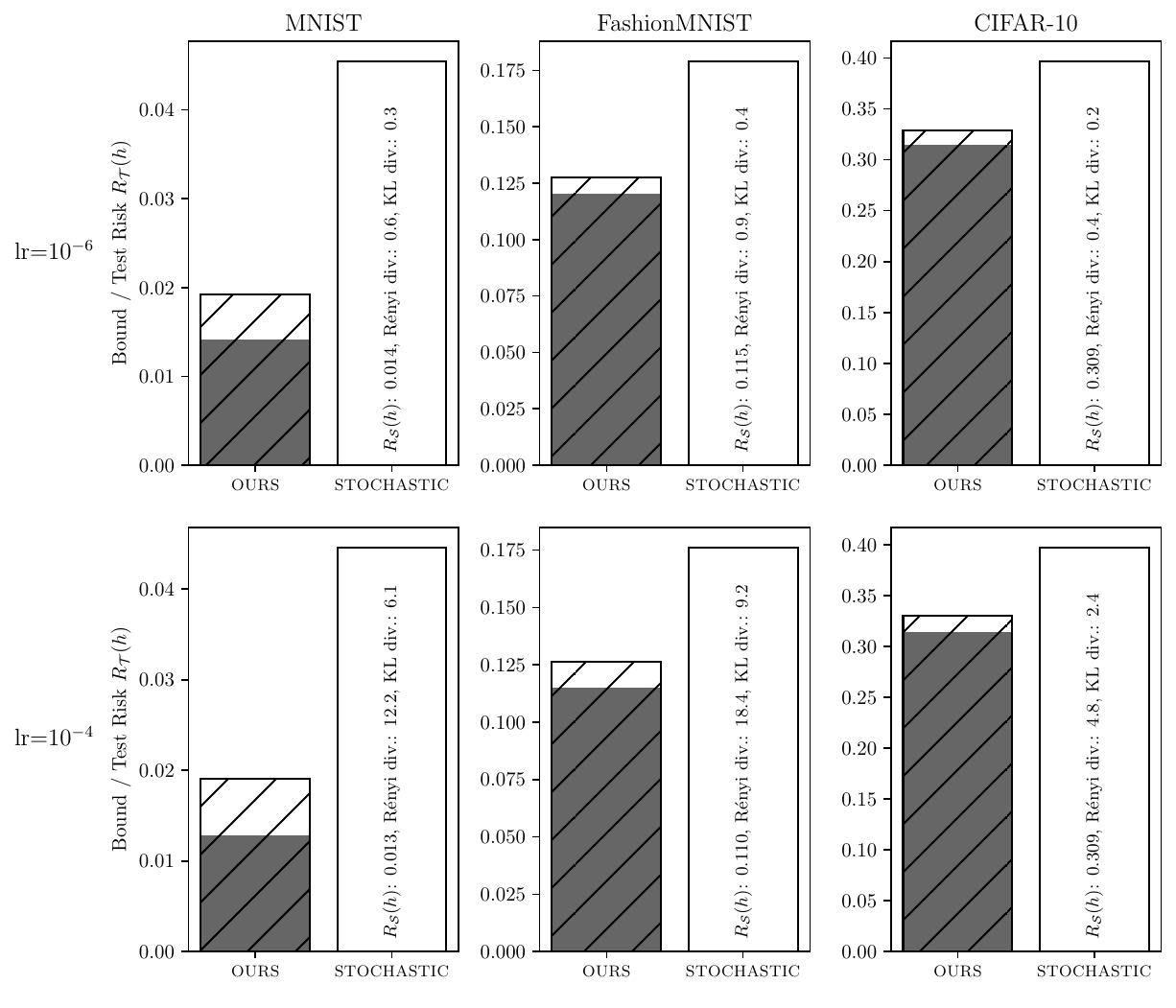}
    \caption{%
    The values of the PAC-Bayes bound (Corollary~\ref{corollary:nn-sto}) and the values of the disintegrated bound (Corollary~\ref{corollary:nn}) for learning rates of $10^{-4}$ and $10^{-6}$, and a split ratio is $0.5$.
    The y-axis shows the values of the bounds (the hatched bar for \ours (Corollary~\ref{corollary:nn}) and the white bar for \stoNN (Corollary~\ref{corollary:nn-sto})) and the test risks $\Risk_{\Tcal}(h)$ (gray shaded bar).
  We also report the values of the empirical risk $\Risk_{\Scal}(h)$, the Rényi divergence (associated with \ours' bound), and the KL divergence (associated with \stoNN's bound). 
}
    \label{figure:exp-2}
\end{figure}

We first compare the ``classic'' PAC-Bayesian bound (Corollary~\ref{corollary:nn-sto}) and our disintegrated PAC-Bayesian bound (Corollary~\ref{corollary:nn}).
To do so, we fix the variance $\sigma^2{=}10^{-3}$ (along with the split ratio equals $0.5$).
We report in Figure~\ref{figure:exp-2}, the mean bound values associated with \ours (\ie, the Training Method that minimizes our bound) and \stoNN 
(we recall that \stoNN is the PAC-Bayesian bound of Corollary~\ref{corollary:nn-sto} on the model learned by \ours).
Actually, \ours leads to more precise bounds than the randomized \stoNN even if the two empirical risks are the same and the KL divergence is smaller than the Rényi one.
This imprecision is due to the non-avoidable sampling according to $\Q$ done in the randomized PAC-Bayesian bound of Corollary~\ref{corollary:nn-sto} (the \mbox{higher $n$}, the tighter the bound).
Thus, using a disintegrated PAC-Bayesian bound avoids sampling a large number of NNs to obtain a low risk.
This confirms that our framework makes sense for practical purposes and has a great advantage in terms of time complexity when computing the bounds.

\begin{figure}[!h]
    \centering
    \includegraphics[width=1.0\linewidth]{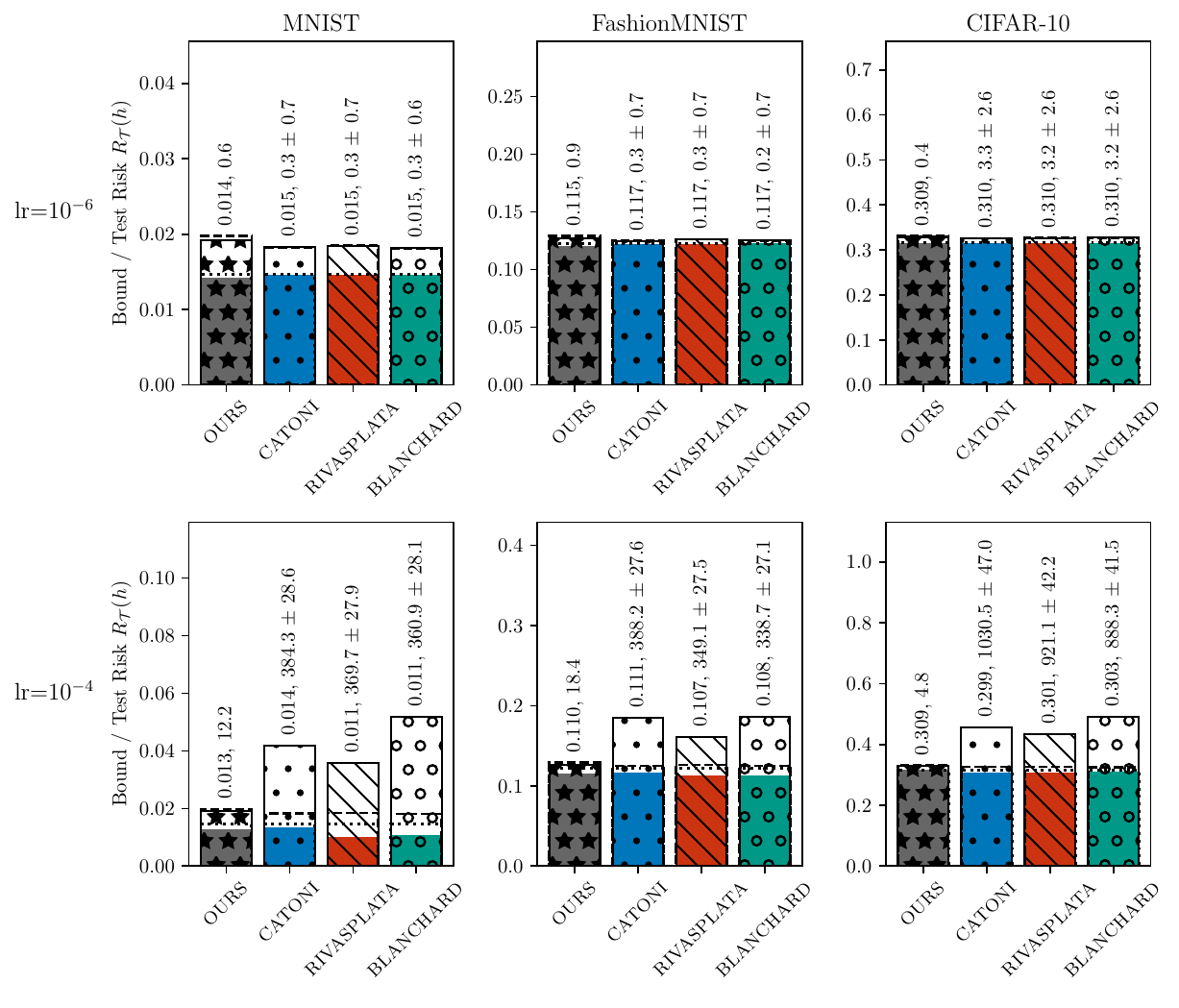}
    \caption{
    The value of the bounds (hatched bars) and the test risks (colored bars) for Corollary~\ref{corollary:nn} (``\ours'') and Corollary~\ref{corollary:nn-rbc} (``\catoni'', ``\rivasplata'' and ``\blanchard'') in two different settings, \ie, with a learning rate of $10^{-6}$ and $10^{-4}$ and with split ratio of $0.5$.
    We also plot the value of the bounds (the dashed lines) and the test risks (the dotted lines) before executing Step {\bf 2)} of our Training Method.
    The y-axis shows the values of the bounds and the test risks $\Risk_{\Tcal}(h)$.
    The empirical risk $\Risk_{\Scal}(h)$ is presented above each bar.
    Moreover, the second value represents the mean value of the divergence (the standard deviations are also given for the disintegrated bounds of the literature).
    }
    \label{figure:exp-1}
\end{figure}

\subsubsection{Analysis of  the tightness of the disintegrated bounds}

We now compare the tightness of the different disintegrated PAC-Bayesian bounds (\ie, our bound and the ones in the literature). 
We study, as before, the case where the split ratio is $0.5$ and the variance $\sigma^2=10^{-3}$.
We report in Figure~\ref{figure:exp-1} for \ours, \rivasplata, \blanchard and \catoni, the mean bounds values; the mean test risk $\Risk_{\Tcal}(h)$ before (\ie, with the prior $\P$) and after applying Step {\bf 2)} (\ie, with the posterior $\AQ$).
Moreover, we report above the bars the mean train risks $\Risk_{\Scal}(h)$ and the mean/standard deviation divergence values obtained after Step {\bf 2)}, \ie, the Rényi divergence $D_2(\AQ\|\P){=}\tfrac{1}{\sigma^2}\|\wbf{-}\vbf_t\|_{2}^{2}$ for \ours and the disintegrated KL divergence $\ln\frac{\AQ(h)}{\P(h)}{=}\tfrac{1}{2\sigma^2}\left[\| \wbf{+}\epsilonbf{-}\vbf_t\|^2_{2} {-}\|\epsilonbf\|^2_{2}\right]$ for the others.
First of all, we can remark that we observe two different behaviors for lr=$10^{-4}$ and lr=$10^{-6}$. 
For lr=$10^{-6}$, the bound values are close to each other, as well as the empirical risks and the divergences (which are close to $0$).
In Figure~\ref{figure:exp-1}, we observe that the bound values and the test risks are close to the one associated with the prior distribution because the divergence is close to $0$.
This is probably due to the fact that the learning rate is too small, implying that the bounds are not optimized.
With a higher learning rate of lr=$10^{-4}$, we observe that our bound remains tight while the disintegrated bounds of the literature are looser.
Hopefully, our bound is improved after performing Step {\bf 2)} of our Training Method.
However, for the bounds of the literature, the value of the disintegrated KL divergence is large, making the bounds looser after executing Step {\bf 2)}.
We now investigate the reasons for the divergence of the bounds
by looking at the influence of the variance $\sigma^2$.

\begin{figure}[t]
    \centering
    \includegraphics[width=1.0\linewidth]{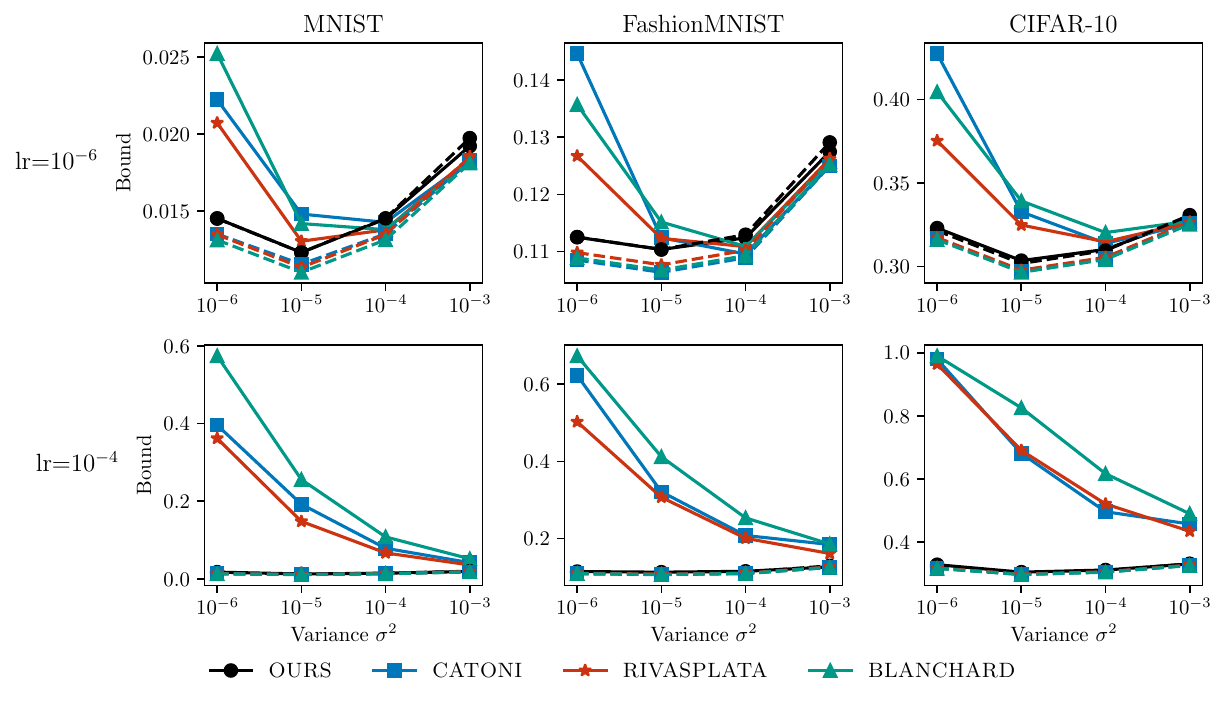}
    \caption{
    We plot the evolution of the mean bound values (the plain lines) in terms of the variance $\sigma^2$ after optimizing the bounds with our Training Method.
    Moreover, we plot the mean bound values (the dashed lines) obtained before executing the Step {\bf 2)} of our Training Method.
    The variance is represented on the x-axis, while the bound values are represented on the y-axis.  
    Furthermore, each row corresponds to a given learning rate ($10^{-6}$ or $10^{-4}$), and each column corresponds to a dataset (either MNIST, FashionMNIST, or CIFAR-10).
    The split ratio considered is $0.5$.
    }
    \label{figure:exp-3}
\end{figure}

\subsubsection{Analysis of the influence of the variance}

Given a split ratio of $0.5$ and lr$\in\{10^{-6}, 10^{-4}\}$, we report in Figure~\ref{figure:exp-3} the evolution of the bound values associated with \ours, \rivasplata, \blanchard, and \catoni when the variance varies from $10^{-6}$ to $10^{-3}$.
First of all, the important point is that \ours behaves differently than \rivasplata, \blanchard, and \catoni.
Indeed, for both learning rates, when $\sigma^2$ decreases, the value of our bound remains low, while the others increase drastically due to the explosion of the disintegrated KL divergence term (see Table~\ref{table:1_prior_0.5} in Appendix~\ref{ap:details} for more details).
Concretely, the disintegrated KL divergence in Corollary~\ref{corollary:nn-rbc} involves the noise $\epsilonbf$  through  $\frac{1}{2\sigma^2}\| \wbf{+}\epsilonbf{-}\vbf_t\|_{2}^2{-}\|\epsilonbf\|^2_{2}
$ compared to our divergence which is $\frac{1}{\sigma^2}\| \wbf{-}\vbf_t\|_{2}^2$ (without noise).
Then, the sampled noise during the optimization procedure $\epsilonbf$ influences the disintegrated KL divergence, making it prone to high variations during training (depending thus $\sigma^2$).
To illustrate the difference during the optimization, we focus on the objective function (detailed in Appendix~\ref{ap:evaluation-minimization}) of Corollary~\ref{corollary:nn} and Corollary~\ref{corollary:nn-rbc} (Equation~\eqref{eq:nn-rivasplata}).
Roughly speaking, the divergence in Corollary~\ref{corollary:nn} does not depend on the sampled hypothesis $h$ (with weights $\omegabf+\epsilonbf$), while the divergence of Equation~\eqref{eq:nn-rivasplata} does. 
In consequence, the derivatives are less dependent on $h$ for Corollary~6 than for Equation~\eqref{eq:nn-rivasplata}.
To be convinced of this, we propose to study the gradient with respect to the current mean vector $\omegabf$.
On the one hand, the gradient $\frac{\partial \Risk_{\Scal}(h)}{\partial \omegabf}$ of the risk \textit{w.r.t.} $\omegabf$ is the same for both bounds; hence, the phenomenon cannot come from this derivative. 
On the other hand, the gradients of the divergence in Equation~\eqref{eq:nn-rivasplata} and Corollary~\ref{corollary:nn} are respectively
\begin{align*}
    \frac{\partial}{\partial \omegabf}\!\!\LB\frac{1}{m}\!\!\LP\!\frac{\LN \omegabf{+}\epsilonbf{-}\vbf_t\RN^2_{2}\!{-}\LN\epsilonbf\RN^2_{2}}{2\sigma^2}\RP\RB &= \frac{\partial}{\partial \omegabf}\LB \frac{1}{m2\sigma^2}\|\omegabf{+}\epsilonbf{-}\vbf_t\|_2^2 \RB\\
    &= \frac{1}{m\sigma^2}\LP \omegabf{+}\epsilonbf{-}\vbf_t\RP = \diamondsuit,\\
    \text{and}\quad\quad  \frac{\partial}{\partial \omegabf}\!\!\LB\frac{1}{m}\!\!\LP \frac{\|\omegabf{-}\vbf_t\|_{2}^{2}}{\sigma^2}\RP\RB= &\frac{\partial}{\partial \omegabf}\LB \frac{1}{m\sigma^2}\|\omegabf{-}\vbf_t\|_2^2 \RB\\
    = &\frac{2}{m\sigma^2}\LP \omegabf{-}\vbf_t\RP = \heartsuit.
\end{align*}
From the two derivatives, we deduce that $\diamondsuit = \frac{1}{2}\heartsuit + \frac{1}{m\sigma^2}\epsilonbf$.
Hence, each gradient step involves a noise in the gradient of the disintegrated KL divergence \mbox{$\frac{1}{m\sigma^2}\epsilonbf \sim \Ncal({\bf 0}, \frac{1}{m}\Ibf_d)$}, which is high for a small $m$.
This randomness causes the disintegrated KL divergence $\frac{1}{2\sigma^2}\LN \omegabf{+}\epsilonbf{-}\vbf_t\RN^2_{2}\!{-}\LN\epsilonbf\RN^2_{2}$ to be larger when $\sigma^2$ decreases since {\it (i)} the divergence is divided by $2m\sigma^2$ and {\it (ii)} the deviation between $\omegabf$ and $\vbf_t$ increases.
In conclusion, this makes the objective function (\ie, the bound) subject to high variations during the optimization, implying higher final bound values.
Thus, the Rényi divergence has a valuable asset over the disintegrated KL divergence since it does not depend on the sampled noise $\epsilonbf$.

\subsubsection{Take-home message from the experiments}
To summarize, our experiments show that our disintegrated bound is, in practice, tighter than the ones in the literature.
This tightness allows us to precisely bound the true risk $\Risk_{\Dcal}(h)$ (or the test risk $\Risk_{\Tcal}(h)$); thus, the model selection from the disintegrated bound is effective.
Moreover, we show that our bound is more easily optimizable than the others. 
This is mainly due to the disintegrated KL divergence, which depends on the sampled hypothesis $h$ with weights $\omegabf{+}\epsilonbf$.
Indeed, the gradients of the disintegrated KL divergence with respect to $\omegabf$ include the noise $\epsilonbf$, making the gradient inaccurate (especially with ``high'' learning rate and small variance $\sigma^2$).

\section{Toward information-theoretic bounds}
\label{sec:info-theoretic}
Before concluding, we discuss another interpretation of the disintegration procedure through Theorem~\ref{theorem:mutual-info} below.
Actually, the Rényi divergence between $\P$ and $\Q$ is sensitive to the choice of the learning \mbox{sample $\Scal$}: when the posterior $\Q$ learned from $\Scal$ differs greatly from the prior $\P$ the divergence is high. 
To avoid such a behavior, we consider Sibson's mutual information~\citep{Verdu2015} which is a measure of dependence between the random variables $\Scal\!\in\!\Zcal^m$ and $h\!\in\!\Hcal$.
It involves an expectation over all the learning samples of a given size $m$ and is defined for a given $\alpha{>}1$ by
\begin{align*}
I_{\alpha}(h{;}\Scal) 
&\triangleq \min_{\P\in\Mcal^{*}(\Hcal)}  \frac{1}{\alpha{-}1}\!\ln\!\LB\EE_{\Scal\sim\Dcal^{m}}\EE_{h\sim \P}\!\LB\!\frac{\AQ(h)}{\P(h)}\!\RB^{\alpha}\RB.
\end{align*}

\noindent The higher $I_{\alpha}(h{;}\Scal)$, the higher the correlation is, meaning that the sampling of $h$ is highly dependent on the choice of $\Scal$. 
This measure has two interesting properties: it generalizes the mutual information~\citep{Verdu2015}, and it can be related to the Rényi divergence.
Indeed, let $\rho(h, \Scal){=} \AQ(h)\Dcal^{m}(\Scal)$, \textit{resp.} $\pi(h, \Scal){=} \P(h)\Dcal^{m}(\Scal)$, be the probability of sampling both  $\Scal{\sim}\Dcal^m$ and  $h{\sim}\AQ$, \textit{resp.}  $\Scal{\sim}\Dcal^m$ and $h{\sim}\P$.
Then we can write: 
\begin{align}
I_{\alpha}(h{;}\Scal) 
 &=\!\!\min_{\P\in\Mcal^{*}(\Hcal)} \frac{1}{\alpha{-}1}\!\ln\!\Bigg[\!\EE_{\Scal\sim\Dcal^{m}}\EE_{h\sim \P}\!\LB\!\frac{\AQ(h)\Dcal^{m}(\Scal)}{\P(h)\Dcal^{m}(\Scal)}\!\RB^{\alpha}\!\!\Bigg]\nonumber\\
 &=\!\!\min_{\P\in\Mcal^{*}(\Hcal)} D_{\alpha}(\rho\|\pi).\label{eq:mutual-info-min}
\end{align}

\noindent From~\citet{Verdu2015} the 
optimal prior $\P^*$ minimizing Equation~\eqref{eq:mutual-info-min} is  a {\it distribution-dependent} prior: 
\begin{align*}
\displaystyle \P^*(h)=\frac{\LB\EE_{\Scal'{\sim}\Dcal^{m}}\AQprime (h)^{\alpha}\RB^{\frac{1}{\alpha}}}{\EE_{h'{\sim}\P}\tfrac{1}{\P(h')}\LB\EE_{\Scal'{\sim}\Dcal^{m}}\AQprime (h')^{\alpha}\RB^{\frac{1}{\alpha}}}.
\end{align*}
This leads to an {\it Information-Theoretic generalization 
bound}\,\footnote{We provide a mutual information-based bound in Appendix~\ref{ap:proof-theorem-mutual-info}.}.

\begin{restatable}[Disintegrated Information-Theoretic Bound]{theorem}{theoremmutualinfo}\label{theorem:mutual-info}
For any distribution $\Dcal$ on $\Zcal$, for any hypothesis set $\Hcal$, for any measurable function $\phi\!:\!\Hcal{\times} \Zcal^{m}{\to}\Rpe$, \mbox{for any $\alpha\!>\!1$}, for any $\delta\in(0,1]$, for any algorithm $A\!:\!\Zcal^{m}\times\Mcal^{*}(\Hcal){\rightarrow} \Mcal(\Hcal)$, we have
\begin{align*}
   \PP_{\substack{\Scal\sim\Dcal^{m},\\h\sim \AQ}} \!\left(
   \frac{\alpha}{\alpha{-}1}\!\ln\!\LP\phi(h,\!\Scal)\RP \le I_{\alpha}(h'{;} \Scal')\!+\!  \ln\left[\frac{1}{\delta^{\frac{\alpha}{\alpha{-}1}}}\!\EE_{\Scal'{\sim}\Dcal^{m}}\EE_{h'{\sim} \P^*}\!\!\LB\phi(h'\!, \Scal')^{\frac{\alpha}{\alpha-1}}\RB\right] 
   \right)\ge 1{-}\delta.
\end{align*}
\end{restatable}

\noindent Note that \citet[Cor.4]{EspositoGastparIssa2020} introduced a bound based on the Sibson's mutual information, but, as discussed in Appendix~\ref{ap:proof-theorem-mutual-info}, Theorem~\ref{theorem:mutual-info} leads to a tighter bound.
From a theoretical view, Theorem~\ref{theorem:mutual-info} brings a different philosophy than the disintegrated PAC-Bayes bounds. 
Indeed, in Theorems~\ref{theorem:disintegrated} and~\ref{theorem:disintegrated-lambda}, given $\Scal$, the Rényi divergence  $D_{\alpha}(\AQ\|\P)$ suggests that the learned posterior $\AQ$ should be close enough to the prior $\P$  to get a low bound.
While in Theorem~\ref{theorem:mutual-info}, the Sibson's mutual information $I_{\alpha}(h'; \Scal')$ suggests that the random variable $h$ has to be {\it not too much correlated} to $\Scal$.
However, the bound of Theorem~\ref{theorem:mutual-info} is not computable in practice due notably to the sample expectation over the unknown distribution $\Dcal$ in $I_{\alpha}$.
An exciting line of future works could be to study how we can make use of  Theorem~\ref{theorem:mutual-info} in practice.
 
\section{Conclusion and future works}
\label{sec:conclu}

We provide a new and general disintegrated PAC-Bayesian bound (Theorem~\ref{theorem:disintegrated}) in the family of Equation~\eqref{eq:disintegration}, \ie, when the derandomization step consists in {\it (i)} learning a posterior distribution $\AQ$ on the classifiers set (given an algorithm, a learning sample $\Scal$ and a prior distribution $\P$) and {\it (ii)} sampling a hypothesis $h$ from this posterior $\AQ$.   
While our bound can be looser than the ones of \citet{RivasplataKuzborskijSzepesvariShaweTaylor2020,BlanchardFleuret2007,Catoni2007}, it provides nice opportunities for learning deterministic classifiers.
Indeed, our bound can be used not only to study the theoretical guarantees of deterministic classifiers but also to derive self-bounding algorithms (based on the bound optimization) that are more stable and efficient than the ones we obtain from the bounds of the literature.
Concretely, the bounds of \citet{RivasplataKuzborskijSzepesvariShaweTaylor2020,BlanchardFleuret2007,Catoni2007} depend on two terms related to the classifier drawn: the risk and the ``disintegrated KL divergence'', while in our bound the (Rényi) divergence term depends on the hypothesis set, implying that the divergence remains the same whatever which classifier is drawn.  
In this sense, our bound is more stable as the learning algorithm seeking to minimize the bound allows, in practice, to choose a better hypothesis than with the bounds of \citet{RivasplataKuzborskijSzepesvariShaweTaylor2020,BlanchardFleuret2007,Catoni2007}.
We have illustrated the interest of our bound on neural networks, but our result could be instantiated to other well-known settings such as linear classifiers~\citep{GermainLacasseLavioletteMarchand2009} or the majority vote classifier~\citep{ZantedeschiViallardMorvantEmonetHabrardGermainGuedj2021}. 

\noindent Our general framework opens the way to the study of other machine learning settings by exploiting the proven \textit{randomized} PAC-Bayesian theorems, for example, for Domain Adaptation~\citep{GermainHabrardLavioletteMorvant2020}, Adversarial Robustness~\citep{ViallardVidotHabrardMorvant2021} or Transductive Learning~\citep{BeginGermainLavioletteRoy2014}.

\noindent Despite being an important step towards the practical use of PAC-Bayes guarantees, our disintegrated bounds arguably have a drawback: we sample a hypothesis from a distribution instead of obtaining a bound for all the possible hypotheses, like for uniform convergence bounds.
While uniform convergence bounds can be vacuous~\citep{NagarajanKolter2019b}, they hold (with high probability on the choice of the learning sample) for all hypotheses including the one with the best guarantee (\ie, the one minimizing the bound). 
In the case of disintegrated bounds, we learn a distribution on the hypothesis set, and then we sample a hypothesis according to this distribution.
Hence, there is a small probability (\ie, less than $\delta$) of sampling a \textit{bad} hypothesis.
An interesting research direction is comparing disintegrated and uniform convergence bounds to understand in which cases using disintegrated bounds can be better than using uniform convergence bounds.
Knowing that there are connections between (agnostic) PAC-learnability and uniform convergence (see, \eg, \citet{ShalevShwartzBenDavid2014}), we believe that defining a new notion of PAC-learnability, which better fits with the disintegrated framework, could help to provide such a comparison.

\backmatter

\bmhead{Acknowledgments}
This work was partially funded by the French ANR Project APRIORI ANR-18-CE23-0015.
Pascal Germain is supported by the Canada CIFAR AI Chair Program, and the NSERC Discovery grant RGPIN-2020-07223.
We would like to thank the reviewers for their valuable comments and their suggestions to improve the paper.

\section*{Declarations}

\noindent\textbf{Funding.} This work was partially funded by the French ANR Project APRIORI ANR-18-CE23-0015.
Pascal Germain is supported by the Canada CIFAR AI Chair Program, and the NSERC Discovery grant RGPIN-2020-07223.\\

\noindent\textbf{Conflict of interest/Competing interests.} The authors have no relevant financial or non-financial interests to disclose.\\

\noindent\textbf{Ethics approval.} Not applicable.\\

\noindent\textbf{Consent to participate.} Not applicable.\\

\noindent\textbf{Consent for publication.} Not applicable.\\

\noindent\textbf{Availability of data and materials.} Not applicable.\\ 

\noindent\textbf{Code availability.} The code is available on Github at \href{https://github.com/paulviallard/MLJ-Disintegrated-PB}{https://github.com/paulviallard/MLJ-Disintegrated-PB}.\\

\noindent\textbf{Authors' contributions.} Conceptualization: {\it Dr. P. Viallard, Dr. E. Morvant, Dr. P. Germain, Pr. A. Habrard}; Formal analysis and investigation: {\it Dr. P. Viallard}; Software: {\it Dr. P. Viallard}; Writing - original draft preparation: {\it Dr. P. Viallard, Dr. E. Morvant}; Writing - review and editing: {\it Dr. P. Germain, Pr. A. Habrard}; Funding acquisition: {\it Dr. E. Morvant, Dr. P. Germain, Pr. A. Habrard}; Supervision: {\it Dr. E. Morvant, Dr. P. Germain, Pr. A. Habrard}.

\bibliography{manuscript}

\newpage
\begin{appendices}

\section*{}

This Appendix is structured as follows. 
We give the proof of Theorem~\ref{theorem:general-classic}, Theorem~\ref{theorem:disintegrated}, Corollary~\ref{corollary:disintegrated}, Theorem~\ref{theorem:disintegrated-lambda}, 
Proposition~\ref{prop:lambda-min}, Corollary~\ref{corollary:nn}, Corollary~\ref{corollary:nn-rbc}, and Corollary~\ref{corollary:nn-sto} in Appendix~\ref{ap:proof-classic}, Appendix~\ref{ap:proof-disintegrated}, Appendix~\ref{ap:proof-corollary-disintegrated}, Appendix~\ref{ap:proof-disintegrated-lambda}, Appendix~\ref{ap:proof-disintegrated-lambda}, Appendix~\ref{ap:proof-corollary-nn}, Appendix~\ref{ap:proof-corollary-nn-rbc}, and Appendix~\ref{ap:proof-nn-sto} respectively.
We also discuss the minimization and the evaluation of the bounds introduced in the different corollaries in Appendix~\ref{ap:evaluation-minimization}.
Additionally, Appendix~\ref{ap:proof-theorem-mutual-info} is devoted to Theorem~\ref{theorem:mutual-info}.
Appendix~\ref{ap:details} provides an exhaustive list of numerical results.

\section{Proof of Theorem~\ref{theorem:general-classic}}
\label{ap:proof-classic}

\theoremclassical*
\begin{proof}
By the Donsker-Varadhan's variational formula \citep[see \eg,][Lemma 3]{BeginGermainLavioletteRoy2016}, we have
\begin{align}
    \forall \Q\in\Mcal(\Hcal),\quad \EE_{h\sim\Q}\ln(\phi(h,\Scal)) &\le \KL(\Q\|\P) + \ln\LB\EE_{h\sim\P}\phi(h,\Scal)\RB.\label{eq:proof-bound-classic-1}
\end{align}
By Markov's inequality and taking the logarithm to both sides, we have 
\begin{align}
    \PP_{\Scal\sim\Dcal^m}\LB \ln\LB\EE_{h\sim\P}\phi(h,\Scal)\RB \le \ln\LB\frac{1}{\delta}\EE_{\Scal\sim\Dcal^m}\EE_{h\sim\P}\phi(h,\Scal)\RB \RB\ge 1-\delta.\label{eq:proof-bound-classic-2}
\end{align}
By merging Equations~\eqref{eq:proof-bound-classic-1} and~\eqref{eq:proof-bound-classic-2}, we obtain Equation~\eqref{eq:kl-bound-classic}.

\noindent{}The proof of Equation~\eqref{eq:renyi-bound-classic} is similar to the one of Equation~\eqref{eq:kl-bound-classic}.
Indeed, from the Rényi change of measure~\citep[see \eg,][Theorem 8]{BeginGermainLavioletteRoy2016}, we have
\begin{align}
    \forall \Q\in\Mcal(\Hcal),\frac{\alpha}{\alpha{-}1}\ln\left[\EE_{h{\sim}\Q}\phi(h,\!\Scal)\right]\le \D_{\alpha}(\Q\|\P) + \ln\left[\EE_{h{\sim}\P}\!\phi(h,\!\Scal)^\frac{\alpha}{\alpha{-}1}\right].\label{eq:proof-bound-classic-3}
\end{align}
By Markov's inequality and taking the logarithm to both sides, we have 
\begin{align}
    \PP_{\Scal\sim\Dcal^m}\LB\ln\left[\EE_{h{\sim}\P}\!\phi(h,\!\Scal)^\frac{\alpha}{\alpha{-}1}\right] \le \ln\left[\frac{1}{\delta}\!\EE_{\Scal{\sim}\Dcal^{m\!}}\EE_{h{\sim}\P}\!\phi(h,\!\Scal)^\frac{\alpha}{\alpha{-}1}\right]\RB\ge 1-\delta.\label{eq:proof-bound-classic-4}
\end{align}
By merging Equations~\eqref{eq:proof-bound-classic-3} and~\eqref{eq:proof-bound-classic-4}, Equation~\eqref{eq:renyi-bound-classic} is obtained.
\end{proof}

\section{Proof of Theorem~\ref{theorem:disintegrated}}
\label{ap:proof-disintegrated}

\theoremdisintegrated*
\begin{proof}
For any sample $\Scal\in\Zcal^m$, prior $\P\in\Mcal^{*}(\Hcal)$ and deterministic algorithm $A$ fixed a priori, let $\AQ=A(\Scal, \P)$ the distribution obtained from the algorithm $A$.
Note that $\phi(h,\!\Scal)$ is a strictly positive random variable.
Hence, from Markov's inequality, we have
\begin{align*}
&\PP_{h\sim \AQ}\LB\phi(h,\!\Scal)\le \frac{2}{\delta}\EE_{h'{\sim} \AQ}\phi(h'\!, \Scal)\RB\ge 1{-}\tfrac{\delta}{2}\\
\iff &\EE_{h\sim \AQ}\Ibf\LB \phi(h,\!\Scal)\le \frac{2}{\delta}\EE_{h'{\sim} \AQ}\phi(h'\!, \Scal) \RB\ge 1{-}\tfrac{\delta}{2}.
\end{align*}
Taking the expectation over $\Scal\sim\Dcal^{m}$ to both sides of the inequality gives 
\begin{align*}
&\EE_{\Scal\sim\Dcal^{m}}\EE_{h\sim \AQ}\Ibf\LB \phi(h,\!\Scal)\le \frac{2}{\delta}\EE_{h'{\sim} \AQ}\phi(h'\!, \Scal) \RB\ge 1-\tfrac{\delta}{2}\\
\iff &\PP_{\Scal\sim\Dcal^{m},h\sim \AQ}\LB \phi(h,\!\Scal)\le \frac{2}{\delta}\EE_{h'{\sim} \AQ}\phi(h'\!, \Scal) \RB\ge 1-\tfrac{\delta}{2}.
\end{align*}
Since both sides of the inequality are strictly positive, we can take the logarithm and multiply by $\frac{\alpha}{\alpha-1}>0$ to obtain
\begin{align*}
\PP_{\Scal\sim\Dcal^{m},h\sim \AQ}\LB\frac{\alpha}{\alpha-1}\ln\LP\phi(h,\!\Scal)\RP \le \frac{\alpha}{\alpha-1}\ln\LP\frac{2}{\delta}\EE_{h'{\sim} \AQ}\phi(h'\!, \Scal)\RP\RB\ge 1-\tfrac{\delta}{2}.
\end{align*}
We develop the right-hand side of the inequality and take the expectation of the hypothesis over the prior distribution $\P$. 
We have for all prior $\P\in\Mcal^{*}(\Hcal)$,
\begin{align*}
 \frac{\alpha}{\alpha-1}\ln\LP\frac{2}{\delta}\EE_{h'{\sim} \AQ}\phi(h'\!, \Scal)\RP
&= \frac{\alpha}{\alpha-1}\ln\LP\frac{2}{\delta}\EE_{h'{\sim}\P}\frac{\AQ(h')}{\P(h')}\phi(h'\!, \Scal)\RP,
\end{align*}
Remark that $\frac{1}{r}+\frac{1}{s}=1$ with $r=\alpha$ and $s=\frac{\alpha}{\alpha-1}$. Hence, we can apply Hölder's inequality:
\begin{align*}
    \EE_{h'{\sim}\P}\frac{\AQ(h')}{\P(h')}\phi(h'\!, \Scal) \le \LB\EE_{h'{\sim}\P}\Bigg(\Bigg[\frac{\AQ(h')}{\P(h')}\Bigg]^{\alpha}\Bigg)\RB^{\frac{1}{\alpha}}\LB\EE_{h'{\sim}\P}\LP\phi(h'\!, \Scal)^{\frac{\alpha}{\alpha-1}}\RP\RB^{\frac{\alpha-1}{\alpha}}.
\end{align*}
Then, since both sides of the inequality are strictly positive, we take the logarithm, add $\ln(\tfrac{2}{\delta})$ and multiply by $\frac{\alpha}{\alpha-1}>0$ to both sides of the inequality, to obtain
\begin{align*}
    &\frac{\alpha}{\alpha{-}1}\ln\!\LP\frac{2}{\delta}\EE_{h'{\sim}\P}\frac{\AQ(h')}{\P(h')}\phi(h'\!, \Scal)\RP\\
    \le &\frac{\alpha}{\alpha{-}1}\ln\!\LP\frac{2}{\delta}\!\LB\EE_{h'{\sim}\P}\LP\Bigg[\frac{\AQ(h')}{\P(h')}\Bigg]^{\alpha}\RP\RB^{\frac{1}{\alpha}}\!\LB\EE_{h'{\sim}\P}\LP\phi(h'\!, \Scal)^{\frac{\alpha}{\alpha-1}}\RP\RB^{\frac{\alpha-1}{\alpha}}\RP\\
    = &\frac{1}{\alpha{-}1}\ln\!\LP\EE_{h'{\sim}\P}\!\LP\Bigg[\frac{\AQ(h')}{\P(h')}\Bigg]^{\alpha}\RP\RP+
    \frac{\alpha}{\alpha{-}1}\ln\frac{2}{\delta} +\ln\!\LP\EE_{h'{\sim}\P}\LP\phi(h'\!, \Scal)^{\frac{\alpha}{\alpha-1}}\RP\!\RP\\
    = &D_{\alpha}(\AQ\|\P) + \frac{\alpha}{\alpha{-}1}\ln\frac{2}{\delta} + \ln\LP\EE_{h'{\sim}\P}\LP\phi(h'\!, \Scal)^{\frac{\alpha}{\alpha-1}}\RP\RP\!.
\end{align*}
From this inequality, we can deduce that
\begin{align}
    \PP_{\Scal\sim\Dcal^{m},h\sim \AQ}\Big[\forall \P\in\Mcal^{*}(\Hcal), &\frac{\alpha}{\alpha-1}\ln\LP\phi(h,\!\Scal)\RP \le
    D_{\alpha}(\AQ\|\P)\nonumber\\
    &+ \frac{\alpha}{\alpha{-}1}\ln\frac{2}{\delta} +\ln\LP\EE_{h'{\sim}\P}\LP\phi(h'\!, \Scal)^{\frac{\alpha}{\alpha-1}}\RP\RP\Big]\ge 1-\tfrac{\delta}{2}.
    \label{eq:disintegrated-proof-1}
\end{align}
Given a prior $\P\in\Mcal^{*}(\Hcal)$, note that $\EE_{h'{\sim}\P}\phi(h'\!, \Scal)^{\frac{\alpha}{\alpha-1}}$ is a strictly positive random variable. 
Hence, we apply Markov's inequality to have
\begin{align*}
    \PP_{\Scal\sim\Dcal^{m}}\LB\EE_{h'{\sim}\P}\LP\phi(h'\!, \Scal)^{\frac{\alpha}{\alpha-1}}\RP \le \frac{2}{\delta}\EE_{\Scal'{\sim}\Dcal^{m}}\EE_{h'{\sim}\P}\LP\phi(h'\!, \Scal')^{\frac{\alpha}{\alpha-1}}\RP\RB\ge 1-\tfrac{\delta}{2}.
\end{align*}
Since the inequality does not depend on the random variable $h\sim \AQ$, we have
\begin{align*}
    &\PP_{\Scal\sim\Dcal^{m}}\!\LB\EE_{h'{\sim}\P}\LP\phi(h'\!, \Scal)^{\!\frac{\alpha}{\alpha-1}}\RP \!\le\! \frac{2}{\delta}\EE_{\Scal'{\sim}\Dcal^{m}}\EE_{h'{\sim}\P}\LP\phi(h'\!, \Scal')^{\!\frac{\alpha}{\alpha-1}}\RP\!\RB\\
    = &\EE_{\Scal\sim\Dcal^{m}}\!\Ibf\!\LB\EE_{h'{\sim}\P}\!\LP\phi(h'\!, \Scal)^{\!\frac{\alpha}{\alpha-1}}\RP \!\le\! \frac{2}{\delta}\EE_{\Scal'{\sim}\Dcal^{m}}\EE_{h'{\sim}\P}\LP\phi(h'\!, \Scal')^{\!\frac{\alpha}{\alpha-1}}\RP\RB\\
    = &\EE_{\Scal\sim\Dcal^{m}}\EE_{h\sim \AQ}\!\Ibf\!\LB\EE_{h'{\sim}\P}\!\LP\phi(h'\!, \Scal)^{\!\frac{\alpha}{\alpha-1}}\RP \!\le\! \frac{2}{\delta}\EE_{\Scal'{\sim}\Dcal^{m}}\EE_{h'{\sim}\P}\LP\phi(h'\!, \Scal')^{\!\frac{\alpha}{\alpha-1}}\RP\RB\\
    = &\PP_{\Scal\sim\Dcal^{m},h\sim \AQ}\!\LB\EE_{h'{\sim}\P}\LP\phi(h'\!, \Scal)^{\!\frac{\alpha}{\alpha-1}}\RP \!\le\! \frac{2}{\delta}\EE_{\Scal'{\sim}\Dcal^{m}}\EE_{h'{\sim}\P}\LP\phi(h'\!, \Scal')^{\!\frac{\alpha}{\alpha-1}}\RP\RB\!.
\end{align*}
Since both sides of the inequality are strictly positive, we take the logarithm to both sides of the inequality, and we add $\frac{\alpha}{\alpha-1}\ln\frac{2}{\delta}$ to have
\begin{align}
    &\PP_{\Scal\sim\Dcal^{m},h\sim \AQ}\LB\EE_{h'{\sim}\P}\LP\phi(h'\!, \Scal)^{\frac{\alpha}{\alpha-1}}\RP \le \frac{2}{\delta}\EE_{\Scal'{\sim}\Dcal^{m}}\EE_{h'{\sim}\P}\LP\phi(h'\!, \Scal')^{\frac{\alpha}{\alpha-1}}\RP\RB\ge 1-\tfrac{\delta}{2} \hspace{0.1cm}\iff\nonumber\\
    &\PP_{\Scal\sim\Dcal^{m},h\sim \AQ}\Bigg[\frac{\alpha}{\alpha{-}1}\ln\frac{2}{\delta}+\ln\LP\EE_{h'{\sim}\P}\LP\phi(h'\!, \Scal)^{\frac{\alpha}{\alpha-1}}\RP\RP
    \le\frac{2\alpha-1}{\alpha{-}1}\ln\frac{2}{\delta}\nonumber\\
    &\hspace{2cm}+\ln\!\LP\EE_{\Scal'{\sim}\Dcal^{m}}\EE_{h'{\sim}\P}\LP\phi(h'\!, \Scal')^{\frac{\alpha}{\alpha-1}}\RP\RP\Bigg]\ge 1{-}\tfrac{\delta}{2}.
    \label{eq:disintegrated-proof-2}
\end{align}
Combining Equations~\eqref{eq:disintegrated-proof-1} and~\eqref{eq:disintegrated-proof-2} with a union bound gives us the desired result.
\end{proof}

\section{Proof of Corollary~\ref{corollary:disintegrated}}
\label{ap:proof-corollary-disintegrated}

\corollarydisintegrated*
\begin{proof} 
Starting from Theorem~\ref{theorem:disintegrated} and rearranging, we have
\begin{align*}
    \PP_{\Scal\sim\Dcal^{m},h\sim \AQ}\!\Bigg[\! &\ln\LP\phi(h,\!\Scal)\RP \le {\frac{2\alpha{-}1}{\alpha}}\ln\frac{2}{\delta}\, +\frac{\alpha{-}1}{\alpha}D_{\alpha}(\AQ\|\P)\\
    &+ \ln\LP\LB\EE_{\Scal'{\sim}\Dcal^{m}}\EE_{h'{\sim}\P}\LP\phi(h'\!, \Scal')^{\frac{\alpha}{\alpha{-}1}}\RP\RB^{\frac{\alpha{-}1}{\alpha}}\RP \!\Bigg]\!\!\ge\! 1{-}\delta.
\end{align*}
Then, we will prove the case when $\alpha\rightarrow 1$ and $\alpha\rightarrow +\infty$ separately.\\

\noindent{}\textbf{When $\alpha\rightarrow 1$.}\\
First, we have $\lim_{\alpha\rightarrow 1^+}\frac{2\alpha{-}1}{\alpha}\ln\frac{2}{\delta} = \ln\frac{2}{\delta}$ and $\lim_{\alpha\rightarrow 1^+}\frac{\alpha-1}{\alpha}D_{\alpha}(\AQ\|\P) = 0$.\\
Furthermore, note that 
\begin{align*}
    \|\phi\|_{\frac{\alpha}{\alpha{-}1}} = \LB\EE_{\Scal'{\sim}\Dcal^{m}}\EE_{h'{\sim}\P}\LP\vert\phi(h'\!, \Scal')\vert^{\frac{\alpha}{\alpha{-}1}}\RP\RB^{\frac{\alpha{-}1}{\alpha}} = \LB\EE_{\Scal'{\sim}\Dcal^{m}}\EE_{h'{\sim}\P}\LP\phi(h'\!, \Scal')^{\frac{\alpha}{\alpha{-}1}}\RP\RB^{\frac{\alpha{-}1}{\alpha}}
\end{align*}
is the $L^{\frac{\alpha}{\alpha{-}1}}$-norm of the function $\phi: \Hcal\times\Zcal^m \rightarrow \Rpe$, where $\lim_{\alpha\rightarrow 1} \|\phi\|_{\frac{\alpha}{\alpha{-}1}} = \lim_{\alpha'\rightarrow +\infty} \|\phi\|_{\alpha'}$ (since we have $\lim_{\alpha\rightarrow 1^+}\frac{\alpha}{\alpha{-}1} = (\lim_{\alpha\rightarrow1}\alpha)(\lim_{\alpha\rightarrow1}\frac{1}{\alpha{-}1}) = +\infty$).
Then, it is well known that
\begin{align*}
    \|\phi\|_{\infty}= \lim_{\alpha'\rightarrow+\infty}\|\phi\|_{\alpha'} = \esssup_{\Scal'\in\Zcal, h'\in\Hcal}\phi(h'{,} \Scal').
\end{align*}
Hence, we have 
\begin{align*}
    &\lim_{\alpha\rightarrow1} \ln\LP\LB\EE_{\Scal'{\sim}\Dcal^{m}}\EE_{h'{\sim}\P}\LP\phi(h'\!, \Scal')^{\frac{\alpha}{\alpha{-}1}}\RP\RB^{\frac{\alpha{-}1}{\alpha}}\RP\\
    = &\ln\LP \lim_{\alpha\rightarrow1} \LB\EE_{\Scal'{\sim}\Dcal^{m}}\EE_{h'{\sim}\P}\LP\phi(h'\!, \Scal')^{\frac{\alpha}{\alpha{-}1}}\RP\RB^{\frac{\alpha{-}1}{\alpha}}\RP\\
    = &\ln\LP \lim_{\alpha\rightarrow1} \| \phi\|_{\frac{\alpha}{\alpha-1}}\RP = \ln\LP \lim_{\alpha'\rightarrow+\infty} \| \phi\|_{\alpha'}\RP \\
    = &\ln\LP \| \phi\|_{\infty}\RP = \ln\LP \esssup_{\Scal'\in\Zcal, h'\in\Hcal}\phi(h'{,} \Scal') \RP.
\end{align*}
Finally, we can deduce that 
\begin{align*}
    &\lim_{\alpha\rightarrow 1}\LB{\frac{2\alpha{-}1}{\alpha}}\ln\frac{2}{\delta}\, +\frac{\alpha{-}1}{\alpha}D_{\alpha}(\AQ\|\P){+} \ln\LP\LB\EE_{\Scal'{\sim}\Dcal^{m}}\EE_{h'{\sim}\P}\LP\phi(h'\!, \Scal')^{\frac{\alpha}{\alpha{-}1}}\RP\RB^{\frac{\alpha{-}1}{\alpha}}\RP\RB\\
    = &\ln\frac{2}{\delta} + \ln\left[\esssup_{\Scal'\in\Zcal, h'\in\Hcal}\phi(h'{,} \Scal')\right].
\end{align*}

~\\
\noindent{}\textbf{When $\alpha\rightarrow +\infty$.}\\
First, we have  
$\lim_{\alpha\rightarrow+\infty}{\frac{2\alpha{-}1}{\alpha}}\ln\frac{2}{\delta} = \ln\frac{2}{\delta}\LB2 -\lim_{\alpha\rightarrow+\infty}\frac{1}{\alpha} \RB = 2\ln\frac{2}{\delta}= \ln\frac{4}{\delta^2}$ and $\lim_{\alpha\rightarrow +\infty} \|\phi\|_{\frac{\alpha}{\alpha{-}1}} = \lim_{\alpha'\rightarrow 1} \|\phi\|_{\alpha'} = \|\phi\|_1$ (since $\lim_{\alpha\rightarrow +\infty}\frac{\alpha}{\alpha-1} = \lim_{\alpha\rightarrow +\infty} \frac{1}{1-\frac{1}{\alpha}}=1$).
Hence, we have 
\begin{align*}
    &\lim_{\alpha\rightarrow+\infty} \ln\LP\LB\EE_{\Scal'{\sim}\Dcal^{m}}\EE_{h'{\sim}\P}\LP\phi(h'\!, \Scal')^{\frac{\alpha}{\alpha{-}1}}\RP\RB^{\frac{\alpha{-}1}{\alpha}}\RP\\
    = &\ln\LP \lim_{\alpha\rightarrow+\infty} \LB\EE_{\Scal'{\sim}\Dcal^{m}}\EE_{h'{\sim}\P}\LP\phi(h'\!, \Scal')^{\frac{\alpha}{\alpha{-}1}}\RP\RB^{\frac{\alpha{-}1}{\alpha}}\RP\\
    = &\ln\LP \lim_{\alpha\rightarrow+\infty} \| \phi\|_{\frac{\alpha}{\alpha-1}}\RP = \ln\LP \lim_{\alpha'\rightarrow1} \| \phi\|_{\alpha'}\RP\\
    = &\ln\LP \| \phi\|_{1}\RP = \ln\LP \EE_{\Scal'{\sim}\Dcal^{m}}\EE_{h'{\sim}\P}\phi(h'\!, \Scal') \RP.
\end{align*}
Moreover, by rearranging the terms in $\frac{\alpha{-}1}{\alpha}D_{\alpha}(\AQ\|\P)$, we have
\begin{align*}
\frac{\alpha{-}1}{\alpha}D_{\alpha}(\AQ\|\P) = &\frac{1}{\alpha}\ln\!\LP \EE_{h{\sim}\P}\!\LP\LB\!\frac{ \AQ(h)}{\P(h)}\RB^{\!\alpha}\RP\RP = \ln\!\LP \LB\EE_{h{\sim}\P}\!\LP\LB\!\frac{ \AQ(h)}{\P(h)}\RB^{\!\alpha}\RP\RB^{\frac{1}{\alpha}}\RP\\
= &\ln\!\LP \LB\EE_{h{\sim}\P}\LP\gamma(h)^{\alpha}\RP\RB^{\frac{1}{\alpha}}\RP = \ln\!\LP \| \gamma\|_{\alpha}\RP,
\end{align*}
where $\| \gamma\|_{\alpha}$ is the $L^{\alpha}$-norm of the function $\gamma$ defined as $\gamma(h)=\tfrac{\AQ(h)}{\P(h)}$.
We have
\begin{align*}
    \lim_{\alpha\rightarrow +\infty}  \frac{\alpha{-}1}{\alpha}D_{\alpha}(\AQ\|\P) =& \lim_{\alpha\rightarrow +\infty}  \ln\!\LP \| \gamma\|_{\alpha}\RP = \ln\LP\lim_{\alpha\rightarrow +\infty} \|\gamma\|_{\alpha}\RP\\
    =& \ln\LP \|\gamma\|_{\infty}\RP = \ln\LP\esssup_{h\in\Hcal}\gamma(h)\RP = \ln\LP\esssup_{h\in\Hcal}\frac{\AQ(h)}{\P(h)}\RP.
\end{align*}
Finally, we can deduce that 
\begin{align*}
    &\lim_{\alpha\rightarrow +\infty}\LB{\frac{2\alpha{-}1}{\alpha}}\ln\frac{2}{\delta}\, +\frac{\alpha{-}1}{\alpha}D_{\alpha}(\AQ\|\P){+} \ln\LP\LB\EE_{\Scal'{\sim}\Dcal^{m}}\EE_{h'{\sim}\P}\LP\phi(h'\!, \Scal')^{\frac{\alpha}{\alpha{-}1}}\RP\RB^{\frac{\alpha{-}1}{\alpha}}\RP\RB\\ 
    = &\ln{\displaystyle\esssup_{h'\in\Hcal}}\,\frac{\AQ(h')}{\P(h')}{+}\ln\!\Big[\frac{4}{\delta^2} {\displaystyle \EE_{\Scal'{\sim}\Dcal^{m}}\EE_{h'{\sim}\P}\phi(h'{,}\Scal')}\Big].\\
\end{align*}
\end{proof}

\section{Proof of Theorem \ref{theorem:disintegrated-lambda}}
\label{ap:proof-disintegrated-lambda}

For the sake of completeness, we first prove an upper bound on $\sqrt{ab}$ \citep[see, \eg,][]{ThiemannIgelWintenbergerSeldin2017}. 
\begin{lemma} For any $a>0, b>0$, we have
\begin{align*}
    \sqrt{\tfrac{a}{b}} = \argmin_{\lambda>0}\LP\frac{a}{\lambda}+\lambda b\RP,\ &\text{and}\ \ 2\sqrt{ab} = \min_{\lambda>0}\LP\frac{a}{\lambda}+\lambda b\RP,\\
    &\ \text{and}\ \ \forall\lambda>0, \sqrt{ab} \le \frac{1}{2}\LP\frac{a}{\lambda}+\lambda b\RP.
\end{align*}
\begin{proof}
Let $f(\lambda) = \LP \tfrac{a}{\lambda}+\lambda b \RP$. The first derivative of $f$ {\it w.r.t.} $\lambda$ is
\begin{align*}
    \frac{\partial f}{\partial\lambda}(\lambda) = \LP b-\frac{a}{\lambda^2}\RP.
\end{align*}
Moreover, from the derivative we can deduce that  we have $\frac{\partial f}{\partial\lambda}(\lambda) < 0 \iff \lambda \in (0, \sqrt{\frac{a}{b}})$, and $\frac{\partial f}{\partial\lambda}(\lambda) > 0 \iff \lambda > \sqrt{\frac{a}{b}}$ and $\frac{\partial f}{\partial\lambda}(\lambda) = 0 \iff \lambda = \sqrt{\frac{a}{b}}$.
It implies that the function is strictly decreasing on $\lambda \in (0, \sqrt{\frac{a}{b}})$, strictly increasing for $\lambda > \sqrt{\frac{a}{b}}$ and admit a unique minimum at $\lambda^* = \sqrt{\frac{a}{b}}$.
Additionally, $f(\lambda^*)=2\sqrt{ab}$ which proves the claim.
\end{proof}
\label{lemma:sqrt}
\end{lemma}
\noindent We can now prove Theorem \ref{theorem:disintegrated-lambda} with Lemma \ref{lemma:sqrt}.\\

\theoremdisintegratedlambda*
\begin{proof}
The proof is similar to the one of Theorem \ref{theorem:disintegrated}. 
Since $\phi(h,\!\Scal)$ is a strictly positive random variable, from Markov's inequality, we have
\begin{align*}
&\PP_{h\sim \AQ}\!\!\LB\phi(h,\!\Scal)\le \frac{2}{\delta}\EE_{h'{\sim} \AQ}\phi(h'\!, \Scal)\RB\!\!\ge 1-\tfrac{\delta}{2}\\
\iff &\EE_{h\sim \AQ}\!\!\Ibf\LB\phi(h,\!\Scal)\le \frac{2}{\delta}\EE_{h'{\sim} \AQ}\phi(h'\!, \Scal)\RB\!\!\ge 1-\tfrac{\delta}{2}.
\end{align*}
Taking the expectation over $\Scal\sim\Dcal^{m}$ to both sides of the inequality gives
\begin{align*}
    &\EE_{\Scal\sim\Dcal^{m}}\EE_{h\sim \AQ}\!\!\Ibf\LB\phi(h,\!\Scal)\le \frac{2}{\delta}\EE_{h'{\sim} \AQ}\phi(h'\!, \Scal)\RB\!\!\ge 1-\tfrac{\delta}{2}\\
    \iff &\PP_{\Scal\sim\Dcal^{m},h\sim \AQ}\!\!\LB\phi(h,\!\Scal) \le \frac{2}{\delta}\EE_{h'{\sim} \AQ}\phi(h'\!, \Scal)\RB\!\!\ge 1-\tfrac{\delta}{2}.
\end{align*}
Using Lemma \ref{lemma:sqrt} with $a=\tfrac{4}{\delta^2}\phi(h'\!, \Scal)^2$ and $b=\tfrac{ \AQ(h')^2}{\P(h')^2}$, we have for all prior $\P{\in}\Mcal^{*}(\Hcal)$
\begin{align*}
    \forall\lambda{>}0,\quad \frac{2}{\delta}\!\EE_{h'{\sim} \AQ}\phi(h'\!, \Scal) &= \EE_{h'{\sim}\P}\sqrt{\frac{ \AQ(h')^2}{\P(h')^2}\frac{4}{\delta^2}\phi(h'\!, \Scal)^2}\\
    &\le \frac{1}{2}\!\LB\lambda\EE_{h'{\sim}\P}\!\!\LP\frac{ \AQ(h')}{\P(h')}\RP^2\!\!{+} \frac{4}{\lambda\delta^2}\EE_{h'{\sim}\P}\LP\phi(h'\!, \Scal)^2\RP\RB\!.
\end{align*}
Then, since both sides of the inequality are strictly positive, we take the logarithm to obtain
\begin{align*}
    \forall\lambda{>}0, \ln\!\LP\frac{2}{\delta}\EE_{h'{\sim} \AQ}\phi(h'\!, \Scal)\RP &\le \ln\!\LP\frac{1}{2}\!\LB\lambda\EE_{h'{\sim}\P}\LP\frac{ \AQ(h')}{\P(h')}\RP^2\!\!{+} \frac{4}{\lambda\delta^2}\EE_{h'{\sim}\P}\LP\phi(h'\!, \Scal)^2\RP\RB\RP\\
    &=  \ln\!\LP\frac{1}{2}\!\LB\lambda \exp(D_2( \AQ\|\P)){+} \frac{4}{\lambda\delta^2}\EE_{h'{\sim}\P}\LP\phi(h'\!, \Scal)^2\RP\RB\RP.
\end{align*}
Hence, we can deduce that
\begin{align}
    \PP_{\Scal\sim\Dcal^{m},h\sim \AQ}\!\!\Bigg[\forall\P{\in}\Mcal^{*}(\Hcal),\ &\forall\lambda>0, \ln\LP\phi(h,\!\Scal)\RP\nonumber\\
    &\le \ln\LP\frac{1}{2}\LB\lambda e^{D_2( \AQ\|\P)}+ \frac{4}{\lambda\delta^2}\EE_{h'{\sim}\P}\LP\phi(h'\!, \Scal)^2\RP\RB\RP\Bigg]\ge 1{-}\tfrac{\delta}{2}.
    \label{eq:disintegrated-param-proof-1}
\end{align}
Given a prior $\P\in\Mcal^{*}(\Hcal)$, note that $\EE_{h'{\sim}\P}\phi(h'\!, \Scal)^{2}$ is a strictly positive random variable. 
Hence, we apply Markov's inequality:
\begin{align*}
    \PP_{\Scal\sim\Dcal^{m}}\LB \EE_{h'{\sim}\P}\phi(h'\!, \Scal)^2 \le \frac{2}{\delta}\EE_{\Scal'{\sim}\Dcal^{m}}\EE_{h'{\sim}\P}\phi(h'\!, \Scal')^2\RB \ge 1-\tfrac{\delta}{2}.
\end{align*}
Since the inequality does not depend on the random variable $h\sim \AQ$, we have
\begin{align*}
    &\PP_{\Scal\sim\Dcal^{m}}\LB\EE_{h'{\sim}\P}\LP\phi(h'\!, \Scal)^2\RP \le\frac{2}{\delta}\EE_{\Scal'{\sim}\Dcal^{m}}\EE_{h'{\sim}\P}\LP\phi(h'\!, \Scal')^2\RP\RB\\
    = &\PP_{\Scal\sim\Dcal^{m},h\sim \AQ}\LB\EE_{h'{\sim}\P}\LP\phi(h'\!, \Scal)^2\RP \le\frac{2}{\delta}\!\EE_{\Scal'{\sim}\Dcal^{m}}\EE_{h'{\sim}\P}\LP\phi(h'\!, \Scal')^2\RP\RB\!.
\end{align*}
Additionally, note that multiplying by $\frac{4}{2\lambda\delta^2}>0$, adding $\frac{\lambda}{2}\exp(D_2( \AQ\|\P))$, and taking the logarithm to both sides of the inequality results in the same indicator function. Indeed,
\begin{align*}
    &\Ibf\LB\EE_{h'{\sim}\P}\LP\phi(h'\!, \Scal)^2\RP \le\frac{2}{\delta}\EE_{\Scal'{\sim}\Dcal^{m}}\EE_{h'{\sim}\P}\LP\phi(h'\!, \Scal')^2\RP\RB\\
    =\ &\Ibf\LB\forall\lambda>0, \tfrac{4}{2\lambda\delta^2}\EE_{h'{\sim}\P}\LP\phi(h'\!, \Scal)^2\RP \le\tfrac{8}{2\lambda\delta^3}\EE_{\Scal'{\sim}\Dcal^{m}}\EE_{h'{\sim}\P}\LP\phi(h'\!, \Scal')^2\RP\RB\\
    =\ &\Ibf\Bigg[\forall\lambda>0, \ln\!\LP\!\tfrac{\lambda}{2}\!\exp(D_2( \AQ\|\P)){+}\tfrac{4}{2\lambda\delta^2}\!\EE_{h'{\sim}\P}\LP\phi(h'\!, \Scal)^2\RP\!\RP\\
    &\hspace{0.3cm}\le \ln\!\LP\!\tfrac{\lambda}{2}\!\exp(D_2( \AQ\|\P)){+}\!\tfrac{8}{2\lambda\delta^3}\EE_{\Scal'{\sim}\Dcal^{m}}\EE_{h'{\sim}\P}\LP\phi(h'\!, \Scal')^2\RP\RP\Bigg]\!.
\end{align*}
Hence, we can deduce that
\begin{align}
    &\PP_{\Scal\sim\Dcal^{m},h\sim \AQ}\Bigg[\forall\lambda{>}0,\,\ln\LP\frac{1}{2}\LB\lambda\exp(D_2( \AQ\|\P)){+} \frac{4}{\lambda\delta^2}\EE_{h'{\sim}\P}\LP\phi(h'\!, \Scal)^2\RP\RB\RP\nonumber\\
    &\hspace{0.2cm}\le \ln\LP\frac{1}{2}\LB\lambda\exp(D_2( \AQ\|\P)){+}\frac{8}{\lambda\delta^3}\EE_{\Scal'{\sim}\Dcal^{m}}\EE_{h'{\sim}\P}\LP\phi(h'\!, \Scal')^2\RP\RB\RP \Bigg]\ge 1-\tfrac{\delta}{2}.
    \label{eq:disintegrated-param-proof-2}
\end{align}
Combining Equations \eqref{eq:disintegrated-param-proof-1} and \eqref{eq:disintegrated-param-proof-2} with a union bound gives us the desired result.
\end{proof}

\section{Proof of Proposition \ref{prop:lambda-min}}
\label{ap:proof-min-lambda}

\proplambdamin*
\begin{proof}
We consider the right-hand side of the inequality of Theorem~\ref{theorem:disintegrated-lambda} (which is strictly positive): we have 
\begin{align}
\ln\LB \frac{\lambda}{2} e^{D_2(\AQ\|\P)} {+} \frac{8}{2\lambda\delta^3}\EE_{\Scal'{\sim}\Dcal^{m}}\EE_{h'{\sim}\P}\LB\phi(h', \Scal')^2\RB\RB.
\label{eq:prop:lambda-min-1}
\end{align}
Since $\ln$ is a strictly increasing function, we have
\begin{align*}
    &\min_{\lambda>0}\LC \ln\LB \frac{\lambda}{2} e^{D_2(\AQ\|\P)} {+} \frac{8}{2\lambda\delta^3}\EE_{\Scal'{\sim}\Dcal^{m}}\EE_{h'{\sim}\P}\LB\phi(h', \Scal')^2\RB\RB \RC\\
    = &\ln\LB\min_{\lambda>0}\LC \frac{\lambda}{2} e^{D_2(\AQ\|\P)} {+} \frac{8}{2\lambda\delta^3}\EE_{\Scal'{\sim}\Dcal^{m}}\EE_{h'{\sim}\P}\LB\phi(h', \Scal')^2\RB\RC\RB.
\end{align*}
Then, we apply Lemma~\ref{lemma:sqrt} by taking $a = \frac{8}{2\delta^3}\EE_{\Scal'{\sim}\Dcal^{m}}\EE_{h'{\sim}\P}\LB\phi(h', \Scal')^2\RB$ and $b=\frac{1}{2}e^{D_2(\AQ\|\P)}$ to obtain $\lambda^*=\sqrt{\frac{a}{b}}= \sqrt{\frac{\EE_{\Scal'{\sim}\Dcal^{m}}{\EE}_{{h'{\sim}\P}}\LB8\phi(h'\!, \Scal')^2\RB}{\delta^3 \exp(D_2(\AQ\|\P))}}$.
Finally, by substituting $\lambda^*$ into Equation~\eqref{eq:prop:lambda-min-1}, we obtain
\begin{align*}
    &\ln\LB \frac{\lambda^*}{2} e^{D_2(\AQ\|\P)} {+} \frac{8}{2\lambda^*\delta^3}\EE_{\Scal'{\sim}\Dcal^{m}}\EE_{h'{\sim}\P}\LB\phi(h', \Scal')^2\RB\RB\\
    = &\frac{1}{2}\LP D_{2}(\AQ\|\P) + \ln\LB\EE_{\Scal'{\sim}\Dcal^{m}}\EE_{h'{\sim} \P}
\LP\frac{8\phi(h'\!, \Scal')^{2}}{\delta^3}\RP\RB\RP, 
\end{align*}
which is the desired result.
\end{proof}

\section{Proof of Corollary~\ref{corollary:nn}}
\label{ap:proof-corollary-nn}
We introduce Theorem~\ref{theorem:disintegrated-union}, which takes into account a set of priors $\Pbf$ while Theorem~\ref{theorem:disintegrated} handles a unique prior $\P$.

\newtheorem*{theorem2p}{Theorem 2'}
\begin{theorem2p} \textit{For any distribution $\Dcal$ on $\Zcal$, for any hypothesis set $\Hcal$,  for any priors set $\Pbf{=}\{\P_t\}_{t=1}^T$ of $T$ prior $\P\in\Mcal^{*}(\Hcal)$, for any measurable function \mbox{$\phi\!:\! \Hcal{\times}\Zcal^{m}{\to} \Rpe$}, for any \mbox{$\alpha\!>\!1$}, for any $\delta\in(0,1]$, for any algorithm \mbox{$A\!:\!\Zcal^{m}{\times}\Mcal^{*}(\Hcal){\to} \Mcal(\Hcal)$}, we have}
\begin{align*}
    \PP_{\Scal\sim\Dcal^{m},h\sim \AQ}\!\Bigg[&\forall \P_t\in\Pbf, \frac{\alpha}{\alpha{-}1}\ln\LP\phi(h,\!\Scal)\RP \le D_{\alpha}(\AQ\|\P){+} \frac{\alpha}{\alpha{-}1}\ln\frac{2}{\delta}\\
    + &\ln\!\frac{2T}{\delta} + \ln\LP\EE_{\Scal'{\sim}\Dcal^{m}}\EE_{h'{\sim}\P}\LP\phi(h'\!, \Scal')^{\frac{\alpha}{\alpha{-}1}}\RP\RP\!\Bigg]\ge 1{-}\delta,
\end{align*}
\mbox{where $\AQ{\triangleq}A(\Scal, \P)$ is output by the deterministic algorithm $A$}.
\customlabel{theorem:disintegrated-union}{2'}
\end{theorem2p}

\begin{proof} The proof is mainly the same as Theorem~\ref{theorem:disintegrated}.
Indeed, we first derive the same equation as Equation~\eqref{eq:disintegrated-proof-1}, we have
\begin{align*}
    \PP_{\Scal\sim\Dcal^{m},h\sim \AQ}\!\Big[&\forall\P{\in}\Mcal^{*}(\Hcal),\,\frac{\alpha}{\alpha{-}1}\ln\!\LP\phi(h,\!\Scal)\RP \le D_{\alpha}( \AQ\|\P)\\
    &+ \frac{\alpha}{\alpha{-}1}\ln\frac{2}{\delta}{+}\ln\!\LP\EE_{h'{\sim}\P}\LP\phi(h'\!, \Scal)^{\frac{\alpha}{\alpha-1}}\RP\RP\Big]\!\!\ge 1{-}\tfrac{\delta}{2}.
\end{align*}
Then, we apply Markov's inequality (as in Theorem~\ref{theorem:disintegrated}) $T$ times with the $T$ priors $\P_t$ belonging to $\Pbf$, however, we set the confidence to $\frac{\delta}{2T}$ instead of $\tfrac{\delta}{2}$, we have 
\begin{align*}
    &\PP_{\Scal\sim\Dcal^{m},h\sim \AQ}\Bigg[\ln\!\LP\EE_{h'{\sim}\P_t}\LB\phi(h'\!, \Scal)^{\frac{\alpha}{\alpha{-}1}}\RB\RP\\
    &\hspace{2cm}\le \ln\!\frac{2T}{\delta}{+}\ln\!\LP\EE_{\Scal'{\sim}\Dcal^{m}}\EE_{h'{\sim}\P_t}\LB\phi(h'\!, \Scal')^{\frac{\alpha}{\alpha-1}}\RB\RP\Bigg]\ge1{-}\tfrac{\delta}{2T}.
\end{align*}
Finally, combining the $T+1$ bounds with a union bound gives us the desired result.
\end{proof}

\noindent We now prove Corollary~\ref{corollary:nn} from Theorem~\ref{theorem:disintegrated-union}.\\

\corollarynn*
\begin{proof}
We instantiate Theorem~\ref{theorem:disintegrated-union} with $\phi(h,\!\Scal)=\exp\!\LB\tfrac{\alpha-1}{\alpha}m\kl(\Risk_{\Scal}(h)\|\Risk_{\Dcal}(h))\RB$ and $\alpha=2$. 
We have with probability at least $1-\delta$ over $\Scal\sim\Dcal^m$ and $h\sim\AQ$, for all prior $\P_t\!\in\! \Pbf$ 
\begin{align*}
    \kl(\Risk_{\Scal}(h)\|\Risk_{\Dcal}(h))\! 
    \le \! \tfrac{1}{m}\!\LB D_{2}(\AQ\|\P_t)+ \ln\LP\frac{8T}{\delta^3}\EE_{\Scal'{\sim}\Dcal^{m}}\EE_{h'{\sim}\P_t}e^{m\kl(\Risk_{\Scal'}(h')\|\Risk_{\Dcal}(h'))}\RP \RB\!.
\end{align*}
From \citet{Maurer2004} we upper-bound  $\EE_{\Scal'{\sim}\Dcal^m}\EE_{h'{\sim}\P_t} e^{m\kl(\Risk_{\Scal'}(h')\|\Risk_{\Dcal}(h'))}$ by $2\sqrt{m}$ for each prior $\P_t$. 
Hence, we have, for all prior $\P_t\!\in\! \Pbf$ 
\begin{align*}
    \kl(\Risk_{\Scal}(h)\|\Risk_{\Dcal}(h))\! 
    \le \! \tfrac{1}{m}\!\LB D_{2}(\AQ\|\P_t)+ \ln\LP\tfrac{16T\sqrt{m}}{\delta^3}\RP \RB\!.
\end{align*}
Additionally, the Rényi divergence $\D_{2}(\AQ\|\P_t)$ between two multivariate Gaussians $\AQ{=}\Ncal(\wbf, \sigma^2{\bf I}_{d})$ and $\P_t{=}\Ncal(\vbf_t, \sigma^2{\bf I}_{d})$ is well known: its closed-form solution is $\D_{2}(\AQ\|\P_t){=}\frac{\|\wbf{-}\vbf_t\|_{2}^{2}}{\sigma^2}$ (see, for example, \citep{GilAlajajiLinder2013}).
\end{proof}

\section{Proof of Corollary~\ref{corollary:nn-rbc}}
\label{ap:proof-corollary-nn-rbc}

We first prove the following lemma in order to prove Corollary~\ref{corollary:nn-rbc}.
\begin{lemma}
If $\AQ=\Ncal(\wbf, \sigma^2\Ibf_d)$ and $\P = \Ncal(\vbf, \sigma^2\Ibf_{d})$, we have 
\begin{align*}
    \ln\frac{\AQ(h)}{\P(h)} = \frac{1}{2\sigma^2}\Big[\|\wbf{+}\epsilonbf-\vbf\|_2^2-\|\epsilonbf\|_2^2\Big],
\end{align*}
where $\epsilonbf{\sim}\Ncal(\zerobf, \sigma^2{\bf I}_{d})$ is a Gaussian noise 
such that  $\wbf{+}\epsilonbf$ are the weights of $h{\sim}\AQ$ with \mbox{$\AQ{=}\Ncal(\wbf, \sigma^2{\bf I}_{d})$}.
\begin{proof}
The probability density functions of $\AQ$ and $\P$ for $h\sim\AQ$ (with the weights $\wbf{+}\epsilonbf$) can be rewritten as
\begin{align*}
    &\AQ(h) = \LB\frac{1}{\sigma\sqrt{2\pi}}\RB^d\!\exp\!\LP\!-\frac{1}{2\sigma^2}\|\wbf{+}\epsilonbf-\wbf\|_2^2\RP = \LB\frac{1}{\sigma\sqrt{2\pi}}\RB^d\!\exp\!\LP\!-\frac{1}{2\sigma^2}\|\epsilonbf\|_2^2\RP\\
    \text{and}\quad &\P(h)=\LB\frac{1}{\sigma\sqrt{2\pi}}\RB^d\!\exp\!\LP\!-\frac{1}{2\sigma^2}\|\wbf{+}\epsilonbf-\vbf\|_2^2\RP.
\end{align*}
We can derive a closed-form expression of $\ln\!\LB\frac{\AQ(h)}{\P(h)}\RB$. 
Indeed, we have
\begin{align*}
    \ln\!\LB\frac{\AQ(h)}{\P(h)}\RB &= \ln\LB\AQ(h)\RB-\ln\LB\P(h)\RB\\
    &= \ln\LP\LB\frac{1}{\sigma\sqrt{2\pi}}\RB^d\!\exp\!\LP\!-\frac{1}{2\sigma^2}\|\epsilonbf\|_2^2\RP\RP\\
    &\hspace{0.4cm}-\ln\LP\LB\frac{1}{\sigma\sqrt{2\pi}}\RB^d\!\exp\!\LP\!-\frac{1}{2\sigma^2}\|\wbf{+}\epsilonbf-\vbf\|_2^2\RP\RP\\
    &= -\frac{1}{2\sigma^2}\|\epsilonbf\|_2^2 + \frac{1}{2\sigma^2}\|\wbf{+}\epsilonbf-\vbf\|_2^2= \frac{1}{2\sigma^2}\Big[\|\wbf{+}\epsilonbf-\vbf\|_2^2-\|\epsilonbf\|_2^2\Big].
\end{align*}
\end{proof}
\label{lemma:disintegrated-kl}
\end{lemma}
\noindent We can now prove Corollary~\ref{corollary:nn-rbc}.

\corollarynnrbc*
\begin{proof}
We will prove the three bounds separately.\\

\noindent\textbf{Equation~\eqref{eq:nn-rivasplata}.} We instantiate Theorem~1{\footnotesize\it (i)} of \citet{RivasplataKuzborskijSzepesvariShaweTaylor2020} with $\phi(h,\!\Scal)=\exp\!\LB m\kl(\Risk_{\Scal}(h)\|\Risk_{\Dcal}(h))\RB$, however, we apply the theorem $T$ times for each prior $\P_t\in\Pbf$ (with a confidence $\frac{\delta}{T}$ instead of $\delta$).
Hence, for each prior $\P_t\in\Pbf$, we have with probability at least $1-\frac{\delta}{T}$ over the random choice of $\Scal\sim\Dcal^m$ and $h\sim\AQ$
\begin{align*}
\kl(\Risk_{\Scal}(h)\|\Risk_{\Dcal}(h))\le \frac{1}{m}\!\LB\ln\!\LB\frac{\AQ(h)}{\P_t(h)}\RB{+}\ln\!\LP\frac{T}{\delta}\!\EE_{\Scal'{\sim}\Dcal^{m}}\EE_{h'{\sim}\P}e^{m\kl(\Risk_{\Scal'}(h')\|\Risk_{\Dcal}(h'))}\!\RP\!\RB\!.
\end{align*}
From \citet{Maurer2004}, we upper-bound  $\EE_{\Scal'{\sim}\Dcal^m}\EE_{h'{\sim}\P_t} e^{m\kl(\Risk_{\Scal'}(h')\|\Risk_{\Dcal}(h'))}$ by $2\sqrt{m}$ and using Lemma~\ref{lemma:disintegrated-kl} we rewrite the disintegrated KL divergence.
Finally, a union-bound argument gives us the claim.\\

\noindent\textbf{Equation~\eqref{eq:nn-blanchard}.}  We apply $T\vert\Bbf\vert$ times Proposition~3.1 of \citet{BlanchardFleuret2007}
with a confidence $\frac{\delta}{T\vert\Bbf\vert}$ instead of $\delta$. 
For each prior $\P_t\in\Pbf$ and hyperparameters $b\in\Bbf$, we have with probability at least $1-\frac{\delta}{T\vert\Bbf\vert}$ over the random choice of $\Scal\sim\Dcal^m$ and $h\sim\AQ$
\begin{align*}
\kl_{+}(\Risk_{\Scal}(h)\|\Risk_{\Dcal}(h))\le \frac{1}{m}\!\LB\frac{b{+}1}{b}\!\!\LB\ln\!\frac{\AQ(h)}{\P_t(h)}\RB_{+}\!\!{+}\ln\!\LP\frac{T\vert\Bbf\vert(b{+}1)}{\delta}\!\RP\!\RB\!.
\end{align*}
From Lemma~\ref{lemma:disintegrated-kl} and a union-bound argument, we obtain the claim.\\

\noindent\textbf{Equation~\eqref{eq:nn-catoni}.} We apply $T\vert\Cbf\vert$ times Theorem 1.2.7 of \citet{Catoni2007} with a confidence $\tfrac{\delta}{T\vert\Cbf\vert}$ instead of $\delta$. 
For each prior $\P_{t}\in\Pbf$ and hyperparameter $c\in\Cbf$, we have with probability at least $1-\tfrac{\delta}{T\vert\Cbf\vert}$ over the random choice of $\Scal\sim\Dcal^m$ and $h\sim\AQ$
\begin{align*}
\Risk_{\Dcal}(h) \!\le\, \frac{1}{1{-}e^{{-}c}}\LB1{-}\exp\LP{-}c\Risk_{\Scal}(h) {-}\frac{1}{m}\!\!\left[\ln\LB\frac{\AQ(h)}{\P_t(h)}\RB {+} \ln\!\frac{T\vert\Cbf\vert}{\delta}\right]\RP\RB\!.
\end{align*}
From Lemma~\ref{lemma:disintegrated-kl} and a union-bound argument, we obtain the claim.
\end{proof}

\section{Proof of Corollary~\ref{corollary:nn-sto}}
\label{ap:proof-nn-sto}

\corollarynnsto*
\begin{proof}
We instantiate Equation~\eqref{eq:seeger} (and apply Jensen's inequality on the left-hand side of the inequation) for each prior $\P_t$ with $\Q{=}\Ncal(\wbf, \sigma^2{\bf I}_{d})$ and $\P_t{=}\Ncal(\vbf_t, \sigma^2{\bf I}_{d})$ with a confidence $\tfrac{\delta}{2T}$ instead of $\delta$.
Indeed, for each prior $\P_t$, with probability at least $1{-}\tfrac{\delta}{2T}$ over the random choice of $\Scal\sim\Dcal^m$, \mbox{we have for all posterior $\Q$ on $\Hcal$},
\begin{align*}
\kl\!\LP\EE_{h{\sim}\Q}\!\!\Risk_{\Scal}(h)\| \EE_{h{\sim}\Q}\!\!\Risk_{\Dcal}(h)\!\RP{\le} \frac{1}{m}\!\LB 
\KL(\Q\|\P_t)
{+}\ln\tfrac{4T\sqrt{m}}{\delta}\RB\!.
\end{align*}
Note that the closed-form solution of the \KL~divergence between the Gaussian distributions $\Q$ and $\P_t$ is well known, we have $\KL(\Q\|\P_t){=}\frac{\|\wbf{-}\vbf_t\|_{2}^{2}}{2\sigma^2}$.
Then, by applying a union-bound argument over the $T$ bounds obtained with the $T$ priors $\P_t$, we have with probability at least $1{-}\frac{\delta}{2}$ over the random choice of $\Scal\sim\Dcal^m$, for all prior $\P_t\in\Pbf$, for all posterior $\Q$
\begin{align*}
\kl\!\LP\EE_{h{\sim}\Q}\!\!\Risk_{\Scal}(h)\| \EE_{h{\sim}\Q}\!\!\Risk_{\Dcal}(h)\!\RP{\le} \frac{1}{m}\!\LB 
\tfrac{\|\wbf{-}\vbf_t\|_{2}^{2}}{2\sigma^2}
{+}\ln\tfrac{4T\sqrt{m}}{\delta}\RB\!.\quad\text{(Equation~\eqref{eq:nn-sto-seeger})}
\end{align*}
Additionally, we obtained Equation~\eqref{eq:nn-sto-sample} by a direct application the Theorem~2.2 of \citet{DziugaiteRoy2017} (with confidence $\frac{\delta}{2}$ instead of $\delta$).
Finally, from a union bound of the two bounds in Equations~\eqref{eq:nn-sto-sample} and~\eqref{eq:nn-sto-seeger} gives the claimed result.
\end{proof}

\section{Evaluation and minimization of the bounds of Corollaries~\ref{corollary:nn}, \ref{corollary:nn-rbc}, \ref{corollary:nn-sto}}
\label{ap:evaluation-minimization}

This appendix presents more details on the optimization and the evaluation of the bounds.

\subsection{Evaluation of the bounds}

Note that, except for Equation~\eqref{eq:nn-catoni}, a generalization gap is upper-bounded instead of the true risk.
Hence, to evaluate the bounds of the corollaries (except for Equation~\eqref{eq:nn-catoni}) we use the invert binary $\kl$ divergence defined as
\begin{align*}
    \kl^{-1} (q \vert \psi){=}\max\Big\{ p\in(0,\! 1) \,\Big\vert\, \kl(q\|p) \le \psi\Big\},
\end{align*}
where $q$ is typically the empirical risk, and $\psi$ is the PAC-Bayesian bound.
Here, the function $\kl^{-1} (q \vert \psi)$ outputs the worst true risk $p$ where the inequality $\kl(q\| p) \le \psi$ holds.
We can actually instantiate $p, q$ and $\psi$ for the different corollaries. 
Indeed, we have for all $\P_t\in\Pbf$
\begin{align*}
&\Risk_{\Dcal}(h) \le \underbrace{\kl^{-1}\!\LP\Risk_{\Scal}(h) \;\middle\vert\; \frac{1}{m}\!\!\left[ \frac{\|\wbf{-}\vbf_t\|_{2}^{2}}{\sigma^2}{+}\ln\frac{16T\sqrt{m}}{\delta^3}\right]\RP}_{\text{Corollary~\ref{corollary:nn}}},\\
&\Risk_{\Dcal}(h) \le \underbrace{\kl^{-1}\!\LP\Risk_{\Scal}(h) \;\middle\vert\; \frac{1}{m}\!\!\LB\!\frac{\LN \wbf{+}\epsilonbf{-}\vbf_t\RN^2_{2}\!{-}\LN\epsilonbf\RN^2_{2}}{2\sigma^2}{+} \ln\!\tfrac{2T\sqrt{m}}{\delta}\RB\RP}_{\text{Equation~\eqref{eq:nn-rivasplata}}},\\
&\Risk_{\Dcal}(h) \le \underbrace{\kl^{-1}\!\LP\Risk_{\Scal}(h) \;\middle\vert\; \frac{1}{m}\!\!\LB\!\frac{b{+}1}{b}\LB\frac{\LN \wbf{+}\epsilonbf{-}\vbf_t\RN^2_{2}\!{-}\LN\epsilonbf\RN^2_{2}}{2\sigma^2}\RB_{+}\!\!{+} \ln\!\tfrac{(b+1)T\vert\Bbf\vert}{\delta}\RB\RP}_{\text{Equation~\eqref{eq:nn-blanchard}}},\\
\text{and }\ &\EE_{h\sim\Q}\Risk_{\Dcal}(h) \le \underbrace{\kl^{-1}\!\LP \spadesuit \;\middle\vert\; \frac{1}{m}\!\LB \frac{\|\wbf{-}\vbf_t\|_{2}^{2}}{2\sigma^2} {+}\ln\frac{4T\sqrt{m}}{\delta}\RB\RP}_{\text{Corollary~\ref{corollary:nn-sto}}},\\
\text{where }\ & \spadesuit = \kl^{-1}\!\LP \frac{1}{n}\sum_{i=1}^{n}\!\Risk_{\Scal}(h_i) \;\middle\vert\; \frac1n \ln\frac{4}{\delta}\RP.
\end{align*}
Hence, $\kl^{-1}$ has to be evaluated in order to obtain the value of the upper-bound on $\Risk_{\Dcal}(h)$ or $\EE_{h\sim\Q}\Risk_{\Dcal}(h)$: the evaluation of $\kl^{-1}(q\vert\psi)$ is performed by the bisection method.
From this new formulation of the bounds, we can remark that the objective is to minimize the function $\kl^{-1} (q \vert \psi)$ in order to minimize the true risk $p$.
To do so, \citet{ReebDoerrGerwinnRakitsch2018} introduced an analytical expression of the derivative of $\kl^{-1}$ with respect to the empirical risk $q$ and the PAC-Bayesian bound $\psi$.
The two partial derivatives are defined in the following way:
\begin{align*}
    &\frac{\partial \kl^{-1}(q\vert\psi)}{\partial q} = \frac{\ln\frac{1-q}{1-\kl^{-1}(q\vert\psi)}-\ln\frac{q}{\kl^{-1}(q\vert\psi)}}{\frac{1-q}{1-\kl^{-1}(q\vert\psi)}-\frac{q}{\kl^{-1}(q\vert\psi)}},\\
    \text{ and } &\frac{\partial \kl^{-1}(q\vert\psi)}{\partial \psi} = \frac{1}{\frac{1-q}{1-\kl^{-1}(q\vert\psi)}-\frac{q}{\kl^{-1}(q\vert\psi)}}.
\end{align*}
Note that these partial derivatives need the evaluation of $\kl^{-1}(q\vert\psi)$ for a given empirical risk $q$ and a PAC-Bayesian bound $\psi$.
Then, by computing the derivatives of $q$ and $\psi$ with respect to the parameters and by using the chain rule of differentiation, a library like PyTorch (see \citet{Pytorch2019}) can automatically compute the derivatives of $\kl^{-1}$ with respect to the parameters.

\subsection{Optimization of the bounds}

The optimization of the bounds associated with the corollaries are presented in Algorithm~\ref{ap:algo}.
This algorithm is divided in two steps:  {\bf 1)} optimizing and chosing the prior $\P$ (Line 6 to 28); and {\bf 2)} optimizing the posterior $\AQ$ (from Line 32 to 39).

\noindent{}In step {\bf 1)}, the prior $\P_t$ is obtained after the epoch $t\in\{1,\dots,T\}$ (line 16) by updating $\omegabf$ (parameterizing the prior $\P_t$) using a mini-batch gradient descent algorithm.
For each epoch $t$ and for each mini-batch $\Ucal\subseteq\Scal_{\text{prior}}$ (Line 8 and 11), we sample a hypothesis $h$ parameterized by $\omegabf+\epsilonbf$ (Line 12 and 13) and update $\omegabf$ with the gradient descent algorithm by minimizing the risk $\Risk_{\Ucal}(h)$ (Line 14).

\noindent{}After each epoch $t$, the prior $\P$ is selected by early stopping on the learning sample $\Scal$.
We first estimate the risk on $\Scal$ (Line 19 to 23) by sampling $h\sim\P_t$ (Line 20 and 21) and computing the losses for each mini-batch $\Ucal$.
Then, we select the prior $\P_t$ if it minimizes the risk (Line 24 to 27).

\noindent{}Given the prior $\P$, we learn a posterior $\AQ$ in step {\bf 2)} during $T'$ epochs. 
For each epoch and each mini-batch $\Ucal\subseteq\Scal$, we sample a hypothesis $h$ associated with the weights $\omegabf+\epsilonbf$ (Line 38 and 39).
At each iteration, the algorithm updates the weights $\omegabf$ (Line 39) by optimizing 
\begin{align}
&\underbrace{\kl^{-1}\!\!\LP\!\Risk_{\Ucal}(h) \middle\vert  \frac{1}{m}\!\!\LB\frac{\|\omegabf{-}\vbf_{t^*}\|_{2}^{2}}{\sigma^2}{+}\ln\frac{16T\sqrt{m}}{\delta^3}\RB\RP}_{\text{Objective function for Corollary~\ref{corollary:nn}}},\label{eq:objective-nn}\\
&\underbrace{\kl^{-1}\!\!\LP\!\Risk_{\Ucal}(h) \middle\vert \frac{1}{m}\!\!\LB\frac{\LN \omegabf{+}\epsilonbf{-}\vbf_{t^*}\RN^2_{2}\!{-}\LN\epsilonbf\RN^2_{2}}{2\sigma^2}{+}\ln\frac{2T\sqrt{m}}{\delta}\RB\RP}_{\text{Objective function for Equation~\eqref{eq:nn-rivasplata}}},\label{eq:objective-rivasplata}\\
&\underbrace{\kl^{-1}\!\LP\Risk_{\Ucal}(h) \;\middle\vert\; \frac{1}{m}\!\!\LB\!\frac{b{+}1}{b}\LB\frac{\LN \omegabf{+}\epsilonbf{-}\vbf_{t^*}\RN^2_{2}\!{-}\LN\epsilonbf\RN^2_{2}}{2\sigma^2}\RB_{+}\!\!\!{+} \ln\!\tfrac{(b{+}1)T\vert\Bbf\vert}{\delta}\RB\RP}_{\text{Objective function for Equation~\eqref{eq:nn-blanchard}}},\label{eq:objective-blanchard}\\
&\underbrace{\displaystyle\frac{1}{1-e^{-c}}\LB1{-}\exp\left({\displaystyle\!\!{-}c\Risk_{\Ucal}(h) {-}\frac{1}{m}\!\!\left[\! \frac{\| \wbf{+}\epsilonbf{-}\vbf_t\|^2_{2} {-}\|\epsilonbf\|^2_{2}}{2\sigma^2} {+} \ln\!\frac{T\vert\Cbf\vert}{\delta}\!\right]\!}\right)\RB.}_{\text{Objective function for Equation~\eqref{eq:nn-catoni}}}\label{eq:objective-catoni}
\end{align}

\noindent{}Note that, as stated in Section~\ref{sec:optimization}, $T'=1$ for MNIST and FashionMNIST while $T'=10$ for CIFAR-10 with a batch size of $32$.
Additionally, the loss is the bounded cross-entropy loss $\ell(h, (\xbf, y)) {=} -\frac{1}{Z}\ln(\Phi(h(\xbf)[y]))$ of \citet{DziugaiteRoy2018} in the risk $\Risk_{\Ucal}(h)$.
The update of the weights $\omegabf$ is done with the Adam optimizer~\citep{KingmaBa2015}.
Concerning the optimization of the hyperparameters $c\in\Cbf$ and $b\in\Bbf$ for Equations~\eqref{eq:nn-blanchard} and~\eqref{eq:nn-catoni}, we {\it (a)} initialize $b\in\Bbf$ or $c\in\Cbf$ with the one that performs best on the first mini-batch and {\it (b)} optimize by gradient descent the hyperparameter.
To evaluate Equations~\eqref{eq:nn-blanchard} and~\eqref{eq:nn-catoni}, we take $b\in\Bbf$ and $c\in\Cbf$ that leads to the tightest bound.

\begin{algorithm}[H]
  \caption{Optimization of the bounds (Training Method)}
  \begin{algorithmic}[1] 
    \State{\centerline{\texttt{Optimizing the prior $\P$} --- Step {\bf 1)} --- Algorithm $A_{\text{prior}}$}}\\
    
    \State{$\omegabf \leftarrow$ Initialize the weights $\omegabf$}
    \State{$r^* \leftarrow +\infty$}
    \State{$t^* \leftarrow +\infty$}
    
    \For{{\bf each} epoch $t\leftarrow 1,\dots,T$}\\
    
        \State{\texttt{Optimizing the prior $\P_t$}}
        \For{{\bf each} mini-batch $\Ucal \subseteq \Scal_{\text{prior}}$}
            \State{Sample a noise $\epsilonbf \sim \Ncal(\zerobf, \sigma^2\Ibf_{d})$}
            \State{$h \leftarrow\ $Hypothesis parameterized by $\omegabf{+}\epsilonbf$}
            \State{$\omegabf \leftarrow $ Update $\omegabf$ with $\Risk_{\Ucal}(h)$}
        \EndFor    
        \State{$\P_t \leftarrow \Ncal(\omegabf, \sigma^2\Ibf_{d})$ where $\P_t{=}\Ncal(\vbf_t, \sigma^2\Ibf_{d})$}\\
        
        \State{\texttt{Selecting the prior $\P$}}
        
        \For{{\bf each} mini-batch $\Ucal \subseteq \Scal$}
            \State{Sample a noise $\epsilonbf \sim \Ncal(\zerobf, \sigma^2\Ibf_{d})$}
            \State{$h \leftarrow\ $Hypothesis parameterized by $\vbf_t{+}\epsilonbf$}
            \State{$r \leftarrow r + \sum_{(\xbf,y)\in\Ucal}\ell(h, (\xbf, y))$}
        \EndFor
        \If{$r < r^*$}
            \State{$r^* \leftarrow r$}
            \State{$\P \leftarrow \P_t$}
            \State{$t^* \leftarrow t$}
        \EndIf
        
    \EndFor\\
    
    \State{\centerline{\texttt{Optimizing the posterior $\AQ$} --- Step {\bf 2)}  --- Algorithm $A$}}\\

    \State{$\AQ \leftarrow \P=\Ncal(\omegabf, \sigma^2\Ibf_{d})=\Ncal(\vbf_{t^*}, \sigma^2\Ibf_{d})$}
    \For{{\bf each} epoch $t'\leftarrow 1,\dots,T'$}
        \For{{\bf each} mini-batch $\Ucal \subseteq \Scal$}
            \State{Sample a noise $\epsilonbf \sim \Ncal(\zerobf, \sigma^2\Ibf_{d})$}
            \State{$h \leftarrow\ $Hypothesis parameterized by $\omegabf{+}\epsilonbf$}
            \State{$\omegabf \leftarrow$ Update $\omegabf$ with either Equation~\eqref{eq:objective-nn},~\eqref{eq:objective-rivasplata},~\eqref{eq:objective-blanchard}, or~\eqref{eq:objective-catoni}}
        \EndFor
    \EndFor
    
    \State{{\bf return} {$\AQ=\Ncal(\omegabf, \sigma^2\Ibf_{d})=\Ncal(\wbf, \sigma^2\Ibf_{d})$} and $\P=\Ncal(\vbf_{t^*}, \sigma^2\Ibf_{d})$}
\end{algorithmic}
\label{ap:algo}
\end{algorithm}~

\section{About Theorem~\ref{theorem:mutual-info}}
\label{ap:proof-theorem-mutual-info}

This section is devoted to {\it (i)} the proof of a bound that is easier to interpret than Theorem~\ref{theorem:mutual-info}, {\it (ii)} the proof of Theorem~\ref{theorem:mutual-info} and {\it (iii)} a discussion about Theorem~\ref{theorem:mutual-info}.

\subsection{A bound easier to interpret}

Since the mutual information is well known, a bound based on this quantity will be more interpretable than the one with the Sibson's.
Hence, we propose a mutual-information-based bound in Theorem~\ref{theorem:mutual-info-kl}.
However, in order to prove this theorem, we need to prove Lemma~\ref{lemma:mutual-info-kl}.

\begin{lemma}
For any distribution $\Dcal$ on $\Zcal$, for any hypothesis set $\Hcal$, for any measurable function $\phi:\Hcal\times \Zcal^{m}\rightarrow [1, +\infty[$, for any $\delta\in(0,1]$, for any deterministic algorithm $A:\Zcal^{m}\times\Mcal^{*}(\Hcal){\rightarrow} \Mcal(\Hcal)$, we have
\begin{align*}
    \PP_{\Scal\sim\D^{m}, h\sim \AQ}\Bigg[\forall\P{\in}\Mcal^{*}(\Hcal),\ \ln\phi(h,\!\Scal)&\le \frac{1}{\delta}\Big[\EE_{\Scal\sim\Dcal^m}\KL(\AQ\|\P)\\
    &+\ln\LP\EE_{\Scal\sim\Dcal^m}\EE_{h\sim\P}\phi(h, \Scal)\RP\Big] \Bigg] \ge 1-\delta.
\end{align*}
\begin{proof}
By developing $\EE_{\Scal\sim\Dcal^m}\EE_{h\sim\AQ}\ln\phi(h, \Scal)$, we have for all prior $\P\in\Mcal^{*}(\Hcal)$
\begin{align*}
\EE_{\Scal\sim\Dcal^m}\EE_{h\sim\AQ}\ln\phi(h, \Scal) &= \EE_{\Scal\sim\Dcal^m}\EE_{h\sim\AQ}\ln\LB\frac{\AQ(h)\P(h)}{\P(h)\AQ(h)}\phi(h, \Scal)\RB\\
&= \EE_{\Scal\sim\Dcal^m}\EE_{h\sim\AQ}\ln\LB\frac{\AQ(h)}{\P(h)}\RB +\EE_{\Scal\sim\Dcal^m}\EE_{h\sim\AQ}\ln\LB\frac{\P(h)}{\AQ(h)}\phi(h, \Scal)\RB\\
&= \EE_{\Scal\sim\Dcal^m}\KL(\AQ\|\P) +\EE_{\Scal\sim\Dcal^m}\EE_{h\sim\AQ}\ln\LB\frac{\P(h)}{\AQ(h)}\phi(h, \Scal)\RB.
\end{align*}
From Jensen's inequality, we have for all prior $\P\in\Mcal^{*}(\Hcal)$
\begin{align}
    &\EE_{\Scal\sim\Dcal^m}\KL(\AQ\|\P) +\EE_{\Scal\sim\Dcal^m}\EE_{h\sim\AQ}\ln\LB\frac{\P(h)}{\AQ(h)}\phi(h, \Scal)\RB\nonumber\\
    \le &\EE_{\Scal\sim\Dcal^m}\KL(\AQ\|\P) +\ln\LB\EE_{\Scal\sim\Dcal^m}\EE_{h\sim\AQ}\frac{\P(h)}{\AQ(h)}\phi(h, \Scal)\RB\nonumber\\
    = &\EE_{\Scal\sim\Dcal^m}\KL(\AQ\|\P) +\ln\LB\EE_{\Scal\sim\Dcal^m}\EE_{h\sim\P}\phi(h, \Scal)\RB.\label{eq:mutual-info-1}
\end{align}
Since we assume in this case that $\phi(h, \Scal) \ge 1$ for all $h\in\Hcal$ and $\Scal\in\Zcal^m$, we have $\ln\phi(h, \Scal) \ge 0$; we can apply Markov's inequality to obtain
\begin{align}
    \PP_{\Scal\sim\D^{m}, h\sim \AQ}\LB \ln\phi(h,\!\Scal)\le \frac{1}{\delta}\EE_{\Scal{\sim}\Dcal^{m}}\EE_{h{\sim} \AQ}\ln\phi(h, \Scal) \RB \ge 1-\delta.\label{eq:mutual-info-2}
\end{align}
Then, from Equations~\eqref{eq:mutual-info-1} and \eqref{eq:mutual-info-2}, we can deduce the stated result.
\end{proof}
\label{lemma:mutual-info-kl}
\end{lemma}
\noindent We are now ready to prove Theorem~\ref{theorem:mutual-info-kl}.
\begin{theorem}
For any distribution $\Dcal$ on $\Zcal$, for any hypothesis set $\Hcal$, for any measurable function $\phi:\Hcal\times \Zcal^{m}\rightarrow [1, +\infty[$, for any $\delta\in(0,1]$, for any deterministic algorithm $A:\Zcal^{m}\times\Mcal^{*}(\Hcal){\rightarrow} \Mcal(\Hcal)$, we have
\begin{align*}
    \PP_{\Scal\sim\D^{m}, h\sim \AQ}\LB\ln\phi(h,\!\Scal)\le \frac{1}{\delta}\LB I(h{;}\Scal) +\ln\LP\EE_{\Scal\sim\Dcal^m}\EE_{h\sim\P^*}\phi(h, \Scal)\RP\RB \RB \ge 1-\delta,
\end{align*}
where $\P^*$ is defined such that $\P^*(h)=\EE_{\Scal\sim\Dcal^m}\AQ(h)$ and $I(h{;} \Scal) = \min_{\P\in\Mcal^{*}(\Hcal)}\EE_{\Scal\sim\Dcal^m}\KL(\AQ\|\P)$.
\begin{proof}
Note that the mutual information is defined by $I(h{;} \Scal) = \min_{\P\in\Mcal^{*}(\Hcal)}\EE_{\Scal\sim\Dcal^m}\KL(\AQ\|\P)$. 
Hence, to prove Theorem~\ref{theorem:mutual-info-kl}, we have to instantiate Lemma~\ref{lemma:mutual-info-kl} with the optimal prior, \ie, the prior $\P$ which minimizes $\EE_{\Scal\sim\Dcal^m}\KL(\AQ\|\P)$.
The optimal prior is well known~\citep[see, \eg,][]{Catoni2007,LeverLavioletteShaweTaylor2013}: for the sake of completeness, we derive it. First, we have
\begin{align*}
    \EE_{\Scal\sim\Dcal^m}\KL(\AQ\|\P) &= \EE_{\Scal\sim\Dcal^m}\EE_{h\sim\AQ}\ln\frac{\AQ(h)}{\P(h)}\\
    &= \EE_{\Scal\sim\Dcal^m}\EE_{h\sim\AQ}\ln\!\LB\frac{\AQ(h)[\EE_{\Scal'\sim\Dcal^m}\Q_{\Scal'}(h)]}{\P(h)[\EE_{\Scal'\sim\Dcal^m}\Q_{\Scal'}(h)]}\RB\\
    &= \EE_{\Scal\sim\Dcal^m}\EE_{h\sim\AQ}\ln\!\LB\frac{\AQ(h)}{\EE_{\Scal'\sim\Dcal^m}\Q_{\Scal'}(h)}\RB{+}\EE_{h\sim\AQ}\ln\!\LB\frac{\EE_{\Scal'\sim\Dcal^m}\Q_{\Scal'}(h)}{\P(h)}\RB.
\end{align*}
Hence, 

\begin{align*}
    \argmin_{\P\in\Mcal^{*}(\Hcal)}\EE_{\Scal\sim\Dcal^m}\KL(\AQ\|\P)= &\argmin_{\P\in\Mcal^{*}(\Hcal)} \Bigg[\EE_{\Scal\sim\Dcal^m}\EE_{h\sim\AQ}\ln\LB\frac{\AQ(h)}{\EE_{\Scal'\sim\Dcal^m}\Q_{\Scal'}(h)}\RB\\
    &\hspace{1.5cm}+ \EE_{h\sim\AQ}\ln\LB\frac{\EE_{\Scal'\sim\Dcal^m}\Q_{\Scal'}(h)}{\P(h)}\RB\Bigg]\\
    =&\argmin_{\P\in\Mcal^{*}(\Hcal)}\LB \EE_{h\sim\AQ}\ln\LB\frac{\EE_{\Scal'\sim\Dcal^m}\Q_{\Scal'}(h)}{\P(h)}\RB \RB=\P^*,
\end{align*}
where $\P^*(h) = \EE_{\Scal'\sim\Dcal^m}\Q_{\Scal'}(h)$.
Note that $\P^*$ is defined from the data distribution $\Dcal$, hence, $\P^*$ is a valid prior when instantiating Lemma~\ref{lemma:mutual-info-kl} with $\P^*$.
Then, we have with probability at least $1{-}\delta$ over $\Scal\sim\Dcal^m$ and $h\sim\AQ$
\begin{align*}
    \ln\phi(h,\!\Scal) &\le \frac{1}{\delta}\LB\EE_{\Scal\sim\Dcal^m}\KL(\AQ\|\P^*) +\ln\LP\EE_{\Scal\sim\Dcal^m}\EE_{h\sim\P^*}\phi(h, \Scal)\RP\RB\\
    &= \frac{1}{\delta}\LB I(h{;} \Scal) +\ln\LP\EE_{\Scal\sim\Dcal^m}\EE_{h\sim\P^*}\phi(h, \Scal)\RP\RB.
\end{align*}
\end{proof}
\label{theorem:mutual-info-kl}
\end{theorem}
\noindent As you can remark, this bound is looser than Theorem~\ref{theorem:mutual-info}, which is based on Sibson's mutual information. 
For example, when we instantiate this bound with $\phi(h,\Scal)=\exp\LB m\kl(\Risk_{\Scal}(h)\|\Risk_{\Dcal}(h))\RB$, the bound will be multiplied by $\frac{1}{\delta m}$, while the bound of Theorem~\ref{theorem:mutual-info} is only multiplied by $\frac{1}{m}$ (but we add the term $\frac{1}{m}\ln\frac{1}{\delta}$ to the bound which is small even for small $m$).

\subsection{Proof of Theorem~\ref{theorem:mutual-info}}
We first introduce Lemma~\ref{lemma:mutual-info} in order to prove Theorem~\ref{theorem:mutual-info}.
\begin{lemma} For any distribution $\Dcal$ on $\Zcal$, for any hypothesis set $\Hcal$, for any prior distribution $\P$ on $\Hcal$, for any measurable function $\phi:\Hcal\times \Zcal^{m}$, for any $\alpha>1$, for any $\delta\in(0,1]$, for any deterministic algorithm $A:\Zcal^{m}\times\Mcal^{*}(\Hcal){\rightarrow} \Mcal(\Hcal)$, we have
\begin{align*}
    \PP_{\Scal\sim\D^{m}, h\sim \AQ}\!\!\Bigg[\forall\P{\in}\Mcal^{*}(\Hcal),& \displaystyle\frac{\alpha}{\alpha{-}1}\!\ln\!\LP\phi(h,\!\Scal)\RP \le D_{\alpha}(\rho\|\pi)\\
    &+\ln\!\LP\!\tfrac{1}{\delta^{\frac{\alpha}{\alpha{-}1}}}\EE_{\Scal'{\sim}\Dcal^{m}}\EE_{h'{\sim} \P}\LP\phi(h'\!, \Scal')^{\frac{\alpha}{\alpha-1}}\RP\!\RP \Bigg]\!\!\ge 1{-}\delta.
\end{align*}
\textit{where $\rho(h, \Scal){=} \AQ(h)\Dcal^{m}(\Scal)$; $\pi(h, \Scal){=} \P(h)\Dcal^{m}(\Scal)$.}
\begin{proof}
Note that $\phi(h,\!\Scal)$ is a non-negative random variable. 
From Markov's inequality, we have
\begin{align*}
    \PP_{\Scal\sim\D^{m}, h\sim \AQ}\LB \phi(h,\!\Scal)\le \frac{1}{\delta}\EE_{\Scal'{\sim}\Dcal^{m}}\EE_{h'{\sim} \AQprime}\phi(h'\!, \Scal') \RB \ge 1-\delta.
\end{align*}
Then, since both sides of the inequality are strictly positive, we take the logarithm to both sides of the equality and multiply by $\frac{\alpha}{\alpha-1}>0$ to obtain
\begin{align*}
\PP_{\Scal\sim\Dcal^{m},h\sim \AQ}\LB\frac{\alpha}{\alpha-1}\ln\LP\phi(h,\!\Scal)\RP \le \frac{\alpha}{\alpha{-}1}\ln\LP\frac{1}{\delta}\EE_{\Scal'{\sim}\Dcal^{m}}\EE_{h'{\sim} \AQprime }\phi(h'\!, \Scal')\RP\RB\ge 1-\delta.
\end{align*}
We develop the right-hand side of the inequality in the indicator function and make the expectation of the hypothesis over the distribution $\P$ appear.
We have for all priors $\P{\in}\Mcal^{*}(\Hcal)$,
\begin{align*}
&\frac{\alpha}{\alpha{-}1}\ln\LP\frac{1}{\delta}\EE_{\Scal'{\sim}\Dcal^{m}}\EE_{h'{\sim} \AQprime }\phi(h'\!, \Scal')\RP
= &\frac{\alpha}{\alpha{-}1}\ln\LP\frac{1}{\delta}\EE_{\Scal'{\sim}\Dcal^{m}}\EE_{h'{\sim} \P}\frac{\AQprime (h')}{\P(h')}\phi(h'\!, \Scal')\RP.
\end{align*}
Then, since $\tfrac{1}{r}+\tfrac{1}{s}=1$ where $r{=}\alpha$ and $s{=}\frac{\alpha}{\alpha-1}$. Hence, Hölder's inequality gives
\begin{align*}
    \EE_{\Scal'{\sim}\Dcal^{m}}\EE_{h'{\sim} \AQprime}\phi(h'\!, \Scal'){\le}\!\LB\EE_{\Scal'{\sim}\Dcal^{m}}\EE_{h'{\sim} \P}\!\LP\Bigg[\frac{\AQprime (h')}{\P(h')}\Bigg]^{\alpha}\RP\!\RB^{\frac{1}{\alpha}}\!\!\LB\EE_{\Scal'{\sim}\Dcal^{m}}\EE_{h'{\sim} \P}\!\LP\phi(h'\!, \Scal')^{\frac{\alpha}{\alpha-1}}\RP\RB^{\frac{\alpha-1}{\alpha}}\!\!\!.
\end{align*}
Since both sides of the inequality are positive, we take the logarithm.
Moreover, we add $\ln(\tfrac{1}{\delta})$, and we multiply by $\frac{\alpha}{\alpha-1}>0$ to both sides of the inequality.
We have
\begin{align*}
    &\frac{\alpha}{\alpha{-}1}\ln\LP\frac{1}{\delta}\EE_{\Scal'{\sim}\Dcal^{m}}\EE_{h'{\sim} \AQprime }\phi(h'\!, \Scal')\RP \\
    \le &\frac{\alpha}{\alpha{-}1}\ln\LP\frac{1}{\delta}\LB\EE_{\Scal'{\sim}\Dcal^{m}}\EE_{h'{\sim} \P}\LP\Bigg[\frac{\AQprime (h')}{\P(h')}\Bigg]^{\alpha}\RP\RB^{\frac{1}{\alpha}}\LB\EE_{\Scal'{\sim}\Dcal^{m}}\EE_{h'{\sim} \P}\LP\phi(h'\!, \Scal')^{\frac{\alpha}{\alpha-1}}\RP\RB^{\frac{\alpha-1}{\alpha}}\RP\\
    = &\frac{1}{\alpha{-}1}\ln\LP\EE_{\Scal'{\sim}\Dcal^{m}}\EE_{h'{\sim} \P}\LP\Bigg[\frac{\AQprime (h')}{\P(h')}\Bigg]^{\alpha}\RP\RP + \ln\LP\frac{1}{\delta^{\frac{\alpha}{\alpha{-}1}}}\EE_{\Scal'{\sim}\Dcal^{m}}\EE_{h'{\sim} \P}\LP\phi(h'\!, \Scal')^{\frac{\alpha}{\alpha-1}}\RP\RP\!.
\end{align*}
Hence, we can deduce that 
\begin{align*}
    \PP_{\Scal\sim\D^{m}, h\sim \AQ}\Bigg[&\forall\P{\in}\Mcal^{*}(\Hcal), \frac{\alpha}{\alpha{-}1}\!\ln\!\LP\phi(h,\!\Scal)\RP \le \frac{1}{\alpha{-}1}\!\ln\!\LP\EE_{\Scal'{\sim}\Dcal^{m}}\EE_{h'{\sim} \P}\!\LP\Bigg[\!\frac{\AQprime (h')}{\P(h')}\!\Bigg]^{\alpha}\RP\RP\\
    &+ \ln\!\LP\!\tfrac{1}{\delta^{\frac{\alpha}{\alpha{-}1}}}\EE_{\Scal'{\sim}\Dcal^{m}}\EE_{h'{\sim} \P}\LP\phi(h'\!, \Scal')^{\frac{\alpha}{\alpha-1}}\RP\!\RP \Bigg]\ge 1{-}\delta,
\end{align*}
where, by definition, we have $D_{\alpha}(\rho\|\pi)=\frac{1}{\alpha{-}1}\!\ln\!\LP\EE_{\Scal'{\sim}\Dcal^{m}}\EE_{h'{\sim} \P}\!\LP\LB\!\frac{\AQprime (h')}{\P(h')}\!\RB^{\alpha}\RP\RP$.
\end{proof}
\label{lemma:mutual-info}
\end{lemma}
\noindent From Lemma~\ref{lemma:mutual-info}, we prove Theorem~\ref{theorem:mutual-info}.

\theoremmutualinfo*
\begin{proof}
Note that Sibson's mutual information is defined as $I_{\alpha}(h{;}\Scal)=\min_{\P\in\Mcal^{*}(\Hcal)}D_{\alpha}(\rho\|\pi)$.
Hence, in order to prove Theorem~\ref{theorem:mutual-info}, we have to instantiate Lemma~\ref{lemma:mutual-info} with the optimal prior, \ie, the prior $\P$ which minimizes $D_{\alpha}(\rho\|\pi)$.
Actually, this optimal prior has a closed-form solution~\citep{Verdu2015}.
For the sake of completeness, we derive it. First, we have
\begin{align*}
    &D_{\alpha}(\rho\|\pi)\\
    = &\frac{1}{\alpha{-}1}\!\ln\!\LP\EE_{\Scal\sim\Dcal^{m}}\EE_{h\sim \P}\!\LP\LB\!\frac{\AQ(h)}{\P(h)}\!\RB^{\alpha}\RP\RP\\
    = &\frac{1}{\alpha{-}1}\!\ln\!\LP\EE_{h\sim \P}\LB\EE_{\Scal\sim\Dcal^{m}}\LP\AQ(h)^{\alpha}\RP\RB\LP\P(h)^{-\alpha}\RP\!\RP\\
    = &\frac{1}{\alpha{-}1}\!\ln\!\LP\!\EE_{h\sim \P}\LB\EE_{\Scal\sim\Dcal^{m}}\LP\AQ (h)^{\alpha}\RP\RB\LP\P(h)^{-\alpha}\RP\!\!\LB\tfrac{\EE_{h'{\sim}\P}\tfrac{1}{\P(h')}\LB\EE_{\Scal'{\sim}\Dcal^{m}}\LP\AQprime (h')^{\alpha}\RP\RB^{\frac{1}{\alpha}}}{\EE_{h'{\sim}\P}\tfrac{1}{\P(h')}\LB\EE_{\Scal'{\sim}\Dcal^{m}}\LP\AQprime (h')^{\alpha}\RP\RB^{\frac{1}{\alpha}}}\RB^{\!\alpha}\RP\\
    = &\frac{\alpha}{\alpha{-}1}\!\ln\!\LP\EE_{h'{\sim}\P}\!\tfrac{1}{\P(h')}\!\!\LB\EE_{\Scal'{\sim}\Dcal^{m}}\LP\AQprime (h')^{\alpha}\RP\RB^{\!\frac{1}{\alpha}}\!\RP\!\\
    &\hspace{0.1cm}+\frac{1}{\alpha{-}1}\!\ln\!\LP\EE_{h\sim\P}\tfrac{1}{\P(h)^{\alpha}}\!\!\LB \!\tfrac{\LB\EE_{\Scal\sim\Dcal^{m}}\LP\AQ (h)^{\alpha}\RP\RB^{\frac{1}{\alpha}}}{\EE_{h'{\sim}\P}\!\!\tfrac{1}{\P(h')}\LB\EE_{\Scal'{\sim}\Dcal^{m}}\LP\AQprime (h')^{\alpha}\RP\RB^{\frac{1}{\alpha}}}\!\RB^{\!\alpha} \RP\\
    = &\frac{\alpha}{\alpha{-}1}\ln\LP\EE_{h'{\sim}\P}\tfrac{1}{\P(h')}\LB\EE_{\Scal'{\sim}\Dcal^{m}}\LP\AQprime (h')^{\alpha}\RP\RB^{\frac{1}{\alpha}}\RP + D_{\alpha}(\P^{*}\| \P),
\end{align*}
where $\P^*(h)=\LB \!\tfrac{\LB\EE_{\Scal\sim\Dcal^{m}}\LP\AQ (h)^{\alpha}\RP\RB^{\frac{1}{\alpha}}}{\EE_{h'{\sim}\P}\tfrac{1}{\P(h')}\LB\EE_{\Scal'{\sim}\Dcal^{m}}\LP\AQprime (h')^{\alpha}\RP\RB^{\frac{1}{\alpha}}}\!\RB$.

\noindent From these equalities and using the fact that $D_{\alpha}(\P^*\| \P)$ is minimal (\ie, equal to zero) when $\P^*=\P$, we can deduce that
\begin{align*}
    &\argmin_{\P\in\Mcal^{*}(\Hcal)}D_{\alpha}(\rho\|\pi)\\
    {=} &\argmin_{\P\in\Mcal^{*}(\Hcal)} \LB\frac{\alpha}{\alpha{-}1}\ln\LP\EE_{h'{\sim}\P}\tfrac{1}{\P(h')}\LB\EE_{\Scal'{\sim}\Dcal^{m}}\LP\AQprime (h')^{\alpha}\RP\RB^{\frac{1}{\alpha}}\RP{+} D_{\alpha}(\P^{*}\| \P)\RB\\
    {=}&\argmin_{\P\in\Mcal^{*}(\Hcal)}D_{\alpha}(\P^{*}\| \P){=}\P^*.
\end{align*}
Note that $\P^*$ is defined from the data distribution $\Dcal$, hence, $\P^*$ is a valid prior when instantiating Lemma~\ref{lemma:mutual-info} with $\P^*$.
Then, we have with probability at least $1{-}\delta$ over $\Scal\sim\Dcal^m$ and $h\sim\AQ$
\begin{align*}
    \frac{\alpha}{\alpha{-}1}\!\ln\!\LP\phi(h,\!\Scal)\RP &\le D_{\alpha}(\rho\|\pi^*) + \ln\!\LP\!\tfrac{1}{\delta^{\frac{\alpha}{\alpha{-}1}}}\EE_{\Scal'{\sim}\Dcal^{m}}\EE_{h'{\sim} \P}\LP\phi(h'\!, \Scal')^{\frac{\alpha}{\alpha-1}}\RP\!\RP\\
    &= I_{\alpha}(h'; \Scal') +\ \ln\!\LP\!\tfrac{1}{\delta^{\frac{\alpha}{\alpha{-}1}}}\EE_{\Scal'{\sim}\Dcal^{m}}\EE_{h'{\sim} \P}\LP\phi(h'\!, \Scal')^{\frac{\alpha}{\alpha-1}}\RP\!\RP.
\end{align*}
where $\pi^*(h, \Scal)=\P^*(h)\Dcal^{m}(\Scal)$.
\end{proof}

\subsection{About Theorem~\ref{theorem:mutual-info}}

For the sake of comparison, we introduce the following corollary of Theorem~\ref{theorem:mutual-info}.

\begin{corollary} Under the assumptions of  Theorem~\ref{theorem:mutual-info}, when $\alpha{\to}1^+$,  with probability at least $1{-}\delta$ we have
\begin{align*}
\ln\phi(h{,}\Scal) \le \ln\frac{1}{\delta} + \ln\left[\esssup_{\Scal'\in\Zcal, h'\in\Hcal}\phi(h'{,} \Scal')\right].
\end{align*}
\textit{When $\alpha{\to}+\infty$, with probability at least $1{-}\delta$ we have}
\begin{align*}
\ln\phi(h{,} \Scal){\le}\ln\LP\esssup_{\Scal\in\Scal, h\in\Hcal}\frac{\AQ(h)}{\P^*(h)}\RP{+}\ln\!\Big[\frac{1}{\delta} {\displaystyle \EE_{\Scal'{\sim}\Dcal^{m}}\EE_{h'{\sim}\P}\phi(h'{,}\Scal')}\Big]\!.  
\end{align*}
\begin{proof} 
The proof is similar to Corollary~\ref{corollary:disintegrated}.
Starting from Theorem~\ref{theorem:mutual-info} and rearranging, we have
\begin{align*}
    \PP_{\substack{\Scal\sim\D^{m}\\ h\sim \AQ}}\Bigg[ &\!\ln\!\LP\phi(h,\!\Scal)\RP \le  \frac{\alpha{-}1}{\alpha}I_{\alpha}(h'; \Scal')\\ 
    &+\ln\frac{1}{\delta} + \ln\!\LP\LB\EE_{\Scal'{\sim}\Dcal^{m}}\EE_{h'{\sim} \P^*}\LP\phi(h'\!, \Scal')^{\frac{\alpha}{\alpha-1}}\RP\RB^{\frac{\alpha{-}1}{\alpha}}\RP \Bigg]\ge 1{-}\delta,
\end{align*}
Then, we will prove separately the case when $\alpha\rightarrow 1$ and $\alpha\rightarrow +\infty$.\\

\noindent{}\textbf{When $\alpha\rightarrow 1$.}\\
First, we have  $\lim_{\alpha\rightarrow 1^+}\frac{\alpha{-}1}{\alpha}I_{\alpha}(h'; \Scal') = 0$.
Furthermore, note that 
\begin{align*}
    \|\phi\|_{\frac{\alpha}{\alpha{-}1}} = \LB\EE_{\Scal'{\sim}\Dcal^{m}}\EE_{h'{\sim}\P^*}\LP\vert\phi(h'\!, \Scal')\vert^{\frac{\alpha}{\alpha{-}1}}\RP\RB^{\frac{\alpha{-}1}{\alpha}} = \LB\EE_{\Scal'{\sim}\Dcal^{m}}\EE_{h'{\sim}\P^*}\LP\phi(h'\!, \Scal')^{\frac{\alpha}{\alpha{-}1}}\RP\RB^{\frac{\alpha{-}1}{\alpha}}
\end{align*}
is the $L^{\frac{\alpha}{\alpha{-}1}}$-norm of the function $\phi: \Hcal\times\Zcal^m \rightarrow \Rpe$, where $\lim_{\alpha\rightarrow 1} \|\phi\|_{\frac{\alpha}{\alpha{-}1}} = \lim_{\alpha'\rightarrow +\infty} \|\phi\|_{\alpha'}$ (since we have $\lim_{\alpha\rightarrow 1^+}\frac{\alpha}{\alpha{-}1} = (\lim_{\alpha\rightarrow1}\alpha)(\lim_{\alpha\rightarrow1}\frac{1}{\alpha{-}1}) = +\infty$).
Then, it is well known that
\begin{align*}
    \|\phi\|_{\infty}= \lim_{\alpha'\rightarrow+\infty}\|\phi\|_{\alpha'} = \esssup_{\Scal'\in\Zcal, h'\in\Hcal}\phi(h'{,} \Scal').
\end{align*}
Hence, we have 
\begin{align*}
    &\lim_{\alpha\rightarrow1} \ln\LP\LB\EE_{\Scal'{\sim}\Dcal^{m}}\EE_{h'{\sim}\P^*}\LP\phi(h'\!, \Scal')^{\frac{\alpha}{\alpha{-}1}}\RP\RB^{\frac{\alpha{-}1}{\alpha}}\RP\\
    = &\ln\LP \lim_{\alpha\rightarrow1} \LB\EE_{\Scal'{\sim}\Dcal^{m}}\EE_{h'{\sim}\P^*}\LP\phi(h'\!, \Scal')^{\frac{\alpha}{\alpha{-}1}}\RP\RB^{\frac{\alpha{-}1}{\alpha}}\RP\\
    = &\ln\LP \lim_{\alpha\rightarrow1} \| \phi\|_{\frac{\alpha}{\alpha-1}}\RP = \ln\LP \lim_{\alpha'\rightarrow+\infty} \| \phi\|_{\alpha'}\RP \\
    = &\ln\LP \| \phi\|_{\infty}\RP = \ln\LP \esssup_{\Scal'\in\Zcal, h'\in\Hcal}\phi(h'{,} \Scal') \RP.
\end{align*}
Finally, we can deduce that 
\begin{align*}
    &\lim_{\alpha\rightarrow 1}\LB \frac{\alpha{-}1}{\alpha}I_{\alpha}(h'; \Scal') +\ \ln\frac{1}{\delta} + \ln\!\LP\LB\EE_{\Scal'{\sim}\Dcal^{m}}\EE_{h'{\sim} \P^*}\LP\phi(h'\!, \Scal')^{\frac{\alpha}{\alpha-1}}\RP\RB^{\frac{\alpha{-}1}{\alpha}}\RP\RB\\
    = &\ln\frac{1}{\delta} + \ln\left[\esssup_{\Scal'\in\Zcal, h'\in\Hcal}\phi(h'{,} \Scal')\right].\\
\end{align*}

~\\
\noindent{}\textbf{When $\alpha\rightarrow +\infty$.}\\
First, we have $\lim_{\alpha\rightarrow +\infty} \|\phi\|_{\frac{\alpha}{\alpha{-}1}} = \lim_{\alpha'\rightarrow 1} \|\phi\|_{\alpha'} = \|\phi\|_1$ 
Hence, we have 
\begin{align*}
    &\lim_{\alpha\rightarrow+\infty} \ln\LP\LB\EE_{\Scal'{\sim}\Dcal^{m}}\EE_{h'{\sim}\P^*}\LP\phi(h'\!, \Scal')^{\frac{\alpha}{\alpha{-}1}}\RP\RB^{\frac{\alpha{-}1}{\alpha}}\RP\\
    = &\ln\LP \lim_{\alpha\rightarrow+\infty} \LB\EE_{\Scal'{\sim}\Dcal^{m}}\EE_{h'{\sim}\P^*}\LP\phi(h'\!, \Scal')^{\frac{\alpha}{\alpha{-}1}}\RP\RB^{\frac{\alpha{-}1}{\alpha}}\RP\\
    = &\ln\LP \lim_{\alpha\rightarrow+\infty} \| \phi\|_{\frac{\alpha}{\alpha-1}}\RP = \ln\LP \lim_{\alpha'\rightarrow1} \| \phi\|_{\alpha'}\RP\\
    = &\ln\LP \| \phi\|_{1}\RP = \ln\LP \EE_{\Scal'{\sim}\Dcal^{m}}\EE_{h'{\sim}\P^*}\phi(h'\!, \Scal') \RP.
\end{align*}
Moreover, by rearranging the terms in $\frac{\alpha{-}1}{\alpha}I_{\alpha}(h'; \Scal')$, we have
\begin{align*}
\frac{\alpha{-}1}{\alpha}I_{\alpha}(h'; \Scal') &= \frac{1}{\alpha}\ln\!\LP \EE_{\Scal{\sim}\Dcal^m}\EE_{h{\sim}\P^*}\!\LP\!\LB\frac{ \AQ(h)}{\P^*(h)}\RB^{\!\alpha}\RP\RP\\
&= \ln\!\LP \LB\EE_{\Scal{\sim}\Dcal^m}\EE_{h{\sim}\P^*}\!\LP\LB\!\frac{ \AQ(h)}{\P^*(h)}\RB^{\!\alpha}\RP\RB^{\frac{1}{\alpha}}\RP\\
&= \ln\!\LP \LB\EE_{h{\sim}\P^*}\LP\gamma(h)^{\alpha}\RP\RB^{\frac{1}{\alpha}}\RP = \ln\!\LP \| \gamma\|_{\alpha}\RP,
\end{align*}
where $\| \gamma\|_{\alpha}$ is the $L^{\alpha}$-norm of the function $\gamma$ defined as $\gamma(h)=\tfrac{\AQ(h)}{\P^*(h)}$.
We have
\begin{align*}
    \lim_{\alpha\rightarrow +\infty}  \frac{\alpha{-}1}{\alpha}I_{\alpha}(h'; \Scal') =& \lim_{\alpha\rightarrow +\infty}  \ln\!\LP \| \gamma\|_{\alpha}\RP = \ln\LP\lim_{\alpha\rightarrow +\infty} \|\gamma\|_{\alpha}\RP\\
    =& \ln\LP \|\gamma\|_{\infty}\RP = \ln\LP\esssup_{\Scal\in\Scal, h\in\Hcal}\gamma(h)\RP = \ln\LP\esssup_{\Scal\in\Scal, h\in\Hcal}\frac{\AQ(h)}{\P^*(h)}\RP.
\end{align*}
Finally, we can deduce that 
\begin{align*}
    & \lim_{\alpha\rightarrow 1}\LB \frac{\alpha{-}1}{\alpha}I_{\alpha}(h'; \Scal') +\ \ln\frac{1}{\delta} + \ln\!\LP\LB\EE_{\Scal'{\sim}\Dcal^{m}}\EE_{h'{\sim} \P^*}\LP\phi(h'\!, \Scal')^{\frac{\alpha}{\alpha-1}}\RP\RB^{\frac{\alpha{-}1}{\alpha}}\RP\RB\\ 
    =\quad &\ln\LP\esssup_{\Scal\in\Scal, h\in\Hcal}\frac{\AQ(h)}{\P^*(h)}\RP{+}\ln\!\Big[\frac{1}{\delta} {\displaystyle \EE_{\Scal'{\sim}\Dcal^{m}}\EE_{h'{\sim}\P}\phi(h'{,}\Scal')}\Big].
\end{align*}
\end{proof}
\end{corollary}
\noindent As for Theorem~\ref{theorem:disintegrated}, this corollary illustrates a trade-off introduced by $\alpha$ between the Sibson's mutual information $I_{\alpha}(h'; \Scal')$ and the term $\ln\!\LP\EE_{\Scal'{\sim}\Dcal^{m}}\EE_{h'{\sim} \P}\LP\phi(h'\!, \Scal')^{\frac{\alpha}{\alpha-1}}\RP\!\RP$.\\

\noindent Furthermore, \citet[Cor.4]{EspositoGastparIssa2020}  introduced a bound involving Sibson's mutual information.
Their bound holds with probability at least $1{-}\delta$ over $\Scal\sim\Dcal^m$ and $h\sim\AQ$:
\begin{align}
2(\Risk_{\Scal}(h){-}\Risk_{\Dcal}(h))^2\le \tfrac{1}{m}\!\LB I_{\alpha}(h'; \Scal') + \ln \tfrac{2}{\delta^{\frac{\alpha}{\alpha{-}1}}}\RB.\label{eq:esposito}
\end{align}
Hence, we compare Equation~\eqref{eq:esposito} with the equations of the following corollary.
\begin{corollary}
For any distribution $\Dcal$ on $\Zcal$, for any hypothesis set $\Hcal$, for any $\alpha\!>\!1$, for any $\delta\in(0,1]$, for any algorithm $A\!:\!\Zcal^{m}\times\Mcal^{*}(\Hcal){\rightarrow} \Mcal(\Hcal)$, with probability at least $1{-}\delta$ over $\Scal\sim\Dcal^m$ and $h\sim\AQ$, we have
\begin{align}
   & \kl(\Risk_{\Scal}(h)\|\Risk_{\Dcal}(h))\! 
    \le \! \tfrac{1}{m}\!\LB I_{\alpha}(h'; \Scal')\! +\! \ln\! \tfrac{2\sqrt{m}}{\delta^{\frac{\alpha}{\alpha{-}1}}} \RB\label{eq:mutual-info-seeger}\\
   \text{\quad and\quad}
    & 2(\Risk_{\Scal}(h){-}\Risk_{\Dcal}(h))^2\! \le\! \tfrac{1}{m}\!\LB I_{\alpha}(h'; \Scal')\! +\! \ln\! \tfrac{2\sqrt{m}}{\delta^{\frac{\alpha}{\alpha{-}1}}}\RB\label{eq:mutual-info-mcallester}\!.
\end{align}
\begin{proof}
First of all, we instantiate Theorem~\ref{theorem:mutual-info} with $\phi(h,\!\Scal)=\exp\!\LB\tfrac{\alpha-1}{\alpha}m\kl(\Risk_{\Scal}(h)\|\Risk_{\Dcal}(h))\RB$, we have (by rearranging the terms)
\begin{align*}
    \kl(\Risk_{\Scal}(h)\|\Risk_{\Dcal}(h))\! 
    \le \! \frac{1}{m}\!\LB I_{\alpha}(h'; \Scal')\! +\! \ln\!\LP\! \tfrac{1}{\delta^{\frac{\alpha}{\alpha{-}1}}}\EE_{\Scal'{\sim}\Dcal^m}\EE_{h'{\sim}\P}e^{m\kl(\Risk_{\Scal'}(h')\|\Risk_{\Dcal}(h'))}\RP \RB\!.
\end{align*}
Then, from \citet{Maurer2004}, we upper-bound  $\EE_{\Scal'{\sim}\Dcal^m}\EE_{h'{\sim}\P} e^{m\kl(\Risk_{\Scal'}(h')\|\Risk_{\Dcal}(h'))}$ by $2\sqrt{m}$ to obtain Equation~\eqref{eq:mutual-info-seeger}.
Finally, to obtain Equation~\eqref{eq:mutual-info-mcallester}, we apply Pinsker's inequality, \ie, $2(\Risk_{\Scal}(h){-}\Risk_{\Dcal}(h))^2\le \kl(\Risk_{\Scal}(h)\|\Risk_{\Dcal}(h))$ on Equation~\eqref{eq:mutual-info-seeger}.
\end{proof}
\label{corollary:mutual-info}
\end{corollary}
\noindent Equation~\eqref{eq:mutual-info-mcallester} is slightly looser than Equation~\eqref{eq:esposito} since it involves an extra term of $\tfrac1m\ln\sqrt{m}$.
However, Equation~\eqref{eq:mutual-info-seeger} is tighter than Equation~\eqref{eq:esposito} when $\kl(\Risk_{\Scal}(h)\|\Risk_{\Dcal}(h)){-}2(\Risk_{\Scal}(h){-}\Risk_{\Dcal}(h))^2 \ge \tfrac1m\ln\sqrt{m}$ (which becomes more frequent as $m$ grows).

\section{Results presented in Section~\ref{sec:expe}}
\label{ap:details}

This appendix presents the details of the results of Section~\ref{sec:expe}.
Tables~\ref{table:1_prior_0.1} to~\ref{table:1_prior_0.9} report empirical results for split ratios going from 0.0 to 0.9 presented in Figures~\ref{figure:exp-4} to~\ref{figure:exp-3}.
More precisely, we report the test risk $\Risk_{\Tcal}(h)$, the empirical risk $\Risk_{\Scal}(h)$, the bound value (Bnd), and the divergence value associated with the network $h$ sampled from the posterior $\AQ$ for each learning rate, variance, dataset, and bound type.
Tables~\ref{table:2_data_mnist} to~\ref{table:2_data_cifar10} report the performances of the prior before applying Step {\bf 2)} outlined in Figures~\ref{figure:exp-1} and~\ref{figure:exp-3}.
In particular, we report the test risk $\Risk_{\Tcal}(h)$, the empirical risk $\Risk_{\Scal}(h)$, the bound values of Corollary~\ref{corollary:nn} and  Equations~\eqref{eq:nn-rivasplata},~\eqref{eq:nn-blanchard},~\eqref{eq:nn-catoni} for each split ratio and variance.\\

\noindent{}Note that for the split 0.0, since Step {\bf 1)} is skipped, the prior distribution $\P$ is only initialized as introduced in Section~\ref{sec:models}. 
Note that in this case, $T=1$ since we have only one prior.
To do the same number of epochs compared to the other splits, we perform 11 epochs (instead of 1) for MNIST and Fashion-MNIST and 110 epochs (instead of 10) for CIFAR-10 during Step {\bf 2)}.
The other parameters are not changed.

\begin{sidewaystable}
\caption{
\looseness=-1
Comparison of \ours, \rivasplata, \blanchard and \catoni based on the disintegrated bounds, and \stoNN based on the randomized bounds learned with two learning rates lr${\ \in}\{10^{-4}, 10^{-6}\}$ and different variances $\sigma^2{\in}\{10^{-3}, 10^{-4}, 10^{-5}, 10^{-6}\}$.
We report the test risk ($\Risk_{\Tcal}(h)$), the bound value (Bnd), the empirical risk ($\Risk_{\Scal}(h)$), and the divergence (Div) associated with each bound (the Rényi divergence for \ours, the KL divergence for \stoNN, and the disintegrated KL divergence for \rivasplata, \blanchard and \catoni).
More precisely, we report the mean $\pm$ the standard deviation for $400$ neural networks sampled from $\AQ$ for \ours, \rivasplata, \blanchard, and \catoni.
We consider, in this figure, that the split ratio is $0.0$.
}
\resizebox{0.83\paperheight}{!}{
\begin{tabular}{rr|clcl|clcl|clcl|clcl}
\toprule
 &  & \multicolumn{4}{c}{$\sigma^2=10^{-6}$} & \multicolumn{4}{c}{$\sigma^2=10^{-5}$} & \multicolumn{4}{c}{$\sigma^2=10^{-4}$} & \multicolumn{4}{c}{$\sigma^2=10^{-3}$} \\
\midrule
 & lr=$10^{-6}$ & $\Risk_{\Tcal}(h)$ & Bnd & $\Risk_{\Scal}(h)$ & Div & $\Risk_{\Tcal}(h)$ & Bnd & $\Risk_{\Scal}(h)$ & Div & $\Risk_{\Tcal}(h)$ & Bnd & $\Risk_{\Scal}(h)$ & Div & $\Risk_{\Tcal}(h)$ & Bnd & $\Risk_{\Scal}(h)$ & Div \\
\midrule
\multirow[c]{5}{*}{\rotatebox[origin=c]{90}{\small{MNIST}}} & \ours & .901 $\pm$ .002 & .908 $\pm$ .002 & .901 $\pm$ .002 & .005 & .897 $\pm$ .013 & .904 $\pm$ .012 & .897 $\pm$ .012 & .009 & .898 $\pm$ .017 & .905 $\pm$ .016 & .898 $\pm$ .016 & .027 & .902 $\pm$ .015 & .908 $\pm$ .014 & .901 $\pm$ .015 & .671 \\
 & \blanchard & .901 $\pm$ .002 & .926 $\pm$ .002 & .901 $\pm$ .002 & 122.846 $\pm$ 15.952 & .897 $\pm$ .013 & .912 $\pm$ .012 & .897 $\pm$ .013 & 39.350 $\pm$ 8.999 & .898 $\pm$ .017 & .907 $\pm$ .016 & .898 $\pm$ .017 & 13.023 $\pm$ 4.818 & .901 $\pm$ .015 & .907 $\pm$ .014 & .901 $\pm$ .014 & 3.041 $\pm$ 2.459 \\
 & \catoni & .901 $\pm$ .002 & .926 $\pm$ .003 & .901 $\pm$ .002 & 121.860 $\pm$ 15.930 & .897 $\pm$ .013 & .909 $\pm$ .012 & .897 $\pm$ .013 & 38.552 $\pm$ 8.872 & .898 $\pm$ .017 & .905 $\pm$ .016 & .898 $\pm$ .017 & 12.474 $\pm$ 4.774 & .901 $\pm$ .014 & .906 $\pm$ .013 & .901 $\pm$ .014 & 3.088 $\pm$ 2.379 \\
 & \rivasplata & .901 $\pm$ .002 & .920 $\pm$ .002 & .901 $\pm$ .002 & 123.301 $\pm$ 15.941 & .896 $\pm$ .014 & .908 $\pm$ .012 & .896 $\pm$ .013 & 39.195 $\pm$ 8.959 & .897 $\pm$ .017 & .905 $\pm$ .016 & .897 $\pm$ .017 & 12.827 $\pm$ 4.858 & .902 $\pm$ .015 & .907 $\pm$ .014 & .901 $\pm$ .015 & 3.232 $\pm$ 2.454 \\
 & \stoNN & \textemdash & .944 & \textemdash & .002 & \textemdash & .941 & \textemdash & .004 & \textemdash & .941 & \textemdash & .014 & \textemdash & .944 & \textemdash & .336 \\
\midrule
\multirow[c]{5}{*}{\rotatebox[origin=c]{90}{\small{Fashion}}} & \ours & .970 $\pm$ .028 & .972 $\pm$ .025 & .970 $\pm$ .027 & .016 & .944 $\pm$ .038 & .949 $\pm$ .035 & .944 $\pm$ .037 & .046 & .910 $\pm$ .027 & .917 $\pm$ .026 & .910 $\pm$ .027 & .140 & .901 $\pm$ .026 & .909 $\pm$ .025 & .901 $\pm$ .026 & 1.255 \\
 & \blanchard & .970 $\pm$ .029 & .978 $\pm$ .019 & .970 $\pm$ .028 & 122.508 $\pm$ 16.085 & .942 $\pm$ .038 & .952 $\pm$ .032 & .943 $\pm$ .038 & 39.957 $\pm$ 8.610 & .910 $\pm$ .031 & .919 $\pm$ .029 & .910 $\pm$ .031 & 12.649 $\pm$ 4.846 & .899 $\pm$ .028 & .905 $\pm$ .027 & .899 $\pm$ .028 & 3.206 $\pm$ 2.566 \\
 & \catoni & .970 $\pm$ .028 & .983 $\pm$ .017 & .970 $\pm$ .027 & 122.364 $\pm$ 15.860 & .945 $\pm$ .038 & .954 $\pm$ .036 & .945 $\pm$ .037 & 38.555 $\pm$ 8.873 & .912 $\pm$ .032 & .919 $\pm$ .031 & .912 $\pm$ .032 & 12.167 $\pm$ 4.762 & .899 $\pm$ .027 & .905 $\pm$ .026 & .899 $\pm$ .027 & 3.122 $\pm$ 2.392 \\
 & \rivasplata & .970 $\pm$ .028 & .977 $\pm$ .021 & .971 $\pm$ .027 & 123.328 $\pm$ 15.929 & .943 $\pm$ .038 & .950 $\pm$ .033 & .943 $\pm$ .038 & 39.300 $\pm$ 8.991 & .908 $\pm$ .031 & .916 $\pm$ .029 & .908 $\pm$ .031 & 12.627 $\pm$ 4.890 & .899 $\pm$ .028 & .905 $\pm$ .027 & .899 $\pm$ .028 & 3.591 $\pm$ 2.610 \\
 & \stoNN & \textemdash & .990 & \textemdash & .008 & \textemdash & .975 & \textemdash & .023 & \textemdash & .950 & \textemdash & .070 & \textemdash & .944 & \textemdash & .627 \\
\midrule
\multirow[c]{5}{*}{\rotatebox[origin=c]{90}{\small{CIFAR-10}}} & \ours & .899 $\pm$ .000 & .907 $\pm$ .000 & .899 $\pm$ .000 & 3.113 & .896 $\pm$ .002 & .914 $\pm$ .002 & .894 $\pm$ .002 & 107.797 & .826 $\pm$ .011 & .885 $\pm$ .009 & .825 $\pm$ .010 & 76.475 & .786 $\pm$ .019 & .851 $\pm$ .015 & .788 $\pm$ .018 & 714.351 \\
 & \blanchard & .899 $\pm$ .000 & .940 $\pm$ .001 & .898 $\pm$ .000 & 314.983 $\pm$ 26.377 & .888 $\pm$ .004 & .927 $\pm$ .002 & .885 $\pm$ .003 & 28.250 $\pm$ 25.255 & .823 $\pm$ .010 & .885 $\pm$ .008 & .822 $\pm$ .010 & 422.401 $\pm$ 29.323 & .798 $\pm$ .019 & .856 $\pm$ .015 & .799 $\pm$ .018 & 292.706 $\pm$ 25.318 \\
 & \catoni & .899 $\pm$ .000 & .941 $\pm$ .000 & .898 $\pm$ .000 & 285.415 $\pm$ 25.085 & .894 $\pm$ .002 & .930 $\pm$ .004 & .892 $\pm$ .002 & 169.713 $\pm$ 19.543 & .857 $\pm$ .010 & .915 $\pm$ .009 & .856 $\pm$ .010 & 273.554 $\pm$ 23.212 & .815 $\pm$ .019 & .864 $\pm$ .017 & .816 $\pm$ .018 & 209.069 $\pm$ 21.230 \\
 & \rivasplata & .899 $\pm$ .001 & .930 $\pm$ .001 & .898 $\pm$ .000 & 362.070 $\pm$ 28.420 & .864 $\pm$ .004 & .933 $\pm$ .002 & .862 $\pm$ .004 & 1568.007 $\pm$ 55.492 & .748 $\pm$ .010 & .837 $\pm$ .007 & .750 $\pm$ .009 & 1219.178 $\pm$ 49.610 & .769 $\pm$ .018 & .828 $\pm$ .015 & .771 $\pm$ .017 & 526.068 $\pm$ 33.837 \\
 & \stoNN & \textemdash & .942 & \textemdash & 1.557 & \textemdash & .945 & \textemdash & 53.898 & \textemdash & .914 & \textemdash & 38.237 & \textemdash & .884 & \textemdash & 357.175 \\
\midrule
 &  & \multicolumn{4}{c}{$\sigma^2=10^{-6}$} & \multicolumn{4}{c}{$\sigma^2=10^{-5}$} & \multicolumn{4}{c}{$\sigma^2=10^{-4}$} & \multicolumn{4}{c}{$\sigma^2=10^{-3}$} \\
\midrule
 & lr=$10^{-4}$ & $\Risk_{\Tcal}(h)$ & Bnd & $\Risk_{\Scal}(h)$ & Div & $\Risk_{\Tcal}(h)$ & Bnd & $\Risk_{\Scal}(h)$ & Div & $\Risk_{\Tcal}(h)$ & Bnd & $\Risk_{\Scal}(h)$ & Div & $\Risk_{\Tcal}(h)$ & Bnd & $\Risk_{\Scal}(h)$ & Div \\
\midrule
\multirow[c]{5}{*}{\rotatebox[origin=c]{90}{\small{MNIST}}} & \ours & .901 $\pm$ .002 & .909 $\pm$ .002 & .901 $\pm$ .002 & 3.767 & .896 $\pm$ .014 & .904 $\pm$ .013 & .896 $\pm$ .014 & .835 & .898 $\pm$ .016 & .905 $\pm$ .015 & .898 $\pm$ .016 & 1.062 & .901 $\pm$ .015 & .909 $\pm$ .014 & .901 $\pm$ .015 & 6.022 \\
 & \blanchard & .900 $\pm$ .003 & .990 $\pm$ .000 & .900 $\pm$ .003 & 12004.196 $\pm$ 152.632 & .894 $\pm$ .017 & .986 $\pm$ .006 & .894 $\pm$ .016 & 3837.785 $\pm$ 93.560 & .888 $\pm$ .021 & .957 $\pm$ .013 & .888 $\pm$ .020 & 1221.198 $\pm$ 49.920 & .898 $\pm$ .015 & .939 $\pm$ .012 & .897 $\pm$ .015 & 391.343 $\pm$ 28.182 \\
 & \catoni & .900 $\pm$ .003 & .997 $\pm$ .002 & .900 $\pm$ .003 & 5694.194 $\pm$ 102.906 & .889 $\pm$ .020 & .967 $\pm$ .012 & .889 $\pm$ .019 & 3331.617 $\pm$ 78.945 & .879 $\pm$ .025 & .941 $\pm$ .016 & .880 $\pm$ .025 & 1481.726 $\pm$ 53.973 & .888 $\pm$ .023 & .937 $\pm$ .015 & .888 $\pm$ .023 & 567.893 $\pm$ 33.441 \\
 & \rivasplata & .900 $\pm$ .004 & .990 $\pm$ .000 & .900 $\pm$ .003 & 1199.818 $\pm$ 152.557 & .892 $\pm$ .017 & .970 $\pm$ .009 & .892 $\pm$ .016 & 3846.699 $\pm$ 84.643 & .886 $\pm$ .020 & .940 $\pm$ .015 & .886 $\pm$ .020 & 1224.463 $\pm$ 49.970 & .897 $\pm$ .018 & .928 $\pm$ .015 & .897 $\pm$ .018 & 393.757 $\pm$ 29.158 \\
 & \stoNN & \textemdash & .944 & \textemdash & 1.884 & \textemdash & .940 & \textemdash & .417 & \textemdash & .941 & \textemdash & .531 & \textemdash & .944 & \textemdash & 3.011 \\
\midrule
\multirow[c]{5}{*}{\rotatebox[origin=c]{90}{\small{Fashion}}} & \ours & .977 $\pm$ .024 & .979 $\pm$ .021 & .977 $\pm$ .023 & 3.926 & .947 $\pm$ .038 & .951 $\pm$ .035 & .947 $\pm$ .038 & 1.623 & .907 $\pm$ .030 & .914 $\pm$ .029 & .907 $\pm$ .030 & 2.947 & .900 $\pm$ .026 & .910 $\pm$ .025 & .900 $\pm$ .026 & 15.978 \\
 & \blanchard & .984 $\pm$ .015 & .990 $\pm$ .000 & .984 $\pm$ .015 & 12019.121 $\pm$ 166.251 & .912 $\pm$ .029 & .988 $\pm$ .004 & .911 $\pm$ .029 & 3846.861 $\pm$ 84.568 & .883 $\pm$ .029 & .953 $\pm$ .019 & .883 $\pm$ .029 & 1232.645 $\pm$ 5.285 & .403 $\pm$ .041 & .648 $\pm$ .038 & .399 $\pm$ .041 & 3853.231 $\pm$ 87.867 \\
 & \catoni & .983 $\pm$ .018 & 1.000 $\pm$ .000 & .983 $\pm$ .017 & 5654.642 $\pm$ 114.040 & .903 $\pm$ .021 & .985 $\pm$ .012 & .902 $\pm$ .021 & 4354.538 $\pm$ 94.427 & .751 $\pm$ .033 & .867 $\pm$ .023 & .750 $\pm$ .033 & 2702.652 $\pm$ 76.863 & .504 $\pm$ .041 & .673 $\pm$ .037 & .502 $\pm$ .041 & 3172.609 $\pm$ 78.698 \\
 & \rivasplata & .983 $\pm$ .016 & .990 $\pm$ .000 & .983 $\pm$ .016 & 11976.720 $\pm$ 165.964 & .905 $\pm$ .023 & .975 $\pm$ .007 & .905 $\pm$ .023 & 3855.872 $\pm$ 84.676 & .855 $\pm$ .035 & .916 $\pm$ .027 & .855 $\pm$ .035 & 125.110 $\pm$ 51.837 & .365 $\pm$ .032 & .559 $\pm$ .032 & .359 $\pm$ .033 & 4823.725 $\pm$ 103.813 \\
 & \stoNN & \textemdash & .990 & \textemdash & 1.963 & \textemdash & .977 & \textemdash & .812 & \textemdash & .948 & \textemdash & 1.473 & \textemdash & .944 & \textemdash & 7.989 \\
\midrule
\multirow[c]{5}{*}{\rotatebox[origin=c]{90}{\small{CIFAR-10}}} & \ours & .899 $\pm$ .000 & .915 $\pm$ .000 & .899 $\pm$ .000 & 63.416 & .890 $\pm$ .003 & .932 $\pm$ .003 & .886 $\pm$ .003 & 68.353 & .786 $\pm$ .011 & .888 $\pm$ .008 & .787 $\pm$ .010 & 2072.610 & .769 $\pm$ .017 & .859 $\pm$ .013 & .770 $\pm$ .017 & 1406.824 \\
 & \blanchard & .869 $\pm$ .002 & .990 $\pm$ .000 & .866 $\pm$ .001 & 27237.938 $\pm$ 251.770 & .813 $\pm$ .004 & .990 $\pm$ .000 & .812 $\pm$ .003 & 12052.733 $\pm$ 159.732 & .697 $\pm$ .011 & .920 $\pm$ .005 & .700 $\pm$ .009 & 5137.799 $\pm$ 103.680 & .674 $\pm$ .020 & .861 $\pm$ .014 & .675 $\pm$ .020 & 2814.450 $\pm$ 76.004 \\
 & \catoni & .928 $\pm$ .001 & 1.000 $\pm$ .000 & .925 $\pm$ .001 & 2145276.795 $\pm$ 2095.160 & .821 $\pm$ .002 & 1.000 $\pm$ .000 & .821 $\pm$ .002 & 375019.277 $\pm$ 896.780 & .689 $\pm$ .011 & .870 $\pm$ .007 & .692 $\pm$ .010 & 5292.535 $\pm$ 106.380 & .629 $\pm$ .019 & .805 $\pm$ .015 & .628 $\pm$ .019 & 4159.131 $\pm$ 96.763 \\
 & \rivasplata & .867 $\pm$ .002 & .990 $\pm$ .000 & .864 $\pm$ .001 & 35956.152 $\pm$ 268.304 & .812 $\pm$ .004 & .976 $\pm$ .001 & .811 $\pm$ .003 & 12135.134 $\pm$ 157.621 & .698 $\pm$ .010 & .874 $\pm$ .006 & .701 $\pm$ .009 & 5191.665 $\pm$ 102.712 & .677 $\pm$ .020 & .819 $\pm$ .015 & .678 $\pm$ .019 & 2839.514 $\pm$ 81.432 \\
 & \stoNN & \textemdash & .947 & \textemdash & 31.708 & \textemdash & .954 & \textemdash & 34.176 & \textemdash & .908 & \textemdash & 1036.305 & \textemdash & .886 & \textemdash & 703.412 \\
\bottomrule
\end{tabular}
}
\label{table:1_prior_0.0}
\end{sidewaystable}

\begin{sidewaystable}
\caption{
\looseness=-1
Comparison of \ours, \rivasplata, \blanchard and \catoni based on the disintegrated bounds, and \stoNN based on the randomized bounds learned with two learning rates lr${\ \in}\{10^{-4}, 10^{-6}\}$ and different variances $\sigma^2{\in}\{10^{-3}, 10^{-4}, 10^{-5}, 10^{-6}\}$.
We report the test risk ($\Risk_{\Tcal}(h)$), the bound value (Bnd), the empirical risk ($\Risk_{\Scal}(h)$), and the divergence (Div) associated with each bound (the Rényi divergence for \ours, the KL divergence for \stoNN, and the disintegrated KL divergence for \rivasplata, \blanchard and \catoni).
More precisely, we report the mean $\pm$ the standard deviation for $400$ neural networks sampled from $\AQ$ for \ours, \rivasplata, \blanchard, and \catoni.
We consider, in this table, that the split ratio is $0.1$.
}
\resizebox{0.83\paperheight}{!}{
\begin{tabular}{rr|clcl|clcl|clcl|clcl}
\toprule
 &  & \multicolumn{4}{c}{$\sigma^2=10^{-6}$} & \multicolumn{4}{c}{$\sigma^2=10^{-5}$} & \multicolumn{4}{c}{$\sigma^2=10^{-4}$} & \multicolumn{4}{c}{$\sigma^2=10^{-3}$} \\
\midrule
 & lr=$10^{-6}$ & $\Risk_{\Tcal}(h)$ & Bnd & $\Risk_{\Scal}(h)$ & Div & $\Risk_{\Tcal}(h)$ & Bnd & $\Risk_{\Scal}(h)$ & Div & $\Risk_{\Tcal}(h)$ & Bnd & $\Risk_{\Scal}(h)$ & Div & $\Risk_{\Tcal}(h)$ & Bnd & $\Risk_{\Scal}(h)$ & Div \\
\midrule
\multirow[c]{5}{*}{\rotatebox[origin=c]{90}{\small{MNIST}}} & \ours & .035 $\pm$ .000 & .044 $\pm$ .000 & .039 $\pm$ .000 & .622 & .024 $\pm$ .000 & .034 $\pm$ .000 & .029 $\pm$ .000 & 2.122 & .029 $\pm$ .002 & .040 $\pm$ .002 & .034 $\pm$ .002 & 12.754 & .034 $\pm$ .004 & .044 $\pm$ .004 & .038 $\pm$ .004 & 7.303 \\
 & \blanchard & .034 $\pm$ .000 & .058 $\pm$ .002 & .038 $\pm$ .000 & 99.876 $\pm$ 14.858 & .024 $\pm$ .000 & .038 $\pm$ .001 & .030 $\pm$ .000 & 21.775 $\pm$ 6.848 & .034 $\pm$ .002 & .043 $\pm$ .002 & .038 $\pm$ .002 & 3.949 $\pm$ 2.877 & .039 $\pm$ .005 & .047 $\pm$ .005 & .043 $\pm$ .005 & .590 $\pm$ 1.085 \\
 & \catoni & .035 $\pm$ .000 & .064 $\pm$ .001 & .039 $\pm$ .000 & 119.663 $\pm$ 15.854 & .024 $\pm$ .000 & .038 $\pm$ .001 & .030 $\pm$ .000 & 26.277 $\pm$ 7.490 & .033 $\pm$ .002 & .041 $\pm$ .002 & .037 $\pm$ .002 & 4.067 $\pm$ 2.882 & .038 $\pm$ .005 & .045 $\pm$ .005 & .042 $\pm$ .004 & .759 $\pm$ 1.217 \\
 & \rivasplata & .034 $\pm$ .000 & .052 $\pm$ .001 & .038 $\pm$ .000 & 104.880 $\pm$ 15.268 & .024 $\pm$ .000 & .036 $\pm$ .001 & .029 $\pm$ .000 & 23.007 $\pm$ 7.187 & .033 $\pm$ .002 & .042 $\pm$ .002 & .037 $\pm$ .002 & 4.116 $\pm$ 2.845 & .038 $\pm$ .005 & .046 $\pm$ .004 & .042 $\pm$ .004 & .775 $\pm$ 1.231 \\
 & \stoNN & \textemdash & .080 & \textemdash & .311 & \textemdash & .067 & \textemdash & 1.061 & \textemdash & .074 & \textemdash & 6.377 & \textemdash & .079 & \textemdash & 3.651 \\
\midrule
\multirow[c]{5}{*}{\rotatebox[origin=c]{90}{\small{Fashion}}} & \ours & .166 $\pm$ .001 & .169 $\pm$ .000 & .159 $\pm$ .000 & .580 & .157 $\pm$ .001 & .160 $\pm$ .001 & .150 $\pm$ .001 & 2.128 & .160 $\pm$ .002 & .161 $\pm$ .003 & .151 $\pm$ .002 & 3.503 & .176 $\pm$ .006 & .179 $\pm$ .006 & .168 $\pm$ .005 & 1.268 \\
 & \blanchard & .165 $\pm$ .001 & .192 $\pm$ .002 & .159 $\pm$ .000 & 96.822 $\pm$ 14.116 & .157 $\pm$ .001 & .166 $\pm$ .002 & .150 $\pm$ .001 & 21.592 $\pm$ 6.681 & .163 $\pm$ .003 & .162 $\pm$ .003 & .153 $\pm$ .003 & 3.846 $\pm$ 2.660 & .178 $\pm$ .005 & .178 $\pm$ .005 & .170 $\pm$ .005 & .463 $\pm$ .954 \\
 & \catoni & .165 $\pm$ .001 & .190 $\pm$ .003 & .159 $\pm$ .000 & 119.927 $\pm$ 15.938 & .157 $\pm$ .001 & .163 $\pm$ .002 & .150 $\pm$ .001 & 26.363 $\pm$ 7.355 & .162 $\pm$ .003 & .161 $\pm$ .003 & .152 $\pm$ .003 & 4.152 $\pm$ 2.945 & .177 $\pm$ .006 & .178 $\pm$ .006 & .169 $\pm$ .006 & .548 $\pm$ 1.032 \\
 & \rivasplata & .165 $\pm$ .001 & .183 $\pm$ .002 & .158 $\pm$ .000 & 101.954 $\pm$ 14.463 & .157 $\pm$ .001 & .163 $\pm$ .002 & .150 $\pm$ .001 & 23.098 $\pm$ 6.977 & .162 $\pm$ .003 & .161 $\pm$ .003 & .153 $\pm$ .003 & 3.852 $\pm$ 2.798 & .177 $\pm$ .006 & .177 $\pm$ .006 & .169 $\pm$ .006 & .516 $\pm$ .985 \\
 & \stoNN & \textemdash & .227 & \textemdash & .290 & \textemdash & .216 & \textemdash & 1.064 & \textemdash & .218 & \textemdash & 1.751 & \textemdash & .237 & \textemdash & .634 \\
\midrule
\multirow[c]{5}{*}{\rotatebox[origin=c]{90}{\small{CIFAR-10}}} & \ours & .479 $\pm$ .000 & .487 $\pm$ .000 & .472 $\pm$ .000 & .052 & .479 $\pm$ .000 & .493 $\pm$ .000 & .477 $\pm$ .000 & .065 & .458 $\pm$ .001 & .479 $\pm$ .000 & .463 $\pm$ .000 & .299 & .480 $\pm$ .002 & .495 $\pm$ .001 & .480 $\pm$ .001 & .793 \\
 & \blanchard & .479 $\pm$ .000 & .550 $\pm$ .003 & .472 $\pm$ .000 & 27.644 $\pm$ 22.868 & .479 $\pm$ .000 & .522 $\pm$ .003 & .477 $\pm$ .000 & 85.476 $\pm$ 12.781 & .458 $\pm$ .001 & .489 $\pm$ .003 & .463 $\pm$ .000 & 24.608 $\pm$ 7.136 & .481 $\pm$ .002 & .495 $\pm$ .002 & .480 $\pm$ .001 & 5.093 $\pm$ 3.299 \\
 & \catoni & .479 $\pm$ .000 & .546 $\pm$ .005 & .472 $\pm$ .000 & 269.855 $\pm$ 22.883 & .479 $\pm$ .000 & .511 $\pm$ .003 & .477 $\pm$ .000 & 85.113 $\pm$ 12.806 & .458 $\pm$ .001 & .483 $\pm$ .002 & .463 $\pm$ .000 & 25.453 $\pm$ 7.155 & .480 $\pm$ .002 & .495 $\pm$ .001 & .480 $\pm$ .001 & 5.468 $\pm$ 3.315 \\
 & \rivasplata & .479 $\pm$ .000 & .528 $\pm$ .002 & .472 $\pm$ .000 & 27.588 $\pm$ 22.859 & .479 $\pm$ .000 & .511 $\pm$ .002 & .477 $\pm$ .000 & 85.745 $\pm$ 13.357 & .458 $\pm$ .001 & .484 $\pm$ .002 & .463 $\pm$ .001 & 25.051 $\pm$ 7.005 & .481 $\pm$ .002 & .494 $\pm$ .001 & .480 $\pm$ .001 & 5.155 $\pm$ 3.260 \\
 & \stoNN & \textemdash & .558 & \textemdash & .026 & \textemdash & .564 & \textemdash & .032 & \textemdash & .550 & \textemdash & .150 & \textemdash & .566 & \textemdash & .397 \\
\midrule
 &  & \multicolumn{4}{c}{$\sigma^2=10^{-6}$} & \multicolumn{4}{c}{$\sigma^2=10^{-5}$} & \multicolumn{4}{c}{$\sigma^2=10^{-4}$} & \multicolumn{4}{c}{$\sigma^2=10^{-3}$} \\
\midrule
 & lr=$10^{-4}$ & $\Risk_{\Tcal}(h)$ & Bnd & $\Risk_{\Scal}(h)$ & Div & $\Risk_{\Tcal}(h)$ & Bnd & $\Risk_{\Scal}(h)$ & Div & $\Risk_{\Tcal}(h)$ & Bnd & $\Risk_{\Scal}(h)$ & Div & $\Risk_{\Tcal}(h)$ & Bnd & $\Risk_{\Scal}(h)$ & Div \\
\midrule
\multirow[c]{5}{*}{\rotatebox[origin=c]{90}{\small{MNIST}}} & \ours & .035 $\pm$ .000 & .048 $\pm$ .000 & .039 $\pm$ .000 & 35.348 & .024 $\pm$ .000 & .037 $\pm$ .001 & .029 $\pm$ .000 & 3.753 & .022 $\pm$ .001 & .042 $\pm$ .001 & .027 $\pm$ .001 & 153.773 & .025 $\pm$ .002 & .041 $\pm$ .002 & .029 $\pm$ .002 & 97.840 \\
 & \blanchard & .032 $\pm$ .000 & .442 $\pm$ .003 & .036 $\pm$ .000 & 1181.482 $\pm$ 14.449 & .022 $\pm$ .000 & .206 $\pm$ .003 & .027 $\pm$ .000 & 3851.110 $\pm$ 84.274 & .019 $\pm$ .001 & .102 $\pm$ .002 & .023 $\pm$ .001 & 1306.371 $\pm$ 51.396 & .024 $\pm$ .002 & .065 $\pm$ .003 & .027 $\pm$ .002 & 411.772 $\pm$ 29.458 \\
 & \catoni & .035 $\pm$ .000 & .362 $\pm$ .003 & .039 $\pm$ .000 & 11925.734 $\pm$ 145.511 & .024 $\pm$ .000 & .152 $\pm$ .002 & .029 $\pm$ .000 & 3841.248 $\pm$ 84.033 & .027 $\pm$ .002 & .084 $\pm$ .002 & .032 $\pm$ .001 & 1235.287 $\pm$ 49.454 & .027 $\pm$ .002 & .059 $\pm$ .003 & .030 $\pm$ .002 & 403.300 $\pm$ 28.587 \\
 & \rivasplata & .030 $\pm$ .000 & .289 $\pm$ .002 & .034 $\pm$ .000 & 12022.576 $\pm$ 151.157 & .021 $\pm$ .000 & .134 $\pm$ .002 & .026 $\pm$ .000 & 3912.803 $\pm$ 85.146 & .018 $\pm$ .000 & .072 $\pm$ .001 & .022 $\pm$ .000 & 1348.169 $\pm$ 53.400 & .023 $\pm$ .002 & .051 $\pm$ .002 & .026 $\pm$ .001 & 424.971 $\pm$ 29.301 \\
 & \stoNN & \textemdash & .084 & \textemdash & 17.674 & \textemdash & .069 & \textemdash & 15.376 & \textemdash & .072 & \textemdash & 76.887 & \textemdash & .072 & \textemdash & 48.920 \\
\midrule
\multirow[c]{5}{*}{\rotatebox[origin=c]{90}{\small{Fashion}}} & \ours & .166 $\pm$ .001 & .172 $\pm$ .000 & .159 $\pm$ .000 & 13.084 & .157 $\pm$ .001 & .163 $\pm$ .001 & .150 $\pm$ .001 & 16.513 & .159 $\pm$ .002 & .164 $\pm$ .002 & .149 $\pm$ .002 & 2.344 & .176 $\pm$ .005 & .181 $\pm$ .005 & .168 $\pm$ .005 & 11.331 \\
 & \blanchard & .160 $\pm$ .001 & .588 $\pm$ .003 & .153 $\pm$ .000 & 1089.829 $\pm$ 137.125 & .150 $\pm$ .001 & .379 $\pm$ .003 & .141 $\pm$ .001 & 3744.491 $\pm$ 83.656 & .155 $\pm$ .002 & .271 $\pm$ .003 & .145 $\pm$ .002 & 1221.062 $\pm$ 49.548 & .173 $\pm$ .005 & .233 $\pm$ .006 & .165 $\pm$ .004 & 369.721 $\pm$ 27.211 \\
 & \catoni & .165 $\pm$ .001 & .500 $\pm$ .003 & .159 $\pm$ .000 & 11954.591 $\pm$ 141.463 & .156 $\pm$ .001 & .311 $\pm$ .002 & .148 $\pm$ .001 & 3826.848 $\pm$ 86.111 & .158 $\pm$ .002 & .248 $\pm$ .003 & .148 $\pm$ .002 & 1226.282 $\pm$ 5.332 & .174 $\pm$ .005 & .252 $\pm$ .006 & .166 $\pm$ .004 & 393.542 $\pm$ 27.890 \\
 & \rivasplata & .158 $\pm$ .001 & .459 $\pm$ .002 & .151 $\pm$ .000 & 11541.128 $\pm$ 14.706 & .149 $\pm$ .001 & .302 $\pm$ .002 & .140 $\pm$ .001 & 3878.145 $\pm$ 85.782 & .154 $\pm$ .002 & .230 $\pm$ .002 & .144 $\pm$ .001 & 1244.035 $\pm$ 49.268 & .172 $\pm$ .005 & .212 $\pm$ .005 & .164 $\pm$ .004 & 378.990 $\pm$ 27.559 \\
 & \stoNN & \textemdash & .229 & \textemdash & 6.542 & \textemdash & .219 & \textemdash & 8.257 & \textemdash & .219 & \textemdash & 1.172 & \textemdash & .239 & \textemdash & 5.666 \\
\midrule
\multirow[c]{5}{*}{\rotatebox[origin=c]{90}{\small{CIFAR-10}}} & \ours & .479 $\pm$ .000 & .489 $\pm$ .000 & .472 $\pm$ .000 & 4.882 & .479 $\pm$ .000 & .496 $\pm$ .000 & .477 $\pm$ .000 & 9.273 & .458 $\pm$ .001 & .480 $\pm$ .000 & .463 $\pm$ .000 & 4.988 & .480 $\pm$ .002 & .497 $\pm$ .001 & .479 $\pm$ .001 & 8.681 \\
 & \blanchard & .479 $\pm$ .000 & .957 $\pm$ .001 & .471 $\pm$ .000 & 22201.935 $\pm$ 218.369 & .479 $\pm$ .000 & .854 $\pm$ .002 & .477 $\pm$ .000 & 8777.551 $\pm$ 125.716 & .457 $\pm$ .001 & .699 $\pm$ .003 & .461 $\pm$ .000 & 2758.075 $\pm$ 77.155 & .474 $\pm$ .001 & .613 $\pm$ .003 & .472 $\pm$ .001 & 903.948 $\pm$ 4.742 \\
 & \catoni & .479 $\pm$ .000 & .995 $\pm$ .000 & .471 $\pm$ .000 & 26347.736 $\pm$ 225.908 & .479 $\pm$ .000 & .771 $\pm$ .002 & .477 $\pm$ .000 & 8566.272 $\pm$ 124.834 & .455 $\pm$ .001 & .650 $\pm$ .002 & .459 $\pm$ .000 & 3117.566 $\pm$ 75.178 & .468 $\pm$ .001 & .621 $\pm$ .001 & .466 $\pm$ .001 & 1481.520 $\pm$ 52.533 \\
 & \rivasplata & .479 $\pm$ .000 & .915 $\pm$ .001 & .471 $\pm$ .000 & 29489.241 $\pm$ 241.010 & .479 $\pm$ .000 & .765 $\pm$ .002 & .477 $\pm$ .000 & 867.264 $\pm$ 126.038 & .456 $\pm$ .001 & .633 $\pm$ .002 & .460 $\pm$ .000 & 2776.052 $\pm$ 72.901 & .472 $\pm$ .001 & .572 $\pm$ .002 & .470 $\pm$ .001 & 937.091 $\pm$ 42.116 \\
 & \stoNN & \textemdash & .559 & \textemdash & 2.441 & \textemdash & .566 & \textemdash & 4.637 & \textemdash & .551 & \textemdash & 2.494 & \textemdash & .567 & \textemdash & 4.340 \\
\bottomrule
\end{tabular}
}
\label{table:1_prior_0.1}
\end{sidewaystable}

\begin{sidewaystable}
\caption{
\looseness=-1
Comparison of \ours, \rivasplata, \blanchard and \catoni based on the disintegrated bounds, and \stoNN based on the randomized bounds learned with two learning rates lr${\ \in}\{10^{-4}, 10^{-6}\}$ and different variances $\sigma^2{\in}\{10^{-3}, 10^{-4}, 10^{-5}, 10^{-6}\}$.
We report the test risk ($\Risk_{\Tcal}(h)$), the bound value (Bnd), the empirical risk ($\Risk_{\Scal}(h)$), and the divergence (Div) associated with each bound (the Rényi divergence for \ours, the KL divergence for \stoNN, and the disintegrated KL divergence for \rivasplata, \blanchard and \catoni).
More precisely, we report the mean $\pm$ the standard deviation for $400$ neural networks sampled from $\AQ$ for \ours, \rivasplata, \blanchard, and \catoni.
We consider, in this table, that the split ratio is $0.2$.
}
\resizebox{0.83\paperheight}{!}{
\begin{tabular}{rr|clcl|clcl|clcl|clcl}
\toprule
 &  & \multicolumn{4}{c}{$\sigma^2=10^{-6}$} & \multicolumn{4}{c}{$\sigma^2=10^{-5}$} & \multicolumn{4}{c}{$\sigma^2=10^{-4}$} & \multicolumn{4}{c}{$\sigma^2=10^{-3}$} \\
\midrule
 & lr=$10^{-6}$ & $\Risk_{\Tcal}(h)$ & Bnd & $\Risk_{\Scal}(h)$ & Div & $\Risk_{\Tcal}(h)$ & Bnd & $\Risk_{\Scal}(h)$ & Div & $\Risk_{\Tcal}(h)$ & Bnd & $\Risk_{\Scal}(h)$ & Div & $\Risk_{\Tcal}(h)$ & Bnd & $\Risk_{\Scal}(h)$ & Div \\
\midrule
\multirow[c]{5}{*}{\rotatebox[origin=c]{90}{\small{MNIST}}} & \ours & .016 $\pm$ .000 & .023 $\pm$ .000 & .019 $\pm$ .000 & .336 & .015 $\pm$ .000 & .023 $\pm$ .000 & .019 $\pm$ .000 & .748 & .014 $\pm$ .001 & .020 $\pm$ .001 & .016 $\pm$ .000 & 2.096 & .019 $\pm$ .002 & .024 $\pm$ .002 & .020 $\pm$ .002 & 2.244 \\
 & \blanchard & .016 $\pm$ .000 & .034 $\pm$ .001 & .019 $\pm$ .000 & 97.590 $\pm$ 14.260 & .015 $\pm$ .000 & .026 $\pm$ .001 & .019 $\pm$ .000 & 21.153 $\pm$ 6.514 & .015 $\pm$ .001 & .020 $\pm$ .001 & .016 $\pm$ .001 & 3.362 $\pm$ 2.569 & .020 $\pm$ .002 & .024 $\pm$ .002 & .021 $\pm$ .002 & .371 $\pm$ .875 \\
 & \catoni & .016 $\pm$ .000 & .034 $\pm$ .001 & .019 $\pm$ .000 & 116.744 $\pm$ 15.447 & .015 $\pm$ .000 & .027 $\pm$ .002 & .019 $\pm$ .000 & 24.135 $\pm$ 7.075 & .015 $\pm$ .001 & .020 $\pm$ .001 & .016 $\pm$ .001 & 3.352 $\pm$ 2.667 & .020 $\pm$ .002 & .024 $\pm$ .002 & .021 $\pm$ .002 & .410 $\pm$ .890 \\
 & \rivasplata & .016 $\pm$ .000 & .030 $\pm$ .001 & .019 $\pm$ .000 & 101.334 $\pm$ 14.728 & .015 $\pm$ .000 & .024 $\pm$ .001 & .019 $\pm$ .000 & 21.663 $\pm$ 6.603 & .015 $\pm$ .001 & .020 $\pm$ .001 & .016 $\pm$ .001 & 3.409 $\pm$ 2.666 & .020 $\pm$ .002 & .024 $\pm$ .002 & .021 $\pm$ .002 & .446 $\pm$ .927 \\
 & \stoNN & \textemdash & .052 & \textemdash & .168 & \textemdash & .051 & \textemdash & .374 & \textemdash & .047 & \textemdash & 1.048 & \textemdash & .053 & \textemdash & 1.122 \\
\midrule
\multirow[c]{5}{*}{\rotatebox[origin=c]{90}{\small{Fashion}}} & \ours & .165 $\pm$ .002 & .169 $\pm$ .001 & .157 $\pm$ .001 & 4.811 & .148 $\pm$ .003 & .155 $\pm$ .002 & .143 $\pm$ .002 & 1.856 & .145 $\pm$ .005 & .153 $\pm$ .006 & .139 $\pm$ .005 & 15.453 & .160 $\pm$ .005 & .166 $\pm$ .005 & .155 $\pm$ .005 & 1.633 \\
 & \blanchard & .163 $\pm$ .002 & .190 $\pm$ .003 & .155 $\pm$ .001 & 96.264 $\pm$ 14.472 & .152 $\pm$ .003 & .163 $\pm$ .003 & .147 $\pm$ .003 & 21.099 $\pm$ 6.507 & .155 $\pm$ .007 & .160 $\pm$ .007 & .151 $\pm$ .007 & 3.929 $\pm$ 2.841 & .163 $\pm$ .006 & .165 $\pm$ .006 & .158 $\pm$ .006 & .340 $\pm$ .885 \\
 & \catoni & .163 $\pm$ .002 & .190 $\pm$ .004 & .156 $\pm$ .001 & 121.542 $\pm$ 16.499 & .150 $\pm$ .002 & .158 $\pm$ .003 & .144 $\pm$ .002 & 27.241 $\pm$ 7.318 & .151 $\pm$ .006 & .155 $\pm$ .006 & .146 $\pm$ .006 & 5.120 $\pm$ 3.150 & .162 $\pm$ .005 & .165 $\pm$ .005 & .157 $\pm$ .005 & .444 $\pm$ .968 \\
 & \rivasplata & .161 $\pm$ .001 & .180 $\pm$ .002 & .153 $\pm$ .001 & 106.403 $\pm$ 14.044 & .150 $\pm$ .002 & .158 $\pm$ .003 & .145 $\pm$ .003 & 23.134 $\pm$ 7.064 & .153 $\pm$ .006 & .157 $\pm$ .006 & .148 $\pm$ .007 & 4.439 $\pm$ 2.924 & .162 $\pm$ .006 & .165 $\pm$ .005 & .157 $\pm$ .005 & .417 $\pm$ .928 \\
 & \stoNN & \textemdash & .226 & \textemdash & 2.405 & \textemdash & .210 & \textemdash & 5.428 & \textemdash & .207 & \textemdash & 7.727 & \textemdash & .223 & \textemdash & .816 \\
\midrule
\multirow[c]{5}{*}{\rotatebox[origin=c]{90}{\small{CIFAR-10}}} & \ours & .390 $\pm$ .000 & .407 $\pm$ .000 & .391 $\pm$ .000 & .040 & .404 $\pm$ .000 & .414 $\pm$ .000 & .398 $\pm$ .000 & .070 & .396 $\pm$ .001 & .411 $\pm$ .000 & .395 $\pm$ .000 & .155 & .416 $\pm$ .002 & .432 $\pm$ .001 & .415 $\pm$ .001 & .970 \\
 & \blanchard & .390 $\pm$ .000 & .473 $\pm$ .004 & .391 $\pm$ .000 & 271.616 $\pm$ 23.555 & .404 $\pm$ .000 & .445 $\pm$ .003 & .398 $\pm$ .000 & 84.868 $\pm$ 13.050 & .396 $\pm$ .001 & .422 $\pm$ .003 & .395 $\pm$ .000 & 23.962 $\pm$ 7.208 & .416 $\pm$ .002 & .432 $\pm$ .002 & .416 $\pm$ .001 & 4.496 $\pm$ 3.018 \\
 & \catoni & .390 $\pm$ .000 & .473 $\pm$ .006 & .391 $\pm$ .000 & 27.502 $\pm$ 23.371 & .404 $\pm$ .000 & .434 $\pm$ .003 & .398 $\pm$ .000 & 84.848 $\pm$ 12.992 & .396 $\pm$ .001 & .415 $\pm$ .002 & .395 $\pm$ .000 & 24.505 $\pm$ 6.942 & .416 $\pm$ .002 & .431 $\pm$ .001 & .415 $\pm$ .001 & 4.859 $\pm$ 3.176 \\
 & \rivasplata & .390 $\pm$ .000 & .450 $\pm$ .002 & .391 $\pm$ .000 & 271.700 $\pm$ 23.586 & .403 $\pm$ .000 & .433 $\pm$ .002 & .398 $\pm$ .000 & 85.027 $\pm$ 13.047 & .396 $\pm$ .001 & .416 $\pm$ .002 & .395 $\pm$ .000 & 23.955 $\pm$ 7.093 & .416 $\pm$ .002 & .431 $\pm$ .002 & .416 $\pm$ .001 & 4.610 $\pm$ 3.084 \\
 & \stoNN & \textemdash & .477 & \textemdash & .020 & \textemdash & .485 & \textemdash & .035 & \textemdash & .482 & \textemdash & .077 & \textemdash & .503 & \textemdash & .485 \\
\midrule
 &  & \multicolumn{4}{c}{$\sigma^2=10^{-6}$} & \multicolumn{4}{c}{$\sigma^2=10^{-5}$} & \multicolumn{4}{c}{$\sigma^2=10^{-4}$} & \multicolumn{4}{c}{$\sigma^2=10^{-3}$} \\
\midrule
 & lr=$10^{-4}$ & $\Risk_{\Tcal}(h)$ & Bnd & $\Risk_{\Scal}(h)$ & Div & $\Risk_{\Tcal}(h)$ & Bnd & $\Risk_{\Scal}(h)$ & Div & $\Risk_{\Tcal}(h)$ & Bnd & $\Risk_{\Scal}(h)$ & Div & $\Risk_{\Tcal}(h)$ & Bnd & $\Risk_{\Scal}(h)$ & Div \\
\midrule
\multirow[c]{5}{*}{\rotatebox[origin=c]{90}{\small{MNIST}}} & \ours & .016 $\pm$ .000 & .025 $\pm$ .000 & .019 $\pm$ .000 & 14.490 & .015 $\pm$ .000 & .024 $\pm$ .000 & .019 $\pm$ .000 & 8.583 & .014 $\pm$ .000 & .021 $\pm$ .001 & .016 $\pm$ .000 & 13.055 & .016 $\pm$ .001 & .023 $\pm$ .001 & .017 $\pm$ .001 & 25.556 \\
 & \blanchard & .016 $\pm$ .000 & .430 $\pm$ .004 & .018 $\pm$ .000 & 11405.062 $\pm$ 153.554 & .014 $\pm$ .000 & .200 $\pm$ .003 & .018 $\pm$ .000 & 3799.912 $\pm$ 89.585 & .013 $\pm$ .000 & .086 $\pm$ .002 & .014 $\pm$ .000 & 1187.859 $\pm$ 48.700 & .015 $\pm$ .001 & .049 $\pm$ .002 & .016 $\pm$ .001 & 38.983 $\pm$ 27.857 \\
 & \catoni & .016 $\pm$ .000 & .355 $\pm$ .002 & .019 $\pm$ .000 & 11954.106 $\pm$ 15.709 & .015 $\pm$ .000 & .149 $\pm$ .003 & .019 $\pm$ .000 & 3828.342 $\pm$ 83.937 & .014 $\pm$ .001 & .064 $\pm$ .002 & .016 $\pm$ .001 & 1218.708 $\pm$ 48.514 & .017 $\pm$ .001 & .041 $\pm$ .002 & .018 $\pm$ .001 & 389.726 $\pm$ 29.076 \\
 & \rivasplata & .015 $\pm$ .000 & .272 $\pm$ .002 & .018 $\pm$ .000 & 1173.953 $\pm$ 149.364 & .013 $\pm$ .000 & .122 $\pm$ .002 & .017 $\pm$ .000 & 3691.345 $\pm$ 82.512 & .012 $\pm$ .000 & .056 $\pm$ .001 & .013 $\pm$ .000 & 1206.615 $\pm$ 5.381 & .015 $\pm$ .001 & .037 $\pm$ .001 & .015 $\pm$ .001 & 391.881 $\pm$ 28.344 \\
 & \stoNN & \textemdash & .053 & \textemdash & 7.245 & \textemdash & .052 & \textemdash & 4.292 & \textemdash & .048 & \textemdash & 6.528 & \textemdash & .051 & \textemdash & 12.778 \\
\midrule
\multirow[c]{5}{*}{\rotatebox[origin=c]{90}{\small{Fashion}}} & \ours & .165 $\pm$ .002 & .172 $\pm$ .001 & .157 $\pm$ .001 & 23.705 & .141 $\pm$ .002 & .156 $\pm$ .002 & .137 $\pm$ .002 & 52.736 & .131 $\pm$ .003 & .147 $\pm$ .003 & .126 $\pm$ .003 & 7.515 & .156 $\pm$ .004 & .165 $\pm$ .004 & .151 $\pm$ .003 & 16.954 \\
 & \blanchard & .136 $\pm$ .001 & .598 $\pm$ .003 & .130 $\pm$ .001 & 11334.327 $\pm$ 145.083 & .125 $\pm$ .001 & .379 $\pm$ .003 & .121 $\pm$ .001 & 3998.068 $\pm$ 88.992 & .124 $\pm$ .001 & .247 $\pm$ .003 & .117 $\pm$ .001 & 126.184 $\pm$ 48.814 & .152 $\pm$ .003 & .216 $\pm$ .004 & .147 $\pm$ .003 & 364.531 $\pm$ 28.029 \\
 & \catoni & .162 $\pm$ .001 & .525 $\pm$ .004 & .154 $\pm$ .001 & 11965.668 $\pm$ 152.681 & .141 $\pm$ .002 & .309 $\pm$ .003 & .137 $\pm$ .002 & 384.802 $\pm$ 84.123 & .132 $\pm$ .003 & .224 $\pm$ .004 & .127 $\pm$ .002 & 1239.918 $\pm$ 49.594 & .155 $\pm$ .004 & .232 $\pm$ .005 & .150 $\pm$ .004 & 394.607 $\pm$ 28.146 \\
 & \rivasplata & .131 $\pm$ .001 & .455 $\pm$ .002 & .127 $\pm$ .001 & 1193.209 $\pm$ 155.390 & .123 $\pm$ .001 & .290 $\pm$ .002 & .119 $\pm$ .001 & 4005.169 $\pm$ 89.793 & .123 $\pm$ .001 & .204 $\pm$ .002 & .116 $\pm$ .001 & 1294.726 $\pm$ 49.874 & .152 $\pm$ .004 & .195 $\pm$ .004 & .146 $\pm$ .003 & 378.905 $\pm$ 27.422 \\
 & \stoNN & \textemdash & .228 & \textemdash & 11.853 & \textemdash & .209 & \textemdash & 26.368 & \textemdash & .198 & \textemdash & 35.258 & \textemdash & .221 & \textemdash & 8.477 \\
\midrule
\multirow[c]{5}{*}{\rotatebox[origin=c]{90}{\small{CIFAR-10}}} & \ours & .390 $\pm$ .000 & .411 $\pm$ .000 & .391 $\pm$ .000 & 13.286 & .404 $\pm$ .000 & .415 $\pm$ .000 & .398 $\pm$ .000 & 3.305 & .396 $\pm$ .001 & .412 $\pm$ .000 & .395 $\pm$ .000 & 3.136 & .415 $\pm$ .001 & .433 $\pm$ .001 & .415 $\pm$ .001 & 6.064 \\
 & \blanchard & .389 $\pm$ .000 & .990 $\pm$ .000 & .391 $\pm$ .000 & 75424.764 $\pm$ 397.521 & .403 $\pm$ .000 & .820 $\pm$ .002 & .397 $\pm$ .000 & 8815.324 $\pm$ 126.764 & .395 $\pm$ .001 & .651 $\pm$ .003 & .394 $\pm$ .000 & 2738.066 $\pm$ 75.053 & .408 $\pm$ .001 & .557 $\pm$ .003 & .405 $\pm$ .001 & 918.500 $\pm$ 42.347 \\
 & \catoni & .390 $\pm$ .000 & .990 $\pm$ .000 & .391 $\pm$ .000 & 26434.787 $\pm$ 228.500 & .403 $\pm$ .000 & .726 $\pm$ .003 & .397 $\pm$ .000 & 8651.380 $\pm$ 126.473 & .394 $\pm$ .001 & .620 $\pm$ .002 & .393 $\pm$ .000 & 4178.302 $\pm$ 9.315 & .401 $\pm$ .001 & .556 $\pm$ .001 & .396 $\pm$ .001 & 1462.235 $\pm$ 55.526 \\
 & \rivasplata & .389 $\pm$ .000 & .902 $\pm$ .001 & .391 $\pm$ .000 & 31497.669 $\pm$ 249.683 & .403 $\pm$ .000 & .715 $\pm$ .002 & .397 $\pm$ .000 & 8707.893 $\pm$ 133.239 & .394 $\pm$ .001 & .578 $\pm$ .003 & .393 $\pm$ .000 & 2741.257 $\pm$ 74.942 & .405 $\pm$ .001 & .512 $\pm$ .002 & .402 $\pm$ .001 & 967.818 $\pm$ 43.629 \\
 & \stoNN & \textemdash & .480 & \textemdash & 6.643 & \textemdash & .486 & \textemdash & 1.653 & \textemdash & .483 & \textemdash & 1.568 & \textemdash & .503 & \textemdash & 3.032 \\
\bottomrule
\end{tabular}
}
\label{table:1_prior_0.2}
\end{sidewaystable}

\begin{sidewaystable}
\caption{
\looseness=-1
Comparison of \ours, \rivasplata, \blanchard and \catoni based on the disintegrated bounds, and \stoNN based on the randomized bounds learned with two learning rates lr${\ \in}\{10^{-4}, 10^{-6}\}$ and different variances $\sigma^2{\in}\{10^{-3}, 10^{-4}, 10^{-5}, 10^{-6}\}$.
We report the test risk ($\Risk_{\Tcal}(h)$), the bound value (Bnd), the empirical risk ($\Risk_{\Scal}(h)$), and the divergence (Div) associated with each bound (the Rényi divergence for \ours, the KL divergence for \stoNN, and the disintegrated KL divergence for \rivasplata, \blanchard and \catoni).
More precisely, we report the mean $\pm$ the standard deviation for $400$ neural networks sampled from $\AQ$ for \ours, \rivasplata, \blanchard, and \catoni.
We consider, in this table, that the split ratio is $0.3$.
}
\resizebox{0.83\paperheight}{!}{
\begin{tabular}{rr|clcl|clcl|clcl|clcl}
\toprule
 &  & \multicolumn{4}{c}{$\sigma^2=10^{-6}$} & \multicolumn{4}{c}{$\sigma^2=10^{-5}$} & \multicolumn{4}{c}{$\sigma^2=10^{-4}$} & \multicolumn{4}{c}{$\sigma^2=10^{-3}$} \\
\midrule
 & lr=$10^{-6}$ & $\Risk_{\Tcal}(h)$ & Bnd & $\Risk_{\Scal}(h)$ & Div & $\Risk_{\Tcal}(h)$ & Bnd & $\Risk_{\Scal}(h)$ & Div & $\Risk_{\Tcal}(h)$ & Bnd & $\Risk_{\Scal}(h)$ & Div & $\Risk_{\Tcal}(h)$ & Bnd & $\Risk_{\Scal}(h)$ & Div \\
\midrule
\multirow[c]{5}{*}{\rotatebox[origin=c]{90}{\small{MNIST}}} & \ours & .012 $\pm$ .000 & .017 $\pm$ .000 & .013 $\pm$ .000 & .181 & .009 $\pm$ .000 & .015 $\pm$ .000 & .011 $\pm$ .000 & .155 & .012 $\pm$ .000 & .020 $\pm$ .000 & .016 $\pm$ .000 & 1.655 & .013 $\pm$ .001 & .019 $\pm$ .001 & .015 $\pm$ .001 & .615 \\
 & \blanchard & .012 $\pm$ .000 & .027 $\pm$ .001 & .013 $\pm$ .000 & 93.915 $\pm$ 14.109 & .012 $\pm$ .000 & .021 $\pm$ .001 & .014 $\pm$ .000 & 19.292 $\pm$ 6.037 & .012 $\pm$ .000 & .020 $\pm$ .001 & .016 $\pm$ .000 & 3.023 $\pm$ 2.430 & .014 $\pm$ .001 & .018 $\pm$ .001 & .015 $\pm$ .001 & .368 $\pm$ .831 \\
 & \catoni & .012 $\pm$ .000 & .025 $\pm$ .001 & .013 $\pm$ .000 & 113.574 $\pm$ 15.436 & .012 $\pm$ .000 & .023 $\pm$ .002 & .014 $\pm$ .000 & 22.347 $\pm$ 6.877 & .012 $\pm$ .000 & .020 $\pm$ .001 & .016 $\pm$ .000 & 2.918 $\pm$ 2.341 & .013 $\pm$ .001 & .018 $\pm$ .001 & .015 $\pm$ .001 & .336 $\pm$ .807 \\
 & \rivasplata & .012 $\pm$ .000 & .023 $\pm$ .001 & .013 $\pm$ .000 & 96.392 $\pm$ 14.300 & .012 $\pm$ .000 & .020 $\pm$ .001 & .014 $\pm$ .000 & 19.905 $\pm$ 6.254 & .012 $\pm$ .000 & .020 $\pm$ .001 & .016 $\pm$ .000 & 2.931 $\pm$ 2.446 & .013 $\pm$ .001 & .018 $\pm$ .001 & .015 $\pm$ .001 & .355 $\pm$ .813 \\
 & \stoNN & \textemdash & .042 & \textemdash & .091 & \textemdash & .039 & \textemdash & .077 & \textemdash & .047 & \textemdash & .827 & \textemdash & .045 & \textemdash & .308 \\
\midrule
\multirow[c]{5}{*}{\rotatebox[origin=c]{90}{\small{Fashion}}} & \ours & .126 $\pm$ .000 & .134 $\pm$ .000 & .124 $\pm$ .000 & .328 & .126 $\pm$ .001 & .130 $\pm$ .001 & .119 $\pm$ .001 & 1.692 & .122 $\pm$ .002 & .126 $\pm$ .002 & .115 $\pm$ .002 & 4.617 & .139 $\pm$ .005 & .145 $\pm$ .005 & .133 $\pm$ .005 & 2.425 \\
 & \blanchard & .126 $\pm$ .000 & .157 $\pm$ .003 & .124 $\pm$ .000 & 88.034 $\pm$ 13.485 & .126 $\pm$ .001 & .136 $\pm$ .002 & .120 $\pm$ .001 & 18.852 $\pm$ 6.115 & .124 $\pm$ .002 & .127 $\pm$ .002 & .118 $\pm$ .002 & 3.014 $\pm$ 2.395 & .142 $\pm$ .006 & .144 $\pm$ .006 & .137 $\pm$ .006 & .370 $\pm$ .819 \\
 & \catoni & .126 $\pm$ .000 & .159 $\pm$ .004 & .124 $\pm$ .000 & 114.259 $\pm$ 15.300 & .126 $\pm$ .001 & .133 $\pm$ .002 & .120 $\pm$ .001 & 22.607 $\pm$ 6.871 & .124 $\pm$ .002 & .126 $\pm$ .002 & .118 $\pm$ .002 & 3.100 $\pm$ 2.513 & .141 $\pm$ .006 & .144 $\pm$ .006 & .136 $\pm$ .006 & .390 $\pm$ .898 \\
 & \rivasplata & .126 $\pm$ .000 & .148 $\pm$ .002 & .124 $\pm$ .000 & 93.107 $\pm$ 13.630 & .126 $\pm$ .001 & .133 $\pm$ .002 & .120 $\pm$ .001 & 19.724 $\pm$ 6.320 & .124 $\pm$ .002 & .126 $\pm$ .002 & .118 $\pm$ .002 & 2.980 $\pm$ 2.451 & .142 $\pm$ .006 & .144 $\pm$ .006 & .136 $\pm$ .006 & .371 $\pm$ .869 \\
 & \stoNN & \textemdash & .187 & \textemdash & .164 & \textemdash & .182 & \textemdash & .846 & \textemdash & .178 & \textemdash & 2.309 & \textemdash & .199 & \textemdash & 1.212 \\
\midrule
\multirow[c]{5}{*}{\rotatebox[origin=c]{90}{\small{CIFAR-10}}} & \ours & .369 $\pm$ .000 & .375 $\pm$ .000 & .358 $\pm$ .000 & .028 & .351 $\pm$ .000 & .368 $\pm$ .000 & .352 $\pm$ .000 & .041 & .359 $\pm$ .001 & .377 $\pm$ .000 & .360 $\pm$ .000 & .183 & .419 $\pm$ .001 & .433 $\pm$ .001 & .416 $\pm$ .001 & .759 \\
 & \blanchard & .369 $\pm$ .000 & .446 $\pm$ .004 & .358 $\pm$ .000 & 269.789 $\pm$ 22.724 & .351 $\pm$ .000 & .401 $\pm$ .004 & .352 $\pm$ .000 & 84.113 $\pm$ 12.530 & .359 $\pm$ .001 & .388 $\pm$ .003 & .360 $\pm$ .000 & 22.878 $\pm$ 6.728 & .419 $\pm$ .001 & .432 $\pm$ .003 & .416 $\pm$ .001 & 4.089 $\pm$ 2.818 \\
 & \catoni & .369 $\pm$ .000 & .450 $\pm$ .007 & .358 $\pm$ .000 & 269.843 $\pm$ 24.225 & .351 $\pm$ .000 & .390 $\pm$ .004 & .352 $\pm$ .000 & 84.500 $\pm$ 12.608 & .359 $\pm$ .001 & .381 $\pm$ .002 & .360 $\pm$ .000 & 23.567 $\pm$ 7.181 & .419 $\pm$ .001 & .432 $\pm$ .001 & .416 $\pm$ .001 & 4.285 $\pm$ 2.942 \\
 & \rivasplata & .369 $\pm$ .000 & .421 $\pm$ .003 & .358 $\pm$ .000 & 27.224 $\pm$ 24.187 & .351 $\pm$ .000 & .388 $\pm$ .002 & .352 $\pm$ .000 & 84.250 $\pm$ 13.274 & .359 $\pm$ .001 & .382 $\pm$ .002 & .360 $\pm$ .000 & 23.053 $\pm$ 6.724 & .419 $\pm$ .001 & .431 $\pm$ .002 & .416 $\pm$ .001 & 4.141 $\pm$ 2.985 \\
 & \stoNN & \textemdash & .445 & \textemdash & .014 & \textemdash & .438 & \textemdash & .020 & \textemdash & .447 & \textemdash & .092 & \textemdash & .504 & \textemdash & .380 \\
\midrule
 &  & \multicolumn{4}{c}{$\sigma^2=10^{-6}$} & \multicolumn{4}{c}{$\sigma^2=10^{-5}$} & \multicolumn{4}{c}{$\sigma^2=10^{-4}$} & \multicolumn{4}{c}{$\sigma^2=10^{-3}$} \\
\midrule
 & lr=$10^{-4}$ & $\Risk_{\Tcal}(h)$ & Bnd & $\Risk_{\Scal}(h)$ & Div & $\Risk_{\Tcal}(h)$ & Bnd & $\Risk_{\Scal}(h)$ & Div & $\Risk_{\Tcal}(h)$ & Bnd & $\Risk_{\Scal}(h)$ & Div & $\Risk_{\Tcal}(h)$ & Bnd & $\Risk_{\Scal}(h)$ & Div \\
\midrule
\multirow[c]{5}{*}{\rotatebox[origin=c]{90}{\small{MNIST}}} & \ours & .012 $\pm$ .000 & .019 $\pm$ .000 & .013 $\pm$ .000 & 24.837 & .012 $\pm$ .000 & .020 $\pm$ .000 & .014 $\pm$ .000 & 12.358 & .012 $\pm$ .000 & .021 $\pm$ .000 & .015 $\pm$ .000 & 13.908 & .013 $\pm$ .001 & .019 $\pm$ .001 & .014 $\pm$ .001 & 16.179 \\
 & \blanchard & .012 $\pm$ .000 & .467 $\pm$ .004 & .013 $\pm$ .000 & 11819.223 $\pm$ 154.992 & .011 $\pm$ .000 & .211 $\pm$ .003 & .014 $\pm$ .000 & 3808.981 $\pm$ 86.014 & .010 $\pm$ .000 & .094 $\pm$ .003 & .014 $\pm$ .000 & 121.397 $\pm$ 51.944 & .012 $\pm$ .001 & .046 $\pm$ .002 & .013 $\pm$ .001 & 372.832 $\pm$ 26.602 \\
 & \catoni & .012 $\pm$ .000 & .339 $\pm$ .002 & .013 $\pm$ .000 & 1196.394 $\pm$ 15.704 & .012 $\pm$ .000 & .159 $\pm$ .003 & .014 $\pm$ .000 & 3838.459 $\pm$ 88.155 & .012 $\pm$ .000 & .070 $\pm$ .002 & .016 $\pm$ .000 & 1218.505 $\pm$ 51.783 & .013 $\pm$ .001 & .037 $\pm$ .001 & .014 $\pm$ .001 & 386.824 $\pm$ 28.233 \\
 & \rivasplata & .012 $\pm$ .000 & .289 $\pm$ .003 & .013 $\pm$ .000 & 1191.037 $\pm$ 152.759 & .011 $\pm$ .000 & .128 $\pm$ .002 & .014 $\pm$ .000 & 3768.785 $\pm$ 9.947 & .010 $\pm$ .000 & .061 $\pm$ .001 & .013 $\pm$ .000 & 1231.638 $\pm$ 49.362 & .011 $\pm$ .001 & .033 $\pm$ .001 & .012 $\pm$ .000 & 382.225 $\pm$ 28.481 \\
 & \stoNN & \textemdash & .044 & \textemdash & 12.418 & \textemdash & .046 & \textemdash & 6.179 & \textemdash & .047 & \textemdash & 6.954 & \textemdash & .045 & \textemdash & 8.089 \\
\midrule
\multirow[c]{5}{*}{\rotatebox[origin=c]{90}{\small{Fashion}}} & \ours & .126 $\pm$ .000 & .137 $\pm$ .000 & .124 $\pm$ .000 & 12.401 & .125 $\pm$ .001 & .132 $\pm$ .001 & .119 $\pm$ .001 & 14.631 & .120 $\pm$ .002 & .128 $\pm$ .002 & .113 $\pm$ .001 & 26.499 & .133 $\pm$ .003 & .143 $\pm$ .003 & .127 $\pm$ .003 & 23.702 \\
 & \blanchard & .123 $\pm$ .000 & .602 $\pm$ .003 & .121 $\pm$ .000 & 10558.872 $\pm$ 139.107 & .119 $\pm$ .001 & .383 $\pm$ .004 & .112 $\pm$ .001 & 3893.091 $\pm$ 86.176 & .113 $\pm$ .001 & .239 $\pm$ .003 & .106 $\pm$ .001 & 1204.211 $\pm$ 5.815 & .132 $\pm$ .003 & .195 $\pm$ .004 & .125 $\pm$ .003 & 362.146 $\pm$ 27.801 \\
 & \catoni & .126 $\pm$ .000 & .531 $\pm$ .004 & .124 $\pm$ .000 & 11966.223 $\pm$ 148.195 & .125 $\pm$ .001 & .299 $\pm$ .003 & .118 $\pm$ .001 & 3829.806 $\pm$ 85.864 & .119 $\pm$ .002 & .209 $\pm$ .002 & .113 $\pm$ .001 & 1225.310 $\pm$ 48.090 & .134 $\pm$ .004 & .202 $\pm$ .005 & .127 $\pm$ .003 & 395.243 $\pm$ 29.182 \\
 & \rivasplata & .123 $\pm$ .000 & .458 $\pm$ .003 & .120 $\pm$ .000 & 11209.156 $\pm$ 143.319 & .118 $\pm$ .001 & .287 $\pm$ .002 & .111 $\pm$ .001 & 3815.804 $\pm$ 85.091 & .112 $\pm$ .001 & .196 $\pm$ .002 & .105 $\pm$ .001 & 126.956 $\pm$ 49.255 & .130 $\pm$ .003 & .173 $\pm$ .004 & .124 $\pm$ .003 & 376.904 $\pm$ 27.549 \\
 & \stoNN & \textemdash & .189 & \textemdash & 6.200 & \textemdash & .184 & \textemdash & 7.316 & \textemdash & .179 & \textemdash & 13.250 & \textemdash & .195 & \textemdash & 11.851 \\
\midrule
\multirow[c]{5}{*}{\rotatebox[origin=c]{90}{\small{CIFAR-10}}} & \ours & .369 $\pm$ .000 & .379 $\pm$ .000 & .358 $\pm$ .000 & 11.657 & .351 $\pm$ .000 & .369 $\pm$ .000 & .352 $\pm$ .000 & 2.267 & .359 $\pm$ .001 & .378 $\pm$ .000 & .360 $\pm$ .000 & 2.616 & .418 $\pm$ .001 & .434 $\pm$ .001 & .415 $\pm$ .001 & 5.675 \\
 & \blanchard & .369 $\pm$ .000 & .990 $\pm$ .000 & .358 $\pm$ .000 & 40152.974 $\pm$ 291.721 & .351 $\pm$ .000 & .809 $\pm$ .003 & .351 $\pm$ .000 & 8753.816 $\pm$ 136.801 & .358 $\pm$ .001 & .635 $\pm$ .004 & .359 $\pm$ .000 & 2728.436 $\pm$ 73.835 & .412 $\pm$ .001 & .568 $\pm$ .004 & .407 $\pm$ .001 & 91.026 $\pm$ 44.096 \\
 & \catoni & .369 $\pm$ .000 & .986 $\pm$ .000 & .358 $\pm$ .000 & 24477.984 $\pm$ 223.367 & .351 $\pm$ .000 & .708 $\pm$ .003 & .351 $\pm$ .000 & 8463.452 $\pm$ 135.001 & .357 $\pm$ .001 & .578 $\pm$ .002 & .357 $\pm$ .000 & 3401.221 $\pm$ 84.878 & .405 $\pm$ .001 & .561 $\pm$ .002 & .399 $\pm$ .001 & 1354.100 $\pm$ 51.315 \\
 & \rivasplata & .369 $\pm$ .000 & .868 $\pm$ .001 & .358 $\pm$ .000 & 24424.968 $\pm$ 223.601 & .351 $\pm$ .000 & .694 $\pm$ .002 & .351 $\pm$ .000 & 8665.339 $\pm$ 136.361 & .358 $\pm$ .001 & .555 $\pm$ .003 & .358 $\pm$ .000 & 274.651 $\pm$ 74.784 & .409 $\pm$ .001 & .521 $\pm$ .003 & .403 $\pm$ .001 & 955.211 $\pm$ 44.609 \\
 & \stoNN & \textemdash & .448 & \textemdash & 5.829 & \textemdash & .439 & \textemdash & 1.134 & \textemdash & .448 & \textemdash & 1.308 & \textemdash & .504 & \textemdash & 2.838 \\
\bottomrule
\end{tabular}
}
\label{table:1_prior_0.3}
\end{sidewaystable}

\begin{sidewaystable}
\caption{
\looseness=-1
Comparison of \ours, \rivasplata, \blanchard and \catoni based on the disintegrated bounds, and \stoNN based on the randomized bounds learned with two learning rates lr${\ \in}\{10^{-4}, 10^{-6}\}$ and different variances $\sigma^2{\in}\{10^{-3}, 10^{-4}, 10^{-5}, 10^{-6}\}$.
We report the test risk ($\Risk_{\Tcal}(h)$), the bound value (Bnd), the empirical risk ($\Risk_{\Scal}(h)$), and the divergence (Div) associated with each bound (the Rényi divergence for \ours, the KL divergence for \stoNN, and the disintegrated KL divergence for \rivasplata, \blanchard and \catoni).
More precisely, we report the mean $\pm$ the standard deviation for $400$ neural networks sampled from $\AQ$ for \ours, \rivasplata, \blanchard, and \catoni.
We consider, in this table, that the split ratio is $0.4$.
}
\resizebox{0.83\paperheight}{!}{
\begin{tabular}{rr|clcl|clcl|clcl|clcl}
\toprule
 &  & \multicolumn{4}{c}{$\sigma^2=10^{-6}$} & \multicolumn{4}{c}{$\sigma^2=10^{-5}$} & \multicolumn{4}{c}{$\sigma^2=10^{-4}$} & \multicolumn{4}{c}{$\sigma^2=10^{-3}$} \\
\midrule
 & lr=$10^{-6}$ & $\Risk_{\Tcal}(h)$ & Bnd & $\Risk_{\Scal}(h)$ & Div & $\Risk_{\Tcal}(h)$ & Bnd & $\Risk_{\Scal}(h)$ & Div & $\Risk_{\Tcal}(h)$ & Bnd & $\Risk_{\Scal}(h)$ & Div & $\Risk_{\Tcal}(h)$ & Bnd & $\Risk_{\Scal}(h)$ & Div \\
\midrule
\multirow[c]{5}{*}{\rotatebox[origin=c]{90}{\small{MNIST}}} & \ours & .010 $\pm$ .000 & .017 $\pm$ .000 & .013 $\pm$ .000 & .194 & .012 $\pm$ .000 & .018 $\pm$ .000 & .014 $\pm$ .000 & .138 & .009 $\pm$ .000 & .015 $\pm$ .000 & .011 $\pm$ .000 & .235 & .014 $\pm$ .001 & .020 $\pm$ .001 & .015 $\pm$ .001 & 1.111 \\
 & \blanchard & .010 $\pm$ .000 & .028 $\pm$ .001 & .013 $\pm$ .000 & 88.323 $\pm$ 13.740 & .012 $\pm$ .000 & .021 $\pm$ .001 & .014 $\pm$ .000 & 16.792 $\pm$ 5.702 & .009 $\pm$ .000 & .014 $\pm$ .001 & .011 $\pm$ .000 & 2.449 $\pm$ 2.313 & .014 $\pm$ .001 & .019 $\pm$ .001 & .016 $\pm$ .001 & .244 $\pm$ .765 \\
 & \catoni & .010 $\pm$ .000 & .026 $\pm$ .001 & .013 $\pm$ .000 & 109.202 $\pm$ 15.634 & .012 $\pm$ .000 & .023 $\pm$ .002 & .014 $\pm$ .000 & 19.918 $\pm$ 6.526 & .009 $\pm$ .000 & .015 $\pm$ .001 & .011 $\pm$ .000 & 2.486 $\pm$ 2.362 & .014 $\pm$ .001 & .019 $\pm$ .001 & .016 $\pm$ .001 & .298 $\pm$ .762 \\
 & \rivasplata & .010 $\pm$ .000 & .024 $\pm$ .001 & .013 $\pm$ .000 & 91.872 $\pm$ 14.470 & .012 $\pm$ .000 & .019 $\pm$ .001 & .014 $\pm$ .000 & 17.002 $\pm$ 5.882 & .009 $\pm$ .000 & .014 $\pm$ .000 & .011 $\pm$ .000 & 2.529 $\pm$ 2.251 & .014 $\pm$ .001 & .019 $\pm$ .001 & .016 $\pm$ .001 & .308 $\pm$ .778 \\
 & \stoNN & \textemdash & .043 & \textemdash & .097 & \textemdash & .044 & \textemdash & .069 & \textemdash & .039 & \textemdash & .117 & \textemdash & .047 & \textemdash & .555 \\
\midrule
\multirow[c]{5}{*}{\rotatebox[origin=c]{90}{\small{Fashion}}} & \ours & .118 $\pm$ .001 & .123 $\pm$ .000 & .112 $\pm$ .000 & .269 & .113 $\pm$ .001 & .118 $\pm$ .001 & .107 $\pm$ .001 & .743 & .117 $\pm$ .002 & .121 $\pm$ .002 & .110 $\pm$ .002 & 2.600 & .131 $\pm$ .004 & .138 $\pm$ .004 & .126 $\pm$ .004 & 1.229 \\
 & \blanchard & .118 $\pm$ .001 & .145 $\pm$ .003 & .112 $\pm$ .000 & 82.403 $\pm$ 13.230 & .113 $\pm$ .001 & .123 $\pm$ .002 & .107 $\pm$ .001 & 16.836 $\pm$ 5.583 & .119 $\pm$ .002 & .121 $\pm$ .003 & .112 $\pm$ .003 & 2.641 $\pm$ 2.369 & .133 $\pm$ .004 & .136 $\pm$ .004 & .128 $\pm$ .004 & .297 $\pm$ .731 \\
 & \catoni & .118 $\pm$ .001 & .151 $\pm$ .004 & .112 $\pm$ .000 & 109.988 $\pm$ 15.347 & .113 $\pm$ .001 & .120 $\pm$ .002 & .107 $\pm$ .001 & 19.889 $\pm$ 6.689 & .118 $\pm$ .002 & .120 $\pm$ .003 & .112 $\pm$ .003 & 2.615 $\pm$ 2.234 & .132 $\pm$ .004 & .136 $\pm$ .004 & .128 $\pm$ .004 & .300 $\pm$ .811 \\
 & \rivasplata & .118 $\pm$ .001 & .137 $\pm$ .002 & .112 $\pm$ .000 & 87.804 $\pm$ 13.640 & .113 $\pm$ .001 & .120 $\pm$ .002 & .107 $\pm$ .001 & 17.491 $\pm$ 6.144 & .118 $\pm$ .002 & .121 $\pm$ .003 & .112 $\pm$ .003 & 2.549 $\pm$ 2.175 & .133 $\pm$ .005 & .137 $\pm$ .004 & .128 $\pm$ .004 & .322 $\pm$ .794 \\
 & \stoNN & \textemdash & .174 & \textemdash & .135 & \textemdash & .168 & \textemdash & .372 & \textemdash & .172 & \textemdash & 1.300 & \textemdash & .191 & \textemdash & .615 \\
\midrule
\multirow[c]{5}{*}{\rotatebox[origin=c]{90}{\small{CIFAR-10}}} & \ours & .334 $\pm$ .000 & .346 $\pm$ .000 & .328 $\pm$ .000 & .025 & .322 $\pm$ .000 & .331 $\pm$ .000 & .313 $\pm$ .000 & .050 & .323 $\pm$ .001 & .334 $\pm$ .000 & .316 $\pm$ .000 & .160 & .333 $\pm$ .001 & .341 $\pm$ .001 & .323 $\pm$ .001 & .461 \\
 & \blanchard & .334 $\pm$ .000 & .421 $\pm$ .004 & .328 $\pm$ .000 & 269.875 $\pm$ 23.982 & .322 $\pm$ .000 & .364 $\pm$ .004 & .313 $\pm$ .000 & 83.082 $\pm$ 13.029 & .323 $\pm$ .001 & .345 $\pm$ .004 & .316 $\pm$ .000 & 21.614 $\pm$ 6.670 & .333 $\pm$ .001 & .340 $\pm$ .002 & .323 $\pm$ .001 & 3.630 $\pm$ 2.750 \\
 & \catoni & .334 $\pm$ .000 & .433 $\pm$ .008 & .328 $\pm$ .000 & 27.270 $\pm$ 24.201 & .322 $\pm$ .000 & .355 $\pm$ .005 & .313 $\pm$ .000 & 84.148 $\pm$ 13.578 & .323 $\pm$ .001 & .338 $\pm$ .002 & .316 $\pm$ .000 & 22.547 $\pm$ 6.801 & .333 $\pm$ .001 & .338 $\pm$ .001 & .323 $\pm$ .001 & 3.831 $\pm$ 2.801 \\
 & \rivasplata & .334 $\pm$ .000 & .394 $\pm$ .003 & .328 $\pm$ .000 & 27.133 $\pm$ 24.109 & .322 $\pm$ .000 & .351 $\pm$ .003 & .313 $\pm$ .000 & 83.438 $\pm$ 13.033 & .323 $\pm$ .001 & .339 $\pm$ .002 & .316 $\pm$ .000 & 21.688 $\pm$ 6.718 & .333 $\pm$ .001 & .339 $\pm$ .002 & .323 $\pm$ .001 & 3.667 $\pm$ 2.757 \\
 & \stoNN & \textemdash & .414 & \textemdash & .013 & \textemdash & .399 & \textemdash & .025 & \textemdash & .403 & \textemdash & .080 & \textemdash & .409 & \textemdash & .230 \\
\midrule
 &  & \multicolumn{4}{c}{$\sigma^2=10^{-6}$} & \multicolumn{4}{c}{$\sigma^2=10^{-5}$} & \multicolumn{4}{c}{$\sigma^2=10^{-4}$} & \multicolumn{4}{c}{$\sigma^2=10^{-3}$} \\
\midrule
 & lr=$10^{-4}$ & $\Risk_{\Tcal}(h)$ & Bnd & $\Risk_{\Scal}(h)$ & Div & $\Risk_{\Tcal}(h)$ & Bnd & $\Risk_{\Scal}(h)$ & Div & $\Risk_{\Tcal}(h)$ & Bnd & $\Risk_{\Scal}(h)$ & Div & $\Risk_{\Tcal}(h)$ & Bnd & $\Risk_{\Scal}(h)$ & Div \\
\midrule
\multirow[c]{5}{*}{\rotatebox[origin=c]{90}{\small{MNIST}}} & \ours & .010 $\pm$ .000 & .019 $\pm$ .000 & .013 $\pm$ .000 & 23.992 & .012 $\pm$ .000 & .019 $\pm$ .000 & .014 $\pm$ .000 & 7.767 & .009 $\pm$ .000 & .015 $\pm$ .000 & .011 $\pm$ .000 & 3.165 & .012 $\pm$ .001 & .019 $\pm$ .001 & .013 $\pm$ .001 & 18.413 \\
 & \blanchard & .010 $\pm$ .000 & .500 $\pm$ .004 & .013 $\pm$ .000 & 1123.328 $\pm$ 151.115 & .012 $\pm$ .000 & .236 $\pm$ .004 & .014 $\pm$ .000 & 381.840 $\pm$ 94.218 & .009 $\pm$ .000 & .096 $\pm$ .003 & .011 $\pm$ .000 & 1184.214 $\pm$ 47.208 & .011 $\pm$ .001 & .048 $\pm$ .002 & .012 $\pm$ .001 & 363.194 $\pm$ 26.547 \\
 & \catoni & .010 $\pm$ .000 & .369 $\pm$ .003 & .013 $\pm$ .000 & 1191.598 $\pm$ 154.180 & .012 $\pm$ .000 & .180 $\pm$ .003 & .014 $\pm$ .000 & 3826.581 $\pm$ 85.362 & .009 $\pm$ .000 & .070 $\pm$ .002 & .011 $\pm$ .000 & 1217.723 $\pm$ 49.984 & .012 $\pm$ .001 & .039 $\pm$ .002 & .014 $\pm$ .001 & 384.476 $\pm$ 29.126 \\
 & \rivasplata & .010 $\pm$ .000 & .316 $\pm$ .003 & .013 $\pm$ .000 & 11557.703 $\pm$ 151.498 & .012 $\pm$ .000 & .142 $\pm$ .002 & .014 $\pm$ .000 & 3751.391 $\pm$ 84.542 & .009 $\pm$ .000 & .061 $\pm$ .002 & .011 $\pm$ .000 & 1172.156 $\pm$ 46.933 & .010 $\pm$ .001 & .035 $\pm$ .001 & .012 $\pm$ .001 & 373.003 $\pm$ 27.844 \\
 & \stoNN & \textemdash & .045 & \textemdash & 11.996 & \textemdash & .045 & \textemdash & 3.884 & \textemdash & .040 & \textemdash & 1.583 & \textemdash & .045 & \textemdash & 9.207 \\
\midrule
\multirow[c]{5}{*}{\rotatebox[origin=c]{90}{\small{Fashion}}} & \ours & .118 $\pm$ .000 & .127 $\pm$ .000 & .112 $\pm$ .000 & 17.987 & .113 $\pm$ .001 & .119 $\pm$ .001 & .107 $\pm$ .001 & 6.361 & .114 $\pm$ .002 & .123 $\pm$ .002 & .107 $\pm$ .002 & 22.582 & .125 $\pm$ .003 & .137 $\pm$ .003 & .122 $\pm$ .003 & 16.872 \\
 & \blanchard & .115 $\pm$ .001 & .659 $\pm$ .004 & .110 $\pm$ .000 & 11835.780 $\pm$ 161.816 & .110 $\pm$ .001 & .395 $\pm$ .004 & .104 $\pm$ .000 & 3828.562 $\pm$ 94.279 & .108 $\pm$ .001 & .244 $\pm$ .004 & .102 $\pm$ .001 & 1185.882 $\pm$ 5.575 & .123 $\pm$ .003 & .192 $\pm$ .004 & .119 $\pm$ .002 & 346.265 $\pm$ 27.827 \\
 & \catoni & .118 $\pm$ .001 & .566 $\pm$ .004 & .112 $\pm$ .000 & 11921.114 $\pm$ 153.739 & .113 $\pm$ .001 & .304 $\pm$ .003 & .107 $\pm$ .000 & 3822.647 $\pm$ 85.225 & .114 $\pm$ .002 & .208 $\pm$ .003 & .107 $\pm$ .002 & 1217.879 $\pm$ 52.353 & .125 $\pm$ .003 & .196 $\pm$ .004 & .121 $\pm$ .002 & 388.473 $\pm$ 29.475 \\
 & \rivasplata & .114 $\pm$ .000 & .476 $\pm$ .003 & .109 $\pm$ .000 & 11206.239 $\pm$ 149.549 & .110 $\pm$ .001 & .292 $\pm$ .003 & .103 $\pm$ .000 & 3745.930 $\pm$ 84.367 & .106 $\pm$ .001 & .197 $\pm$ .003 & .101 $\pm$ .001 & 1229.005 $\pm$ 51.052 & .122 $\pm$ .003 & .170 $\pm$ .004 & .118 $\pm$ .003 & 361.652 $\pm$ 28.452 \\
 & \stoNN & \textemdash & .177 & \textemdash & 8.994 & \textemdash & .169 & \textemdash & 3.180 & \textemdash & .172 & \textemdash & 11.291 & \textemdash & .189 & \textemdash & 8.436 \\
\midrule
\multirow[c]{5}{*}{\rotatebox[origin=c]{90}{\small{CIFAR-10}}} & \ours & .334 $\pm$ .000 & .350 $\pm$ .000 & .328 $\pm$ .000 & 12.067 & .322 $\pm$ .000 & .332 $\pm$ .000 & .313 $\pm$ .000 & 4.172 & .323 $\pm$ .001 & .336 $\pm$ .000 & .316 $\pm$ .000 & 3.382 & .332 $\pm$ .001 & .343 $\pm$ .001 & .322 $\pm$ .001 & 6.855 \\
 & \blanchard & .334 $\pm$ .000 & .977 $\pm$ .001 & .328 $\pm$ .000 & 28565.558 $\pm$ 245.568 & .322 $\pm$ .000 & .803 $\pm$ .003 & .313 $\pm$ .000 & 8479.553 $\pm$ 126.804 & .321 $\pm$ .001 & .614 $\pm$ .004 & .315 $\pm$ .000 & 2727.786 $\pm$ 7.572 & .327 $\pm$ .001 & .487 $\pm$ .004 & .317 $\pm$ .001 & 887.578 $\pm$ 42.449 \\
 & \catoni & .334 $\pm$ .000 & .983 $\pm$ .000 & .328 $\pm$ .000 & 24136.528 $\pm$ 211.963 & .322 $\pm$ .000 & .694 $\pm$ .004 & .313 $\pm$ .000 & 7928.671 $\pm$ 122.159 & .320 $\pm$ .001 & .515 $\pm$ .002 & .314 $\pm$ .000 & 237.703 $\pm$ 65.952 & .323 $\pm$ .001 & .468 $\pm$ .002 & .312 $\pm$ .001 & 1157.073 $\pm$ 47.283 \\
 & \rivasplata & .334 $\pm$ .000 & .922 $\pm$ .001 & .328 $\pm$ .000 & 33282.032 $\pm$ 246.654 & .322 $\pm$ .000 & .680 $\pm$ .003 & .312 $\pm$ .000 & 8493.458 $\pm$ 128.894 & .320 $\pm$ .001 & .527 $\pm$ .003 & .314 $\pm$ .000 & 2739.108 $\pm$ 7.556 & .325 $\pm$ .001 & .436 $\pm$ .003 & .314 $\pm$ .001 & 91.066 $\pm$ 43.389 \\
 & \stoNN & \textemdash & .417 & \textemdash & 6.033 & \textemdash & .400 & \textemdash & 2.086 & \textemdash & .403 & \textemdash & 1.691 & \textemdash & .410 & \textemdash & 3.427 \\
\bottomrule
\end{tabular}
}
\label{table:1_prior_0.4}
\end{sidewaystable}

\begin{sidewaystable}
\caption{
\looseness=-1
Comparison of \ours, \rivasplata, \blanchard and \catoni based on the disintegrated bounds, and \stoNN based on the randomized bounds learned with two learning rates lr${\ \in}\{10^{-4}, 10^{-6}\}$ and different variances $\sigma^2{\in}\{10^{-3}, 10^{-4}, 10^{-5}, 10^{-6}\}$.
We report the test risk ($\Risk_{\Tcal}(h)$), the bound value (Bnd), the empirical risk ($\Risk_{\Scal}(h)$), and the divergence (Div) associated with each bound (the Rényi divergence for \ours, the KL divergence for \stoNN, and the disintegrated KL divergence for \rivasplata, \blanchard and \catoni).
More precisely, we report the mean $\pm$ the standard deviation for $400$ neural networks sampled from $\AQ$ for \ours, \rivasplata, \blanchard, and \catoni.
We consider, in this table, that the split ratio is $0.5$.
}
\resizebox{0.83\paperheight}{!}{
\begin{tabular}{rr|clcl|clcl|clcl|clcl}
\toprule
 &  & \multicolumn{4}{c}{$\sigma^2=10^{-6}$} & \multicolumn{4}{c}{$\sigma^2=10^{-5}$} & \multicolumn{4}{c}{$\sigma^2=10^{-4}$} & \multicolumn{4}{c}{$\sigma^2=10^{-3}$} \\
\midrule
 & lr=$10^{-6}$ & $\Risk_{\Tcal}(h)$ & Bnd & $\Risk_{\Scal}(h)$ & Div & $\Risk_{\Tcal}(h)$ & Bnd & $\Risk_{\Scal}(h)$ & Div & $\Risk_{\Tcal}(h)$ & Bnd & $\Risk_{\Scal}(h)$ & Div & $\Risk_{\Tcal}(h)$ & Bnd & $\Risk_{\Scal}(h)$ & Div \\
\midrule
\multirow[c]{5}{*}{\rotatebox[origin=c]{90}{\small{MNIST}}} & \ours & .008 $\pm$ .000 & .015 $\pm$ .000 & .010 $\pm$ .000 & .084 & .006 $\pm$ .000 & .012 $\pm$ .000 & .009 $\pm$ .000 & .053 & .008 $\pm$ .000 & .014 $\pm$ .000 & .010 $\pm$ .000 & .179 & .014 $\pm$ .001 & .019 $\pm$ .001 & .014 $\pm$ .001 & .576 \\
 & \blanchard & .008 $\pm$ .000 & .025 $\pm$ .001 & .010 $\pm$ .000 & 81.167 $\pm$ 12.801 & .006 $\pm$ .000 & .014 $\pm$ .001 & .009 $\pm$ .000 & 15.518 $\pm$ 5.438 & .009 $\pm$ .000 & .014 $\pm$ .001 & .010 $\pm$ .000 & 2.140 $\pm$ 2.072 & .015 $\pm$ .001 & .018 $\pm$ .001 & .015 $\pm$ .001 & .284 $\pm$ .649 \\
 & \catoni & .008 $\pm$ .000 & .022 $\pm$ .001 & .010 $\pm$ .000 & 104.063 $\pm$ 14.662 & .006 $\pm$ .000 & .015 $\pm$ .000 & .009 $\pm$ .000 & 17.676 $\pm$ 5.963 & .008 $\pm$ .000 & .014 $\pm$ .001 & .010 $\pm$ .000 & 2.152 $\pm$ 2.085 & .015 $\pm$ .001 & .018 $\pm$ .001 & .015 $\pm$ .001 & .252 $\pm$ .680 \\
 & \rivasplata & .008 $\pm$ .000 & .021 $\pm$ .001 & .010 $\pm$ .000 & 84.581 $\pm$ 13.035 & .006 $\pm$ .000 & .013 $\pm$ .001 & .009 $\pm$ .000 & 15.545 $\pm$ 5.594 & .008 $\pm$ .000 & .014 $\pm$ .000 & .010 $\pm$ .000 & 2.185 $\pm$ 1.992 & .015 $\pm$ .001 & .018 $\pm$ .001 & .015 $\pm$ .001 & .276 $\pm$ .693 \\
 & \stoNN & \textemdash & .039 & \textemdash & .042 & \textemdash & .035 & \textemdash & .026 & \textemdash & .038 & \textemdash & .090 & \textemdash & .045 & \textemdash & .288 \\
\midrule
\multirow[c]{5}{*}{\rotatebox[origin=c]{90}{\small{Fashion}}} & \ours & .106 $\pm$ .000 & .113 $\pm$ .000 & .101 $\pm$ .000 & .133 & .104 $\pm$ .001 & .110 $\pm$ .000 & .099 $\pm$ .000 & .327 & .108 $\pm$ .002 & .112 $\pm$ .001 & .101 $\pm$ .001 & .903 & .120 $\pm$ .004 & .127 $\pm$ .003 & .115 $\pm$ .003 & .868 \\
 & \blanchard & .106 $\pm$ .000 & .136 $\pm$ .003 & .101 $\pm$ .000 & 77.573 $\pm$ 12.564 & .104 $\pm$ .001 & .115 $\pm$ .003 & .099 $\pm$ .000 & 15.278 $\pm$ 5.599 & .109 $\pm$ .002 & .111 $\pm$ .002 & .102 $\pm$ .001 & 2.153 $\pm$ 2.081 & .122 $\pm$ .004 & .126 $\pm$ .004 & .117 $\pm$ .004 & .248 $\pm$ .715 \\
 & \catoni & .106 $\pm$ .000 & .145 $\pm$ .005 & .101 $\pm$ .000 & 104.356 $\pm$ 14.712 & .104 $\pm$ .001 & .112 $\pm$ .002 & .099 $\pm$ .000 & 17.566 $\pm$ 5.996 & .109 $\pm$ .002 & .110 $\pm$ .001 & .102 $\pm$ .001 & 2.217 $\pm$ 2.084 & .122 $\pm$ .004 & .125 $\pm$ .004 & .117 $\pm$ .004 & .262 $\pm$ .699 \\
 & \rivasplata & .106 $\pm$ .000 & .127 $\pm$ .002 & .101 $\pm$ .000 & 82.150 $\pm$ 12.955 & .104 $\pm$ .001 & .112 $\pm$ .001 & .099 $\pm$ .000 & 15.509 $\pm$ 5.629 & .109 $\pm$ .002 & .111 $\pm$ .001 & .102 $\pm$ .001 & 2.178 $\pm$ 2.060 & .122 $\pm$ .004 & .126 $\pm$ .004 & .117 $\pm$ .004 & .264 $\pm$ .704 \\
 & \stoNN & \textemdash & .162 & \textemdash & .066 & \textemdash & .159 & \textemdash & .164 & \textemdash & .162 & \textemdash & .451 & \textemdash & .179 & \textemdash & .434 \\
\midrule
\multirow[c]{5}{*}{\rotatebox[origin=c]{90}{\small{CIFAR-10}}} & \ours & .312 $\pm$ .000 & .323 $\pm$ .000 & .304 $\pm$ .000 & .027 & .281 $\pm$ .000 & .304 $\pm$ .000 & .285 $\pm$ .000 & .035 & .298 $\pm$ .001 & .310 $\pm$ .000 & .291 $\pm$ .000 & .101 & .315 $\pm$ .001 & .329 $\pm$ .001 & .309 $\pm$ .001 & .368 \\
 & \blanchard & .312 $\pm$ .000 & .405 $\pm$ .004 & .304 $\pm$ .000 & 268.149 $\pm$ 22.835 & .281 $\pm$ .000 & .339 $\pm$ .004 & .285 $\pm$ .000 & 8.690 $\pm$ 12.628 & .298 $\pm$ .001 & .320 $\pm$ .004 & .291 $\pm$ .000 & 19.648 $\pm$ 6.249 & .315 $\pm$ .001 & .327 $\pm$ .003 & .310 $\pm$ .001 & 3.213 $\pm$ 2.590 \\
 & \catoni & .312 $\pm$ .000 & .428 $\pm$ .009 & .304 $\pm$ .000 & 269.415 $\pm$ 22.884 & .281 $\pm$ .000 & .333 $\pm$ .005 & .285 $\pm$ .000 & 83.414 $\pm$ 13.018 & .298 $\pm$ .001 & .314 $\pm$ .003 & .291 $\pm$ .000 & 2.711 $\pm$ 6.481 & .315 $\pm$ .001 & .326 $\pm$ .001 & .310 $\pm$ .001 & 3.273 $\pm$ 2.597 \\
 & \rivasplata & .312 $\pm$ .000 & .375 $\pm$ .003 & .304 $\pm$ .000 & 268.589 $\pm$ 22.845 & .281 $\pm$ .000 & .325 $\pm$ .003 & .285 $\pm$ .000 & 81.532 $\pm$ 12.712 & .298 $\pm$ .001 & .315 $\pm$ .002 & .291 $\pm$ .000 & 19.813 $\pm$ 6.288 & .315 $\pm$ .001 & .327 $\pm$ .002 & .310 $\pm$ .001 & 3.233 $\pm$ 2.599 \\
 & \stoNN & \textemdash & .391 & \textemdash & .013 & \textemdash & .370 & \textemdash & .017 & \textemdash & .377 & \textemdash & .050 & \textemdash & .397 & \textemdash & .184 \\
\midrule
 &  & \multicolumn{4}{c}{$\sigma^2=10^{-6}$} & \multicolumn{4}{c}{$\sigma^2=10^{-5}$} & \multicolumn{4}{c}{$\sigma^2=10^{-4}$} & \multicolumn{4}{c}{$\sigma^2=10^{-3}$} \\
\midrule
 & lr=$10^{-4}$ & $\Risk_{\Tcal}(h)$ & Bnd & $\Risk_{\Scal}(h)$ & Div & $\Risk_{\Tcal}(h)$ & Bnd & $\Risk_{\Scal}(h)$ & Div & $\Risk_{\Tcal}(h)$ & Bnd & $\Risk_{\Scal}(h)$ & Div & $\Risk_{\Tcal}(h)$ & Bnd & $\Risk_{\Scal}(h)$ & Div \\
\midrule
\multirow[c]{5}{*}{\rotatebox[origin=c]{90}{\small{MNIST}}} & \ours & .008 $\pm$ .000 & .017 $\pm$ .000 & .010 $\pm$ .000 & 29.993 & .006 $\pm$ .000 & .013 $\pm$ .000 & .009 $\pm$ .000 & 3.162 & .008 $\pm$ .000 & .015 $\pm$ .000 & .010 $\pm$ .000 & 1.418 & .013 $\pm$ .001 & .019 $\pm$ .001 & .013 $\pm$ .001 & 12.231 \\
 & \blanchard & .008 $\pm$ .000 & .574 $\pm$ .005 & .010 $\pm$ .000 & 11894.556 $\pm$ 155.958 & .006 $\pm$ .000 & .256 $\pm$ .004 & .009 $\pm$ .000 & 3826.515 $\pm$ 86.973 & .008 $\pm$ .000 & .108 $\pm$ .003 & .010 $\pm$ .000 & 1184.777 $\pm$ 48.158 & .010 $\pm$ .001 & .052 $\pm$ .002 & .011 $\pm$ .000 & 36.865 $\pm$ 28.054 \\
 & \catoni & .008 $\pm$ .000 & .396 $\pm$ .003 & .010 $\pm$ .000 & 11986.455 $\pm$ 15.722 & .006 $\pm$ .000 & .192 $\pm$ .002 & .009 $\pm$ .000 & 3824.971 $\pm$ 85.072 & .008 $\pm$ .000 & .079 $\pm$ .002 & .010 $\pm$ .000 & 1213.611 $\pm$ 48.751 & .013 $\pm$ .001 & .042 $\pm$ .002 & .014 $\pm$ .001 & 384.275 $\pm$ 28.556 \\
 & \rivasplata & .008 $\pm$ .000 & .362 $\pm$ .003 & .010 $\pm$ .000 & 11905.971 $\pm$ 15.609 & .006 $\pm$ .000 & .148 $\pm$ .003 & .009 $\pm$ .000 & 377.259 $\pm$ 84.127 & .008 $\pm$ .000 & .067 $\pm$ .002 & .010 $\pm$ .000 & 118.841 $\pm$ 5.043 & .010 $\pm$ .001 & .036 $\pm$ .001 & .011 $\pm$ .000 & 369.675 $\pm$ 27.947 \\
 & \stoNN & \textemdash & .041 & \textemdash & 14.996 & \textemdash & .035 & \textemdash & 1.581 & \textemdash & .039 & \textemdash & .709 & \textemdash & .045 & \textemdash & 6.116 \\
\midrule
\multirow[c]{5}{*}{\rotatebox[origin=c]{90}{\small{Fashion}}} & \ours & .106 $\pm$ .000 & .114 $\pm$ .000 & .101 $\pm$ .000 & 6.310 & .103 $\pm$ .001 & .113 $\pm$ .000 & .099 $\pm$ .000 & 9.312 & .106 $\pm$ .002 & .115 $\pm$ .001 & .100 $\pm$ .001 & 14.924 & .115 $\pm$ .003 & .126 $\pm$ .003 & .110 $\pm$ .002 & 18.364 \\
 & \blanchard & .105 $\pm$ .000 & .674 $\pm$ .004 & .101 $\pm$ .000 & 10795.464 $\pm$ 143.426 & .102 $\pm$ .000 & .412 $\pm$ .004 & .098 $\pm$ .000 & 3685.940 $\pm$ 82.481 & .103 $\pm$ .001 & .253 $\pm$ .004 & .097 $\pm$ .001 & 1178.401 $\pm$ 48.359 & .113 $\pm$ .002 & .186 $\pm$ .004 & .108 $\pm$ .002 & 338.697 $\pm$ 27.104 \\
 & \catoni & .106 $\pm$ .000 & .623 $\pm$ .005 & .101 $\pm$ .000 & 11971.564 $\pm$ 15.589 & .104 $\pm$ .001 & .321 $\pm$ .004 & .099 $\pm$ .000 & 3825.370 $\pm$ 87.728 & .107 $\pm$ .002 & .208 $\pm$ .003 & .100 $\pm$ .001 & 1214.976 $\pm$ 48.846 & .116 $\pm$ .003 & .184 $\pm$ .004 & .111 $\pm$ .003 & 388.197 $\pm$ 27.580 \\
 & \rivasplata & .105 $\pm$ .000 & .503 $\pm$ .003 & .100 $\pm$ .000 & 11139.304 $\pm$ 15.540 & .102 $\pm$ .000 & .307 $\pm$ .003 & .097 $\pm$ .000 & 381.075 $\pm$ 87.924 & .102 $\pm$ .001 & .201 $\pm$ .003 & .096 $\pm$ .001 & 1201.832 $\pm$ 48.877 & .112 $\pm$ .002 & .161 $\pm$ .003 & .107 $\pm$ .002 & 349.146 $\pm$ 27.482 \\
 & \stoNN & \textemdash & .163 & \textemdash & 3.155 & \textemdash & .161 & \textemdash & 4.656 & \textemdash & .163 & \textemdash & 7.462 & \textemdash & .176 & \textemdash & 9.182 \\
\midrule
\multirow[c]{5}{*}{\rotatebox[origin=c]{90}{\small{CIFAR-10}}} & \ours & .312 $\pm$ .000 & .328 $\pm$ .000 & .304 $\pm$ .000 & 12.006 & .281 $\pm$ .000 & .304 $\pm$ .000 & .285 $\pm$ .000 & 1.802 & .297 $\pm$ .001 & .311 $\pm$ .000 & .291 $\pm$ .000 & 2.056 & .314 $\pm$ .001 & .330 $\pm$ .001 & .309 $\pm$ .001 & 4.782 \\
 & \blanchard & .312 $\pm$ .000 & .990 $\pm$ .000 & .304 $\pm$ .000 & 48007.471 $\pm$ 31.730 & .280 $\pm$ .000 & .825 $\pm$ .003 & .284 $\pm$ .000 & 8824.774 $\pm$ 134.331 & .296 $\pm$ .001 & .617 $\pm$ .004 & .290 $\pm$ .000 & 2723.775 $\pm$ 66.832 & .309 $\pm$ .001 & .490 $\pm$ .004 & .303 $\pm$ .001 & 888.277 $\pm$ 41.530 \\
 & \catoni & .312 $\pm$ .000 & .980 $\pm$ .000 & .304 $\pm$ .000 & 21278.808 $\pm$ 207.839 & .280 $\pm$ .000 & .681 $\pm$ .004 & .284 $\pm$ .000 & 6951.932 $\pm$ 118.540 & .296 $\pm$ .001 & .496 $\pm$ .003 & .290 $\pm$ .000 & 2145.470 $\pm$ 6.045 & .305 $\pm$ .001 & .457 $\pm$ .002 & .299 $\pm$ .001 & 103.494 $\pm$ 47.021 \\
 & \rivasplata & .312 $\pm$ .000 & .964 $\pm$ .001 & .304 $\pm$ .000 & 42834.626 $\pm$ 284.116 & .280 $\pm$ .000 & .690 $\pm$ .003 & .284 $\pm$ .000 & 8675.531 $\pm$ 136.658 & .296 $\pm$ .001 & .521 $\pm$ .003 & .290 $\pm$ .000 & 2718.415 $\pm$ 66.664 & .307 $\pm$ .001 & .434 $\pm$ .003 & .301 $\pm$ .001 & 921.068 $\pm$ 42.158 \\
 & \stoNN & \textemdash & .394 & \textemdash & 6.003 & \textemdash & .371 & \textemdash & .901 & \textemdash & .378 & \textemdash & 1.028 & \textemdash & .397 & \textemdash & 2.391 \\
\bottomrule
\end{tabular}
}
\label{table:1_prior_0.5}
\end{sidewaystable}

\begin{sidewaystable}
\caption{
\looseness=-1
Comparison of \ours, \rivasplata, \blanchard and \catoni based on the disintegrated bounds, and \stoNN based on the randomized bounds learned with two learning rates lr${\ \in}\{10^{-4}, 10^{-6}\}$ and different variances $\sigma^2{\in}\{10^{-3}, 10^{-4}, 10^{-5}, 10^{-6}\}$.
We report the test risk ($\Risk_{\Tcal}(h)$), the bound value (Bnd), the empirical risk ($\Risk_{\Scal}(h)$), and the divergence (Div) associated with each bound (the Rényi divergence for \ours, the KL divergence for \stoNN, and the disintegrated KL divergence for \rivasplata, \blanchard and \catoni).
More precisely, we report the mean $\pm$ the standard deviation for $400$ neural networks sampled from $\AQ$ for \ours, \rivasplata, \blanchard, and \catoni.
We consider, in this table, that the split ratio is $0.6$.
}
\resizebox{0.83\paperheight}{!}{
\begin{tabular}{rr|clcl|clcl|clcl|clcl}
\toprule
 &  & \multicolumn{4}{c}{$\sigma^2=10^{-6}$} & \multicolumn{4}{c}{$\sigma^2=10^{-5}$} & \multicolumn{4}{c}{$\sigma^2=10^{-4}$} & \multicolumn{4}{c}{$\sigma^2=10^{-3}$} \\
\midrule
 & lr=$10^{-6}$ & $\Risk_{\Tcal}(h)$ & Bnd & $\Risk_{\Scal}(h)$ & Div & $\Risk_{\Tcal}(h)$ & Bnd & $\Risk_{\Scal}(h)$ & Div & $\Risk_{\Tcal}(h)$ & Bnd & $\Risk_{\Scal}(h)$ & Div & $\Risk_{\Tcal}(h)$ & Bnd & $\Risk_{\Scal}(h)$ & Div \\
\midrule
\multirow[c]{5}{*}{\rotatebox[origin=c]{90}{\small{MNIST}}} & \ours & .008 $\pm$ .000 & .014 $\pm$ .000 & .010 $\pm$ .000 & .040 & .007 $\pm$ .000 & .014 $\pm$ .000 & .009 $\pm$ .000 & .068 & .008 $\pm$ .000 & .013 $\pm$ .000 & .009 $\pm$ .000 & .092 & .008 $\pm$ .000 & .014 $\pm$ .001 & .009 $\pm$ .000 & .128 \\
 & \blanchard & .008 $\pm$ .000 & .026 $\pm$ .002 & .010 $\pm$ .000 & 75.043 $\pm$ 11.586 & .007 $\pm$ .000 & .016 $\pm$ .001 & .009 $\pm$ .000 & 13.220 $\pm$ 4.956 & .008 $\pm$ .000 & .012 $\pm$ .001 & .009 $\pm$ .000 & 1.774 $\pm$ 1.772 & .008 $\pm$ .000 & .012 $\pm$ .001 & .009 $\pm$ .000 & .190 $\pm$ .594 \\
 & \catoni & .008 $\pm$ .000 & .022 $\pm$ .001 & .010 $\pm$ .000 & 96.561 $\pm$ 13.980 & .007 $\pm$ .000 & .016 $\pm$ .000 & .009 $\pm$ .000 & 15.107 $\pm$ 5.370 & .008 $\pm$ .000 & .013 $\pm$ .001 & .009 $\pm$ .000 & 1.835 $\pm$ 1.837 & .008 $\pm$ .000 & .013 $\pm$ .000 & .009 $\pm$ .000 & .219 $\pm$ .619 \\
 & \rivasplata & .008 $\pm$ .000 & .021 $\pm$ .001 & .010 $\pm$ .000 & 76.898 $\pm$ 12.301 & .007 $\pm$ .000 & .014 $\pm$ .001 & .009 $\pm$ .000 & 13.370 $\pm$ 4.931 & .008 $\pm$ .000 & .013 $\pm$ .000 & .009 $\pm$ .000 & 1.695 $\pm$ 1.741 & .008 $\pm$ .000 & .013 $\pm$ .001 & .009 $\pm$ .000 & .183 $\pm$ .580 \\
 & \stoNN & \textemdash & .038 & \textemdash & .020 & \textemdash & .037 & \textemdash & .034 & \textemdash & .037 & \textemdash & .046 & \textemdash & .037 & \textemdash & .064 \\
\midrule
\multirow[c]{5}{*}{\rotatebox[origin=c]{90}{\small{Fashion}}} & \ours & .109 $\pm$ .000 & .115 $\pm$ .000 & .102 $\pm$ .000 & .128 & .114 $\pm$ .001 & .117 $\pm$ .001 & .104 $\pm$ .001 & .436 & .101 $\pm$ .001 & .108 $\pm$ .001 & .096 $\pm$ .001 & .452 & .110 $\pm$ .003 & .116 $\pm$ .003 & .103 $\pm$ .003 & .438 \\
 & \blanchard & .109 $\pm$ .000 & .139 $\pm$ .003 & .102 $\pm$ .000 & 7.878 $\pm$ 11.599 & .114 $\pm$ .001 & .121 $\pm$ .003 & .104 $\pm$ .001 & 13.041 $\pm$ 5.012 & .102 $\pm$ .001 & .106 $\pm$ .002 & .096 $\pm$ .001 & 1.840 $\pm$ 1.864 & .111 $\pm$ .003 & .113 $\pm$ .003 & .104 $\pm$ .002 & .184 $\pm$ .600 \\
 & \catoni & .109 $\pm$ .000 & .152 $\pm$ .006 & .102 $\pm$ .000 & 96.732 $\pm$ 13.464 & .114 $\pm$ .001 & .119 $\pm$ .002 & .104 $\pm$ .001 & 15.103 $\pm$ 5.363 & .102 $\pm$ .001 & .105 $\pm$ .001 & .096 $\pm$ .001 & 1.825 $\pm$ 1.886 & .111 $\pm$ .003 & .112 $\pm$ .003 & .104 $\pm$ .003 & .224 $\pm$ .610 \\
 & \rivasplata & .109 $\pm$ .000 & .129 $\pm$ .002 & .102 $\pm$ .000 & 75.029 $\pm$ 11.918 & .114 $\pm$ .001 & .118 $\pm$ .002 & .104 $\pm$ .001 & 13.495 $\pm$ 5.112 & .102 $\pm$ .001 & .106 $\pm$ .001 & .096 $\pm$ .001 & 1.798 $\pm$ 1.859 & .111 $\pm$ .003 & .114 $\pm$ .003 & .104 $\pm$ .002 & .219 $\pm$ .610 \\
 & \stoNN & \textemdash & .164 & \textemdash & .064 & \textemdash & .167 & \textemdash & .218 & \textemdash & .157 & \textemdash & .226 & \textemdash & .165 & \textemdash & .219 \\
\midrule
\multirow[c]{5}{*}{\rotatebox[origin=c]{90}{\small{CIFAR-10}}} & \ours & .277 $\pm$ .000 & .297 $\pm$ .000 & .276 $\pm$ .000 & .021 & .288 $\pm$ .000 & .307 $\pm$ .000 & .286 $\pm$ .000 & .027 & .273 $\pm$ .001 & .284 $\pm$ .000 & .263 $\pm$ .000 & .079 & .281 $\pm$ .001 & .302 $\pm$ .001 & .281 $\pm$ .001 & .227 \\
 & \blanchard & .277 $\pm$ .000 & .386 $\pm$ .005 & .276 $\pm$ .000 & 262.952 $\pm$ 24.385 & .288 $\pm$ .000 & .346 $\pm$ .005 & .286 $\pm$ .000 & 76.609 $\pm$ 12.923 & .273 $\pm$ .001 & .293 $\pm$ .004 & .263 $\pm$ .000 & 17.724 $\pm$ 6.241 & .281 $\pm$ .001 & .299 $\pm$ .002 & .281 $\pm$ .001 & 2.580 $\pm$ 2.299 \\
 & \catoni & .277 $\pm$ .000 & .398 $\pm$ .001 & .276 $\pm$ .000 & 268.083 $\pm$ 24.567 & .288 $\pm$ .000 & .343 $\pm$ .007 & .286 $\pm$ .000 & 82.887 $\pm$ 13.493 & .273 $\pm$ .001 & .287 $\pm$ .003 & .263 $\pm$ .000 & 18.978 $\pm$ 6.437 & .281 $\pm$ .001 & .297 $\pm$ .001 & .281 $\pm$ .001 & 2.661 $\pm$ 2.317 \\
 & \rivasplata & .277 $\pm$ .000 & .354 $\pm$ .004 & .276 $\pm$ .000 & 263.581 $\pm$ 24.435 & .288 $\pm$ .000 & .330 $\pm$ .003 & .286 $\pm$ .000 & 77.488 $\pm$ 12.464 & .273 $\pm$ .001 & .288 $\pm$ .002 & .263 $\pm$ .000 & 17.704 $\pm$ 5.927 & .281 $\pm$ .001 & .299 $\pm$ .002 & .281 $\pm$ .001 & 2.619 $\pm$ 2.297 \\
 & \stoNN & \textemdash & .363 & \textemdash & .010 & \textemdash & .374 & \textemdash & .014 & \textemdash & .349 & \textemdash & .040 & \textemdash & .368 & \textemdash & .113 \\
\midrule
 &  & \multicolumn{4}{c}{$\sigma^2=10^{-6}$} & \multicolumn{4}{c}{$\sigma^2=10^{-5}$} & \multicolumn{4}{c}{$\sigma^2=10^{-4}$} & \multicolumn{4}{c}{$\sigma^2=10^{-3}$} \\
\midrule
 & lr=$10^{-4}$ & $\Risk_{\Tcal}(h)$ & Bnd & $\Risk_{\Scal}(h)$ & Div & $\Risk_{\Tcal}(h)$ & Bnd & $\Risk_{\Scal}(h)$ & Div & $\Risk_{\Tcal}(h)$ & Bnd & $\Risk_{\Scal}(h)$ & Div & $\Risk_{\Tcal}(h)$ & Bnd & $\Risk_{\Scal}(h)$ & Div \\
\midrule
\multirow[c]{5}{*}{\rotatebox[origin=c]{90}{\small{MNIST}}} & \ours & .008 $\pm$ .000 & .016 $\pm$ .000 & .010 $\pm$ .000 & 9.520 & .007 $\pm$ .000 & .014 $\pm$ .000 & .009 $\pm$ .000 & 3.594 & .008 $\pm$ .000 & .014 $\pm$ .000 & .009 $\pm$ .000 & 1.877 & .008 $\pm$ .000 & .014 $\pm$ .001 & .009 $\pm$ .000 & 6.589 \\
 & \blanchard & .008 $\pm$ .000 & .657 $\pm$ .005 & .010 $\pm$ .000 & 1209.158 $\pm$ 157.539 & .007 $\pm$ .000 & .304 $\pm$ .005 & .009 $\pm$ .000 & 3795.285 $\pm$ 88.141 & .008 $\pm$ .000 & .124 $\pm$ .004 & .009 $\pm$ .000 & 1183.704 $\pm$ 5.113 & .007 $\pm$ .000 & .052 $\pm$ .002 & .009 $\pm$ .000 & 347.860 $\pm$ 25.275 \\
 & \catoni & .008 $\pm$ .000 & .452 $\pm$ .004 & .010 $\pm$ .000 & 12032.708 $\pm$ 157.184 & .007 $\pm$ .000 & .225 $\pm$ .003 & .009 $\pm$ .000 & 3834.246 $\pm$ 89.809 & .008 $\pm$ .000 & .093 $\pm$ .003 & .009 $\pm$ .000 & 1225.575 $\pm$ 51.027 & .007 $\pm$ .000 & .039 $\pm$ .002 & .008 $\pm$ .000 & 39.374 $\pm$ 26.987 \\
 & \rivasplata & .008 $\pm$ .000 & .423 $\pm$ .004 & .010 $\pm$ .000 & 11943.688 $\pm$ 156.365 & .007 $\pm$ .000 & .179 $\pm$ .003 & .009 $\pm$ .000 & 3787.407 $\pm$ 87.968 & .008 $\pm$ .000 & .075 $\pm$ .002 & .009 $\pm$ .000 & 1173.457 $\pm$ 49.846 & .007 $\pm$ .000 & .035 $\pm$ .002 & .008 $\pm$ .000 & 348.717 $\pm$ 26.495 \\
 & \stoNN & \textemdash & .039 & \textemdash & 4.760 & \textemdash & .038 & \textemdash & 1.797 & \textemdash & .037 & \textemdash & .938 & \textemdash & .038 & \textemdash & 3.294 \\
\midrule
\multirow[c]{5}{*}{\rotatebox[origin=c]{90}{\small{Fashion}}} & \ours & .109 $\pm$ .000 & .119 $\pm$ .000 & .102 $\pm$ .000 & 16.776 & .114 $\pm$ .001 & .119 $\pm$ .001 & .104 $\pm$ .001 & 7.869 & .101 $\pm$ .001 & .111 $\pm$ .001 & .095 $\pm$ .001 & 14.224 & .109 $\pm$ .002 & .116 $\pm$ .002 & .101 $\pm$ .002 & 9.187 \\
 & \blanchard & .108 $\pm$ .000 & .743 $\pm$ .004 & .101 $\pm$ .000 & 11048.501 $\pm$ 146.969 & .112 $\pm$ .001 & .468 $\pm$ .005 & .101 $\pm$ .001 & 3798.865 $\pm$ 87.270 & .099 $\pm$ .001 & .268 $\pm$ .005 & .093 $\pm$ .001 & 1144.740 $\pm$ 49.199 & .106 $\pm$ .002 & .183 $\pm$ .004 & .099 $\pm$ .002 & 328.466 $\pm$ 24.435 \\
 & \catoni & .109 $\pm$ .000 & .712 $\pm$ .005 & .102 $\pm$ .000 & 1191.096 $\pm$ 15.212 & .114 $\pm$ .001 & .367 $\pm$ .005 & .104 $\pm$ .001 & 3831.104 $\pm$ 88.371 & .101 $\pm$ .001 & .216 $\pm$ .003 & .095 $\pm$ .001 & 1221.392 $\pm$ 5.970 & .108 $\pm$ .002 & .175 $\pm$ .003 & .101 $\pm$ .002 & 386.528 $\pm$ 26.498 \\
 & \rivasplata & .108 $\pm$ .000 & .557 $\pm$ .003 & .101 $\pm$ .000 & 11148.085 $\pm$ 145.818 & .111 $\pm$ .001 & .340 $\pm$ .003 & .100 $\pm$ .001 & 3757.976 $\pm$ 83.965 & .098 $\pm$ .001 & .209 $\pm$ .003 & .092 $\pm$ .001 & 1176.081 $\pm$ 49.829 & .106 $\pm$ .002 & .156 $\pm$ .003 & .098 $\pm$ .002 & 34.716 $\pm$ 24.874 \\
 & \stoNN & \textemdash & .168 & \textemdash & 8.388 & \textemdash & .168 & \textemdash & 3.935 & \textemdash & .159 & \textemdash & 7.112 & \textemdash & .165 & \textemdash & 4.594 \\
\midrule
\multirow[c]{5}{*}{\rotatebox[origin=c]{90}{\small{CIFAR-10}}} & \ours & .277 $\pm$ .000 & .301 $\pm$ .000 & .276 $\pm$ .000 & 8.466 & .288 $\pm$ .000 & .308 $\pm$ .000 & .286 $\pm$ .000 & 2.415 & .273 $\pm$ .001 & .285 $\pm$ .000 & .263 $\pm$ .000 & 2.256 & .280 $\pm$ .001 & .303 $\pm$ .001 & .280 $\pm$ .001 & 2.747 \\
 & \blanchard & .277 $\pm$ .000 & .990 $\pm$ .000 & .276 $\pm$ .000 & 58878.209 $\pm$ 356.845 & .288 $\pm$ .000 & .868 $\pm$ .003 & .286 $\pm$ .000 & 8858.838 $\pm$ 134.545 & .272 $\pm$ .001 & .625 $\pm$ .005 & .262 $\pm$ .000 & 2709.659 $\pm$ 76.197 & .278 $\pm$ .001 & .480 $\pm$ .005 & .277 $\pm$ .001 & 86.940 $\pm$ 43.864 \\
 & \catoni & .277 $\pm$ .000 & .974 $\pm$ .000 & .276 $\pm$ .000 & 17581.286 $\pm$ 185.476 & .288 $\pm$ .000 & .662 $\pm$ .005 & .286 $\pm$ .000 & 5118.582 $\pm$ 105.636 & .272 $\pm$ .001 & .456 $\pm$ .003 & .262 $\pm$ .000 & 1548.107 $\pm$ 58.565 & .277 $\pm$ .001 & .426 $\pm$ .002 & .274 $\pm$ .001 & 783.103 $\pm$ 41.593 \\
 & \rivasplata & .277 $\pm$ .000 & .990 $\pm$ .000 & .276 $\pm$ .000 & 82459.214 $\pm$ 398.763 & .288 $\pm$ .000 & .733 $\pm$ .003 & .286 $\pm$ .000 & 8674.850 $\pm$ 13.468 & .272 $\pm$ .001 & .518 $\pm$ .004 & .262 $\pm$ .000 & 2709.173 $\pm$ 77.205 & .277 $\pm$ .001 & .418 $\pm$ .004 & .275 $\pm$ .001 & 874.307 $\pm$ 44.089 \\
 & \stoNN & \textemdash & .366 & \textemdash & 4.233 & \textemdash & .374 & \textemdash & 1.207 & \textemdash & .350 & \textemdash & 1.128 & \textemdash & .369 & \textemdash & 1.374 \\
\bottomrule
\end{tabular}
}
\label{table:1_prior_0.6}
\end{sidewaystable}

\begin{sidewaystable}
\caption{
\looseness=-1
Comparison of \ours, \rivasplata, \blanchard and \catoni based on the disintegrated bounds, and \stoNN based on the randomized bounds learned with two learning rates lr${\ \in}\{10^{-4}, 10^{-6}\}$ and different variances $\sigma^2{\in}\{10^{-3}, 10^{-4}, 10^{-5}, 10^{-6}\}$.
We report the test risk ($\Risk_{\Tcal}(h)$), the bound value (Bnd), the empirical risk ($\Risk_{\Scal}(h)$), and the divergence (Div) associated with each bound (the Rényi divergence for \ours, the KL divergence for \stoNN, and the disintegrated KL divergence for \rivasplata, \blanchard and \catoni).
More precisely, we report the mean $\pm$ the standard deviation for $400$ neural networks sampled from $\AQ$ for \ours, \rivasplata, \blanchard, and \catoni.
We consider, in this table, that the split ratio is $0.7$.
}
\resizebox{0.83\paperheight}{!}{
\begin{tabular}{rr|clcl|clcl|clcl|clcl}
\toprule
 &  & \multicolumn{4}{c}{$\sigma^2=10^{-6}$} & \multicolumn{4}{c}{$\sigma^2=10^{-5}$} & \multicolumn{4}{c}{$\sigma^2=10^{-4}$} & \multicolumn{4}{c}{$\sigma^2=10^{-3}$} \\
\midrule
 & lr=$10^{-6}$ & $\Risk_{\Tcal}(h)$ & Bnd & $\Risk_{\Scal}(h)$ & Div & $\Risk_{\Tcal}(h)$ & Bnd & $\Risk_{\Scal}(h)$ & Div & $\Risk_{\Tcal}(h)$ & Bnd & $\Risk_{\Scal}(h)$ & Div & $\Risk_{\Tcal}(h)$ & Bnd & $\Risk_{\Scal}(h)$ & Div \\
\midrule
\multirow[c]{5}{*}{\rotatebox[origin=c]{90}{\small{MNIST}}} & \ours & .011 $\pm$ .000 & .019 $\pm$ .000 & .013 $\pm$ .000 & .047 & .010 $\pm$ .000 & .018 $\pm$ .000 & .012 $\pm$ .000 & .125 & .010 $\pm$ .000 & .017 $\pm$ .000 & .012 $\pm$ .000 & .116 & .010 $\pm$ .001 & .018 $\pm$ .001 & .012 $\pm$ .001 & .132 \\
 & \blanchard & .011 $\pm$ .000 & .032 $\pm$ .002 & .013 $\pm$ .000 & 65.017 $\pm$ 11.099 & .010 $\pm$ .000 & .019 $\pm$ .001 & .012 $\pm$ .000 & 1.819 $\pm$ 4.995 & .010 $\pm$ .000 & .016 $\pm$ .001 & .012 $\pm$ .000 & 1.551 $\pm$ 1.635 & .010 $\pm$ .001 & .016 $\pm$ .001 & .012 $\pm$ .001 & .115 $\pm$ .560 \\
 & \catoni & .011 $\pm$ .000 & .028 $\pm$ .001 & .013 $\pm$ .000 & 84.529 $\pm$ 13.023 & .010 $\pm$ .000 & .021 $\pm$ .000 & .012 $\pm$ .000 & 11.910 $\pm$ 5.053 & .010 $\pm$ .000 & .017 $\pm$ .001 & .012 $\pm$ .000 & 1.228 $\pm$ 1.637 & .010 $\pm$ .001 & .017 $\pm$ .001 & .012 $\pm$ .001 & .173 $\pm$ .512 \\
 & \rivasplata & .011 $\pm$ .000 & .026 $\pm$ .001 & .013 $\pm$ .000 & 68.055 $\pm$ 11.606 & .010 $\pm$ .000 & .018 $\pm$ .001 & .012 $\pm$ .000 & 1.637 $\pm$ 4.962 & .010 $\pm$ .000 & .016 $\pm$ .000 & .012 $\pm$ .000 & 1.408 $\pm$ 1.639 & .010 $\pm$ .001 & .016 $\pm$ .001 & .012 $\pm$ .001 & .160 $\pm$ .529 \\
 & \stoNN & \textemdash & .044 & \textemdash & .023 & \textemdash & .043 & \textemdash & .062 & \textemdash & .042 & \textemdash & .058 & \textemdash & .043 & \textemdash & .066 \\
\midrule
\multirow[c]{5}{*}{\rotatebox[origin=c]{90}{\small{Fashion}}} & \ours & .099 $\pm$ .000 & .112 $\pm$ .000 & .098 $\pm$ .000 & .067 & .107 $\pm$ .001 & .115 $\pm$ .001 & .100 $\pm$ .001 & .542 & .098 $\pm$ .002 & .107 $\pm$ .001 & .093 $\pm$ .001 & .353 & .108 $\pm$ .003 & .117 $\pm$ .002 & .102 $\pm$ .002 & .312 \\
 & \blanchard & .099 $\pm$ .000 & .138 $\pm$ .004 & .098 $\pm$ .000 & 61.733 $\pm$ 1.862 & .107 $\pm$ .001 & .119 $\pm$ .003 & .101 $\pm$ .001 & 1.651 $\pm$ 4.230 & .099 $\pm$ .001 & .104 $\pm$ .002 & .094 $\pm$ .001 & 1.342 $\pm$ 1.664 & .108 $\pm$ .003 & .113 $\pm$ .003 & .103 $\pm$ .002 & .143 $\pm$ .534 \\
 & \catoni & .099 $\pm$ .000 & .155 $\pm$ .007 & .098 $\pm$ .000 & 83.929 $\pm$ 12.212 & .107 $\pm$ .001 & .116 $\pm$ .003 & .101 $\pm$ .001 & 11.543 $\pm$ 4.870 & .099 $\pm$ .002 & .103 $\pm$ .002 & .094 $\pm$ .001 & 1.437 $\pm$ 1.594 & .108 $\pm$ .003 & .112 $\pm$ .003 & .103 $\pm$ .002 & .153 $\pm$ .545 \\
 & \rivasplata & .099 $\pm$ .000 & .128 $\pm$ .002 & .098 $\pm$ .000 & 65.737 $\pm$ 11.733 & .107 $\pm$ .001 & .116 $\pm$ .002 & .101 $\pm$ .001 & 1.958 $\pm$ 4.794 & .099 $\pm$ .002 & .105 $\pm$ .002 & .094 $\pm$ .001 & 1.491 $\pm$ 1.618 & .108 $\pm$ .003 & .114 $\pm$ .003 & .103 $\pm$ .002 & .155 $\pm$ .546 \\
 & \stoNN & \textemdash & .161 & \textemdash & .034 & \textemdash & .164 & \textemdash & .271 & \textemdash & .155 & \textemdash & .177 & \textemdash & .166 & \textemdash & .156 \\
\midrule
\multirow[c]{5}{*}{\rotatebox[origin=c]{90}{\small{CIFAR-10}}} & \ours & .277 $\pm$ .000 & .296 $\pm$ .000 & .272 $\pm$ .000 & .016 & .266 $\pm$ .000 & .281 $\pm$ .000 & .257 $\pm$ .000 & .022 & .253 $\pm$ .001 & .272 $\pm$ .000 & .248 $\pm$ .000 & .069 & .236 $\pm$ .001 & .258 $\pm$ .001 & .235 $\pm$ .001 & .118 \\
 & \blanchard & .277 $\pm$ .000 & .399 $\pm$ .006 & .272 $\pm$ .000 & 257.371 $\pm$ 23.327 & .266 $\pm$ .000 & .322 $\pm$ .005 & .257 $\pm$ .000 & 7.190 $\pm$ 11.685 & .253 $\pm$ .001 & .281 $\pm$ .005 & .248 $\pm$ .000 & 15.214 $\pm$ 5.838 & .236 $\pm$ .001 & .255 $\pm$ .002 & .235 $\pm$ .001 & 2.223 $\pm$ 2.016 \\
 & \catoni & .277 $\pm$ .000 & .399 $\pm$ .002 & .272 $\pm$ .000 & 269.048 $\pm$ 23.489 & .266 $\pm$ .000 & .328 $\pm$ .009 & .257 $\pm$ .000 & 81.217 $\pm$ 13.476 & .253 $\pm$ .001 & .275 $\pm$ .004 & .248 $\pm$ .000 & 16.576 $\pm$ 6.087 & .236 $\pm$ .001 & .253 $\pm$ .002 & .235 $\pm$ .001 & 2.248 $\pm$ 2.082 \\
 & \rivasplata & .277 $\pm$ .000 & .362 $\pm$ .004 & .272 $\pm$ .000 & 258.993 $\pm$ 23.725 & .266 $\pm$ .000 & .305 $\pm$ .004 & .257 $\pm$ .000 & 72.737 $\pm$ 12.750 & .253 $\pm$ .001 & .275 $\pm$ .003 & .248 $\pm$ .000 & 15.342 $\pm$ 5.948 & .236 $\pm$ .001 & .255 $\pm$ .002 & .235 $\pm$ .001 & 2.220 $\pm$ 2.180 \\
 & \stoNN & \textemdash & .362 & \textemdash & .008 & \textemdash & .345 & \textemdash & .011 & \textemdash & .336 & \textemdash & .034 & \textemdash & .322 & \textemdash & .059 \\
\midrule
 &  & \multicolumn{4}{c}{$\sigma^2=10^{-6}$} & \multicolumn{4}{c}{$\sigma^2=10^{-5}$} & \multicolumn{4}{c}{$\sigma^2=10^{-4}$} & \multicolumn{4}{c}{$\sigma^2=10^{-3}$} \\
\midrule
 & lr=$10^{-4}$ & $\Risk_{\Tcal}(h)$ & Bnd & $\Risk_{\Scal}(h)$ & Div & $\Risk_{\Tcal}(h)$ & Bnd & $\Risk_{\Scal}(h)$ & Div & $\Risk_{\Tcal}(h)$ & Bnd & $\Risk_{\Scal}(h)$ & Div & $\Risk_{\Tcal}(h)$ & Bnd & $\Risk_{\Scal}(h)$ & Div \\
\midrule
\multirow[c]{5}{*}{\rotatebox[origin=c]{90}{\small{MNIST}}} & \ours & .011 $\pm$ .000 & .025 $\pm$ .000 & .013 $\pm$ .000 & 45.094 & .010 $\pm$ .000 & .019 $\pm$ .000 & .012 $\pm$ .000 & 7.479 & .010 $\pm$ .000 & .018 $\pm$ .000 & .012 $\pm$ .000 & 5.269 & .010 $\pm$ .000 & .018 $\pm$ .001 & .011 $\pm$ .001 & 6.510 \\
 & \blanchard & .011 $\pm$ .000 & .737 $\pm$ .004 & .013 $\pm$ .000 & 11285.050 $\pm$ 147.363 & .010 $\pm$ .000 & .381 $\pm$ .006 & .012 $\pm$ .000 & 3785.071 $\pm$ 85.889 & .010 $\pm$ .000 & .160 $\pm$ .005 & .011 $\pm$ .000 & 1181.043 $\pm$ 46.219 & .009 $\pm$ .000 & .067 $\pm$ .003 & .011 $\pm$ .001 & 34.267 $\pm$ 26.244 \\
 & \catoni & .011 $\pm$ .000 & .547 $\pm$ .004 & .013 $\pm$ .000 & 11965.668 $\pm$ 153.481 & .010 $\pm$ .000 & .283 $\pm$ .004 & .012 $\pm$ .000 & 3811.642 $\pm$ 88.111 & .010 $\pm$ .000 & .120 $\pm$ .004 & .011 $\pm$ .000 & 1212.373 $\pm$ 48.835 & .009 $\pm$ .000 & .050 $\pm$ .002 & .010 $\pm$ .000 & 383.387 $\pm$ 27.059 \\
 & \rivasplata & .011 $\pm$ .000 & .509 $\pm$ .004 & .013 $\pm$ .000 & 11555.623 $\pm$ 15.287 & .010 $\pm$ .000 & .226 $\pm$ .004 & .012 $\pm$ .000 & 3695.054 $\pm$ 9.289 & .009 $\pm$ .000 & .096 $\pm$ .003 & .011 $\pm$ .000 & 1171.892 $\pm$ 47.812 & .009 $\pm$ .000 & .044 $\pm$ .002 & .010 $\pm$ .000 & 343.025 $\pm$ 25.804 \\
 & \stoNN & \textemdash & .050 & \textemdash & 22.547 & \textemdash & .044 & \textemdash & 3.740 & \textemdash & .043 & \textemdash & 2.634 & \textemdash & .043 & \textemdash & 3.255 \\
\midrule
\multirow[c]{5}{*}{\rotatebox[origin=c]{90}{\small{Fashion}}} & \ours & .099 $\pm$ .000 & .116 $\pm$ .000 & .098 $\pm$ .000 & 11.922 & .107 $\pm$ .001 & .117 $\pm$ .001 & .101 $\pm$ .001 & 6.556 & .097 $\pm$ .001 & .109 $\pm$ .001 & .092 $\pm$ .001 & 9.235 & .105 $\pm$ .002 & .118 $\pm$ .002 & .100 $\pm$ .002 & 1.362 \\
 & \blanchard & .098 $\pm$ .000 & .795 $\pm$ .004 & .098 $\pm$ .000 & 10179.790 $\pm$ 138.889 & .101 $\pm$ .001 & .524 $\pm$ .006 & .096 $\pm$ .001 & 3752.748 $\pm$ 9.952 & .095 $\pm$ .001 & .291 $\pm$ .005 & .090 $\pm$ .001 & 1091.018 $\pm$ 47.577 & .104 $\pm$ .002 & .195 $\pm$ .005 & .098 $\pm$ .002 & 309.857 $\pm$ 24.422 \\
 & \catoni & .099 $\pm$ .000 & .808 $\pm$ .002 & .098 $\pm$ .000 & 11999.071 $\pm$ 158.418 & .107 $\pm$ .001 & .425 $\pm$ .006 & .100 $\pm$ .001 & 3817.800 $\pm$ 91.674 & .098 $\pm$ .001 & .235 $\pm$ .004 & .093 $\pm$ .001 & 1216.042 $\pm$ 5.641 & .106 $\pm$ .002 & .182 $\pm$ .004 & .101 $\pm$ .002 & 376.493 $\pm$ 27.018 \\
 & \rivasplata & .098 $\pm$ .000 & .619 $\pm$ .004 & .097 $\pm$ .000 & 10768.160 $\pm$ 146.634 & .099 $\pm$ .001 & .369 $\pm$ .004 & .094 $\pm$ .001 & 3565.270 $\pm$ 88.164 & .094 $\pm$ .001 & .224 $\pm$ .004 & .089 $\pm$ .001 & 1137.876 $\pm$ 48.421 & .103 $\pm$ .002 & .164 $\pm$ .003 & .097 $\pm$ .002 & 318.512 $\pm$ 24.741 \\
 & \stoNN & \textemdash & .164 & \textemdash & 5.961 & \textemdash & .166 & \textemdash & 3.278 & \textemdash & .156 & \textemdash & 4.618 & \textemdash & .166 & \textemdash & 5.181 \\
\midrule
\multirow[c]{5}{*}{\rotatebox[origin=c]{90}{\small{CIFAR-10}}} & \ours & .277 $\pm$ .000 & .303 $\pm$ .000 & .272 $\pm$ .000 & 12.803 & .266 $\pm$ .000 & .282 $\pm$ .000 & .257 $\pm$ .000 & 2.312 & .253 $\pm$ .001 & .272 $\pm$ .000 & .248 $\pm$ .000 & 1.641 & .236 $\pm$ .001 & .259 $\pm$ .001 & .235 $\pm$ .001 & 1.929 \\
 & \blanchard & .277 $\pm$ .000 & .990 $\pm$ .000 & .272 $\pm$ .000 & 2577.092 $\pm$ 236.075 & .266 $\pm$ .000 & .901 $\pm$ .003 & .257 $\pm$ .000 & 8788.732 $\pm$ 134.680 & .253 $\pm$ .001 & .662 $\pm$ .005 & .247 $\pm$ .000 & 2683.054 $\pm$ 73.139 & .235 $\pm$ .001 & .464 $\pm$ .006 & .233 $\pm$ .001 & 85.586 $\pm$ 41.917 \\
 & \catoni & .277 $\pm$ .000 & 1.000 $\pm$ .000 & .272 $\pm$ .000 & 177807.417 $\pm$ 546.892 & .266 $\pm$ .000 & .601 $\pm$ .005 & .257 $\pm$ .000 & 331.757 $\pm$ 83.561 & .253 $\pm$ .001 & .416 $\pm$ .003 & .247 $\pm$ .000 & 85.973 $\pm$ 4.961 & .234 $\pm$ .001 & .369 $\pm$ .003 & .233 $\pm$ .001 & 485.863 $\pm$ 31.335 \\
 & \rivasplata & .277 $\pm$ .000 & .990 $\pm$ .000 & .272 $\pm$ .000 & 48522.489 $\pm$ 309.735 & .266 $\pm$ .000 & .762 $\pm$ .003 & .257 $\pm$ .000 & 850.968 $\pm$ 131.507 & .252 $\pm$ .001 & .542 $\pm$ .004 & .247 $\pm$ .000 & 2696.074 $\pm$ 73.062 & .234 $\pm$ .001 & .393 $\pm$ .004 & .232 $\pm$ .001 & 858.936 $\pm$ 41.972 \\
 & \stoNN & \textemdash & .366 & \textemdash & 6.401 & \textemdash & .346 & \textemdash & 1.156 & \textemdash & .336 & \textemdash & .821 & \textemdash & .322 & \textemdash & .965 \\
\bottomrule
\end{tabular}
}
\label{table:1_prior_0.7}
\end{sidewaystable}

\begin{sidewaystable}
\caption{
\looseness=-1
Comparison of \ours, \rivasplata, \blanchard and \catoni based on the disintegrated bounds, and \stoNN based on the randomized bounds learned with two learning rates lr${\ \in}\{10^{-4}, 10^{-6}\}$ and different variances $\sigma^2{\in}\{10^{-3}, 10^{-4}, 10^{-5}, 10^{-6}\}$.
We report the test risk ($\Risk_{\Tcal}(h)$), the bound value (Bnd), the empirical risk ($\Risk_{\Scal}(h)$), and the divergence (Div) associated with each bound (the Rényi divergence for \ours, the KL divergence for \stoNN, and the disintegrated KL divergence for \rivasplata, \blanchard and \catoni).
More precisely, we report the mean $\pm$ the standard deviation for $400$ neural networks sampled from $\AQ$ for \ours, \rivasplata, \blanchard, and \catoni.
We consider, in this table, that the split ratio is $0.8$.
}
\resizebox{0.83\paperheight}{!}{
\begin{tabular}{rr|clcl|clcl|clcl|clcl}
\toprule
 &  & \multicolumn{4}{c}{$\sigma^2=10^{-6}$} & \multicolumn{4}{c}{$\sigma^2=10^{-5}$} & \multicolumn{4}{c}{$\sigma^2=10^{-4}$} & \multicolumn{4}{c}{$\sigma^2=10^{-3}$} \\
\midrule
 & lr=$10^{-6}$ & $\Risk_{\Tcal}(h)$ & Bnd & $\Risk_{\Scal}(h)$ & Div & $\Risk_{\Tcal}(h)$ & Bnd & $\Risk_{\Scal}(h)$ & Div & $\Risk_{\Tcal}(h)$ & Bnd & $\Risk_{\Scal}(h)$ & Div & $\Risk_{\Tcal}(h)$ & Bnd & $\Risk_{\Scal}(h)$ & Div \\
\midrule
\multirow[c]{5}{*}{\rotatebox[origin=c]{90}{\small{MNIST}}} & \ours & .011 $\pm$ .000 & .020 $\pm$ .000 & .013 $\pm$ .000 & .064 & .008 $\pm$ .000 & .017 $\pm$ .000 & .010 $\pm$ .000 & .050 & .011 $\pm$ .000 & .018 $\pm$ .000 & .011 $\pm$ .000 & .112 & .010 $\pm$ .001 & .016 $\pm$ .001 & .009 $\pm$ .001 & .073 \\
 & \blanchard & .011 $\pm$ .000 & .034 $\pm$ .003 & .013 $\pm$ .000 & 49.248 $\pm$ 1.541 & .008 $\pm$ .000 & .018 $\pm$ .001 & .010 $\pm$ .000 & 8.031 $\pm$ 3.654 & .011 $\pm$ .000 & .016 $\pm$ .001 & .011 $\pm$ .000 & .810 $\pm$ 1.248 & .010 $\pm$ .001 & .014 $\pm$ .001 & .010 $\pm$ .001 & .102 $\pm$ .448 \\
 & \catoni & .011 $\pm$ .000 & .030 $\pm$ .002 & .013 $\pm$ .000 & 66.244 $\pm$ 11.961 & .008 $\pm$ .000 & .018 $\pm$ .001 & .010 $\pm$ .000 & 8.685 $\pm$ 3.987 & .011 $\pm$ .000 & .019 $\pm$ .001 & .011 $\pm$ .000 & 1.011 $\pm$ 1.283 & .010 $\pm$ .001 & .016 $\pm$ .001 & .010 $\pm$ .001 & .131 $\pm$ .422 \\
 & \rivasplata & .011 $\pm$ .000 & .028 $\pm$ .002 & .013 $\pm$ .000 & 5.344 $\pm$ 1.600 & .008 $\pm$ .000 & .017 $\pm$ .001 & .010 $\pm$ .000 & 7.757 $\pm$ 4.187 & .011 $\pm$ .000 & .017 $\pm$ .001 & .011 $\pm$ .000 & .861 $\pm$ 1.361 & .010 $\pm$ .001 & .014 $\pm$ .001 & .010 $\pm$ .001 & .090 $\pm$ .460 \\
 & \stoNN & \textemdash & .046 & \textemdash & .032 & \textemdash & .041 & \textemdash & .025 & \textemdash & .043 & \textemdash & .056 & \textemdash & .040 & \textemdash & .037 \\
\midrule
\multirow[c]{5}{*}{\rotatebox[origin=c]{90}{\small{Fashion}}} & \ours & .103 $\pm$ .000 & .117 $\pm$ .000 & .099 $\pm$ .000 & .068 & .098 $\pm$ .001 & .114 $\pm$ .001 & .096 $\pm$ .001 & .178 & .104 $\pm$ .001 & .117 $\pm$ .002 & .099 $\pm$ .002 & .587 & .107 $\pm$ .004 & .119 $\pm$ .004 & .101 $\pm$ .003 & .328 \\
 & \blanchard & .103 $\pm$ .000 & .144 $\pm$ .004 & .099 $\pm$ .000 & 5.069 $\pm$ 9.537 & .098 $\pm$ .001 & .116 $\pm$ .003 & .096 $\pm$ .001 & 8.105 $\pm$ 3.874 & .104 $\pm$ .001 & .113 $\pm$ .002 & .100 $\pm$ .002 & .990 $\pm$ 1.435 & .109 $\pm$ .004 & .114 $\pm$ .004 & .102 $\pm$ .004 & .102 $\pm$ .444 \\
 & \catoni & .103 $\pm$ .000 & .168 $\pm$ .009 & .099 $\pm$ .000 & 66.761 $\pm$ 1.939 & .098 $\pm$ .001 & .115 $\pm$ .004 & .096 $\pm$ .001 & 8.698 $\pm$ 3.974 & .104 $\pm$ .001 & .112 $\pm$ .002 & .100 $\pm$ .002 & .934 $\pm$ 1.413 & .109 $\pm$ .004 & .113 $\pm$ .004 & .102 $\pm$ .004 & .100 $\pm$ .457 \\
 & \rivasplata & .103 $\pm$ .000 & .132 $\pm$ .003 & .099 $\pm$ .000 & 52.096 $\pm$ 1.745 & .098 $\pm$ .001 & .113 $\pm$ .002 & .096 $\pm$ .001 & 7.820 $\pm$ 4.154 & .104 $\pm$ .001 & .114 $\pm$ .002 & .100 $\pm$ .002 & .939 $\pm$ 1.417 & .108 $\pm$ .004 & .115 $\pm$ .004 & .102 $\pm$ .004 & .100 $\pm$ .464 \\
 & \stoNN & \textemdash & .165 & \textemdash & .034 & \textemdash & .162 & \textemdash & .089 & \textemdash & .166 & \textemdash & .294 & \textemdash & .168 & \textemdash & .164 \\
\midrule
\multirow[c]{5}{*}{\rotatebox[origin=c]{90}{\small{CIFAR-10}}} & \ours & .249 $\pm$ .000 & .265 $\pm$ .000 & .237 $\pm$ .000 & .014 & .247 $\pm$ .000 & .271 $\pm$ .000 & .243 $\pm$ .000 & .018 & .259 $\pm$ .001 & .281 $\pm$ .001 & .252 $\pm$ .001 & .055 & .249 $\pm$ .001 & .274 $\pm$ .001 & .245 $\pm$ .001 & .072 \\
 & \blanchard & .249 $\pm$ .000 & .384 $\pm$ .007 & .237 $\pm$ .000 & 24.108 $\pm$ 22.114 & .247 $\pm$ .000 & .316 $\pm$ .006 & .243 $\pm$ .000 & 59.096 $\pm$ 1.459 & .259 $\pm$ .001 & .289 $\pm$ .006 & .252 $\pm$ .001 & 11.804 $\pm$ 5.001 & .249 $\pm$ .001 & .269 $\pm$ .003 & .245 $\pm$ .001 & 1.578 $\pm$ 1.705 \\
 & \catoni & .249 $\pm$ .000 & .368 $\pm$ .003 & .237 $\pm$ .000 & 27.284 $\pm$ 23.618 & .247 $\pm$ .000 & .337 $\pm$ .012 & .243 $\pm$ .000 & 73.833 $\pm$ 11.692 & .259 $\pm$ .001 & .285 $\pm$ .005 & .252 $\pm$ .001 & 12.808 $\pm$ 5.089 & .249 $\pm$ .001 & .266 $\pm$ .002 & .245 $\pm$ .001 & 1.635 $\pm$ 1.773 \\
 & \rivasplata & .249 $\pm$ .000 & .341 $\pm$ .005 & .237 $\pm$ .000 & 244.258 $\pm$ 22.339 & .247 $\pm$ .000 & .298 $\pm$ .004 & .243 $\pm$ .000 & 61.907 $\pm$ 11.135 & .259 $\pm$ .001 & .283 $\pm$ .003 & .252 $\pm$ .001 & 11.818 $\pm$ 4.923 & .249 $\pm$ .001 & .269 $\pm$ .002 & .245 $\pm$ .001 & 1.629 $\pm$ 1.763 \\
 & \stoNN & \textemdash & .328 & \textemdash & .007 & \textemdash & .334 & \textemdash & .009 & \textemdash & .344 & \textemdash & .028 & \textemdash & .337 & \textemdash & .036 \\
\midrule
 &  & \multicolumn{4}{c}{$\sigma^2=10^{-6}$} & \multicolumn{4}{c}{$\sigma^2=10^{-5}$} & \multicolumn{4}{c}{$\sigma^2=10^{-4}$} & \multicolumn{4}{c}{$\sigma^2=10^{-3}$} \\
\midrule
 & lr=$10^{-4}$ & $\Risk_{\Tcal}(h)$ & Bnd & $\Risk_{\Scal}(h)$ & Div & $\Risk_{\Tcal}(h)$ & Bnd & $\Risk_{\Scal}(h)$ & Div & $\Risk_{\Tcal}(h)$ & Bnd & $\Risk_{\Scal}(h)$ & Div & $\Risk_{\Tcal}(h)$ & Bnd & $\Risk_{\Scal}(h)$ & Div \\
\midrule
\multirow[c]{5}{*}{\rotatebox[origin=c]{90}{\small{MNIST}}} & \ours & .011 $\pm$ .000 & .030 $\pm$ .000 & .013 $\pm$ .000 & 53.875 & .008 $\pm$ .000 & .018 $\pm$ .000 & .010 $\pm$ .000 & 4.369 & .011 $\pm$ .000 & .019 $\pm$ .000 & .011 $\pm$ .000 & 5.063 & .009 $\pm$ .001 & .016 $\pm$ .001 & .009 $\pm$ .001 & 4.854 \\
 & \blanchard & .011 $\pm$ .000 & .828 $\pm$ .004 & .013 $\pm$ .000 & 10014.066 $\pm$ 14.769 & .008 $\pm$ .000 & .491 $\pm$ .007 & .010 $\pm$ .000 & 3707.758 $\pm$ 86.461 & .011 $\pm$ .000 & .211 $\pm$ .007 & .011 $\pm$ .000 & 1151.660 $\pm$ 47.238 & .009 $\pm$ .001 & .076 $\pm$ .005 & .009 $\pm$ .001 & 303.402 $\pm$ 25.072 \\
 & \catoni & .011 $\pm$ .000 & .684 $\pm$ .004 & .013 $\pm$ .000 & 12238.359 $\pm$ 158.595 & .008 $\pm$ .000 & .343 $\pm$ .005 & .010 $\pm$ .000 & 3834.114 $\pm$ 88.516 & .011 $\pm$ .000 & .168 $\pm$ .006 & .011 $\pm$ .000 & 121.777 $\pm$ 48.460 & .009 $\pm$ .001 & .060 $\pm$ .003 & .008 $\pm$ .001 & 356.740 $\pm$ 25.649 \\
 & \rivasplata & .011 $\pm$ .000 & .662 $\pm$ .005 & .013 $\pm$ .000 & 1207.265 $\pm$ 161.842 & .008 $\pm$ .000 & .305 $\pm$ .005 & .010 $\pm$ .000 & 3785.930 $\pm$ 87.976 & .011 $\pm$ .000 & .125 $\pm$ .004 & .010 $\pm$ .000 & 1141.437 $\pm$ 46.910 & .009 $\pm$ .001 & .048 $\pm$ .002 & .008 $\pm$ .000 & 305.573 $\pm$ 23.629 \\
 & \stoNN & \textemdash & .055 & \textemdash & 26.937 & \textemdash & .042 & \textemdash & 2.185 & \textemdash & .044 & \textemdash & 2.532 & \textemdash & .040 & \textemdash & 2.427 \\
\midrule
\multirow[c]{5}{*}{\rotatebox[origin=c]{90}{\small{Fashion}}} & \ours & .102 $\pm$ .000 & .121 $\pm$ .000 & .099 $\pm$ .000 & 1.120 & .098 $\pm$ .001 & .115 $\pm$ .001 & .096 $\pm$ .001 & 3.956 & .102 $\pm$ .001 & .118 $\pm$ .002 & .098 $\pm$ .001 & 7.830 & .103 $\pm$ .003 & .118 $\pm$ .003 & .097 $\pm$ .002 & 8.797 \\
 & \blanchard & .101 $\pm$ .000 & .990 $\pm$ .000 & .098 $\pm$ .000 & 27936.970 $\pm$ 235.840 & .096 $\pm$ .001 & .585 $\pm$ .007 & .094 $\pm$ .001 & 321.105 $\pm$ 81.006 & .098 $\pm$ .001 & .348 $\pm$ .007 & .094 $\pm$ .001 & 1045.641 $\pm$ 44.087 & .101 $\pm$ .002 & .208 $\pm$ .006 & .095 $\pm$ .002 & 273.641 $\pm$ 24.046 \\
 & \catoni & .103 $\pm$ .000 & .865 $\pm$ .002 & .099 $\pm$ .000 & 12143.837 $\pm$ 161.857 & .098 $\pm$ .001 & .536 $\pm$ .008 & .096 $\pm$ .001 & 3802.871 $\pm$ 87.750 & .103 $\pm$ .001 & .286 $\pm$ .006 & .098 $\pm$ .001 & 1202.907 $\pm$ 47.928 & .105 $\pm$ .003 & .191 $\pm$ .005 & .098 $\pm$ .003 & 354.246 $\pm$ 25.507 \\
 & \rivasplata & .102 $\pm$ .000 & .746 $\pm$ .004 & .098 $\pm$ .000 & 11305.448 $\pm$ 149.693 & .096 $\pm$ .001 & .438 $\pm$ .005 & .093 $\pm$ .001 & 3458.977 $\pm$ 83.715 & .097 $\pm$ .001 & .264 $\pm$ .004 & .093 $\pm$ .001 & 1101.567 $\pm$ 44.816 & .099 $\pm$ .002 & .172 $\pm$ .004 & .094 $\pm$ .002 & 285.588 $\pm$ 24.451 \\
 & \stoNN & \textemdash & .168 & \textemdash & 5.060 & \textemdash & .163 & \textemdash & 1.978 & \textemdash & .166 & \textemdash & 3.915 & \textemdash & .166 & \textemdash & 4.399 \\
\midrule
\multirow[c]{5}{*}{\rotatebox[origin=c]{90}{\small{CIFAR-10}}} & \ours & .249 $\pm$ .000 & .274 $\pm$ .000 & .237 $\pm$ .000 & 14.083 & .247 $\pm$ .000 & .273 $\pm$ .000 & .243 $\pm$ .000 & 1.770 & .259 $\pm$ .001 & .282 $\pm$ .001 & .252 $\pm$ .001 & 1.098 & .248 $\pm$ .001 & .275 $\pm$ .001 & .245 $\pm$ .001 & 1.461 \\
 & \blanchard & .249 $\pm$ .000 & .990 $\pm$ .000 & .237 $\pm$ .000 & 26575.507 $\pm$ 218.278 & .247 $\pm$ .000 & .925 $\pm$ .002 & .243 $\pm$ .000 & 7135.143 $\pm$ 117.030 & .259 $\pm$ .001 & .739 $\pm$ .006 & .251 $\pm$ .001 & 2581.211 $\pm$ 74.799 & .247 $\pm$ .001 & .526 $\pm$ .007 & .243 $\pm$ .001 & 831.790 $\pm$ 4.592 \\
 & \catoni & .249 $\pm$ .000 & 1.000 $\pm$ .000 & .237 $\pm$ .000 & 154168.585 $\pm$ 539.590 & .247 $\pm$ .000 & .677 $\pm$ .008 & .243 $\pm$ .000 & 3148.174 $\pm$ 83.069 & .259 $\pm$ .001 & .549 $\pm$ .006 & .252 $\pm$ .001 & 1735.530 $\pm$ 57.888 & .248 $\pm$ .001 & .425 $\pm$ .005 & .244 $\pm$ .001 & 675.780 $\pm$ 38.306 \\
 & \rivasplata & .249 $\pm$ .000 & .990 $\pm$ .000 & .237 $\pm$ .000 & 35062.089 $\pm$ 246.257 & .247 $\pm$ .000 & .824 $\pm$ .003 & .243 $\pm$ .000 & 8092.236 $\pm$ 125.162 & .259 $\pm$ .001 & .610 $\pm$ .005 & .251 $\pm$ .001 & 2652.857 $\pm$ 75.369 & .247 $\pm$ .001 & .441 $\pm$ .005 & .242 $\pm$ .001 & 84.056 $\pm$ 4.952 \\
 & \stoNN & \textemdash & .334 & \textemdash & 7.041 & \textemdash & .335 & \textemdash & .885 & \textemdash & .345 & \textemdash & .549 & \textemdash & .337 & \textemdash & .731 \\
\bottomrule
\end{tabular}
}
\label{table:1_prior_0.8}
\end{sidewaystable}

\begin{sidewaystable}
\caption{
\looseness=-1
Comparison of \ours, \rivasplata, \blanchard and \catoni based on the disintegrated bounds, and \stoNN based on the randomized bounds learned with two learning rates lr${\ \in}\{10^{-4}, 10^{-6}\}$ and different variances $\sigma^2{\in}\{10^{-3}, 10^{-4}, 10^{-5}, 10^{-6}\}$.
We report the test risk ($\Risk_{\Tcal}(h)$), the bound value (Bnd), the empirical risk ($\Risk_{\Scal}(h)$), and the divergence (Div) associated with each bound (the Rényi divergence for \ours, the KL divergence for \stoNN, and the disintegrated KL divergence for \rivasplata, \blanchard and \catoni).
More precisely, we report the mean $\pm$ the standard deviation for $400$ neural networks sampled from $\AQ$ for \ours, \rivasplata, \blanchard, and \catoni.
We consider, in this table, that the split ratio is $0.9$.
}
\resizebox{0.83\paperheight}{!}{
\begin{tabular}{rr|clcl|clcl|clcl|clcl}
\toprule
 &  & \multicolumn{4}{c}{$\sigma^2=10^{-6}$} & \multicolumn{4}{c}{$\sigma^2=10^{-5}$} & \multicolumn{4}{c}{$\sigma^2=10^{-4}$} & \multicolumn{4}{c}{$\sigma^2=10^{-3}$} \\
\midrule
 & lr=$10^{-6}$ & $\Risk_{\Tcal}(h)$ & Bnd & $\Risk_{\Scal}(h)$ & Div & $\Risk_{\Tcal}(h)$ & Bnd & $\Risk_{\Scal}(h)$ & Div & $\Risk_{\Tcal}(h)$ & Bnd & $\Risk_{\Scal}(h)$ & Div & $\Risk_{\Tcal}(h)$ & Bnd & $\Risk_{\Scal}(h)$ & Div \\
\midrule
\multirow[c]{5}{*}{\rotatebox[origin=c]{90}{\small{MNIST}}} & \ours & .008 $\pm$ .000 & .018 $\pm$ .000 & .008 $\pm$ .000 & .029 & .011 $\pm$ .000 & .020 $\pm$ .000 & .010 $\pm$ .000 & .052 & .009 $\pm$ .000 & .018 $\pm$ .001 & .009 $\pm$ .000 & .059 & .008 $\pm$ .000 & .019 $\pm$ .001 & .009 $\pm$ .001 & .023 \\
 & \blanchard & .008 $\pm$ .000 & .033 $\pm$ .004 & .009 $\pm$ .000 & 35.446 $\pm$ 8.610 & .011 $\pm$ .000 & .020 $\pm$ .002 & .010 $\pm$ .000 & 4.933 $\pm$ 2.958 & .009 $\pm$ .000 & .015 $\pm$ .001 & .009 $\pm$ .000 & .490 $\pm$ .960 & .008 $\pm$ .001 & .016 $\pm$ .001 & .009 $\pm$ .001 & .059 $\pm$ .299 \\
 & \catoni & .008 $\pm$ .000 & .026 $\pm$ .002 & .009 $\pm$ .000 & 41.267 $\pm$ 9.234 & .011 $\pm$ .000 & .019 $\pm$ .001 & .010 $\pm$ .000 & 4.564 $\pm$ 3.263 & .009 $\pm$ .000 & .016 $\pm$ .001 & .009 $\pm$ .000 & .581 $\pm$ .989 & .008 $\pm$ .001 & .017 $\pm$ .001 & .009 $\pm$ .001 & .078 $\pm$ .320 \\
 & \rivasplata & .008 $\pm$ .000 & .025 $\pm$ .002 & .009 $\pm$ .000 & 35.856 $\pm$ 8.648 & .011 $\pm$ .000 & .019 $\pm$ .001 & .010 $\pm$ .000 & 4.620 $\pm$ 2.983 & .009 $\pm$ .000 & .015 $\pm$ .001 & .009 $\pm$ .000 & .448 $\pm$ 1.045 & .008 $\pm$ .000 & .016 $\pm$ .001 & .009 $\pm$ .001 & .041 $\pm$ .330 \\
 & \stoNN & \textemdash & .041 & \textemdash & .014 & \textemdash & .045 & \textemdash & .026 & \textemdash & .042 & \textemdash & .030 & \textemdash & .043 & \textemdash & .012 \\
\midrule
\multirow[c]{5}{*}{\rotatebox[origin=c]{90}{\small{Fashion}}} & \ours & .094 $\pm$ .000 & .113 $\pm$ .000 & .089 $\pm$ .000 & .029 & .091 $\pm$ .001 & .119 $\pm$ .001 & .095 $\pm$ .001 & .107 & .092 $\pm$ .002 & .113 $\pm$ .001 & .089 $\pm$ .001 & .097 & .103 $\pm$ .003 & .124 $\pm$ .003 & .099 $\pm$ .003 & .045 \\
 & \blanchard & .094 $\pm$ .000 & .140 $\pm$ .006 & .089 $\pm$ .000 & 32.563 $\pm$ 8.007 & .091 $\pm$ .001 & .119 $\pm$ .004 & .095 $\pm$ .001 & 4.567 $\pm$ 2.912 & .092 $\pm$ .002 & .106 $\pm$ .002 & .089 $\pm$ .001 & .468 $\pm$ 1.101 & .104 $\pm$ .003 & .116 $\pm$ .003 & .099 $\pm$ .003 & .063 $\pm$ .300 \\
 & \catoni & .094 $\pm$ .000 & .146 $\pm$ .002 & .089 $\pm$ .000 & 4.355 $\pm$ 9.121 & .091 $\pm$ .001 & .120 $\pm$ .005 & .095 $\pm$ .001 & 4.895 $\pm$ 3.064 & .092 $\pm$ .002 & .106 $\pm$ .002 & .089 $\pm$ .001 & .473 $\pm$ 1.052 & .103 $\pm$ .003 & .117 $\pm$ .003 & .099 $\pm$ .003 & .079 $\pm$ .319 \\
 & \rivasplata & .094 $\pm$ .000 & .127 $\pm$ .004 & .089 $\pm$ .000 & 33.175 $\pm$ 8.710 & .091 $\pm$ .001 & .117 $\pm$ .002 & .095 $\pm$ .001 & 4.774 $\pm$ 3.003 & .092 $\pm$ .002 & .107 $\pm$ .002 & .089 $\pm$ .001 & .479 $\pm$ .924 & .103 $\pm$ .003 & .118 $\pm$ .003 & .099 $\pm$ .002 & .045 $\pm$ .330 \\
 & \stoNN & \textemdash & .159 & \textemdash & .015 & \textemdash & .166 & \textemdash & .053 & \textemdash & .159 & \textemdash & .048 & \textemdash & .172 & \textemdash & .023 \\
\midrule
\multirow[c]{5}{*}{\rotatebox[origin=c]{90}{\small{CIFAR-10}}} & \ours & .231 $\pm$ .000 & .268 $\pm$ .000 & .228 $\pm$ .000 & .011 & .235 $\pm$ .000 & .267 $\pm$ .000 & .227 $\pm$ .000 & .009 & .218 $\pm$ .001 & .253 $\pm$ .001 & .214 $\pm$ .001 & .024 & .231 $\pm$ .001 & .264 $\pm$ .002 & .224 $\pm$ .002 & .036 \\
 & \blanchard & .231 $\pm$ .000 & .418 $\pm$ .010 & .228 $\pm$ .000 & 193.922 $\pm$ 19.216 & .235 $\pm$ .000 & .312 $\pm$ .009 & .227 $\pm$ .000 & 39.705 $\pm$ 8.929 & .218 $\pm$ .000 & .256 $\pm$ .007 & .214 $\pm$ .001 & 6.919 $\pm$ 3.722 & .231 $\pm$ .001 & .255 $\pm$ .003 & .224 $\pm$ .002 & .878 $\pm$ 1.248 \\
 & \catoni & .231 $\pm$ .000 & .388 $\pm$ .005 & .228 $\pm$ .000 & 255.538 $\pm$ 22.306 & .235 $\pm$ .000 & .337 $\pm$ .003 & .227 $\pm$ .000 & 53.736 $\pm$ 1.302 & .218 $\pm$ .000 & .257 $\pm$ .007 & .214 $\pm$ .001 & 7.060 $\pm$ 3.626 & .231 $\pm$ .001 & .255 $\pm$ .003 & .224 $\pm$ .002 & .857 $\pm$ 1.264 \\
 & \rivasplata & .231 $\pm$ .000 & .364 $\pm$ .007 & .228 $\pm$ .000 & 202.026 $\pm$ 19.688 & .235 $\pm$ .000 & .293 $\pm$ .006 & .227 $\pm$ .000 & 42.458 $\pm$ 9.250 & .218 $\pm$ .001 & .251 $\pm$ .004 & .214 $\pm$ .001 & 6.780 $\pm$ 3.575 & .231 $\pm$ .001 & .256 $\pm$ .002 & .224 $\pm$ .002 & .854 $\pm$ 1.275 \\
 & \stoNN & \textemdash & .328 & \textemdash & .005 & \textemdash & .327 & \textemdash & .005 & \textemdash & .312 & \textemdash & .012 & \textemdash & .324 & \textemdash & .018 \\
\midrule
 &  & \multicolumn{4}{c}{$\sigma^2=10^{-6}$} & \multicolumn{4}{c}{$\sigma^2=10^{-5}$} & \multicolumn{4}{c}{$\sigma^2=10^{-4}$} & \multicolumn{4}{c}{$\sigma^2=10^{-3}$} \\
\midrule
 & lr=$10^{-4}$ & $\Risk_{\Tcal}(h)$ & Bnd & $\Risk_{\Scal}(h)$ & Div & $\Risk_{\Tcal}(h)$ & Bnd & $\Risk_{\Scal}(h)$ & Div & $\Risk_{\Tcal}(h)$ & Bnd & $\Risk_{\Scal}(h)$ & Div & $\Risk_{\Tcal}(h)$ & Bnd & $\Risk_{\Scal}(h)$ & Div \\
\midrule
\multirow[c]{5}{*}{\rotatebox[origin=c]{90}{\small{MNIST}}} & \ours & .008 $\pm$ .000 & .018 $\pm$ .000 & .009 $\pm$ .000 & 2.107 & .011 $\pm$ .000 & .021 $\pm$ .000 & .010 $\pm$ .000 & 1.329 & .008 $\pm$ .000 & .019 $\pm$ .001 & .008 $\pm$ .000 & 3.598 & .008 $\pm$ .001 & .020 $\pm$ .001 & .009 $\pm$ .001 & 4.216 \\
 & \blanchard & .008 $\pm$ .000 & .982 $\pm$ .001 & .008 $\pm$ .000 & 11722.999 $\pm$ 157.452 & .011 $\pm$ .000 & .706 $\pm$ .008 & .010 $\pm$ .000 & 3475.807 $\pm$ 77.708 & .008 $\pm$ .000 & .331 $\pm$ .011 & .008 $\pm$ .000 & 1076.767 $\pm$ 46.059 & .008 $\pm$ .000 & .108 $\pm$ .008 & .009 $\pm$ .001 & 242.819 $\pm$ 23.775 \\
 & \catoni & .008 $\pm$ .000 & 1.000 $\pm$ .000 & .008 $\pm$ .000 & 60838.120 $\pm$ 346.289 & .011 $\pm$ .000 & .515 $\pm$ .007 & .010 $\pm$ .000 & 3728.586 $\pm$ 86.803 & .008 $\pm$ .000 & .243 $\pm$ .007 & .008 $\pm$ .000 & 1166.491 $\pm$ 48.086 & .008 $\pm$ .000 & .087 $\pm$ .006 & .009 $\pm$ .001 & 277.823 $\pm$ 25.431 \\
 & \rivasplata & .008 $\pm$ .000 & .879 $\pm$ .003 & .008 $\pm$ .000 & 12257.175 $\pm$ 152.738 & .010 $\pm$ .000 & .481 $\pm$ .007 & .010 $\pm$ .000 & 3602.529 $\pm$ 78.717 & .008 $\pm$ .000 & .201 $\pm$ .007 & .008 $\pm$ .000 & 1126.882 $\pm$ 47.430 & .008 $\pm$ .001 & .067 $\pm$ .004 & .009 $\pm$ .001 & 242.366 $\pm$ 22.337 \\
 & \stoNN & \textemdash & .042 & \textemdash & 1.053 & \textemdash & .045 & \textemdash & .664 & \textemdash & .042 & \textemdash & 1.799 & \textemdash & .043 & \textemdash & 2.108 \\
\midrule
\multirow[c]{5}{*}{\rotatebox[origin=c]{90}{\small{Fashion}}} & \ours & .094 $\pm$ .000 & .115 $\pm$ .000 & .089 $\pm$ .000 & 2.501 & .091 $\pm$ .001 & .121 $\pm$ .001 & .095 $\pm$ .001 & 2.925 & .092 $\pm$ .002 & .114 $\pm$ .001 & .088 $\pm$ .001 & 3.069 & .102 $\pm$ .002 & .125 $\pm$ .003 & .098 $\pm$ .002 & 3.159 \\
 & \blanchard & .094 $\pm$ .000 & .990 $\pm$ .000 & .089 $\pm$ .000 & 19455.864 $\pm$ 19.460 & .089 $\pm$ .001 & .792 $\pm$ .007 & .093 $\pm$ .001 & 3402.546 $\pm$ 86.590 & .090 $\pm$ .001 & .461 $\pm$ .010 & .087 $\pm$ .001 & 1002.861 $\pm$ 44.393 & .102 $\pm$ .002 & .244 $\pm$ .009 & .098 $\pm$ .002 & 206.177 $\pm$ 2.051 \\
 & \catoni & .094 $\pm$ .000 & 1.000 $\pm$ .000 & .089 $\pm$ .000 & 60888.029 $\pm$ 346.501 & .091 $\pm$ .001 & .813 $\pm$ .012 & .095 $\pm$ .001 & 3756.375 $\pm$ 9.419 & .092 $\pm$ .002 & .390 $\pm$ .010 & .089 $\pm$ .001 & 1161.884 $\pm$ 52.073 & .103 $\pm$ .003 & .215 $\pm$ .007 & .099 $\pm$ .002 & 277.284 $\pm$ 25.479 \\
 & \rivasplata & .094 $\pm$ .000 & .990 $\pm$ .000 & .089 $\pm$ .000 & 27137.315 $\pm$ 227.934 & .088 $\pm$ .001 & .597 $\pm$ .007 & .093 $\pm$ .001 & 3371.321 $\pm$ 86.352 & .090 $\pm$ .001 & .331 $\pm$ .007 & .086 $\pm$ .001 & 1003.481 $\pm$ 48.362 & .101 $\pm$ .002 & .195 $\pm$ .006 & .097 $\pm$ .002 & 207.442 $\pm$ 21.896 \\
 & \stoNN & \textemdash & .160 & \textemdash & 1.250 & \textemdash & .167 & \textemdash & 1.463 & \textemdash & .160 & \textemdash & 1.535 & \textemdash & .172 & \textemdash & 1.579 \\
\midrule
\multirow[c]{5}{*}{\rotatebox[origin=c]{90}{\small{CIFAR-10}}} & \ours & .231 $\pm$ .000 & .279 $\pm$ .000 & .228 $\pm$ .000 & 12.925 & .235 $\pm$ .000 & .268 $\pm$ .000 & .227 $\pm$ .000 & 1.371 & .218 $\pm$ .001 & .254 $\pm$ .001 & .214 $\pm$ .001 & .715 & .231 $\pm$ .001 & .264 $\pm$ .002 & .224 $\pm$ .002 & 1.019 \\
 & \blanchard & .231 $\pm$ .000 & .990 $\pm$ .000 & .228 $\pm$ .000 & 26032.808 $\pm$ 222.475 & .235 $\pm$ .000 & .986 $\pm$ .001 & .227 $\pm$ .000 & 6875.633 $\pm$ 112.137 & .217 $\pm$ .000 & .831 $\pm$ .006 & .214 $\pm$ .001 & 2292.053 $\pm$ 68.347 & .230 $\pm$ .001 & .606 $\pm$ .010 & .222 $\pm$ .001 & 76.644 $\pm$ 39.246 \\
 & \catoni & .231 $\pm$ .000 & 1.000 $\pm$ .000 & .228 $\pm$ .000 & 17684.651 $\pm$ 576.711 & .235 $\pm$ .000 & .980 $\pm$ .000 & .227 $\pm$ .000 & 8265.727 $\pm$ 123.941 & .218 $\pm$ .000 & .834 $\pm$ .011 & .214 $\pm$ .001 & 2664.069 $\pm$ 73.915 & .231 $\pm$ .001 & .517 $\pm$ .009 & .224 $\pm$ .002 & 85.593 $\pm$ 41.022 \\
 & \rivasplata & .231 $\pm$ .000 & .988 $\pm$ .001 & .228 $\pm$ .000 & 14284.846 $\pm$ 169.166 & .235 $\pm$ .000 & .919 $\pm$ .002 & .227 $\pm$ .000 & 7121.350 $\pm$ 114.645 & .217 $\pm$ .000 & .699 $\pm$ .006 & .213 $\pm$ .001 & 2502.412 $\pm$ 68.728 & .229 $\pm$ .001 & .494 $\pm$ .007 & .221 $\pm$ .001 & 776.237 $\pm$ 39.540 \\
 & \stoNN & \textemdash & .335 & \textemdash & 6.462 & \textemdash & .328 & \textemdash & .685 & \textemdash & .313 & \textemdash & .358 & \textemdash & .324 & \textemdash & .510 \\
\bottomrule
\end{tabular}
}
\label{table:1_prior_0.9}
\end{sidewaystable}

\begin{sidewaystable}
\caption{
Comparison of the bound values before performing Step {\bf 2)} of our Training Method for \ours, \rivasplata, \blanchard and \catoni. 
More precisely, for each split and each variance $\sigma^2{\in}\{10^{-3}, 10^{-4}, 10^{-5}, 10^{-6}\}$, we report the mean $\pm$ the standard deviation (for $400$ neural networks sampled from $\P$) of the test risk ($\Risk_{\Tcal}(h)$), the empirical risk ($\Risk_{\Scal}(h)$), and the value of the bounds of Corollaries~\ref{corollary:nn} and~\ref{corollary:nn-rbc}.
We consider in this table that the dataset is MNIST.
}
\begin{subtable}[!ht]{0.50\textwidth}
    \centering
    \resizebox{0.40\paperheight}{!}{
    \begin{tabular}{cccccccc}
\toprule
 & Split & $\Risk_{\Tcal}(h)$ & $\Risk_{\Scal}(h)$ & Cor.~\ref{corollary:nn} & Eq.~\eqref{eq:nn-rivasplata} & Eq.~\eqref{eq:nn-blanchard} & Eq.~\eqref{eq:nn-catoni} \\
\midrule
\multirow[c]{10}{*}{\rotatebox[origin=c]{90}{\small{$\sigma^2=10^{-6}$}}} & .0 & .901 $\pm$ .002 & .901 $\pm$ .002 & .908 $\pm$ .002 & .906 $\pm$ .002 & .905 $\pm$ .002 & .906 $\pm$ .002 \\
 & .1 & .035 $\pm$ .000 & .039 $\pm$ .000 & .045 $\pm$ .000 & .043 $\pm$ .000 & .043 $\pm$ .000 & .042 $\pm$ .000 \\
 & .2 & .016 $\pm$ .000 & .019 $\pm$ .000 & .023 $\pm$ .000 & .022 $\pm$ .000 & .022 $\pm$ .000 & .022 $\pm$ .000 \\
 & .3 & .012 $\pm$ .000 & .013 $\pm$ .000 & .017 $\pm$ .000 & .016 $\pm$ .000 & .015 $\pm$ .000 & .015 $\pm$ .000 \\
 & .4 & .010 $\pm$ .000 & .013 $\pm$ .000 & .017 $\pm$ .000 & .016 $\pm$ .000 & .016 $\pm$ .000 & .016 $\pm$ .000 \\
 & .5 & .008 $\pm$ .000 & .010 $\pm$ .000 & .015 $\pm$ .000 & .013 $\pm$ .000 & .013 $\pm$ .000 & .014 $\pm$ .000 \\
 & .6 & .008 $\pm$ .000 & .010 $\pm$ .000 & .014 $\pm$ .000 & .013 $\pm$ .000 & .013 $\pm$ .000 & .014 $\pm$ .000 \\
 & .7 & .011 $\pm$ .000 & .013 $\pm$ .000 & .019 $\pm$ .000 & .017 $\pm$ .000 & .017 $\pm$ .000 & .018 $\pm$ .000 \\
 & .8 & .011 $\pm$ .000 & .013 $\pm$ .000 & .020 $\pm$ .000 & .018 $\pm$ .000 & .018 $\pm$ .000 & .020 $\pm$ .000 \\
 & .9 & .008 $\pm$ .000 & .009 $\pm$ .000 & .018 $\pm$ .000 & .015 $\pm$ .000 & .014 $\pm$ .000 & .015 $\pm$ .000 \\
\midrule
\multirow[c]{10}{*}{\rotatebox[origin=c]{90}{\small{$\sigma^2=10^{-5}$}}} & .0 & .897 $\pm$ .013 & .897 $\pm$ .012 & .904 $\pm$ .012 & .902 $\pm$ .012 & .902 $\pm$ .012 & .903 $\pm$ .012 \\
 & .1 & .024 $\pm$ .000 & .030 $\pm$ .001 & .035 $\pm$ .001 & .034 $\pm$ .001 & .033 $\pm$ .001 & .033 $\pm$ .001 \\
 & .2 & .015 $\pm$ .000 & .019 $\pm$ .000 & .023 $\pm$ .000 & .022 $\pm$ .000 & .021 $\pm$ .000 & .021 $\pm$ .000 \\
 & .3 & .009 $\pm$ .000 & .011 $\pm$ .000 & .015 $\pm$ .000 & .014 $\pm$ .000 & .013 $\pm$ .000 & .013 $\pm$ .000 \\
 & .4 & .012 $\pm$ .000 & .014 $\pm$ .000 & .018 $\pm$ .000 & .017 $\pm$ .000 & .017 $\pm$ .000 & .017 $\pm$ .000 \\
 & .5 & .006 $\pm$ .000 & .009 $\pm$ .000 & .012 $\pm$ .000 & .011 $\pm$ .000 & .011 $\pm$ .000 & .012 $\pm$ .000 \\
 & .6 & .007 $\pm$ .000 & .009 $\pm$ .000 & .014 $\pm$ .000 & .013 $\pm$ .000 & .012 $\pm$ .000 & .013 $\pm$ .000 \\
 & .7 & .010 $\pm$ .000 & .012 $\pm$ .000 & .018 $\pm$ .000 & .016 $\pm$ .000 & .016 $\pm$ .000 & .017 $\pm$ .000 \\
 & .8 & .008 $\pm$ .000 & .010 $\pm$ .000 & .017 $\pm$ .000 & .015 $\pm$ .000 & .014 $\pm$ .000 & .017 $\pm$ .000 \\
 & .9 & .011 $\pm$ .000 & .010 $\pm$ .000 & .020 $\pm$ .000 & .017 $\pm$ .000 & .017 $\pm$ .000 & .018 $\pm$ .000 \\
\bottomrule
\end{tabular}
}
\end{subtable}
\begin{subtable}[!ht]{0.40\textwidth}
    \centering
    \resizebox{0.40\paperheight}{!}{
    \begin{tabular}{cccccccc}
\toprule
 & Split & $\Risk_{\Tcal}(h)$ & $\Risk_{\Scal}(h)$ & Cor.~\ref{corollary:nn} & Eq.~\eqref{eq:nn-rivasplata} & Eq.~\eqref{eq:nn-blanchard} & Eq.~\eqref{eq:nn-catoni} \\
\midrule
\multirow[c]{10}{*}{\rotatebox[origin=c]{90}{\small{$\sigma^2=10^{-4}$}}} & .0 & .898 $\pm$ .017 & .898 $\pm$ .017 & .905 $\pm$ .016 & .903 $\pm$ .016 & .902 $\pm$ .016 & .903 $\pm$ .016 \\
 & .1 & .035 $\pm$ .003 & .039 $\pm$ .002 & .045 $\pm$ .002 & .044 $\pm$ .002 & .043 $\pm$ .002 & .043 $\pm$ .002 \\
 & .2 & .015 $\pm$ .001 & .016 $\pm$ .001 & .020 $\pm$ .001 & .019 $\pm$ .001 & .019 $\pm$ .001 & .019 $\pm$ .001 \\
 & .3 & .012 $\pm$ .000 & .016 $\pm$ .000 & .020 $\pm$ .001 & .019 $\pm$ .001 & .019 $\pm$ .001 & .019 $\pm$ .001 \\
 & .4 & .009 $\pm$ .000 & .011 $\pm$ .000 & .015 $\pm$ .000 & .014 $\pm$ .000 & .014 $\pm$ .000 & .014 $\pm$ .000 \\
 & .5 & .008 $\pm$ .000 & .010 $\pm$ .000 & .015 $\pm$ .000 & .013 $\pm$ .000 & .013 $\pm$ .000 & .014 $\pm$ .000 \\
 & .6 & .008 $\pm$ .000 & .009 $\pm$ .000 & .013 $\pm$ .000 & .012 $\pm$ .000 & .012 $\pm$ .000 & .013 $\pm$ .000 \\
 & .7 & .010 $\pm$ .000 & .012 $\pm$ .000 & .017 $\pm$ .000 & .016 $\pm$ .000 & .015 $\pm$ .000 & .016 $\pm$ .000 \\
 & .8 & .011 $\pm$ .000 & .011 $\pm$ .000 & .018 $\pm$ .000 & .016 $\pm$ .000 & .016 $\pm$ .000 & .018 $\pm$ .000 \\
 & .9 & .009 $\pm$ .000 & .009 $\pm$ .000 & .018 $\pm$ .001 & .015 $\pm$ .001 & .015 $\pm$ .001 & .016 $\pm$ .001 \\
\midrule
\multirow[c]{10}{*}{\rotatebox[origin=c]{90}{\small{$\sigma^2=10^{-3}$}}} & .0 & .903 $\pm$ .014 & .902 $\pm$ .014 & .909 $\pm$ .013 & .907 $\pm$ .013 & .907 $\pm$ .013 & .907 $\pm$ .013 \\
 & .1 & .041 $\pm$ .005 & .045 $\pm$ .005 & .050 $\pm$ .005 & .049 $\pm$ .005 & .048 $\pm$ .005 & .048 $\pm$ .005 \\
 & .2 & .020 $\pm$ .002 & .022 $\pm$ .002 & .026 $\pm$ .002 & .025 $\pm$ .002 & .025 $\pm$ .002 & .024 $\pm$ .002 \\
 & .3 & .014 $\pm$ .001 & .015 $\pm$ .001 & .019 $\pm$ .001 & .018 $\pm$ .001 & .018 $\pm$ .001 & .018 $\pm$ .001 \\
 & .4 & .015 $\pm$ .001 & .016 $\pm$ .001 & .021 $\pm$ .001 & .020 $\pm$ .001 & .019 $\pm$ .001 & .019 $\pm$ .001 \\
 & .5 & .015 $\pm$ .001 & .015 $\pm$ .001 & .020 $\pm$ .001 & .019 $\pm$ .001 & .018 $\pm$ .001 & .018 $\pm$ .001 \\
 & .6 & .008 $\pm$ .000 & .010 $\pm$ .000 & .014 $\pm$ .001 & .013 $\pm$ .001 & .012 $\pm$ .000 & .013 $\pm$ .000 \\
 & .7 & .010 $\pm$ .001 & .012 $\pm$ .001 & .018 $\pm$ .001 & .016 $\pm$ .001 & .016 $\pm$ .001 & .017 $\pm$ .001 \\
 & .8 & .010 $\pm$ .001 & .010 $\pm$ .001 & .016 $\pm$ .001 & .014 $\pm$ .001 & .014 $\pm$ .001 & .016 $\pm$ .001 \\
 & .9 & .008 $\pm$ .000 & .009 $\pm$ .001 & .019 $\pm$ .001 & .016 $\pm$ .001 & .015 $\pm$ .001 & .017 $\pm$ .001 \\
\bottomrule
\end{tabular}
    }
\end{subtable}
\label{table:2_data_mnist}
\end{sidewaystable}

\begin{sidewaystable}
\caption{
Comparison of the bound values before performing Step {\bf 2)} of our Training Method for \ours, \rivasplata, \blanchard and \catoni. 
More precisely, for each split and each variance $\sigma^2{\in}\{10^{-3}, 10^{-4}, 10^{-5}, 10^{-6}\}$, we report the mean $\pm$ the standard deviation (for $400$ neural networks sampled from $\P$) of the test risk ($\Risk_{\Tcal}(h)$), the empirical risk ($\Risk_{\Scal}(h)$), and the value of the bounds of Corollaries~\ref{corollary:nn} and~\ref{corollary:nn-rbc}.
We consider in this table that the dataset is Fashion-MNIST.
}
\begin{subtable}[!ht]{0.50\textwidth}
    \centering
    \resizebox{0.40\paperheight}{!}{
    \begin{tabular}{cccccccc}
\toprule
 & Split & $\Risk_{\Tcal}(h)$ & $\Risk_{\Scal}(h)$ & Cor.~\ref{corollary:nn} & Eq.~\eqref{eq:nn-rivasplata} & Eq.~\eqref{eq:nn-blanchard} & Eq.~\eqref{eq:nn-catoni} \\
\midrule
\multirow[c]{10}{*}{\rotatebox[origin=c]{90}{\small{$\sigma^2=10^{-6}$}}} & .0 & .970 $\pm$ .028 & .970 $\pm$ .027 & .972 $\pm$ .025 & .971 $\pm$ .025 & .971 $\pm$ .026 & .972 $\pm$ .026 \\
 & .1 & .166 $\pm$ .001 & .159 $\pm$ .000 & .169 $\pm$ .000 & .167 $\pm$ .000 & .166 $\pm$ .000 & .167 $\pm$ .000 \\
 & .2 & .168 $\pm$ .002 & .160 $\pm$ .001 & .170 $\pm$ .001 & .168 $\pm$ .001 & .167 $\pm$ .001 & .168 $\pm$ .001 \\
 & .3 & .126 $\pm$ .000 & .124 $\pm$ .000 & .134 $\pm$ .000 & .132 $\pm$ .000 & .131 $\pm$ .000 & .131 $\pm$ .000 \\
 & .4 & .118 $\pm$ .001 & .112 $\pm$ .000 & .123 $\pm$ .000 & .120 $\pm$ .000 & .119 $\pm$ .000 & .119 $\pm$ .000 \\
 & .5 & .106 $\pm$ .000 & .101 $\pm$ .000 & .113 $\pm$ .000 & .110 $\pm$ .000 & .109 $\pm$ .000 & .109 $\pm$ .000 \\
 & .6 & .109 $\pm$ .000 & .102 $\pm$ .000 & .115 $\pm$ .000 & .112 $\pm$ .000 & .110 $\pm$ .000 & .110 $\pm$ .000 \\
 & .7 & .099 $\pm$ .000 & .098 $\pm$ .000 & .112 $\pm$ .000 & .109 $\pm$ .000 & .108 $\pm$ .000 & .107 $\pm$ .000 \\
 & .8 & .103 $\pm$ .000 & .099 $\pm$ .000 & .117 $\pm$ .000 & .112 $\pm$ .000 & .111 $\pm$ .000 & .110 $\pm$ .000 \\
 & .9 & .094 $\pm$ .000 & .089 $\pm$ .000 & .113 $\pm$ .000 & .107 $\pm$ .000 & .105 $\pm$ .000 & .106 $\pm$ .000 \\
\midrule
\multirow[c]{10}{*}{\rotatebox[origin=c]{90}{\small{$\sigma^2=10^{-5}$}}} & .0 & .945 $\pm$ .038 & .945 $\pm$ .037 & .949 $\pm$ .035 & .948 $\pm$ .035 & .948 $\pm$ .036 & .948 $\pm$ .036 \\
 & .1 & .158 $\pm$ .001 & .151 $\pm$ .001 & .161 $\pm$ .001 & .159 $\pm$ .001 & .158 $\pm$ .001 & .159 $\pm$ .001 \\
 & .2 & .157 $\pm$ .003 & .151 $\pm$ .003 & .162 $\pm$ .003 & .159 $\pm$ .003 & .158 $\pm$ .003 & .159 $\pm$ .003 \\
 & .3 & .126 $\pm$ .001 & .121 $\pm$ .001 & .131 $\pm$ .001 & .128 $\pm$ .001 & .127 $\pm$ .001 & .128 $\pm$ .001 \\
 & .4 & .114 $\pm$ .001 & .107 $\pm$ .001 & .118 $\pm$ .001 & .115 $\pm$ .001 & .114 $\pm$ .001 & .114 $\pm$ .001 \\
 & .5 & .104 $\pm$ .001 & .099 $\pm$ .000 & .110 $\pm$ .000 & .108 $\pm$ .000 & .107 $\pm$ .000 & .106 $\pm$ .000 \\
 & .6 & .115 $\pm$ .001 & .104 $\pm$ .001 & .117 $\pm$ .001 & .114 $\pm$ .001 & .113 $\pm$ .001 & .112 $\pm$ .001 \\
 & .7 & .107 $\pm$ .001 & .101 $\pm$ .001 & .115 $\pm$ .001 & .111 $\pm$ .001 & .110 $\pm$ .001 & .109 $\pm$ .001 \\
 & .8 & .098 $\pm$ .001 & .096 $\pm$ .001 & .114 $\pm$ .001 & .109 $\pm$ .001 & .108 $\pm$ .001 & .107 $\pm$ .001 \\
 & .9 & .091 $\pm$ .001 & .095 $\pm$ .001 & .119 $\pm$ .001 & .113 $\pm$ .001 & .111 $\pm$ .001 & .112 $\pm$ .001 \\
\bottomrule
\end{tabular}
}
\end{subtable}
\begin{subtable}[!ht]{0.40\textwidth}
    \centering
    \resizebox{0.40\paperheight}{!}{
    \begin{tabular}{cccccccc}
\toprule
 & Split & $\Risk_{\Tcal}(h)$ & $\Risk_{\Scal}(h)$ & Cor.~\ref{corollary:nn} & Eq.~\eqref{eq:nn-rivasplata} & Eq.~\eqref{eq:nn-blanchard} & Eq.~\eqref{eq:nn-catoni} \\
\midrule
\multirow[c]{10}{*}{\rotatebox[origin=c]{90}{\small{$\sigma^2=10^{-4}$}}} & .0 & .912 $\pm$ .027 & .912 $\pm$ .027 & .918 $\pm$ .026 & .916 $\pm$ .027 & .916 $\pm$ .027 & .916 $\pm$ .026 \\
 & .1 & .164 $\pm$ .003 & .154 $\pm$ .003 & .164 $\pm$ .003 & .162 $\pm$ .003 & .161 $\pm$ .003 & .162 $\pm$ .004 \\
 & .2 & .164 $\pm$ .009 & .160 $\pm$ .009 & .170 $\pm$ .010 & .168 $\pm$ .010 & .167 $\pm$ .010 & .168 $\pm$ .010 \\
 & .3 & .125 $\pm$ .002 & .119 $\pm$ .002 & .129 $\pm$ .002 & .126 $\pm$ .002 & .126 $\pm$ .002 & .126 $\pm$ .002 \\
 & .4 & .119 $\pm$ .003 & .113 $\pm$ .003 & .124 $\pm$ .003 & .121 $\pm$ .003 & .120 $\pm$ .003 & .120 $\pm$ .003 \\
 & .5 & .109 $\pm$ .002 & .102 $\pm$ .001 & .113 $\pm$ .001 & .110 $\pm$ .001 & .109 $\pm$ .001 & .109 $\pm$ .001 \\
 & .6 & .102 $\pm$ .001 & .096 $\pm$ .001 & .109 $\pm$ .001 & .105 $\pm$ .001 & .105 $\pm$ .001 & .104 $\pm$ .001 \\
 & .7 & .099 $\pm$ .002 & .094 $\pm$ .001 & .108 $\pm$ .001 & .104 $\pm$ .001 & .103 $\pm$ .001 & .102 $\pm$ .001 \\
 & .8 & .104 $\pm$ .001 & .100 $\pm$ .002 & .118 $\pm$ .002 & .113 $\pm$ .002 & .112 $\pm$ .002 & .111 $\pm$ .002 \\
 & .9 & .092 $\pm$ .002 & .089 $\pm$ .001 & .113 $\pm$ .001 & .107 $\pm$ .001 & .105 $\pm$ .001 & .106 $\pm$ .001 \\
\midrule
\multirow[c]{10}{*}{\rotatebox[origin=c]{90}{\small{$\sigma^2=10^{-3}$}}} & .0 & .899 $\pm$ .026 & .899 $\pm$ .027 & .906 $\pm$ .026 & .904 $\pm$ .026 & .904 $\pm$ .026 & .905 $\pm$ .025 \\
 & .1 & .178 $\pm$ .006 & .170 $\pm$ .006 & .181 $\pm$ .006 & .178 $\pm$ .006 & .177 $\pm$ .006 & .179 $\pm$ .006 \\
 & .2 & .164 $\pm$ .006 & .159 $\pm$ .006 & .169 $\pm$ .006 & .167 $\pm$ .006 & .166 $\pm$ .006 & .167 $\pm$ .006 \\
 & .3 & .143 $\pm$ .007 & .138 $\pm$ .007 & .148 $\pm$ .007 & .146 $\pm$ .007 & .145 $\pm$ .007 & .145 $\pm$ .007 \\
 & .4 & .133 $\pm$ .005 & .129 $\pm$ .005 & .140 $\pm$ .005 & .137 $\pm$ .005 & .137 $\pm$ .005 & .137 $\pm$ .005 \\
 & .5 & .122 $\pm$ .004 & .117 $\pm$ .004 & .129 $\pm$ .004 & .126 $\pm$ .004 & .125 $\pm$ .004 & .125 $\pm$ .004 \\
 & .6 & .111 $\pm$ .003 & .104 $\pm$ .003 & .117 $\pm$ .003 & .114 $\pm$ .003 & .113 $\pm$ .003 & .112 $\pm$ .003 \\
 & .7 & .109 $\pm$ .003 & .103 $\pm$ .003 & .118 $\pm$ .003 & .114 $\pm$ .003 & .113 $\pm$ .003 & .112 $\pm$ .003 \\
 & .8 & .108 $\pm$ .004 & .102 $\pm$ .004 & .120 $\pm$ .004 & .115 $\pm$ .004 & .114 $\pm$ .004 & .113 $\pm$ .004 \\
 & .9 & .103 $\pm$ .003 & .099 $\pm$ .002 & .124 $\pm$ .003 & .118 $\pm$ .003 & .116 $\pm$ .003 & .116 $\pm$ .003 \\
\bottomrule
\end{tabular}
    }
\end{subtable}
\label{table:2_data_fashion}
\end{sidewaystable}

\begin{sidewaystable}
\caption{
Comparison of the bound values before performing Step {\bf 2)} of our Training Method for \ours, \rivasplata, \blanchard and \catoni. 
More precisely, for each split and each variance $\sigma^2{\in}\{10^{-3}, 10^{-4}, 10^{-5}, 10^{-6}\}$, we report the mean $\pm$ the standard deviation (for $400$ neural networks sampled from $\P$) of the test risk ($\Risk_{\Tcal}(h)$), the empirical risk ($\Risk_{\Scal}(h)$), and the value of the bounds of Corollaries~\ref{corollary:nn} and~\ref{corollary:nn-rbc}.
We consider in this table that the dataset is CIFAR-10.
}
\begin{subtable}[!ht]{0.50\textwidth}
    \centering
    \resizebox{0.40\paperheight}{!}{
    \begin{tabular}{cccccccc}
\toprule
 & Split & $\Risk_{\Tcal}(h)$ & $\Risk_{\Scal}(h)$ & Cor.~\ref{corollary:nn} & Eq.~\eqref{eq:nn-rivasplata} & Eq.~\eqref{eq:nn-blanchard} & Eq.~\eqref{eq:nn-catoni} \\
\midrule
\multirow[c]{10}{*}{\rotatebox[origin=c]{90}{\small{$\sigma^2=10^{-6}$}}} & .0 & .899 $\pm$ .000 & .899 $\pm$ .000 & .906 $\pm$ .000 & .904 $\pm$ .000 & .903 $\pm$ .000 & .904 $\pm$ .000 \\
 & .1 & .476 $\pm$ .000 & .470 $\pm$ .000 & .486 $\pm$ .000 & .482 $\pm$ .000 & .481 $\pm$ .000 & .485 $\pm$ .000 \\
 & .2 & .390 $\pm$ .000 & .389 $\pm$ .000 & .406 $\pm$ .000 & .402 $\pm$ .000 & .401 $\pm$ .000 & .404 $\pm$ .000 \\
 & .3 & .370 $\pm$ .000 & .358 $\pm$ .000 & .374 $\pm$ .000 & .371 $\pm$ .000 & .370 $\pm$ .000 & .372 $\pm$ .000 \\
 & .4 & .334 $\pm$ .000 & .328 $\pm$ .000 & .346 $\pm$ .000 & .342 $\pm$ .000 & .341 $\pm$ .000 & .342 $\pm$ .000 \\
 & .5 & .307 $\pm$ .000 & .302 $\pm$ .000 & .321 $\pm$ .000 & .317 $\pm$ .000 & .316 $\pm$ .000 & .317 $\pm$ .000 \\
 & .6 & .274 $\pm$ .000 & .276 $\pm$ .000 & .297 $\pm$ .000 & .293 $\pm$ .000 & .291 $\pm$ .000 & .291 $\pm$ .000 \\
 & .7 & .275 $\pm$ .000 & .272 $\pm$ .000 & .296 $\pm$ .000 & .290 $\pm$ .000 & .289 $\pm$ .000 & .288 $\pm$ .000 \\
 & .8 & .249 $\pm$ .000 & .237 $\pm$ .000 & .265 $\pm$ .000 & .259 $\pm$ .000 & .257 $\pm$ .000 & .256 $\pm$ .000 \\
 & .9 & .227 $\pm$ .000 & .230 $\pm$ .000 & .269 $\pm$ .000 & .260 $\pm$ .000 & .258 $\pm$ .000 & .258 $\pm$ .000 \\
\midrule
\multirow[c]{10}{*}{\rotatebox[origin=c]{90}{\small{$\sigma^2=10^{-5}$}}} & .0 & .899 $\pm$ .001 & .899 $\pm$ .000 & .906 $\pm$ .000 & .904 $\pm$ .000 & .904 $\pm$ .000 & .904 $\pm$ .000 \\
 & .1 & .476 $\pm$ .000 & .478 $\pm$ .000 & .494 $\pm$ .000 & .490 $\pm$ .000 & .489 $\pm$ .000 & .493 $\pm$ .000 \\
 & .2 & .403 $\pm$ .000 & .398 $\pm$ .000 & .414 $\pm$ .000 & .410 $\pm$ .000 & .409 $\pm$ .000 & .412 $\pm$ .000 \\
 & .3 & .349 $\pm$ .000 & .350 $\pm$ .000 & .367 $\pm$ .000 & .363 $\pm$ .000 & .362 $\pm$ .000 & .364 $\pm$ .000 \\
 & .4 & .322 $\pm$ .000 & .313 $\pm$ .000 & .330 $\pm$ .000 & .327 $\pm$ .000 & .326 $\pm$ .000 & .327 $\pm$ .000 \\
 & .5 & .281 $\pm$ .000 & .283 $\pm$ .000 & .302 $\pm$ .000 & .298 $\pm$ .000 & .297 $\pm$ .000 & .297 $\pm$ .000 \\
 & .6 & .290 $\pm$ .000 & .286 $\pm$ .000 & .307 $\pm$ .000 & .303 $\pm$ .000 & .301 $\pm$ .000 & .301 $\pm$ .000 \\
 & .7 & .266 $\pm$ .000 & .257 $\pm$ .000 & .281 $\pm$ .000 & .276 $\pm$ .000 & .274 $\pm$ .000 & .274 $\pm$ .000 \\
 & .8 & .247 $\pm$ .000 & .243 $\pm$ .000 & .271 $\pm$ .000 & .265 $\pm$ .000 & .263 $\pm$ .000 & .262 $\pm$ .000 \\
 & .9 & .236 $\pm$ .000 & .227 $\pm$ .000 & .266 $\pm$ .000 & .257 $\pm$ .000 & .255 $\pm$ .000 & .255 $\pm$ .000 \\
\bottomrule
\end{tabular}
}
\end{subtable}
\begin{subtable}[!ht]{0.40\textwidth}
    \centering
    \resizebox{0.40\paperheight}{!}{
    \begin{tabular}{cccccccc}
\toprule
 & Split & $\Risk_{\Tcal}(h)$ & $\Risk_{\Scal}(h)$ & Cor.~\ref{corollary:nn} & Eq.~\eqref{eq:nn-rivasplata} & Eq.~\eqref{eq:nn-blanchard} & Eq.~\eqref{eq:nn-catoni}\\
\midrule
\multirow[c]{10}{*}{\rotatebox[origin=c]{90}{\small{$\sigma^2=10^{-4}$}}} & .0 & .900 $\pm$ .004 & .900 $\pm$ .003 & .907 $\pm$ .003 & .905 $\pm$ .003 & .905 $\pm$ .003 & .905 $\pm$ .003 \\
 & .1 & .458 $\pm$ .001 & .464 $\pm$ .001 & .479 $\pm$ .001 & .476 $\pm$ .001 & .475 $\pm$ .001 & .478 $\pm$ .001 \\
 & .2 & .395 $\pm$ .001 & .396 $\pm$ .000 & .412 $\pm$ .000 & .409 $\pm$ .000 & .408 $\pm$ .000 & .411 $\pm$ .000 \\
 & .3 & .361 $\pm$ .001 & .361 $\pm$ .000 & .378 $\pm$ .000 & .375 $\pm$ .000 & .373 $\pm$ .000 & .376 $\pm$ .000 \\
 & .4 & .323 $\pm$ .001 & .316 $\pm$ .000 & .334 $\pm$ .000 & .330 $\pm$ .000 & .329 $\pm$ .000 & .331 $\pm$ .000 \\
 & .5 & .296 $\pm$ .001 & .291 $\pm$ .000 & .310 $\pm$ .000 & .306 $\pm$ .000 & .304 $\pm$ .000 & .305 $\pm$ .000 \\
 & .6 & .271 $\pm$ .001 & .263 $\pm$ .000 & .284 $\pm$ .000 & .279 $\pm$ .000 & .278 $\pm$ .000 & .278 $\pm$ .000 \\
 & .7 & .253 $\pm$ .001 & .246 $\pm$ .000 & .270 $\pm$ .000 & .265 $\pm$ .000 & .263 $\pm$ .000 & .262 $\pm$ .000 \\
 & .8 & .259 $\pm$ .001 & .252 $\pm$ .001 & .281 $\pm$ .001 & .275 $\pm$ .001 & .273 $\pm$ .001 & .272 $\pm$ .001 \\
 & .9 & .217 $\pm$ .000 & .216 $\pm$ .001 & .255 $\pm$ .001 & .246 $\pm$ .001 & .243 $\pm$ .001 & .244 $\pm$ .001 \\
\midrule
\multirow[c]{10}{*}{\rotatebox[origin=c]{90}{\small{$\sigma^2=10^{-3}$}}} & .0 & .905 $\pm$ .012 & .904 $\pm$ .012 & .911 $\pm$ .011 & .909 $\pm$ .011 & .909 $\pm$ .011 & .909 $\pm$ .011 \\
 & .1 & .479 $\pm$ .002 & .480 $\pm$ .001 & .496 $\pm$ .001 & .493 $\pm$ .001 & .491 $\pm$ .001 & .495 $\pm$ .001 \\
 & .2 & .415 $\pm$ .002 & .415 $\pm$ .001 & .432 $\pm$ .001 & .428 $\pm$ .001 & .427 $\pm$ .001 & .430 $\pm$ .001 \\
 & .3 & .417 $\pm$ .001 & .416 $\pm$ .001 & .434 $\pm$ .001 & .430 $\pm$ .001 & .429 $\pm$ .001 & .431 $\pm$ .001 \\
 & .4 & .333 $\pm$ .001 & .323 $\pm$ .001 & .341 $\pm$ .001 & .337 $\pm$ .001 & .336 $\pm$ .001 & .338 $\pm$ .001 \\
 & .5 & .316 $\pm$ .001 & .311 $\pm$ .001 & .331 $\pm$ .001 & .327 $\pm$ .001 & .325 $\pm$ .001 & .326 $\pm$ .001 \\
 & .6 & .280 $\pm$ .001 & .281 $\pm$ .001 & .302 $\pm$ .001 & .298 $\pm$ .001 & .296 $\pm$ .001 & .296 $\pm$ .001 \\
 & .7 & .239 $\pm$ .001 & .234 $\pm$ .001 & .257 $\pm$ .001 & .252 $\pm$ .001 & .250 $\pm$ .001 & .250 $\pm$ .001 \\
 & .8 & .249 $\pm$ .001 & .245 $\pm$ .001 & .274 $\pm$ .001 & .268 $\pm$ .001 & .266 $\pm$ .001 & .264 $\pm$ .001 \\
 & .9 & .233 $\pm$ .001 & .232 $\pm$ .002 & .272 $\pm$ .002 & .263 $\pm$ .002 & .260 $\pm$ .002 & .260 $\pm$ .002 \\
\bottomrule
\end{tabular}
    }
\end{subtable}
\label{table:2_data_cifar10}
\end{sidewaystable}

\end{appendices}

\end{document}